\documentclass[10pt,journal,compsoc]{IEEEtran}
\ifCLASSOPTIONcompsoc
  \usepackage[nocompress]{cite}
\else
  \usepackage{cite}
\fi
\ifCLASSINFOpdf
   \usepackage[pdftex]{graphicx}
   \graphicspath{{fig}}
\else
   \usepackage[dvips]{graphicx}
\fi
\usepackage[export]{adjustbox}
\usepackage{amsmath}
\usepackage{amssymb}
\usepackage{array}
\usepackage{booktabs}

\ifCLASSOPTIONcompsoc
 \usepackage[caption=false,font=footnotesize,labelfont=sf,textfont=sf, justification=justified]{subfig}
\else
 \usepackage[caption=false,font=footnotesize, justification=justified]{subfig}
\fi
\usepackage[font=footnotesize,labelfont=sf,
   justification=justified,]{caption} 
\usepackage{fixltx2e}

\usepackage{stfloats}
\usepackage{url}

\usepackage{dcolumn}
\usepackage{multirow}
\newcolumntype{d}[1]{D{.}{.}{#1}}
\usepackage{hyperref}

\newcommand{\modelname}{AvatarMe}
\newcommand{\modelnameplus}{AvatarMe\textsuperscript{++}}
\newcommand{\datasetname}{RealFaceDB}
\newcommand{\threedmm}{3DMM}
\newcommand{\threed}{3D}

\newcommand{\renderready}{render-ready}

\makeatletter
\long\def\@IEEEtitleabstractindextextbox#1{\parbox{0.922\textwidth}{#1}}
\makeatother

\begin{document}
%
\title{\modelnameplus{}: Facial Shape and BRDF Inference 
with Photorealistic Rendering-Aware GANs}
%
%
%

\author{
    Alexandros Lattas, \and
    Stylianos Moschoglou, \and
    Stylianos Ploumpis, \and\protect\\
    Baris Gecer, \and
    Abhijeet Ghosh, \and
    Stefanos Zafeiriou\thanks{
        All authors are with the Department of Computing, 
        Imperial College London, South Kensington Campus, London SW7 2AZ, UK.
        }\thanks{
        Corresponding author is A.L. (a.lattas@imperial.ac.uk). Other emails available at \protect{\url{https://ibug.doc.ic.ac.uk/people}}.
        }\thanks{
        The dataset, project page and supplemental materials are available at\protect\\
        \protect{\url{https://github.com/lattas/avatarme}}.
        }
}


%
%

\markboth{IEEE Transactions on Pattern Analysis and Machine Intelligence}%
{Shell \MakeLowercase{\textit{et al.}}: Bare Demo of IEEEtran.cls for Computer Society Journals}
%


\IEEEtitleabstractindextext{%
\begin{abstract}
Over the last years, 
with the advent of Generative Adversarial Networks (GANs), 
many face analysis tasks have accomplished astounding performance, 
with applications including, but not limited to, face generation 
and 3D face reconstruction from a single ``in-the-wild'' image. 
Nevertheless, to the best of our knowledge, there is no method
which can produce render-ready high-resolution 3D faces
from ``in-the-wild'' images and this can be attributed to the:
(a) scarcity of available data for training, 
and (b) lack of robust methodologies that can successfully be applied on very
high-resolution data. 
In this paper, we introduce the first method that is able 
to reconstruct photorealistic render-ready 3D facial
geometry and BRDF from a single ``in-the-wild'' image.
To achieve this,
we capture a large dataset of facial shape and reflectance, which we have made public.
Moreover, we define a fast and photorealistic differentiable rendering methodology
with accurate facial skin diffuse and specular reflection, 
self-occlusion and subsurface scattering approximation.
With this, we train a network that disentangles the facial diffuse and specular reflectance components
from a mesh and texture with baked illumination,
scanned or reconstructed with a 3DMM fitting method.
As we demonstrate in a series of qualitative and quantitative experiments, 
our method outperforms the existing arts by a significant margin and reconstructs authentic,
4K by 6K-resolution 3D faces from a single low-resolution
image, that are ready to be rendered in various applications
and bridge the uncanny valley.
\end{abstract}

}

\IEEEdisplaynontitleabstractindextext

%
\IEEEpeerreviewmaketitle
\maketitle
\begin{figure*}[htp!]
    \centering
    \includegraphics[width=\linewidth]{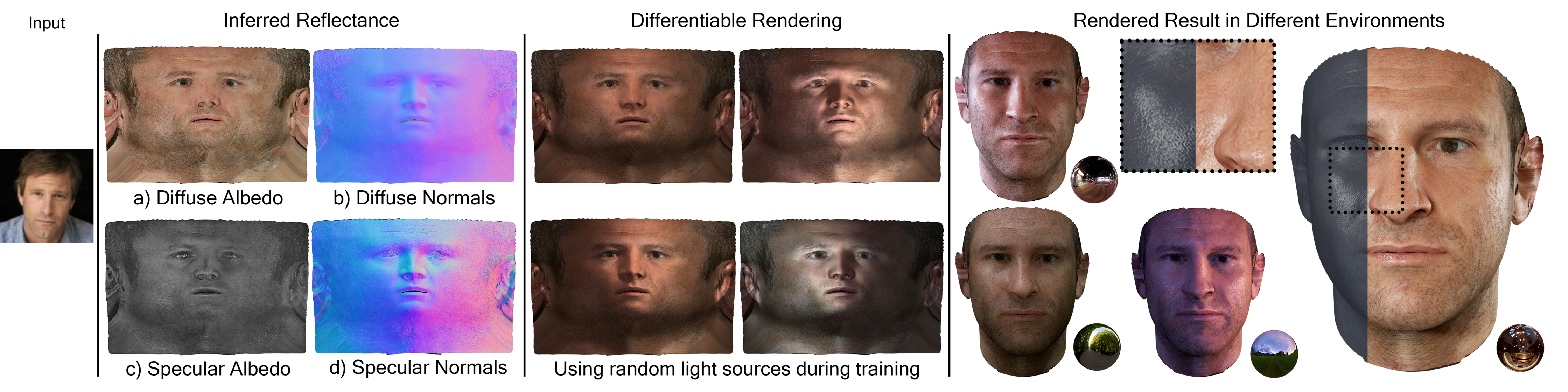}
    \caption{
        From left to right:
        Input image from LFW \cite{huang2008labeled};
        AvatarMe\textsuperscript{++} predicted reflectance (diffuse albedo, diffuse normals,
        specular albedo and specular normals);
        Rendered predictions with our photorealistic differentiable rendering;
        Rendered reconstruction in different environments.
    }
    \label{fig:intro}
\end{figure*}
\IEEEraisesectionheading{\section{Introduction}}
\label{sec:introduction}
\IEEEPARstart{3D} face reconstruction from a single image is one of the most popular and well-studied problems 
in the intersection of computer vision, graphics and machine learning. 
Apart from its countless applications,
it demonstrates the power of recent developments
in scanning, learning, and synthesizing \threed{} objects
\cite{blanz_morphable_1999, zhou_dense_2019}. 
Recently, mainly due to the advent of deep learning, 
tremendous progress has been made in 3D face reconstruction from images captured even in arbitrary recording conditions 
(also referred to as ``in-the-wild'') 
\cite{gecer_ganfit_2019, sela_unrestricted_2017, tewari_self-supervised_2018, tran_learning_2019}. 
Nevertheless, even though the geometry can be inferred somewhat accurately, 
in order to render a reconstructed face in arbitrary virtual environments, 
much more information than a \threed{} smooth geometry is required, 
i.e., skin reflectance as well as high-frequency surface normals. 
In this paper, we propose a meticulously designed pipeline 
for the reconstruction of high-resolution \renderready{} faces from ``in-the-wild'' images
captured in arbitrary poses, lighting conditions, and occlusions. 
A result from our pipeline is showcased in Fig.~\ref{fig:intro}. 

The seminal work in the field is the \threed{} Morphable Model (\threedmm{}) fitting algorithm~\cite{blanz_morphable_1999}. 
The facial texture and shape that is reconstructed by the \threedmm{} algorithm 
always lies in a space that is spanned by a linear basis,
which is learned by Principal Component Analysis (PCA). 
The linear basis, 
even though remarkable in representing the basic characteristics of the reconstructed face, 
fails in reconstructing high-frequency details in texture and geometry. 
Furthermore, the PCA model fails in representing the complex structure of facial texture captured in ``in-the-wild'' conditions. 
Therefore, \threedmm{} fitting usually fails in ``in-the-wild'' images. 
In the years that followed, \threedmm{} fitting was extended so that it could use a PCA model on robust features, 
i.e., Histogram of Oriented Gradients (HOGs) \cite{dalal2005histogram}, 
for representing facial texture \cite{booth_3d_2016},
with improved results in ``in-the-wild'' images.
The recently proposed, Morphable Face Albedo model \cite{smith_morphable_2020}
additionally reconstructs diffuse and specular albedo with PCA.
Nevertheless, these methods cannot reconstruct high-resolution facial textures.
Finally, the non-linear facial and head albedo and normals models
\cite{gecer_synthesizing_2020, li_learning_2020},
generate high-resolution facial textures but have not been shown in ``in-the-wild'' fitting.

With the advent of deep learning, 
many regression methods using an encoder-decoder structure 
have been proposed to infer \threed{} geometry, reflectance and illumination
\cite{chen_photo-realistic_2019, gecer_ganfit_2019, sela_unrestricted_2017, shu_neural_2017, tewari_self-supervised_2018, tran_learning_2019, zhou_dense_2019}.
Some of the methods demonstrate that it is possible to reconstruct shape and texture, 
even in real-time on a CPU \cite{zhou_dense_2019}.
However, the methods \cite{sela_unrestricted_2017, shu_neural_2017, tewari_self-supervised_2018, tran_learning_2019, zhou_dense_2019}
fail to reconstruct highly-detailed texture and shape,
due to various factors
such as the use of basic reflectance models 
(e.g., the Lambertian reflectance model), 
the use of synthetic data, or mesh-convolutions on colored meshes.
Their results are not \renderready{},
and cannot be used directly in industrial rendering applications
for photorealistic results.
Furthermore, in many of the above methods,
the reconstructed texture and shape lose 
many of the identity characteristics of the original image.

Arguably, the first generic method which demonstrated 
that it is possible to reconstruct high-quality texture and shape 
from single ``in-the-wild'' images 
is the recently proposed GANFIT method \cite{gecer_ganfit_2019}. 
GANFIT can be described as an extension of the original \threedmm{} fitting strategy 
but with the following differences: 
(a) instead of a PCA texture model, it uses a GAN \cite{karras_progressive_2018} trained on high-resolution UV-maps, 
and (b) in order to preserve the identity in the reconstructed texture and shape, 
it uses features from a state-of-the-art face recognition network \cite{deng_arcface_2019}.
However, the reconstructed texture and shape is not \renderready{} due to 
(a) the texture containing baked illumination, 
and (b) not being able to reconstruct high-frequency normals or specular reflectance. 

Inverse rendering of an ``in-the-wild'' image or 3DMM-reconstructed texture
to acquire its shape and its Bidirectional Reflectance Distribution Function (BRDF) parameters
is an ill-posed problem and hence statistical priors are needed.
Numerous works have proposed the use differentiable rendering loss while reconstructing 3DMMs 
\cite{gecer_ganfit_2019, tewari_fml_2019, shu_neural_2017, Zhu_2020_CVPR, genova_unsupervised_2018}.
However none of them photorealistically render
the reconstructed face and acquire its relfectance properties,
but project the reconstructed shape and use a simplistic shading model.
This is mostly because of the low availability in facial reflectance data,
and the inherent challenges in differentiable rendering.
Recent works in differentiable rendering
\cite{nikhila_ravi_pytorch3d_2020, genova_unsupervised_2018}
enables us to capitalize on our large reflectance dataset
to implement a fast photorealistic facial differentiable rendering framework
and use it in reconstructing high-resolution facial shapes and BRDF.

Early attempts to infer photorealistic \renderready{} information 
from single ``in-the-wild'' images have been made in some works
\cite{chen_photo-realistic_2019, huynh_mesoscopic_2018, saito_photorealistic_2017, yamaguchi_high-fidelity_2018}.
Arguably, some of the results showcased in the above papers are of decent quality.
Nevertheless, the methods do not generalize since
they directly manipulate and augment 
the low-quality and potentially occluded input facial texture.
As a result,
the quality of the final reconstruction always depends on the input image.
Even more, the employed \threed{} model may not be very representative, 
and a very small number of subjects 
(e.g., $25$ \cite{yamaguchi_high-fidelity_2018}, $122$ \cite{chen_photo-realistic_2019})
were available for training for the high-frequency details of the face. 
While closest to our work, 
these approaches focus on easily creating a digital avatar 
rather than high-quality \renderready{} face reconstruction 
from ``in-the-wild'' images, which is the goal of our work. 

We present an elaborate methodology for high-quality \threed{} facial geometry and reflectance reconstruction
from a single ``in-the-wild'' image.
In particularly, we collect a big dataset of facial reflectance,
and use an end-to-end reflectance inference network with a photorealistic differentiable rendering loss.
Our method builds upon recent reconstruction methods (e.g.,~GANFIT \cite{gecer_ganfit_2019})
and applies super-resolution and domain-adaption algorithms
to GAN-generated generated high-quality facial textures.
We show that this methodology is superior to the previous state-of-the-art (e.g.~\cite{chen_photo-realistic_2019, yamaguchi_high-fidelity_2018}),
who apply algorithms for high-frequency estimation of the original input, 
which could be of low quality and are affected by environment illumination.
We demonstrate that it is possible to produce \renderready{} 
faces from arbitrary faces (pose, occlusion, etc.) including paintings,
which can be realistically re-rendered in any environment.
Specifically, our contributions are:
\begin{itemize}
    \item A dataset of facial reflectance and geometry collected 
    using state-of-the-art methods from over 200 subjects, 
    which is now available to the community.
    \item A differentiable rendering framework
    that fully utilizes both diffuse and specular reflectance data,
    and enables the fast approximation of subsurface scattering
    and occlusion shadows.
    \item An image-translation network that
    transforms a facial geometry with baked-in illumination
    to diffuse and specular albedo and normals using the above differentiable rendering framework.
    \item An end-to-end algorithm for reconstructing 
    high-resolution 3D faces
    including their shape and BRDF,
    from a single ``in-the-wild'' image.
\end{itemize}

This work is an extension of \modelname \cite{lattas_avatarme_2020}.
Compared to the conference paper, \modelnameplus{} adds:
a) a photorealistic differentiable rendering method;
b) online data augmentation of the training data in randomized illumination environments;
c) a single image-translation network for BRDF inference, that utilizes a stochastic rendering loss,
geometrical and global information;
d) an extensive ablation study and experiments for the aforementioned additions.

\section{Related Work}
\label{sec:related_work}
\subsection{Facial Geometry and Reflectance Capture}
\label{sec:related_work_capture}
Debevec et al.~\cite{debevec_acquiring_2000} first proposed to employ
a specialized light stage setup to acquire a reflectance field 
of a human face for photorealistic image-based relighting. 
Weyrich et al.~\cite{weyrich_analysis_2006} used an LED sphere 
and 16 cameras to densely record facial reflectance 
and computed view-independent estimates of facial reflectance 
from the acquired data, including per-pixel diffuse and specular albedos, 
and per-region specular roughness parameters. 
These initial works required cumbersome and impractical dense capturing.

Ma et al.~\cite{ma_rapid_2007} introduced polarized spherical gradient illumination 
(using an LED sphere) for efficient acquisition of separated diffuse and specular albedos 
and photometric normals of a face using just eight photographs.
They demonstrated high quality facial geometry, 
including skin mesostructure as well as realistic rendering with the acquired data. 
However, the method was restricted to frontal viewpoint acquisition,
as the polarization pattern used on the LED sphere was view-dependent.
Subsequently, Ghosh et al.~\cite{ghosh_multiview_2011} extended polarized spherical gradient
illumination for multi-view facial acquisition 
by employing two orthogonal spherical polarization patterns. 
This allows capturing separated diffuse and specular reflectance 
and photometric normals from any viewpoint around the equator of the LED sphere.
Until today, it can be considered the state-of-the art in terms of high quality facial capture.
In the recent years, significant progress has also been made
in passive facial capture,
from high quality facial geometry capture~\cite{beeler_coupled_2012} 
to even detailed facial appearance estimation~\cite{gotardo_practical_2018}. 
However, the quality of the acquired data with such methods 
is lower compared to active illumination techniques.

Recently, Kampouris et al.~\cite{kampouris_diffuse-specular_2018} 
demonstrated how to utilize unpolarized binary spherical gradient illumination 
for estimating separated diffuse and specular albedo and photometric normals 
using color-space analysis. 
The method does not require polarization 
and hence needs half the number of photographs 
compared to polarized spherical gradients.
Moreover, it enables completely view-independent reflectance separation, 
making it faster and more robust for high quality facial capture~\cite{lattas_multi-view_2019}.
For our work, we use the unpolarised active illumination-based multi-view facial capture method~\cite{kampouris_diffuse-specular_2018,lattas_multi-view_2019}
for acquiring high quality facial reflectance data in order to build our training data.

\subsection{Facial Geometry and Texture Estimation}
\label{sec:related_work_3dmm}
Over the years, numerous methods have been introduced in the literature 
that tackle the problem of \threed{} facial reconstruction from a single input image \cite{blanz_morphable_1999, bfm09, booth_3d_2016, egger_3d_2020, tuan_tran_regressing_2017, cole_synthesizing_2017, guo_cnn-based_2019, richardson20163d}.
Early methods required a statistical \threedmm{}
both for shape and appearance,
usually encoded in a low dimensional space constructed by PCA \cite{blanz_morphable_1999, booth_3d_2016, bfm09}.
A 3DMM is typically fitted on a 2D image
using an energy based cost optimization with respect to
the model's identity and expression parameters
as well as the parameters of the camera and scene illumination,
as thoroughly explained in the \threedmm{} review of \cite{egger_3d_2020}.
Moreover, many approaches have tried to leverage the power of Convolutional Neural Networks (CNNs) 
to either regress the latent parameters of a PCA model \cite{tuan_tran_regressing_2017, cole_synthesizing_2017}
or utilize a \threedmm{} to synthesize images 
and formulate an image-to-image translation problem using CNNs \cite{guo_cnn-based_2019, richardson20163d}.
Similar works have also modeled complete head topologies \cite{FLAME:SiggraphAsia2017, Ploumpis_2019_CVPR, ploumpis_towards_2020, luo2021normalized}.
Finally, the recent Morphable Face Albedo Model \cite{smith_morphable_2020}
separately models diffuse and specular albedo with a PCA model.

\subsection{Image-to-Image Translation}
\label{sec:related_work_translation}
Image-to-image translation refers to the task of translating an input image 
to a designated target domain (e.g., turning sketches into images, or day into night scenes). 
With the introduction of GANs \cite{goodfellow_generative_2014}, 
image-to-image translation improved dramatically \cite{isola_image--image_2017, zhu_unpaired_2017}. Recently, with the increasing capabilities in the hardware, 
image-to-image translation has also been successfully attempted 
in high-resolution data \cite{wang_high-resolution_2018}.
In this work, we improve on pix2pixHD~\cite{wang_high-resolution_2018},
by building an image-translation network that incorporates
the photorealistic differentiable rendering of its results,
so that a rendering loss can be introduced.
Our model successfully learns to disentangle relfectance components
from textures with rendered illumination and occlusion shadows.

\subsection{Differentiable Rendering}
\label{sec:related_work_rendering}
Multiple works in the past have attempted
to differentiate rendering models to solve inverse rendering problems
with limited success \cite{patow_survey_2003}.
OpenDR \cite{loper_opendr_2014}
was the first complete framework in Python
for differentiable rendering,
built using an auto-differentiation framework.
Neural 3D Mesh Renderer \cite{kato_neural_2018}
introduced a rasterization approximation which enables differentiation
for non-occluded gradients.
TF Mesh Renderer \cite{genova_unsupervised_2018} introduced a differentiable rasterizer
which interpolates per-vertex attributes in view-space,
using positive and negative barycentric coordinates to overcome discontinuities.
SoftRasterizer \cite{liu_soft_2019}
introduced a fully differentiable rendering framework,
by using a probabilistic rasterization function
with an aggregation function for z-buffering.
It significantly improves the gradient flow 
over \cite{loper_opendr_2014, kato_neural_2018}
and can be used with differentiable local-illumination models.
Similarly, \cite{chen_learning_2019}
separate the fore- and background rasterization
and use barycentric coordinates to propagate the gradient only for the foreground pixels.
For global illumination models,
Li et al.~\cite{li_differentiable_2018}
introduced a differentiable ray tracer,
which uses an edge-sampling-based method to provide a continuous rendering function
and importance sampling to improve on performance.
Moreover, Loubet et al.~\cite{loubet_reparameterizing_2019}
introduced spherical rotations that remove the discontinuities, 
with respect to visibility, cameras, lights, and geometry.

Several complete frameworks exist that combine deep learning with differentiable rendering.
TF Mesh Renderer \cite{genova_unsupervised_2018}
is integrated with Tensorflow,
which also includes a library for graphics and differentiable rendering.
Kaolin \cite{jatavallabhula_kaolin_2019}
is a library of Pytorch implementations
including \cite{shu_neural_2017, liu_soft_2019}.
Finally, Pytorch3D \cite{nikhila_ravi_pytorch3d_2020}
is a complete modular differentiable rendering framework, based on SoftRasterizer,
with additional modules for shading, performance, and compatibility improvements.
To the best of our knowledge, our method is the first
to show fast photorealistic differentiable rendering for human skin.
We extend Pytorch3D framework \cite{nikhila_ravi_pytorch3d_2020}
for accurate facial skin diffuse and specular reflection, 
self-occlusion and subsurface scattering approximation, and integrate it with a high-resolution image-translation GAN.

\subsection{Facial BRDF Estimation}
\label{sec:related_work_svbrdf_estimation}
Many approaches have been successful in acquiring the reflectance of materials from a single image,
using deep networks with an encoder-decoder architecture
\cite{deschaintre_single-image_2018, li_modeling_2017, li_materials_2018, asselin2020deep}.
However, they only explore planar surfaces in a constrained environment,
typically assuming a single point-light source.
Similar principles have also been successfully applied
to ``in-the-wild'' outdoor images \cite{yu2019inverserendernet}.
Early applications on human faces \cite{sengupta_sfsnet_2018, shu_neural_2017}
used image translation networks to infer facial reflection from an ``in-the-wild'' image,
producing low-resolution results.
Recent approaches attempt to incorporate additional facial normal and displacement mappings
resulting in representations with high frequency details \cite{chen_photo-realistic_2019}.
Although this method demonstrates impressive results in geometry inference,
it tends to fail in conditions with challenging illumination 
and extreme head poses, and does not produce re-lightable results.
Saito et al.~\cite{saito_photorealistic_2017} proposed a deep learning approach 
for data-driven inference of high resolution facial texture map of an entire face 
for realistic rendering,
using an input of a single low-resolution face image with partial facial coverage.
This has been extended to inference of facial mesostructure, given a diffuse albedo
\cite{huynh_mesoscopic_2018},
and even complete facial reflectance and displacement maps besides albedo texture, given a partial facial image as input~\cite{yamaguchi_high-fidelity_2018}.
The above methods are the most related to our work
and achieve the creation of digital avatars from ``in-the-wild'' images.
In this work, we show high quality facial reflectance reconstruction,
from such images and existing models and datasets.

Various alternative paradigms that produce renderable human faces
have been recently proposed.
\cite{dib2021practical} introduce a coarse-to-fine optimization
utilizing differentiable ray-tracing for facial geometry and albedo reconstruction.
\cite{R_2021_CVPR} reconstruct neural face reflectance fields 
that enable rendering with complex physical effects.
Finally, \cite{bi2021deep} generate dynamic head textures and shapes
with an encoder-decoder, that are relightable and animatable
from VR-headset views.

The state-of-the-art facial \threedmm{} fitting method GANFIT \cite{gecer_ganfit_2019}
uses a GAN-generated texture with an iterative optimization method
of the \threedmm{}'s weights. Each iteration utilizes lambertian differentiable rendering
and a deep face recognition network, achieving high quality texture
with fine identity characteristics.
In this work, we use an image-translation network
that learns to disentangle the reflectance components reconstructed from a \threedmm{} fitting method,
guided by our high-resolution photorealistic differentiable rendering loss.
Our facial geometry and spatially-varying BRDF textures
are high-resolution and ready to be rendered with high-quality photorealistic results.

\section{Data Acquisition}
\label{sec:method_data}
\subsection{Training Data Capturing Setup}
\label{sec:method_data_capturing_setup}
\begin{figure}[h]
    \centering
    \captionsetup[subfigure]{labelformat=empty}
    \captionsetup[subfloat]{justification=centering}
    \subfloat[
        Diff.Alb.~$\mathbf{A_D}$]{
        \includegraphics[width=0.19\linewidth]{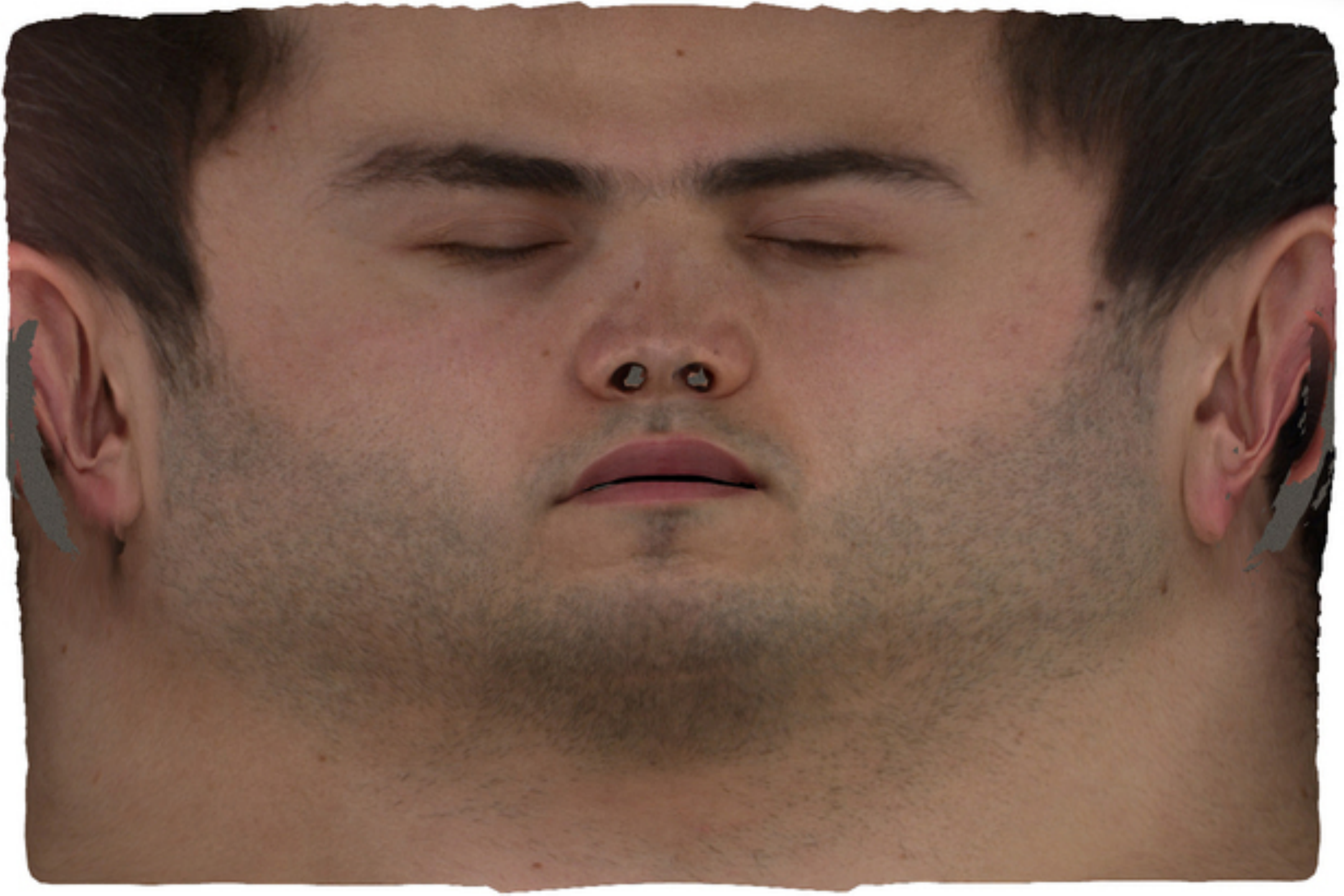}}
    \subfloat[
        Spec.Alb.~$\mathbf{A_S}$]{
        \includegraphics[width=0.19\linewidth]{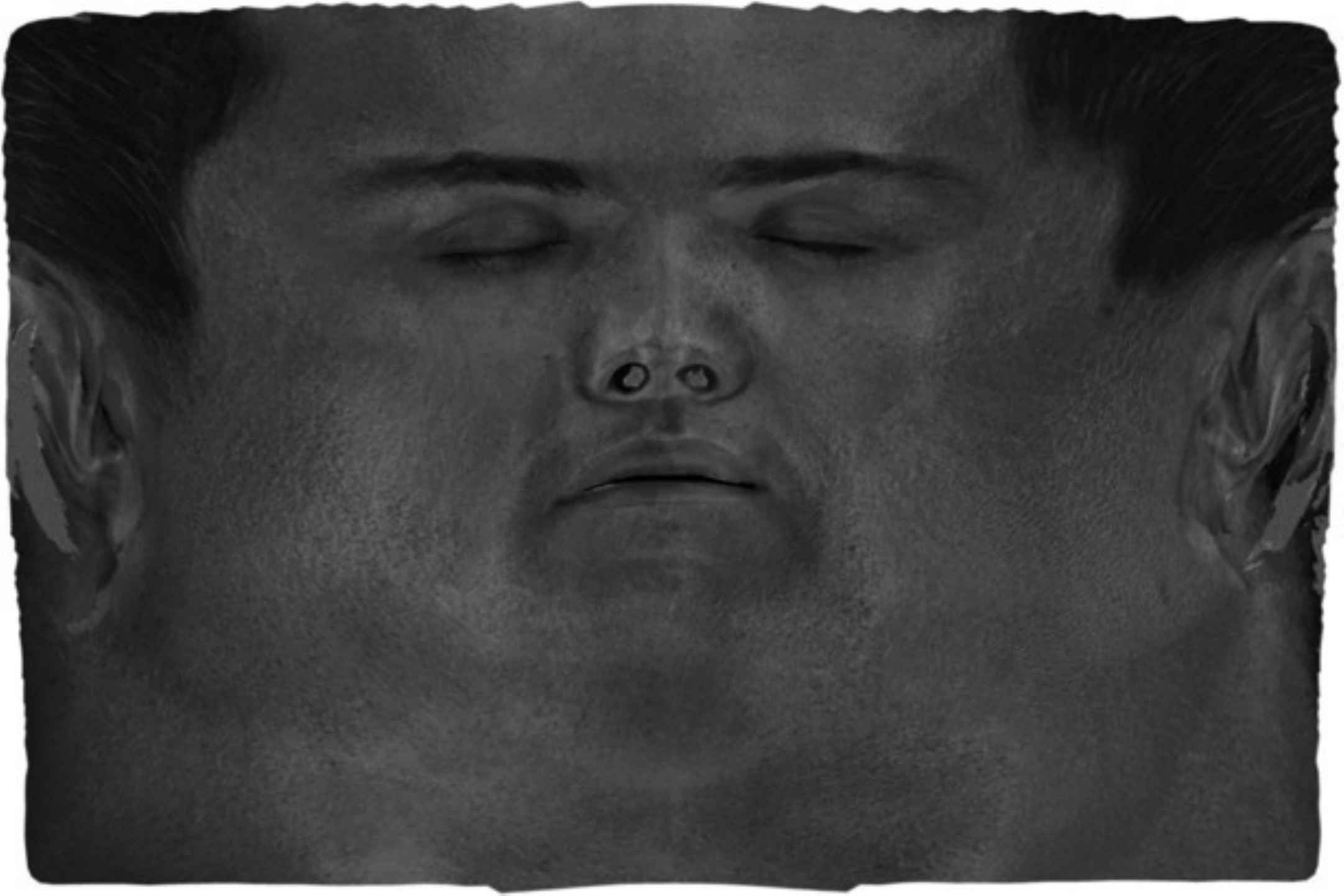}}
    \subfloat[
        Diff.Nor.~$\mathbf{N_D}$]{
        \includegraphics[width=0.19\linewidth]{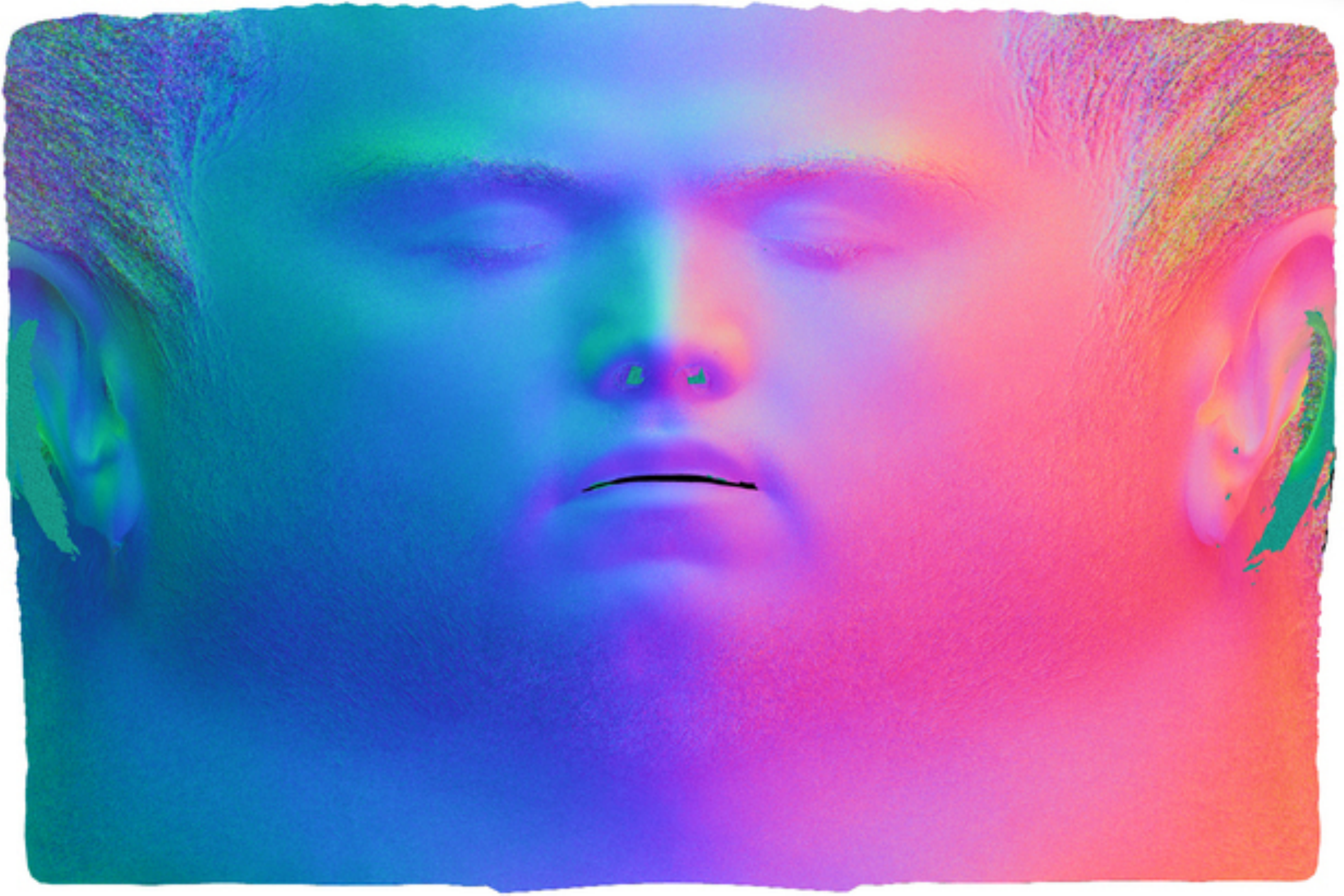}}
    \subfloat[
        Spec.Nor.~$\mathbf{N_S}$]{
        \includegraphics[width=0.19\linewidth]{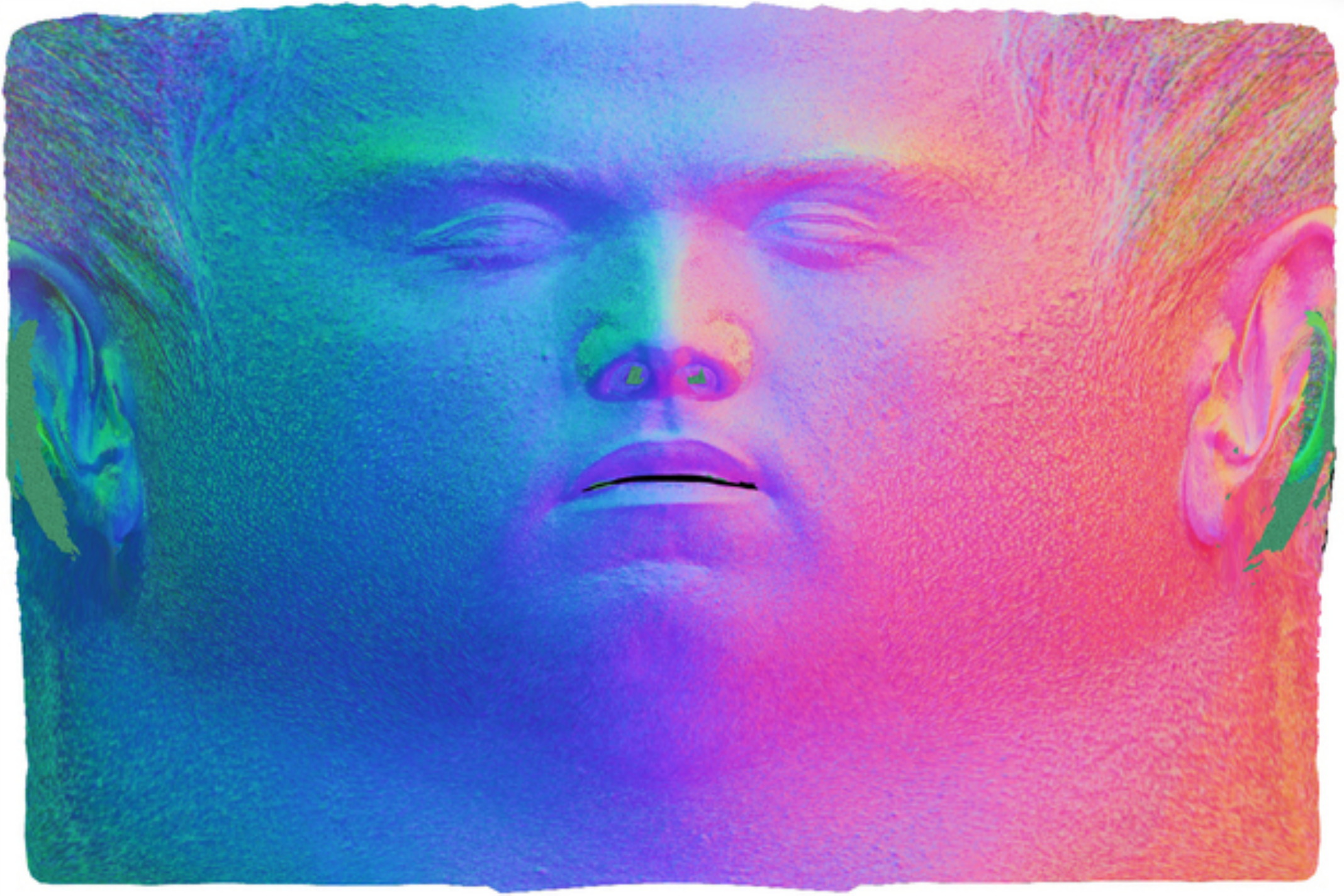}}
    \subfloat[
        Shape~$\mathbf{S}$]{
        \includegraphics[width=0.19\linewidth]{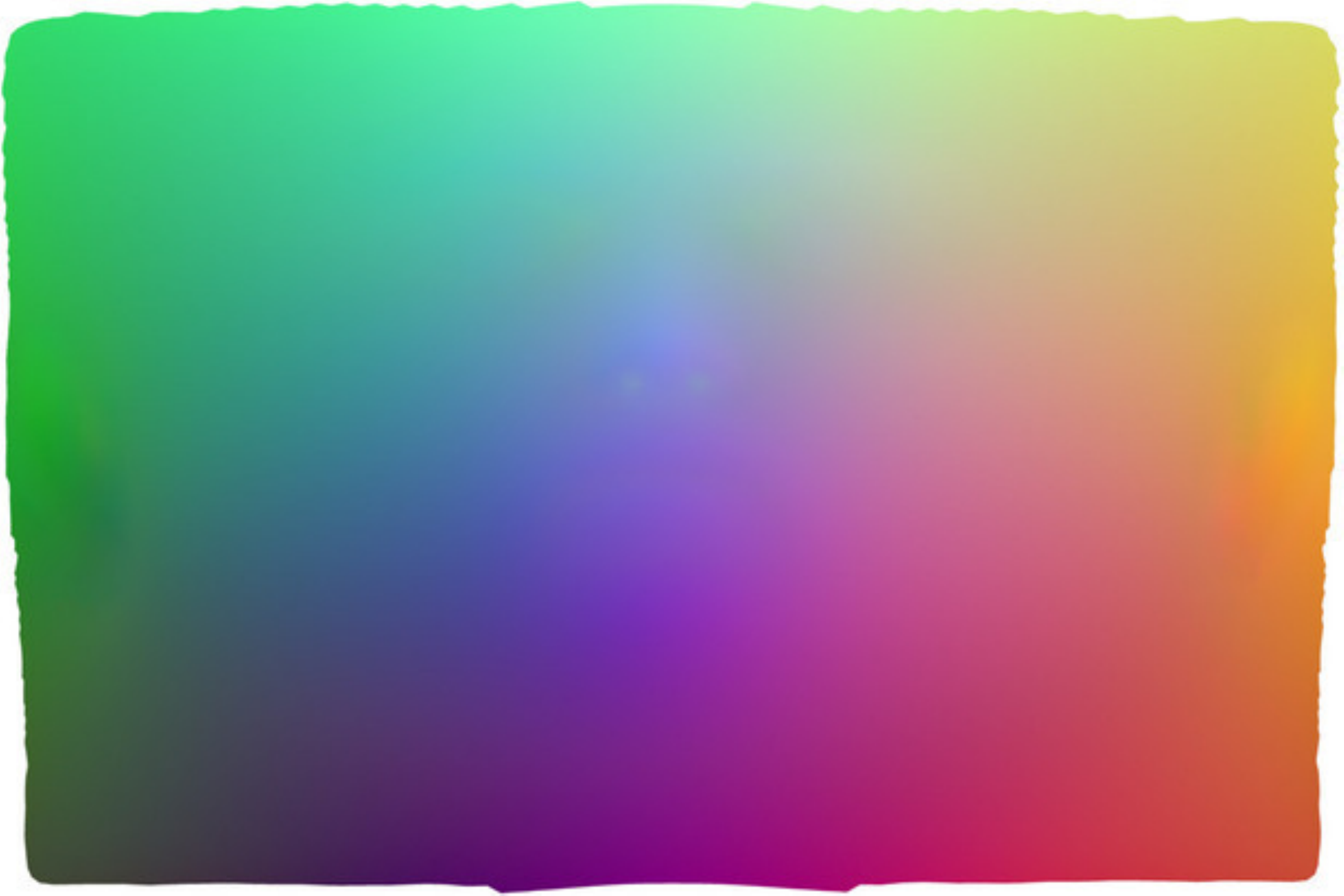}}
    \caption{
        Example of a captured subject data using
        \cite{kampouris_diffuse-specular_2018, lattas_multi-view_2019},
        registered and projected to a standard UV topology.
    }
    \label{fig:data_captured_data}
\end{figure}

We use our facial capturing system \cite{kampouris_icl_2018},
comprising of an LED sphere with 168 lights 
(partitioned into two polarization banks) and 9 DSLR cameras. 
Half of the LEDs on the sphere are vertically polarized 
(for parallel polarization), 
and the other half are horizontally polarized 
(for cross-polarization) in an interleaved pattern.
On this setup, we can employ the state-of-the-art method of \cite{ghosh_multiview_2011}
for capturing high resolution pore-level reflectance maps of faces.
On the same apparatus, we remove the polarizers
and use the color-space analysis for diffuse-specular separation 
and multi-view facial capture \cite{kampouris_diffuse-specular_2018, lattas_multi-view_2019},
to acquire reflectance of similar quality (Fig. \ref{fig:data_captured_data}).
This is our preferred method, 
since it requires less than half of the data captured 
(hence reduced capture time) 
and provides a simpler setup without polarizers,
enabling the acquisition of larger datasets.

Following the capture and diffuse-specular separation with \cite{kampouris_diffuse-specular_2018},
we develop a pipeline to prepare the training data as follows: A base geometry is reconstructed from the full-on images using 
structure-from-motion \cite{schoenberger2016sfm} and multi-view stereo \cite{schoenberger2016mvs}.
The geometry is fitted with landmarks \cite{sagonas_semi-automatic_2013} using its rendering
and registered to a template mesh \cite{amberg_optimal_2007}.
Finally, the camera-space reflectance is projected
to a uniform UV map, manually constructed for minimal distortion,
at a resolution of $\hat{H},\hat{W}=6144,4096$ pixels,
as shown in Fig.~\ref{fig:data_captured_data}.

\subsection{Data Collection}
\label{sec:method_data_collection}
In this work, we capture faces of over 200 individuals 
of different ages and characteristics under 7 different expressions.
We curate a dataset called \datasetname,
by sampling square $512\times512$ pixels patches, of 
(a) diffuse albedo $\mathbf{A_D}$, (b) specular albedo $\mathbf{A_S}$,
(c) diffuse normals $\mathbf{N_D}$, (d) specular normals $\mathbf{N_S}$ 
and (e) shape $\mathbf{S}$ in UV space.
The patches are anonymized, shuffled,
and only the correspondence between same patches of different type is kept.
We have made \datasetname{} public for the research community
\footnote{
Dataset available at
\protect{\url{https://github.com/lattas/avatarme}} .
}.
The captured subjects are $63.2\%$ male and $36.8\%$ female;
$61.1\%$ White, $26.3\%$ Asian, $5.5\%$ Black and $7.1\%$ other; $55.8\%$ 0-25 years old, $37.3\%$ 25-40 years old, $6.9\%$ over 40 years old.

\subsection{Data Augmentation}
\label{sec:method_data_augmentation}
In our initial method \modelname{}\cite{lattas_avatarme_2020},
we only rendered our dataset in the environment of the target 3DMM.
In \modelnameplus{} (Sec.~\ref{sec:method_recon_renderme_renderloss}),
we can augment the training data by rendering them in random environments,
centered around the target environment.
This does not only improve the model's accuracy on the target environment,
but enables it to successfully generalize to other domains (Figs.~\ref{fig:results_others},\ref{fig:results_ablation_barplot},\ref{fig:results_ablation_images}).
Moreover, our captured dataset is imbalanced on race,
due to the demographic limitations in our area 
and the immobility of our capturing device.
In an attempt to balance the dataset, we use the albedo measurements of \cite{donner2006spectral}
to augment our acquired albedos.
Specifically, we use a patch from the forehead of the captured diffuse albedo,
match it to the closest albedo from \cite{donner2006spectral}
and then apply a transformation to another albedo from their chart.
Since all our albedos are in the same UV space,
a manually constructed ``skin'' mask ensures that common non-skin areas remain unchanged.

\section{Method}
\subsection{Overview}
\label{sec:method_recon}
\begin{figure*}[ht]
    \centering
    \includegraphics[width=\linewidth]{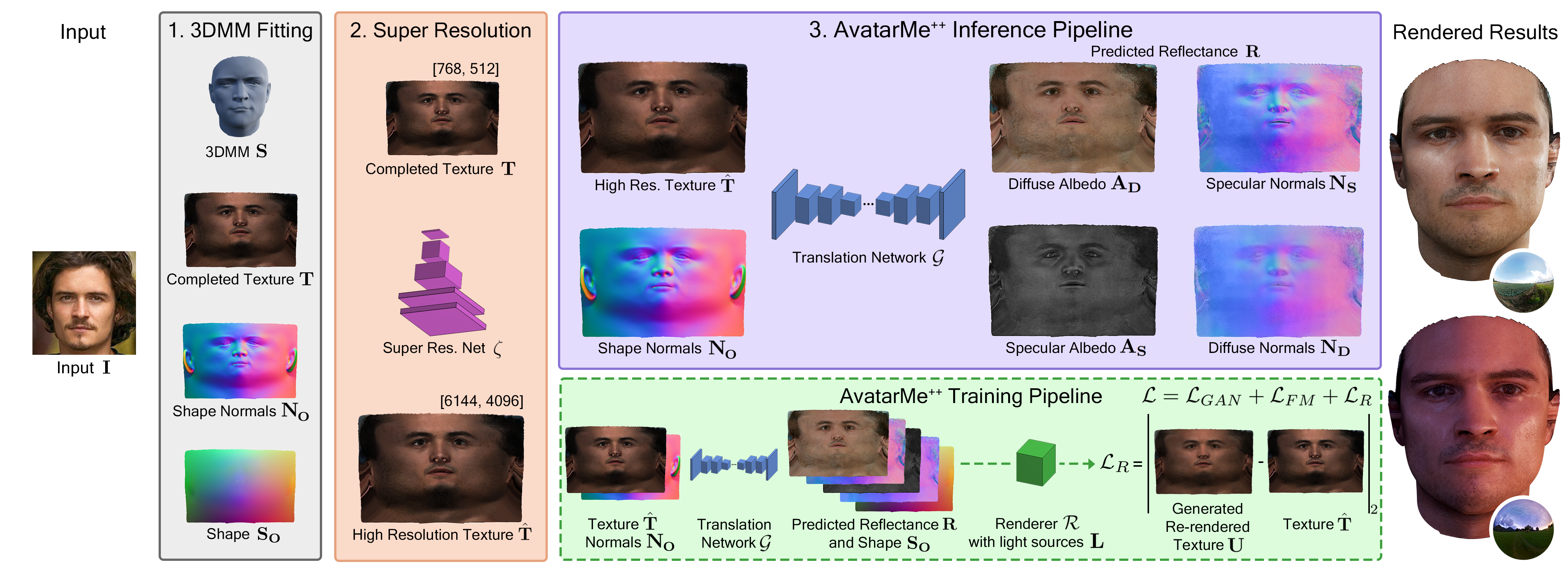}
    \caption{
        Summary of the AvatarMe\textsuperscript{++} method:
        Given an ``in-the-wild'' image $\mathbf{I}$,
        we first fit a 3D Morphable Model (3DMM) to acquire the shape $\mathbf{S_O}$, 
        texture $\mathbf{T}$ and shape normals $\mathbf{N_O}$ in UV space.
        Then, we upscale the texture $\mathbf{T}$ using a state-of-the-art super resolution
        network $\zeta$, trained on synthetic data rendered in the texture's $\mathbf{T}$ domain.
        A deep network $\mathcal{G}$ is then used to transform the upscaled texture $\mathbf{\hat{T}}$
        and normals $\mathbf{N_O)}$ to reflectance maps,
        namely the diffuse albedo $\mathbf{A_D}$, specular albedo $\mathbf{A_S}$,
        diffuse normals $\mathbf{N_D}$ and specular normals $\mathbf{N_S}$.
        The deep image-translation network is trained on high-resolution captured facial BRDF,
        which we have made public as RealFaceDB.
        To train \modelnameplus{}, we define a photorealistic differentiable rendering module $\mathcal{R}$,
        with subsurface-scattering and self-occlusion approximation.
        During training, $\mathcal{R}$ is used to create synthetic data pairs,
        by rendering the captured data in the target's environment $\mathbf{L}$ and random ones.
        The loss $\mathcal{L}$ used during training, is comprised of an adversarial loss $\mathcal{L}_{GAN}$,
        a feature-matching loss $\mathcal{L}_{FM}$ and our photorealistic differentiable loss $\mathcal{L}_R$.
        The complete high resolution (up to $6\text{k}\times4$k) BRDF maps can be used for photorealistic rendering, while the specular normals $\mathbf{N_S}$ can be used to enhance the 3DMM's geometry.
    }
    \label{fig:method_overview}
\end{figure*}

We aim to reconstruct the shape and reflectance properties of a subject from a single ``in-the-wild'' image,
that can be used for photorealistic rendering.
These are the shape $\mathbf{S}$,
diffuse albedo $\mathbf{A_D}$, specular albedo $\mathbf{A_S}$,
diffuse normals $\mathbf{N_D}$ and specular normals $\mathbf{N_S}$,
facial reflectance components that can be used for photorealistic rendering
(i.e.~\cite{ma_rapid_2007, ghosh_multiview_2011}).

As shown in Fig.~\ref{fig:method_overview},
we fit a 3DMM to an ``in-the-wild'' image 
(Sec.~\ref{sec:method_recon_ganfit}),
obtaining a 3D facial geometry $\mathbf{S}$
with a texture $\mathbf{T}$,
which is typically of low-resolution and contains baked-in illumination and shadows.
We upsample $\mathbf{T}$ using a deep super-resolution network
trained on textures of the same domain as the ones from the 3DMM
(Sec.~\ref{sec:method_recon_super_res}).
Then, the \modelname{} models (Sec.~\ref{sec:method_recon_avatarme})
or the \modelnameplus{} model (Sec.~\ref{sec:method_recon_renderme})
transform the upsampled texture $\mathbf{\hat{T}}$ 
into the reflectance components 
$\mathbf{A_D}, \mathbf{A_S}, \mathbf{N_D}, \mathbf{N_S}$.
\modelname{} utilizes four image-translation networks for the above transformation.
\modelnameplus{}, an extension to \modelname{},
uses a combined image-translation network and
incorporates a photorealistic differentiable renderer (Sec.~\ref{sec:method_diffrender}),
achieving improved results, generalization and computational speed.

\subsection{Initial Geometry and Texture Estimation}
\label{sec:method_recon_ganfit}
The first step of our method
is the acquisition of a facial shape $\mathbf{S}$
and texture $\mathbf{T}$ from a single image.
In our approach, we adopt the 3DMM-based fitting method
GANFIT \cite{gecer_ganfit_2019}.
Apart from the usage of deep identity features,
GANFIT synthesizes consistent realistic UV texture maps,
using a GAN as a statistical representation of the facial texture.
Alternatively, our method and training can easily be modified
to use other methods (i.e.~\cite{chen_learning_2019, gecer2020ostec})
as long as they produce a consistent shape and texture
(results in Fig.~\ref{fig:results_others}).
We reconstruct the initial 
base shape $\mathbf{S} \in \mathbb{R}^{n\times3}$ of $n$ vertices
and texture $\mathbf{T} \in \mathbb{R}^{W\times H\times3}$
from the input image $\mathbf{I}$ as follows:
\begin{equation}
    \mathbf{T},\mathbf{S} = \mathcal{F}(\mathbf{I})
    \label{eq:method_recon_ganfit}
\end{equation}
where 
$\mathcal{F}:  \mathbb{R}^{k \times m \times 3} \mapsto \mathbb{R}^{W\times H\times3}, \mathbb{R}^{n\times3}$ 
denotes the GANFIT reconstruction method for an 
$\mathbf{I} \in \mathbb{R}^{k \times m \times 3}$ arbitrary sized image,
and $n$ number of vertices on a fixed topology.

The acquired shape is of adequate quality for rendering,
however the texture $\mathbf{T}$ is of limited resolution
and most importantly, 
contains significant baked-in illumination and self-shadows.
We proceed by upsampling $\mathbf{T}$ in the next section,
and then discuss the ways to learn the disentanglement
of the baked-in illumination in $\mathbf{T}$,
into high-resolution spatially-varying reflectance parameter UV maps.

\subsection{3DMM Capturing Environment Estimation}
\label{sec:method_data_env_estimation}
A drawback of the texture modeled by typical \threedmm{}s
is that they reproduce the environment conditions 
of their training data (i.e. reflection and shadows),
which inhibits rendering.
In our case, the textures generated by \cite{gecer_ganfit_2019}
contain sharp highlights and shadows,
made by point-light sources, as well as environment illumination.
In order to alleviate this problem, 
we model the illumination conditions of the dataset used in \cite{gecer_ganfit_2019}
and synthesize UV maps with the same illumination, 
in order to train a transformation between texture 
with baked-illumination $\mathbf{T}$
and reflectance maps $\mathbf{R}$. 

Initially, we acquire random texture and mesh outputs from GANFIT,
by fitting random facial images.
Using a cornea model \cite{nishino_eyes_2004}, 
we estimate the average view direction $\mathbf{v_d}$,
the direction for the apparent 3 point light sources $\mathbf{l_d}$
and their intensity $\mathbf{c_d}$ used
in the 3DMM texture data, including an environment color defined as $\mathbf{e_{d}}$.
Then, we render our acquired subjects (Section \ref{sec:method_data_collection}), 
as if they were samples from the dataset used in the training of
the 3DMM used, in our case \cite{gecer_ganfit_2019}.
In this way, we also have accurate ground truth of their reflectance. 
We compute a rendering $\rho$ 
for each subject with reflectance
$\mathbf{R} = \{\mathbf{A_D}, \mathbf{A_S}, \mathbf{N_D}, \mathbf{N_S}\}$, 
directly in a UV map $\mathbf{T_d}$,
using the predicted environment parameters.
We denote this rendering process by  
$ \rho: \mathbf{R, c_d, l_d, v_d, e_d} \mapsto \mathbf{T_d} \in \mathbb{R}^{\hat{W}\times\hat{W}\times3}$ 
which renders the captured reflectance to the domain of the 3DMM textures with baked illumination.

The above estimation of the target 3DMM environment can be further improved,
after an initial training of the \modelnameplus{} network $\mathcal{G}$.
We acquire a number of random 3DMM-generated textures in the target environment
and use $\mathcal{G}$ to acquire their reflectance.
Then, we initialize our differentiable renderer (Sec.~\ref{sec:method_diffrender}) 
with the calculated parameters ($\mathbf{v_d}, \mathbf{l_d}, \mathbf{c_d}$).
In an iterative process, we render the acquired reflectance maps in the current best environment parameters and compare the rendering with the initial 3DMM-generated textures,
using an L1 rendering loss.
Since the renderer is differentiable,
we use gradient descent to further optimize the estimated parameters $\mathbf{v_d}, \mathbf{l_d}, \mathbf{c_d}$.
Then, we can re-train $\mathcal{G}$ in the optimized environment estimation.

\subsection{Super Resolution}
\label{sec:method_recon_super_res}
Although the texture 
$\mathbf{T} \in \mathbb{R}^{W\times H\times3}$ 
from GANFIT \cite{gecer_ganfit_2019} has reasonably good quality
and resolution ($H,W=768,512$)
it is below par compared to artist-made 
render-ready 3D faces.
On the contrary, the facial reflectance textures 
we capture in (Sec.~\ref{sec:method_data_capturing_setup})
are in a resolution of $\hat{W},\hat{H}=6144,4096$.
Therefore, we train a state-of-the-art super resolution network, 
RCAN \cite{zhang_image_2018},
that increases the resolution of
$\mathbf{T} \in \mathbb{R}^{W\times H\times3}$
to $\mathbf{T} \in \mathbb{R}^{\hat{W}\times\hat{H}\times3}$,
using $\times4$ upsampling twice.
We define the super-resolution network 
$(\zeta : \mathbb{R}^{H_p\times H_p\times3} \mapsto \mathbb{R}^{\hat{H}_p\times\hat{H}_p\times3})$,
which is trained on square patches of $H_p=64 \mapsto \hat{H}_p=512$,
given the large size of the results
and the low number of available training data.
At testing time, the whole texture from GANFIT $\mathbf{T}$ 
is up-scaled by the following:
\begin{equation}
    \hat{\mathbf{T}} = \zeta(\mathbf{T})
    \label{eq:method_recon_super_res}
\end{equation}
To train the super-resolution $\zeta$
to upsample the textures generated from Eq.~\ref{eq:method_recon_ganfit},
we use our estimation of GANFIT's illumination environment (Sec.\ref{sec:method_data_env_estimation})
to render our captured data with the same environment.

\subsection{Reflectance Inference with \modelname{}}
\label{sec:method_recon_avatarme}
The significant issue of the texture $\mathbf{T}$ produced by typical \threedmm{}s
is that they are trained on data with ambient illumination 
(i.e. reflection, shadows), which they reproduce.
GANFIT-produced textures contain sharp highlights and shadows,
made by strong point-light sources, as well as baked environment illumination,
which prohibits photorealistic rendering.
In order to alleviate this problem, 
we first model the illumination conditions of the dataset used in \cite{gecer_ganfit_2019}
and then synthesize UV maps with the same illumination 
$\mathbf{T_d}$ (Sec.~\ref{sec:method_data_env_estimation}).
We can then use the pairs of $\mathbf{T_d}$
with the ground truth reflectance data
($\mathbf{A_D}, \mathbf{A_S}, \mathbf{N_D}, \mathbf{N_S}$)
to train image-translation networks in a supervised manner.
Finally, following \cite{wang_high-resolution_2018}, 
we formulate the networks' objective as:
\begin{equation}
    \min_G \left(
        \max_D \mathcal{L}_{GAN}(G, D_k)
        +
        \lambda_{FM}
        \mathcal{L}_{FM}(G, D_k)
    \right)
    \label{eq:recon_avatarme_loss}
\end{equation}
where $\mathcal{L}_{GAN}(G, D_k)$ is the sum of adversarial loss
and $\mathcal{L}_{FM}(G, D_k)$ is the feature matching loss,
for all 3 discriminators of pix2pixHD \cite{wang_high-resolution_2018}.
The feature matching term is controlled by $\lambda_{FM}$.

We find that we can improve learning
by giving the network an insight into the geometry of the reconstructed shape.
In this manner, for each training data pair,
we extract the shape $\mathbf{S_O}$
and its normals $\mathbf{N_O}$ in the same UV parameterization as the textures,
and complete the 2D RGB texture, by using bilinear interpolation in the 2D UV space.
Below we describe the baseline pipeline, \textit{AvatarMe},
in which 4 image-translation networks are used,
to first acquire the diffuse albedo $\mathbf{A_D}$
from the upsampled reconstructed texture $\mathbf{\hat{T}}$
and then
the specular albedo $\mathbf{A_S}$,
diffuse normals $\mathbf{N_D}$
and specular normals $\mathbf{N_S}$
from the diffuse albedo $\mathbf{A_D}$.
The better performing pipeline \textit{\modelnameplus{}},
with a single rendering-aware network
is described in Sec.~\ref{sec:method_recon_renderme}.

\subsubsection{Diffuse Albedo Extraction}
\label{sec:method_recon_avatarme_delight}
We formulate de-lighting as a domain adaptation problem and train an image-to-image translation network. 
To do this, we follow two strategies different from the standard image translation approaches.
Firstly, the shading and occlusion on the skin surface is geometry dependent
and thus use both the texture and geometry of the 3DMM as input to the network.
We find that this improves not only the network's accuracy,
but also the consistency between patches.
Instead of using the 3-channel shape texture $\mathbf{S_O}$,
we define the 1-channel texture $\mathbf{D_O}$,
that contains only the $Z$ axis of $\mathbf{S_O}$.
To do so, we concatenate the texture $\mathbf{T_d}$
with the UV map of the depth of the mesh in object space $\mathbf{D_O}$.
We feed the network with a 4D tensor of
$[\mathbf{T_{d_R}},\mathbf{T_{d_G}},\mathbf{T_{d_B}},\mathbf{D_O}]$
and predict the resulting 3-channel albedo $\mathbf{A_D}$.
Instead of $\mathbf{D_O}$, the shape normals ($\mathbf{N_O}$) can also be used.
Secondly, we split the original high-resolution data into overlapping patches
of $\hat{H}_p\times\hat{H}_p$ pixels, in order to augment the number of data samples
and fit the data into the available GPU memory.

Therefore, in order to de-light $\hat{\mathbf{T}}$,
and acquire the diffuse albedo $\mathbf{A_D}$,
we train an image-to-image translation network $\mathcal{G}_{A_D}:
\mathbf{T_d}, \mathbf{D_O} \mapsto \mathbf{A_D} \in
\mathbb{R}^{\hat{H}_p\times\hat{H}_p\times3}$ 
and then extract the diffuse albedo $\mathbf{A_D}$ by the following:
\begin{align}
    \mathbf{A_D} = \mathcal{G}_{A_D}(\hat{\mathbf{T}}, \mathbf{D_O})
    \label{eq:recon_avatarme_diffalbedo}
\end{align}

\subsubsection{Specular Albedo Extraction}
\label{sec:method_recon_avatarme_specAlb}
Predicting the entire specular BRDF
and the per-pixel specular roughness from the illuminated texture $\mathbf{\hat{T}}$ 
or the inferred diffuse albedo $\mathbf{A_D}$, 
poses an unnecessary challenge.
As shown in \cite{ghosh_multiview_2011, kampouris_diffuse-specular_2018}
a subject can be realistically rendered 
using only the intensity of the specular reflection (specular albedo) $\mathbf{A_S}$,
which is consistent on a face due to the skin's refractive index. 
The spatial variation is correlated to facial skin structures 
such as skin pores, wrinkles, or hair, 
which are apparent in both the baked texture $\mathbf{T}$ 
and the diffuse albedo $\mathbf{A_D}$.
Both can be used as input to the network,
and we empirically found that our predicted high quality diffuse albedo $\mathbf{A_D}$
produces more accurate and consistent results.
Therefore, having inferred $\mathbf{A_D}$ with $\mathcal{G_{A_D}}$, 
we infer the specular albedo $\mathbf{A_S}$
by a similar patch-based image-to-image translation network 
from the diffuse albedo
($\mathcal{G}_{A_S}: \mathbf{A_D} \mapsto \mathbf{A_S} \in
\mathbb{R}^{\hat{H}_p\times\hat{H}_p\times1}$):
\begin{equation}
    \mathbf{A_S} = \mathcal{G}_{A_S}(\mathbf{A_D})
    \label{eq:recon_avatarme_specalbedo}
\end{equation}

\subsubsection{Specular and Diffuse Normals Extraction}
\label{sec:method_recon_avatarme_specNormals}
The specular normals exhibit sharp surface details,
such as fine wrinkles and skin pores, 
and are challenging to estimate,
as the appearance of some high-frequency details 
is dependent on the lighting conditions and viewpoint of the texture.
Therefore, much detail may not be apparent in the input image or reconstructed texture.
Previous works fail to predict high-frequency details \cite{chen_photo-realistic_2019}, 
or rely on separating the mid- and high-frequency information in two separate maps,
as a generator network may discard the high-frequency as noise \cite{yamaguchi_high-fidelity_2018}. 
Instead, we show that it is possible to employ 
an image-to-image translation network 
with feature matching loss \cite{wang_high-resolution_2018}
on a large high-resolution training dataset, 
which produces more detailed and accurate results.

Similarly to the specular albedo inference with $\mathcal{G}_{A_S}$,
we feed the network with the predicted diffuse albedo $\mathbf{A_D}$.
Using $\mathbf{A_D}$ instead of the 3DMM texture $\mathbf{T}$
produces more consistent results.
Even though $\mathbf{T}$ contains some specular highlights,
these are always concentrated on a small subset of the image,
since they're reconstructed using \cite{gecer_ganfit_2019}.
We can also luma-transform (in sRGB) the diffuse albedo
to grayscale $\mathbf{A_D^{(gray)}}$, in order to reduce the number of channels.
Moreover, the consistency of the results is greatly improved
when also feeding the network with the 3DMM geometry,
in this case, the shape normals.
Finally, we also transform the shape normals $\mathbf{N_O}$
in tangent space $\mathbf{N_T}$,
where the basis is a vector pointing to $[0, 0, 1]$.
We find that in this multiple-network approach,
the inferred specular normals details
are better accentuated, when using both the input shape normals $\mathbf{N_T}$
and the predicted specular normals $\mathbf{N_S}$ in the tangent space.

Therefore, we train another image-translation network 
$\mathcal{G}_{N_S}: \mathbf{A_D^{gray}}, \mathbf{N_T} \mapsto \mathbf{N_S}$,
$\in \mathbb{R}^{\hat{H}_p\times\hat{H}_p\times3}$ 
to transform the concatenation of the grayscale diffuse albedo $\mathbf{A_D^{gray}}$
and the shape normals in tangent space $\mathbf{N_T}$
to the specular normals $\mathbf{N_S}$. 
The specular normals are extracted by the following:
\begin{align}
    \mathbf{N_S} = \mathcal{G}_{N_S}(\mathbf{A_D^{gray}}, \mathbf{N_T})
    \label{eq:recon_avatarme_specNormals}
\end{align}

The diffuse normals $\mathbf{N_D}$ are 
highly correlated with the 3DMM-reconstructed shape normals $\mathbf{N_O}$,
as the evenly scattered light blurs most skin details.
Similarly fpr $\mathcal{G_{N_S}}$, we train a network
$\mathcal{G}_{N_D}: \mathbf{A_D^{gray}}, \mathbf{N_O} \mapsto \mathbf{N_D} 
\in \mathbb{R}^{\hat{H}_p\times\hat{H}_p\times3}$ 
to map the concatenation of the grayscale diffuse albedo $\mathbf{A_D^{gray}}$ 
and the shape normals in object space $\mathbf{N_O}$ 
to the diffuse normals $\mathbf{N_D}$. 
The diffuse normals are extracted as:
\begin{align}
    \mathbf{N_D} = \mathcal{G}_{N_D}(\mathbf{A_D^{gray}}, \mathbf{N_O})
    \label{eq:recon_avatarme_diffNormals}
\end{align}

Finally, the inferred specular normals can enhance 
the mesoscopic structure of the reconstructed geometry $\mathbf{S}$,
by refining its features and adding plausible details.
Based on \cite{nehab_efficiently_2005},
we integrate the specular normals in the tangent space $\mathbf{N_S}$
to produce a height UV map, which describes high-resolution per-pixel surface elevation.
The height map can be then be embossed on a subdivided 3DMM-reconstructed geometry $\mathbf{S}$,
to produce a higher-resolution shape.

\subsection{Photorealistic Differentiable Facial Rendering}
\label{sec:method_diffrender}
Here, we formulate a photorealistic differentiable rendering methodology, 
that can be incorporated in our image-translation networks during training
and render the training data and results in different illumination environments.

Shading can be modeled as \textit{local illumination},
which only models the surface reflection of light sources on objects
and \textit{global illumination}, which models light propagation in a scene,
including indirect illumination,
and produces more realistic results at a higher computational cost.
We choose to rely on local illumination shading,
since most such models are differentiable
and much faster to compute than global illumination.
Despite producing more realistic results,
rendering high-resolution human skin with global illumination takes several minutes,
and would be impractical to use while training a deep neural network like ours.
There exist various local illumination models
appropriate for rendering human skin,
on which we capitalize on to compose the following methodology
for photorealistic facial rendering.
We achieve fast and differentiable photorealistic rendering,
by using a local illumination model and approximations
for self-occlusion and subsurface scattering.
Additionally, ambient occlusion is inherently baked in the captured diffuse albedo
and is thus reproduced during rendering.

\subsubsection{Shading Model}
\label{sec:method_diffrender_shading_model}
We use Lambertian shading for the diffuse component $\mathbf{U_{D}}$
and Blinn-Phong \cite{blinn_texture_1976} shading for 
the specular component $\mathbf{U_{S}}$,
given their photorealistic results for human skin and their cheap computation.
For the specular exponent $s$, 
we use a common spatially varying UV map. 
For a reflectance and shape set
($\mathbf{A_D, A_S, N_D, N_S, S}$),
a camera with view direction $\mathbf{v}$
and a set of $n_l$ light sources 
with $\mathbf{l_j}$ direction and $\mathbf{c_j}$ intensity and
an ambient illumination intensity $\mathbf{c_a}$,
we evaluate the shading for each pixel $i$ as follows:
\begin{equation}
    \mathbf{U_{D_i}} = 
        \mathbf{c_a}
        \mathbf{A_{D_i}}
        \sum_{j = 1}^{n_l}
        (
            \mathbf{N_{D_i}} \cdot \mathbf{l_j}
        ) \mathbf{c_j}
\label{eq:render_diff_lambertian}
\end{equation}
\begin{equation}
    \mathbf{U_{S_i}} = 
        \mathbf{A_{S_i}}
        \sum_{j = 1}^{n_l}
        (
            \chi^{+}
            (
                \mathbf{N_{S_i}} \cdot \mathbf{h_j}
            )
        )^{s} \mathbf{c_j}
        , \quad \quad
        \mathbf{h_j}
        =
        \frac{
            \mathbf{l_j+v_j}
        }{
            \left|\left|\mathbf{l_j+v_j}\right|\right|
        }
\label{eq:render_spec_blinnphong}
\end{equation}
where $\chi^{+}(x)$ a piece-wise function that returns $\max{\{0, x\}}$,
since negative angles between the normals and light source direction
do not contribute to specular reflection.

\subsubsection{Rendering Directly in UV Space}
\label{sec:method_diffrender_uv_space}
Rasterization, the process of transforming the geometrical shape and texture
to pixels visible by a camera, is traditionally non-differentiable,
and also computationally expensive.
The various methods that have been proposed for differentiable rasterization
are based on sub-optimal approximations
\cite{liu_soft_2019, chen_learning_2019, kato_neural_2018} 
or are very expensive \cite{loubet_reparameterizing_2019, li_differentiable_2018}.
This motivated our pipeline to completely avoid rasterization,
by rendering directly on the UV space.

The geometry shape vertices $\mathbf{S}$ are projected and interpolated in the same UV space 
as the reflectance textures $\mathbf{S_O}$, creating texels
with a one-to-one correspondence
between normals, shape and reflectance pixels.
Hence, each texel's $S_{O_{u,v}}$ are used to evaluate
the view direction $\mathbf{v}$ used in Eq.~\ref{eq:render_spec_blinnphong}.
This way
(a) is faster than using rasterization,
(b) is differentiable,
(c) can be used with small patches of a larger texture and shape and
(d) creates a pixel-to-pixel correspondence
between reflectance and rendering.

\subsubsection{Fast Differentiable Facial Subsurface Scattering}
\label{sec:method_diffrender_sss}
Subsurface scattering (SSS) describes the light that exits a translucent medium
at a different point from where it had entered.
Human skin, a dielectric material, exhibits such properties
and the travel distance can be further than 
that covered by the lambertian model (Eq.\ref{eq:render_diff_lambertian}).
Subsurface scattering in the skin has a smoothing effect,
with predominantly red color bleed 
and is required for the photorealistic rendering of skin \cite{borshukov_realistic_nodate}.
These effects also vary across the skin 
and are stronger in more translucent areas such as the nose.
Accurate SSS requires the expensive measurements of light transport, 
however we find that the following modifications to our renderer
produce a photorealistic approximation
that improves the results of our method.

\begin{figure}[h]
    \centering
    
    \subfloat{
        \includegraphics[width=0.24\linewidth]{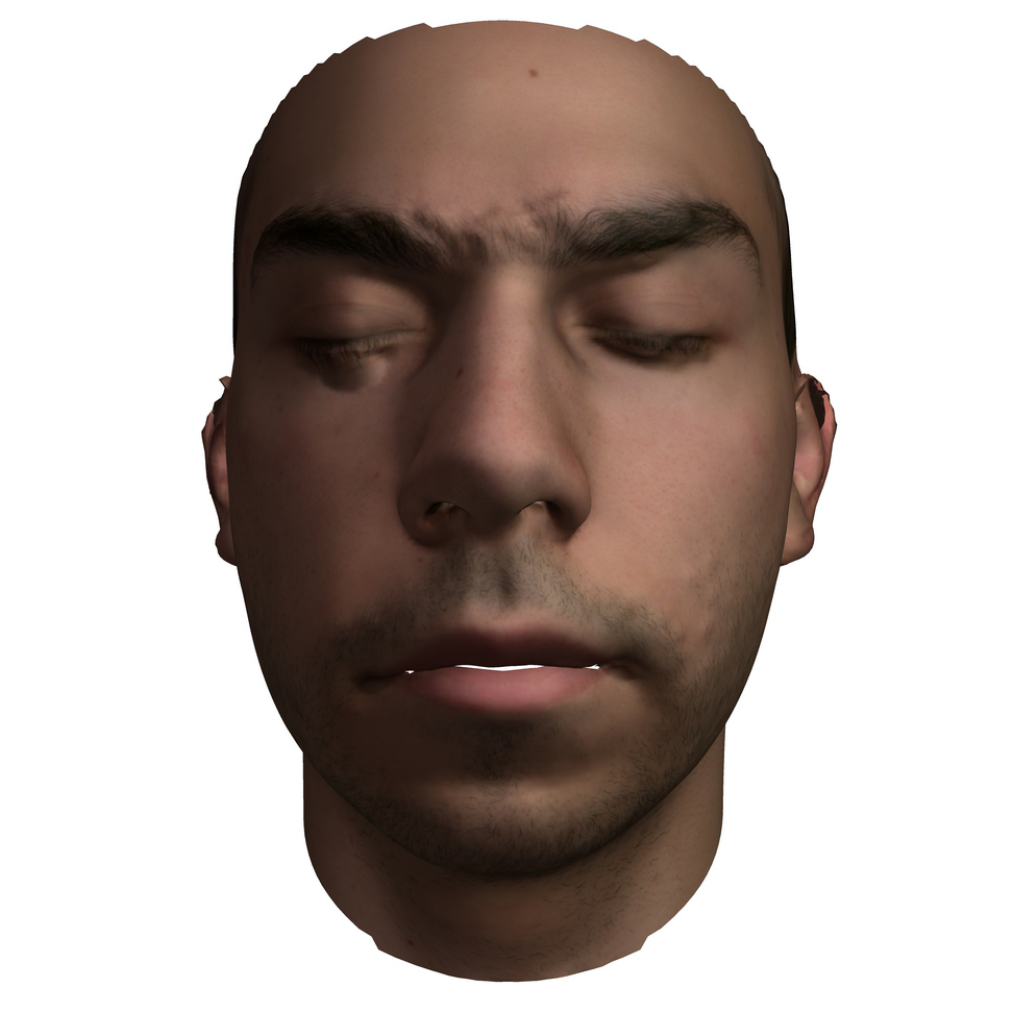}}
    \subfloat{
        \includegraphics[width=0.24\linewidth]{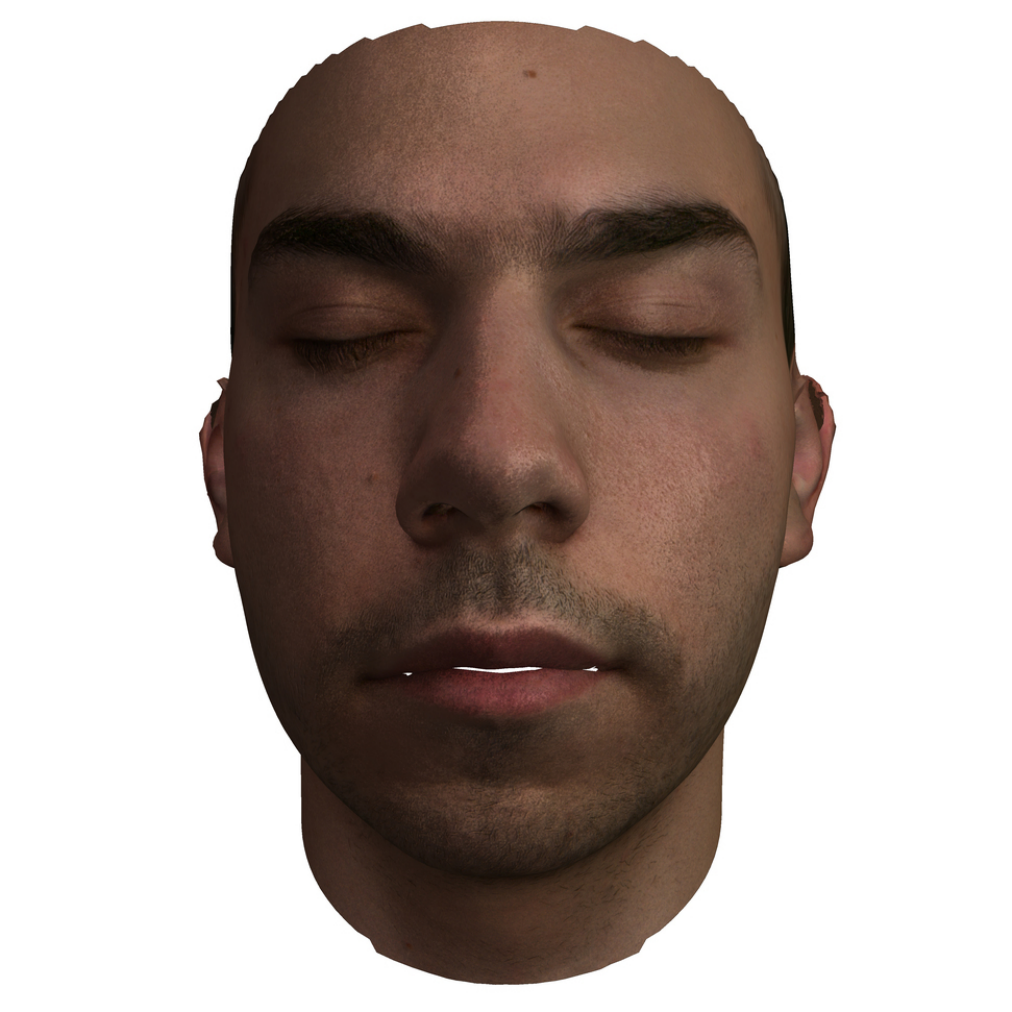}}
    \subfloat{
        \includegraphics[width=0.24\linewidth]{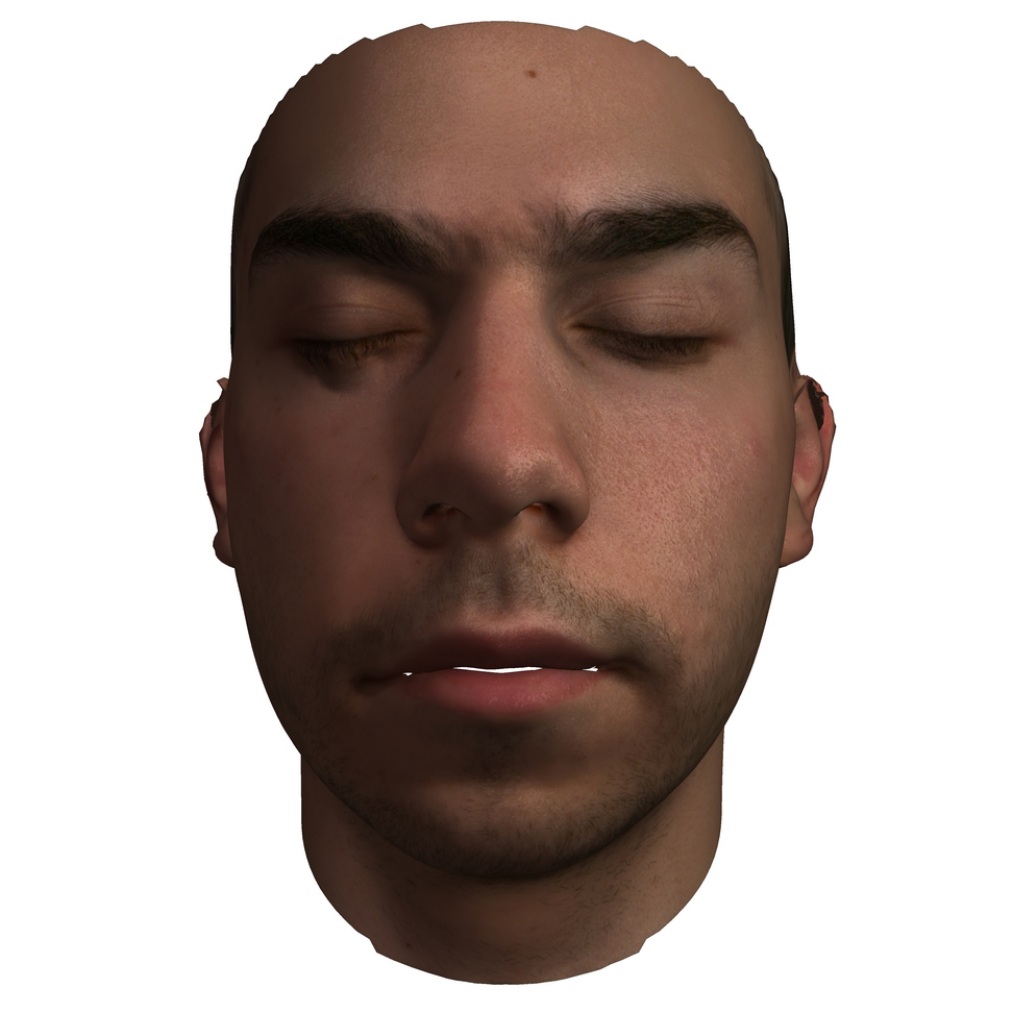}}
    \subfloat{
        \includegraphics[width=0.24\linewidth]{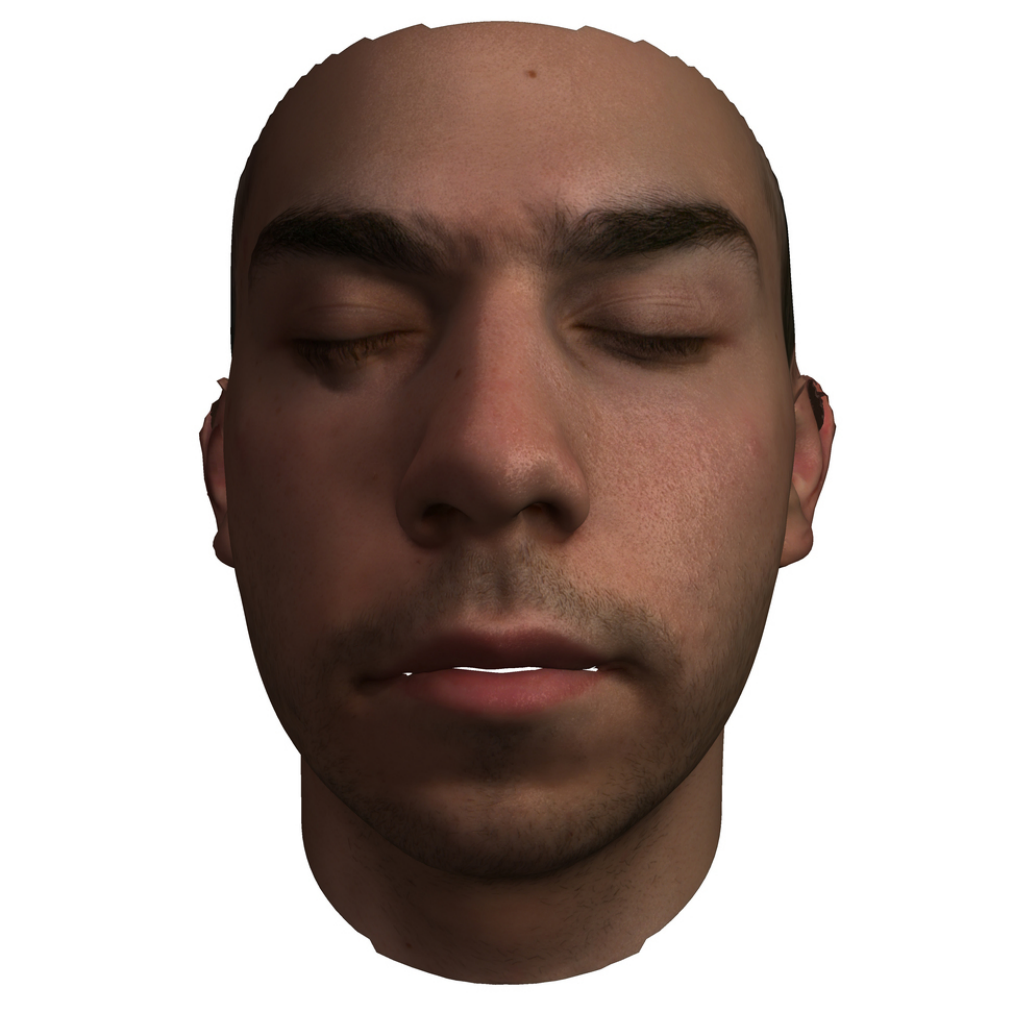}} \\
    \setcounter{subfigure}{0}
    \captionsetup[subfloat]{justification=centering}
    \subfloat[
        Pytorch3D, \\using $\mathbf{S, A_D, N}$]{
        \includegraphics[width=0.24\linewidth]{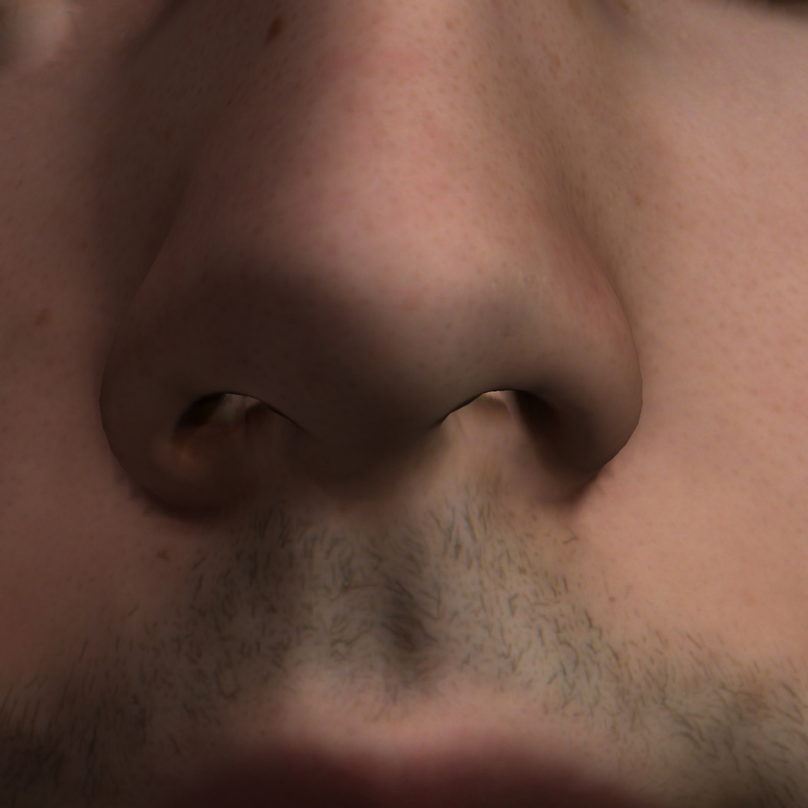}}
    \subfloat[
        Blinn-Phong, \\using $\mathbf{S, A_D,}$\\$\mathbf{A_S,N_S}$]{
        \includegraphics[width=0.24\linewidth]{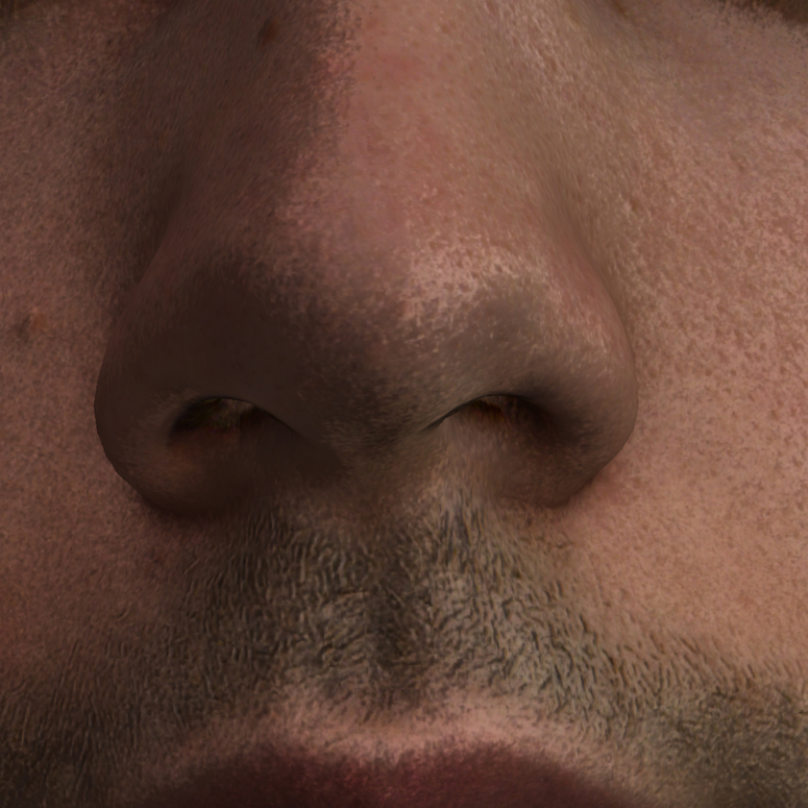}}
    \subfloat[
        b) with Subsurface Scattering]{
        \includegraphics[width=0.24\linewidth]{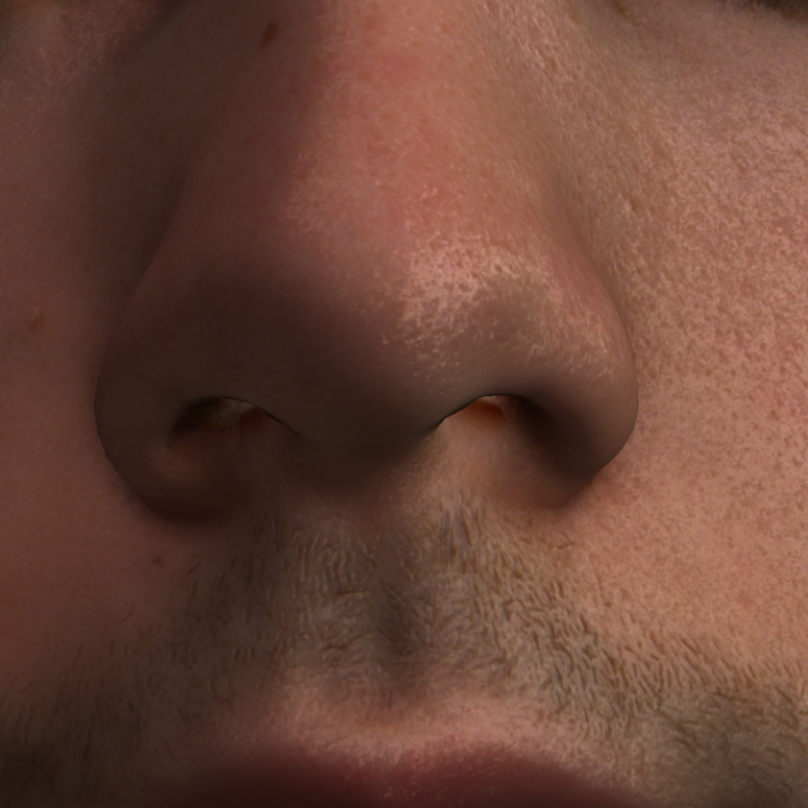}}
    \subfloat[
        c) with self-occlusion AE (in \& left of nose)]{
        \includegraphics[width=0.24\linewidth]{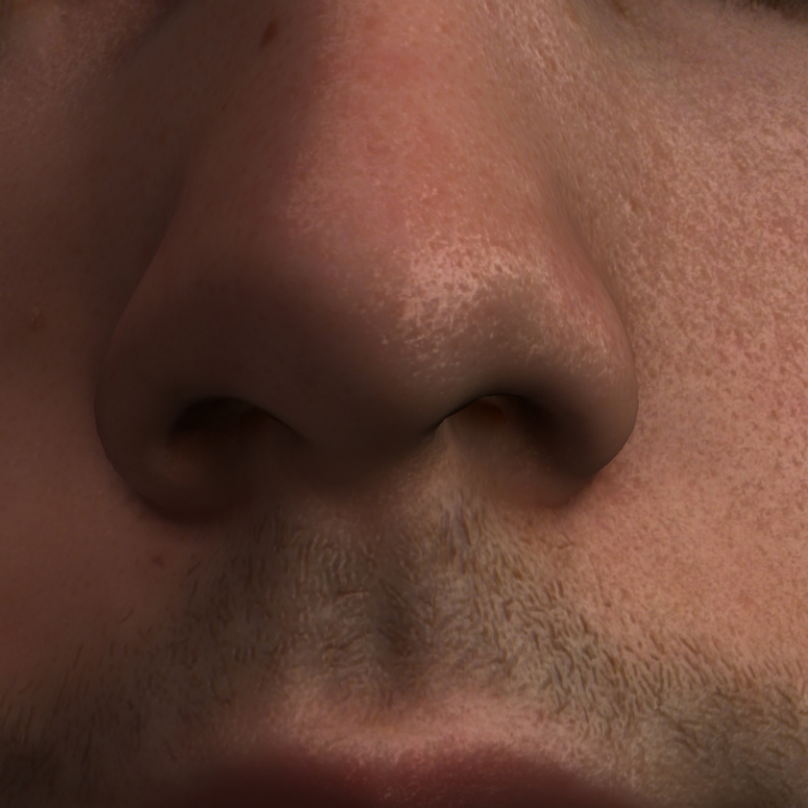}}
    
    \caption{
        The impact of our rendering modifications,
        to the Pytorch3D Mesh Renderer \cite{nikhila_ravi_pytorch3d_2020}.
        Top row: rasterized mesh with rendered texture,
        bottom: detail.
        Our improvements in realisticity, also
        improve the network's ability to recover the albedo and specular highlights.
    }
    \label{fig:results_ablation_rendering}
\end{figure}

A local illumination BRDF 
as described in Eq.~\ref{eq:render_diff_lambertian}
cannot model light scattered over large areas.
However, the scattering occurring in human skin travels only a few millimeters
and can be modeled
by separately modeling the normals for diffuse reflection $\mathbf{N_D}$ \cite{ma_rapid_2007}.
We separately capture (Sec.~\ref{sec:method_data_capturing_setup})
and infer both $\mathbf{N_D}$ and $\mathbf{N_S}$,
which are then separately used to evaluate the diffuse
(Eq.~\ref{eq:render_diff_lambertian})
and specular (Eq.~\ref{eq:render_spec_blinnphong}) components.
Both normals are wavelength dependent \cite{kampouris_diffuse-specular_2018}
and we acquire $\mathbf{N_D}$ from the red channel and 
$\mathbf{N_S}$ from the blue channel of our captures.
This method accurately models the SSS angular blur 
(Fig.~\ref{fig:results_ablation_rendering})
and does not impose a computational overhead during rendering or training.

Nevertheless, the above does not reproduce the spectrally-dependent spatial blur
produced by SSS, which results in red-dominated color bleed and shadow smoothing.
These can be accurately approximated by
texture-space SSS \cite{borshukov_realistic_nodate},
which blurs the diffuse component $\mathbf{U_{D}}$ under multiple
kernels, based on the wavelength associated with the $R,G,B$ channels.
\cite{borshukov_realistic_nodate} 
uses the weighted combinations per channel for 6 different kernels,
which is too expensive for our training requirements.
In line with \cite{hable2010uncharted},
we find that a single kernel is adequate and much faster.
Empirically, for a gaussian filter $g()$,
we calculate the mean kernel at $k=1.4mm$
or $21$ pixels in our standard topology
and define a weighted combination of 
$\mathbf{w_{D}}=diag(0.5, 0.85, 0.95)$ 
for the diffuse component and
$\mathbf{w_{SSS}}=diag(0.5, 0.15, 0.05)$ 
for the subsurface scattering.
We use a manually created standard translucency map $\mathbf{C}$,
which describes the amount of light absorbed 
and scattered on different facial areas.
Finally, since we are using $N_D$, 
we localize this effect only on the darker areas,
by multiplying the translucency map with an inverse brightness mask.
Therefore, the diffuse component $\mathbf{U_{D}}$
with subsurface scattering $\mathcal{S}$ is defined as:
\begin{equation}
    \mathcal{S}(\mathbf{U_{D_i}}) =
    (\mathbf{1}-\mathbf{C'_i}) 
    \circ
    \mathbf{w_{D}} \mathbf{U_{D_i}} 
    +
    \mathbf{C'_i} \circ \mathbf{w_{SSS}}\
    g(\mathbf{U_{D_i}})
\label{eq:diff_render_sss}
\end{equation}
\begin{equation*}
    \mathbf{C'} = 
    \mathbf{C} \circ \left(
    \mathbf{1} - \sum\nolimits_{j=1}^{n_L} (\mathbf{N_D} \cdot \mathbf{l}_j)\mathbf{c_j}
    \right)
\end{equation*}
where $\circ$ is the Hadamard product. As shown in Fig.~\ref{fig:results_ablation_rendering}, 
the usage of both SSS methods provides realistic results,
with the minimum computational overhead.

\subsubsection{Differentiable Shadows Simulation}
\label{sec:method_diffrender_shadows}

The rendering framework so far does not include self-occlusion shadows,
whose computation entails several challenges.
Pytorch3D \cite{nikhila_ravi_pytorch3d_2020} does not support self-occlusion tracing,
and efficient local illumination models, 
such as Blinn-Phong, inherently do not model it.
On the other hand, differentiable global illumination algorithms 
\cite{li_differentiable_2018, loubet_reparameterizing_2019},
that compensate for self-occlusion,
are too expensive while training.
Finally, the patches being rendered are often unaware
of the geometry that causes self-occlusion,
as it may appear on other patches
(i.e.~patches of nose and cheek).

As an efficient and simple solution,
we propose a simple autoencoder network,
that generates self-occlusion UV maps based on light source direction and intensity.
We train the autoencoder on UV maps baked only with self-occlusion.
We also train a linear regressor that maps light sources features $\mathbf{L}$
to the autoencoder's learned latent features $\mathbf{H_\mathcal{O}}$.
For each training example, 
we create a set of $n_l$ random light sources,
with direction $\mathbf{l_j}$ and luminosity $\mathbf{c_{l_j}}$,
which we stack in a matrix 
$\mathbf{L} = [ 
    \mathbf{l_j}\ \dots \ \mathbf{l_{N_l}}\ \mathbf{c_{l_j}}\ \dots \ \mathbf{c_{l_{n_l}}}
]^\top$.
We acquire self-occlusion UV-maps using a global illumination method, 
at the same topology of our main dataset,
for the mean geometry of our dataset.
We then pre-train the autoencoder and regressor on these data.
On each rendering step of the main network training,
the rendering environment's light source features $\mathbf{L}$
are regressed to the latent space of the autoencoder,
using the learned weights $\mathbf{W_\mathcal{O}}$
from which the decoder $\mathcal{O}(\mathbf{L}\mathbf{W_\mathcal{O}})$
generates the self-occlusion map.
The result is multiplied with the rendered diffuse and specular components,
to produce the final rendering:
\begin{equation}
    \mathcal{R}(\mathbf{R}, \mathbf{L})
    =
    \mathcal{O}(\mathbf{L}\mathbf{W_\mathcal{O}})
    \circ
    \left(
        \mathcal{S}(\mathbf{U_D})
        +
        \mathbf{U_S}
    \right)
    \label{eq:diff_render_shadow}
\end{equation}

\begin{figure}[h]
    \vspace{-0.5cm}
    \centering
    \captionsetup[subfloat]{justification=centering}
    \subfloat[
        Generated]{
        \includegraphics[width=0.21\linewidth,frame]{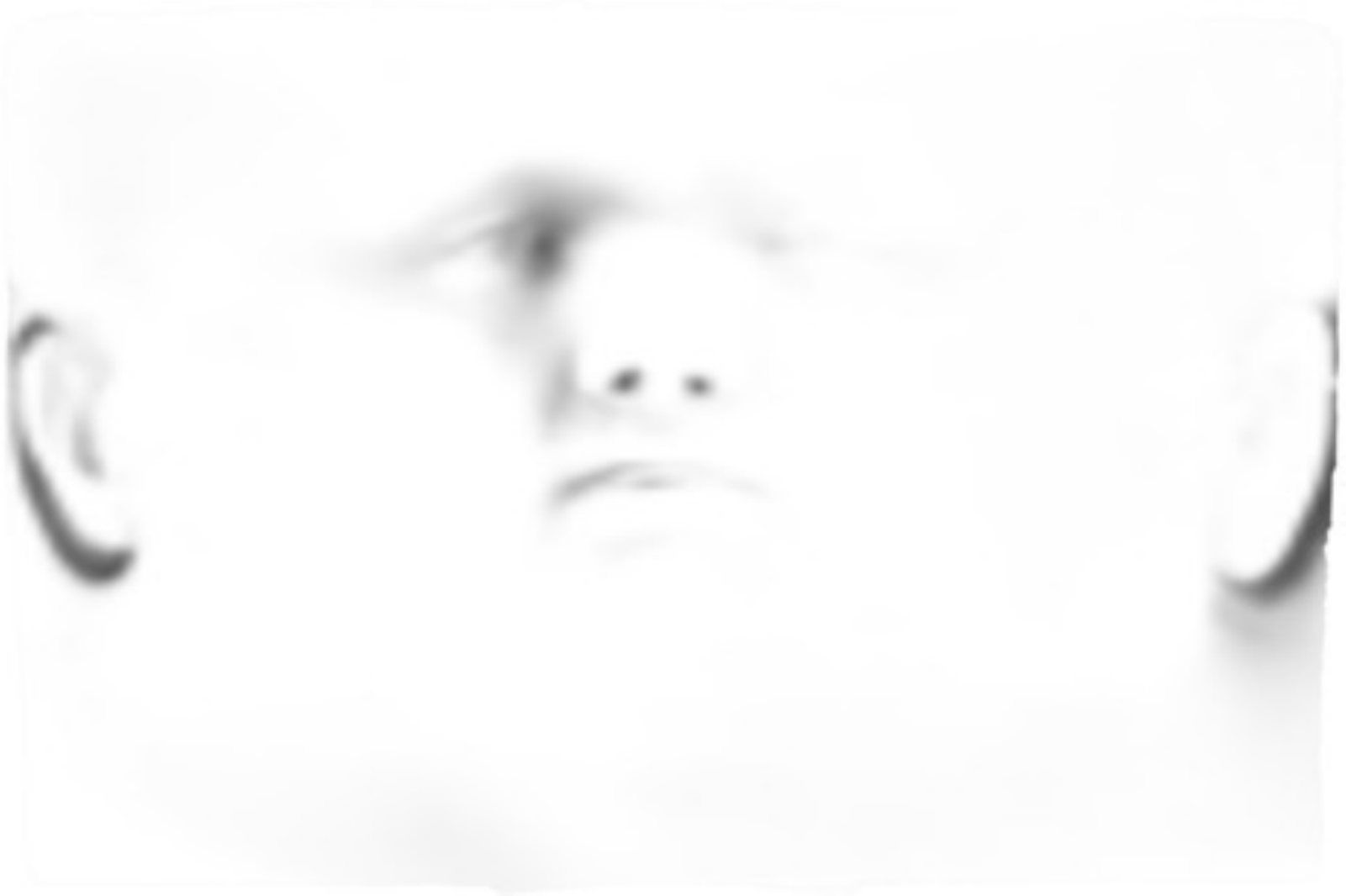}}
    \subfloat[
        Truth]{
        \includegraphics[width=0.21\linewidth,frame]{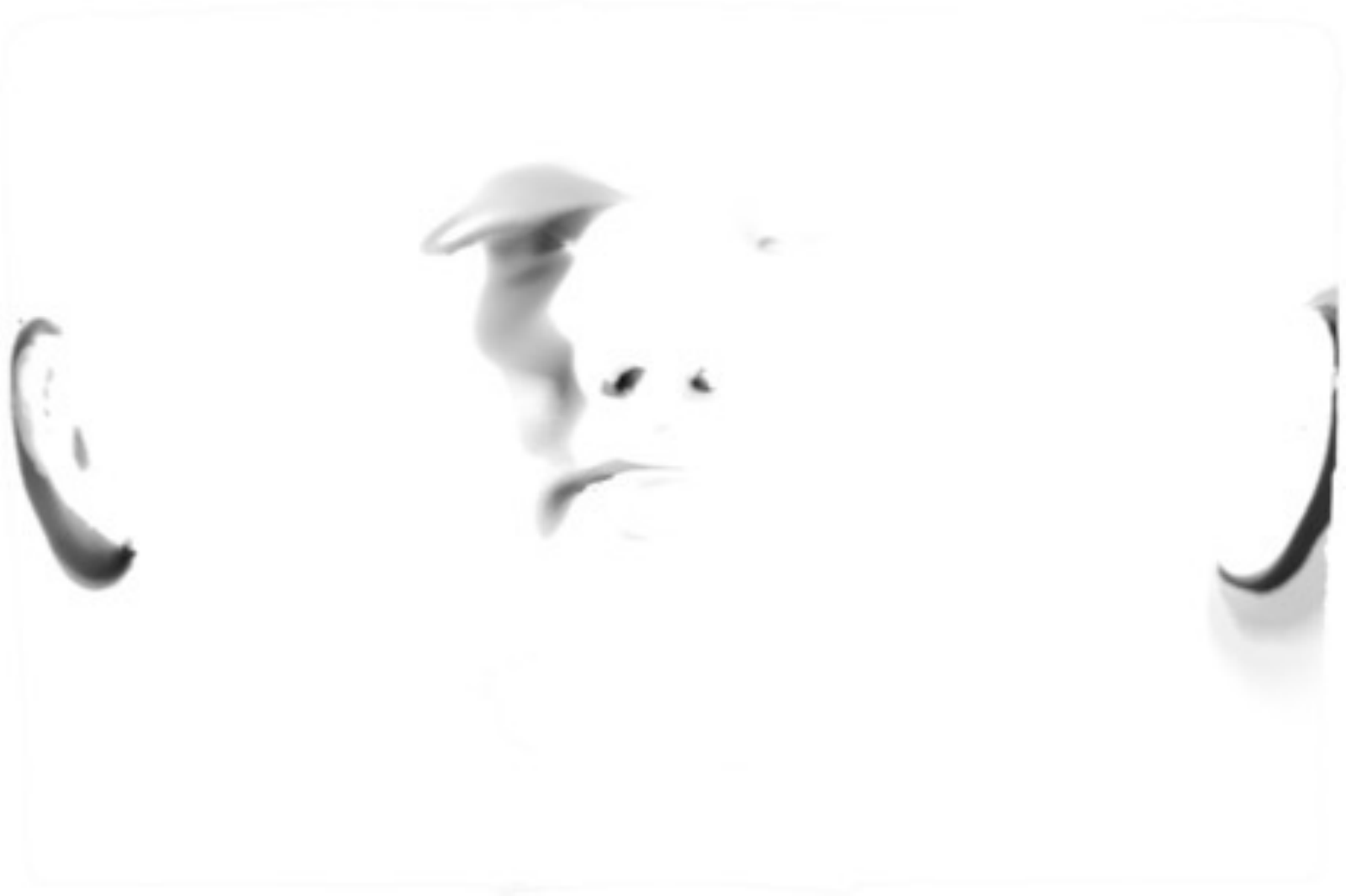}}
        \quad
    \subfloat[
        Generated]{
        \includegraphics[width=0.21\linewidth,frame]{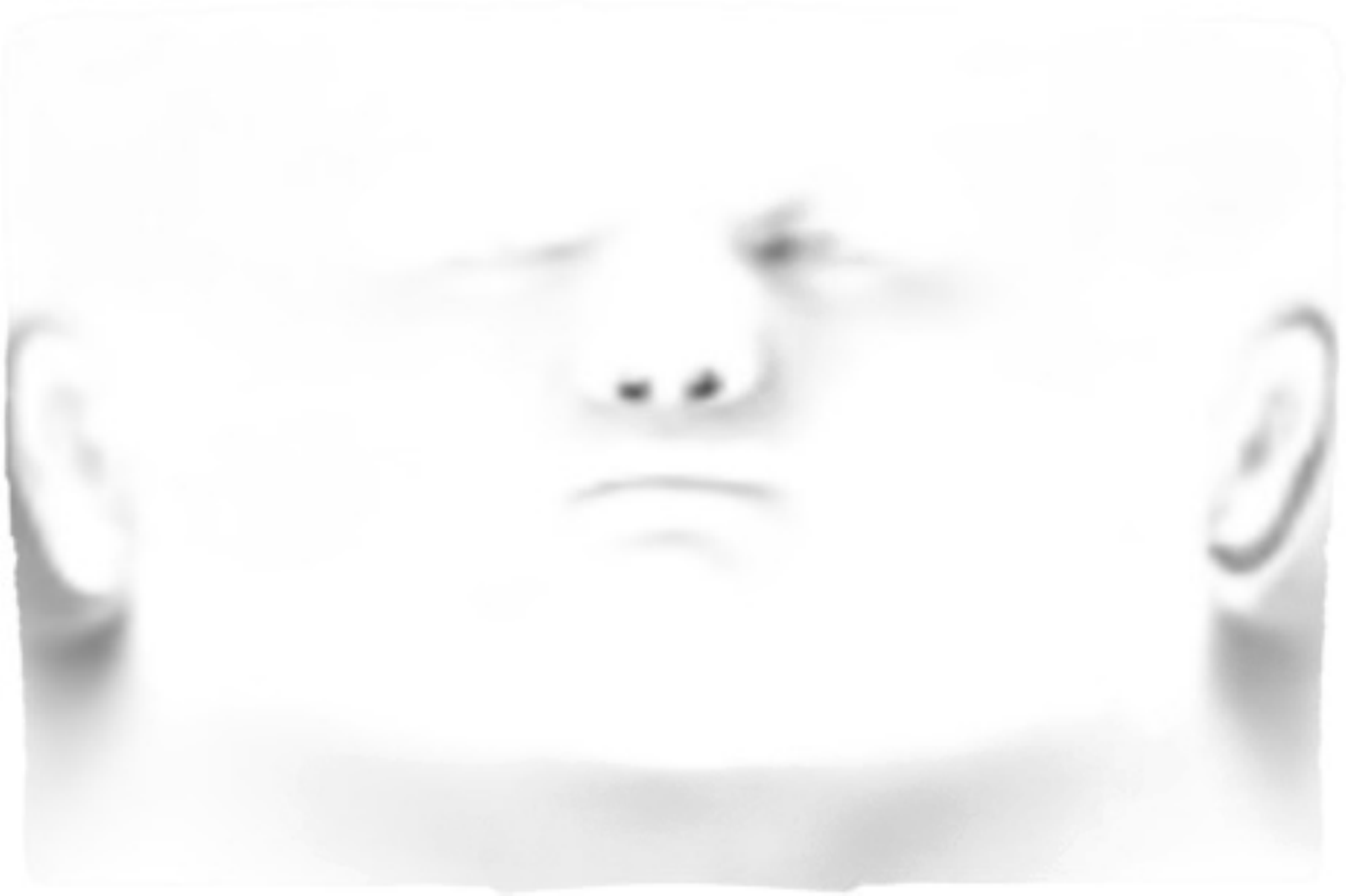}}
    \subfloat[
        Truth]{
        \includegraphics[width=0.21\linewidth,frame]{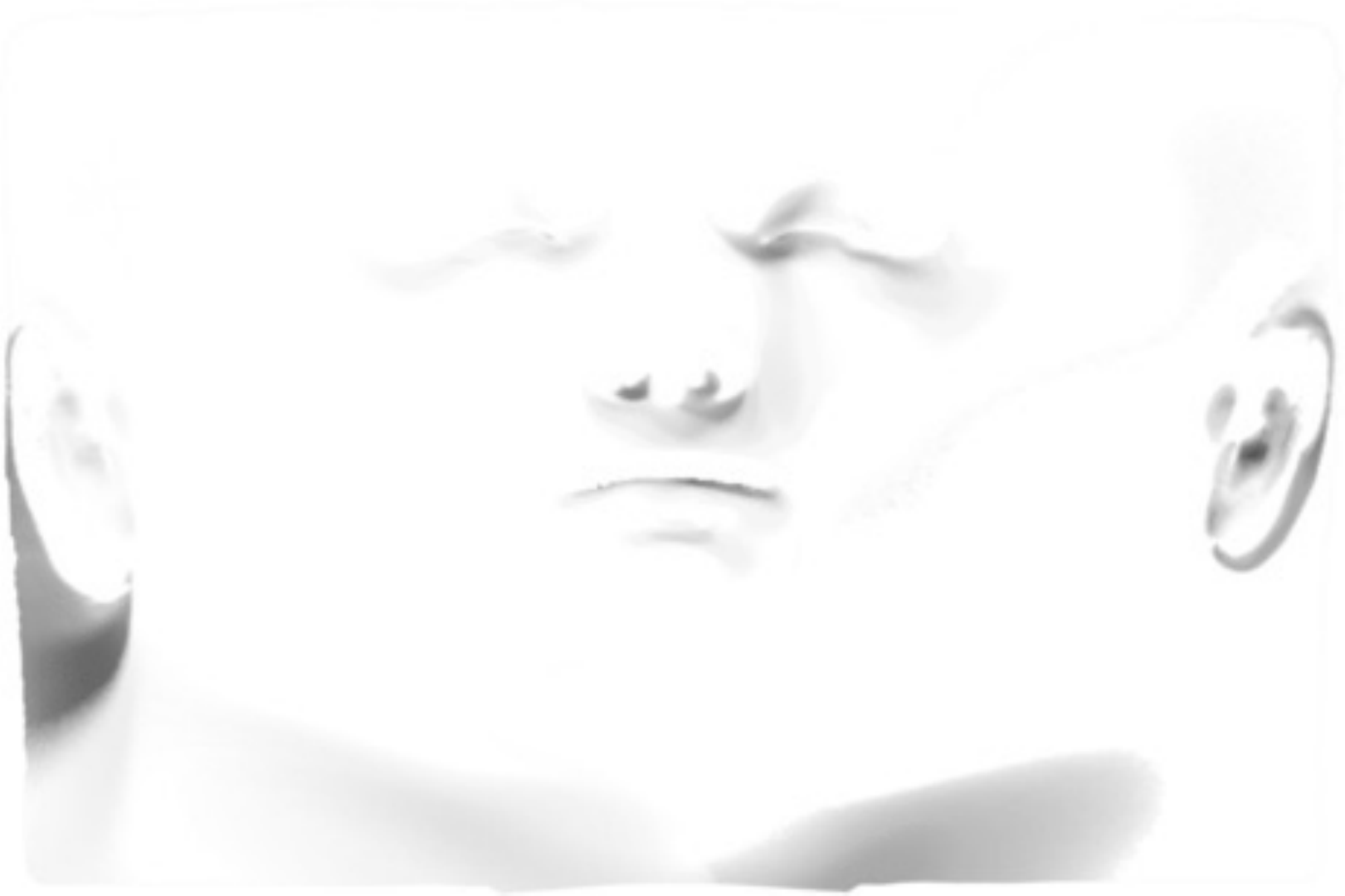}}
    
    \caption{
        Prediction and ground truth
        for our self-occlusion autoencoder $\mathcal{O}$
        (Sec.~\ref{sec:method_diffrender_shadows})
        for randomly sampled sets of 3 light sources,
        as input.
    }
    \label{fig:results_ablation_shadows}
\end{figure}
Self-occlusion on human faces (i.e.~Fig.~\ref{fig:results_ablation_shadows})
does not exhibit sharp edges
and is similar between different facial geometries.
Therefore, we make the following simplifications:
a) We use self-occlusion UV maps rendered from our dataset's mean shape, 
which enables the decoder $\mathcal{O}$ to correctly learn 
a meaningful latent representation,
b) We train the autoencoder on low-resolution inputs
and upsample the needed cropped patch.
Thus, we have fast and differentiable self-occlusion generation 
during training, with a minimal footprint,
so that our main network can learn to remove it.

\subsection{Reflectance Inference with AvatarMe\textsuperscript{++}}

\begin{figure*}[ht]
    \centering
    \includegraphics[width=\linewidth]{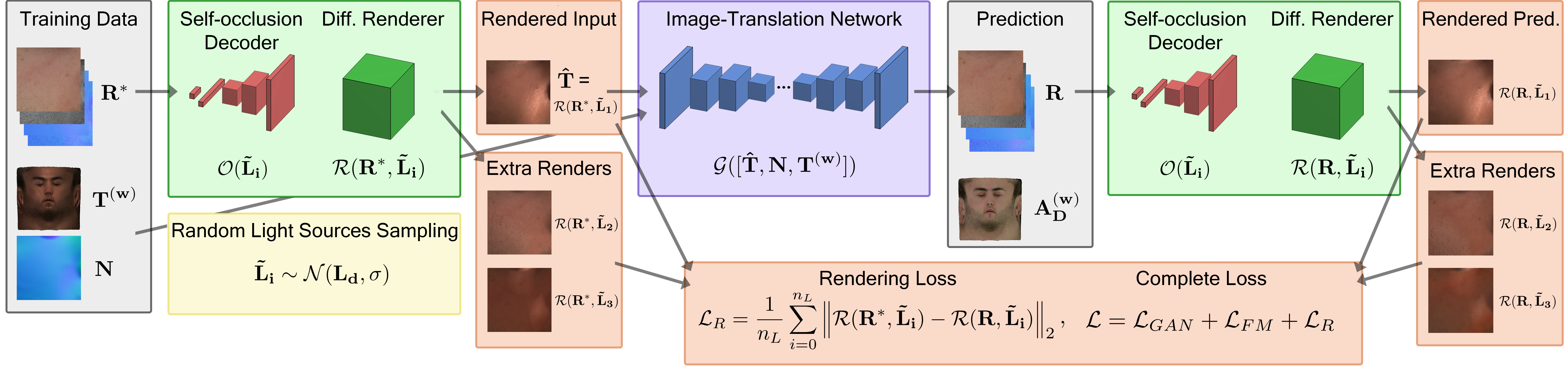}
    \caption{
        \modelnameplus{} training methodology:
        For each iteration
        we render reflectance patches $\mathbf{R^*=[A_D,A_S,N_D,N_S]}$
        from the captured data (RealFaceDB),
        using our differentiable renderer $\mathcal{R}$ (Sec.~\ref{sec:method_diffrender})
        and occlusion autoencoder (Sec.~\ref{sec:method_diffrender_shadows}).
        The rendering parameters
        $\mathbf{\tilde{L}_i} \sim \mathcal{N}(\mathbf{L_d}, \sigma), \quad i=1\dots n_L$ are sampled 
        from a distribution with mean the target 3DMM environment 
        $\mathbf{L_d}$.
        We pass one rendered patch $\mathbf{\hat{T}=\mathcal{R}(\mathbf{R, \tilde{L}_i}})$ to 
        our main network $\mathcal{G}$ (Sec.~\ref{sec:method_recon_renderme})
        and produce the reflectance patches $\mathbf{R}$.
        The training $\mathbf{R^{*}}$ and generated $\mathbf{R}$ reflectance patches are used for the adversarial loss $\mathcal{L}_{GAN}$
        and feature-matching loss $\mathcal{L}_{FM}$.
        Moreover, the consistency of the predicted patches
        is improved by including the down-scaled input texture
        $\mathbf{T^{(w)}}$ in $\mathcal{G}$'s input,
        and the down-scaled diffuse albedo $\mathbf{A_D^{(w)}}$
        in the $\mathcal{G}$'s target.
        Additionally, we render the training and predicted data,
        with each $\mathbf{\tilde{L}_i}$
        and define the rendering loss $\mathcal{L}_R$, as the average loss for each random environment.
    }
    \label{fig:renderme_training_overview}
\end{figure*}

\label{sec:method_recon_renderme}
Here, we introduce a single rendering-aware image-translation network
which jointly generates all reflectance components
($\mathbf{A_D, A_S, N_D, N_S}$) from the upsampled 3DMM texture $\mathbf{\hat{T}}$.
The motivation behind a single network is three-fold:
jointly generating the above reflectance components:
a) enables the introduction of a stochastic photorealistic rendering loss,
which we show that improves the accuracy and generalization of the network,
b) enables the network to share learned parameters between components,
decreasing the size and memory requirement of the network,
c) reduces the required inference time,
since only a single forward pass is required.
Fig.~\ref{fig:renderme_training_overview} shows an overview of this approach.
In this manner, we introduce \modelnameplus{} with a novel fast photorealistic 
differentiable rendering methodology (Sec.~\ref{sec:method_diffrender}),
an updated model architecture (Sec.~\ref{sec:method_recon_renderme_combined})
and a stochastic rendering loss (Sec.~\ref{sec:method_recon_renderme_renderloss})
which greatly improves the the results of \modelname.

\subsubsection{Combined Image-to-Image Translation Model}
\label{sec:method_recon_renderme_combined}
We formulate the reflectance acquisition problem,
as a domain adaptation problem and train a single image-translation network,
The network $\mathcal{G}$ learns an inverse rendering function,
on a UV texture $\mathbf{\hat{T}}$ with baked illumination.

An important challenge of this patch-based inference
is producing consistent patches (especially of diffuse albedo $\mathbf{A_D}$)
that can be seamlessly stitched together.
We find that we can alleviate this issue by including
the whole texture $\mathbf{T^{(w)}} = \mathbf{T}_{\downarrow_{\hat{H}_p\times\hat{H}_p}}$ 
to $\mathcal{G}$'s input 
and the diffuse albedo $\mathbf{A_D^{(w)}} = \mathbf{A_D}_{\downarrow_{\hat{H}_p\times\hat{H}_p}}$
to $\mathcal{G}$'s output,
both downsampled ($\downarrow$) to the same size as the training patches.
By including $\mathbf{T^{(w)}}, \mathbf{A_D^{(w)}}$ in the adversarial loss,
we show that the network can learn the albedo color from the $\mathbf{T^{(w)}}$
and apply it when generating the high-resolution albedo patches.

The input to the generator $\mathcal{G}$ is formulated as the 9D tensor 
$\mathbf{\hat{T}^+} = [\mathbf{\hat{T}}, \mathbf{N_O}, \mathbf{\hat{T}^{(w)}}] \in \mathcal{R}^{\hat{W}\times\hat{H}\times9}$.
The output is a 13D tensor 
$\mathbf{R^+} = [\mathbf{A_D}, \mathbf{A_S}, \mathbf{N_D}, \mathbf{N_S}, \mathbf{A_D^{(w)}}]  \in \mathcal{R}^{\hat{W}\times\hat{H}\times13}$.
Due to the resolution of our textures,
we train the network on randomized $\hat{H}_p\times\hat{H}_p$ patches of 
$\mathbf{\hat{T}^+}$ and $\mathbf{R}$. 
$\mathbf{A_D^{(w)}}$ is only used for the adversarial loss and ignored at testing.
Therefore, the reflectance $\mathbf{R}$ is extracted by the following:
\begin{equation}
    \mathbf{R}
    =
    \left[\mathbf{A_D}, \mathbf{A_S}, \mathbf{N_D}, \mathbf{N_S}\right]
    =
    \mathcal{G}([
        \mathbf{\hat{T}}, \mathbf{N}, \mathbf{\hat{T}^{(w)}}
    ])
    \label{eq:method_renderme_net}
\end{equation}

\subsubsection{Stochastic Rendering Loss}
\label{sec:method_recon_renderme_renderloss}
So far, we have described a single-network $\mathcal{G}$
that generates the facial reflectance $\mathbf{R = [A_D, A_S, N_D, N_S]}$
(Sec.~\ref{sec:method_recon_renderme_combined})
and a fast photorealistic differentiable rendering method (Sec.~\ref{sec:method_diffrender}).
Therefore, we can now introduce a rendering loss in the training of $\mathcal{G}$,
where the input texture $\mathbf{T_d}$ is compared with the predicted reflectance,
rendered in the domain $\mathbf{L_d}$ of the input texture.
The photorealistic rendering loss is defined as 
$\mathcal{L}_R = \|\mathbf{T_d} - \mathcal{R}(\mathbf{R, L_d})\|_2$,
using Eq.~\ref{eq:diff_render_shadow}.
Please note that $\mathcal{R}$ also uses the shape $\mathbf{S_O}$
which remains static during training and we omit it to avoid cluttering.
A similar approach has been proposed for 
planar surfaces under flash illumination in 
\cite{deschaintre_single-image_2018, li_materials_2018}.
To the best of our knowledge, this is the first attempt on facial BRDF
and the first included in a GAN-based image-translation network.
Other facial acquisition methods (i.e.~\cite{gecer_ganfit_2019, genova_unsupervised_2018, shu_neural_2017, sengupta_sfsnet_2018, tewari_mofa_2017})
that use a differentiable rendering loss, 
merely re-project the inferred texture or use the unrealistic Lambertian model.

The introduction of the above rendering loss 
leaves the network unaware of specular features across the whole face,
while it motivates the network 
to include shading elements in the diffuse albedo,
which could be accurately reproduced in the rendering.
This is mainly because all the training data are rendered in the same domain of 
the 3DMM textures used (Sec.~\ref{sec:method_data_env_estimation}) 
and the network can always expect shadows
and highlights at the same places and intensities.
Therefore we take 2 steps to introduce stochasticity to our training data,
similar to \cite{deschaintre_single-image_2018},
which improves our network's accuracy and generalization outside the target 3DMM domain.

Firstly, we sample random variations of the estimated target environment parameters
$\mathbf{\tilde{L}} \sim \mathcal{N}(\mathbf{L_d}, \sigma)$ in each training iteration.
We use it to render both the input to the network 
$\mathbf{\tilde{T}} = \mathcal{R}(\mathbf{R^{*}, \tilde{L})}$
and the predicted reflectance $\mathcal{R}(\mathbf{R}, \mathbf{\tilde{L})}$
for the rendering loss.
Secondly, for each training iteration,
we sample additional $n_L$ environment parameters,
$\mathbf{\tilde{L}_i} \sim \mathcal{N}(\mathbf{L_d}, \sigma), i = 1\dots n_L$
which are not fed to the network, 
but are only used to compute an average rendering loss
for these environments.
In this manner, we stochastically approximate all light source directions within the allowed variation,
while also penalizing the network for over-fitting the target 3DMM environment.
Then, for the ground truth $\mathbf{R^{*}}$
and the inferred reflectance $\mathbf{R}$, for each training iteration,
the rendering loss $\mathcal{L}_{R}$ is defined as:
\begin{equation}
    \mathcal{L}_{R}
    =
    \frac{1}{n_L}
    \sum_{i = 0}^{n_L}
    \left\| 
        \mathcal{R}(\mathbf{R^*}, \mathbf{\tilde{L}_i})
        -
        \mathcal{R}(\mathbf{R}, \mathbf{\tilde{L}_i})
    \right\|_2,
    \mathbf{\tilde{L}_i} \sim \mathcal{N}(\mathbf{L_d}, \sigma)
    \label{eq:recon_renderme_renderloss}
\end{equation}

Overall,
the objective of our image-translation network,
which is based on pix2pixHD \cite{wang_high-resolution_2018} is defined as:
\begin{equation}
    \min_G \left(
        \max_D \mathcal{L}_{GAN}(G, D_k)
        +
        \lambda_{FM}
        \mathcal{L}_{FM}(G, D_k)
        +
        \lambda_{R}
        \mathcal{L}_{R}
    \right)
    \label{eq:recon_renderme_loss}
\end{equation}
where $\mathcal{L}_{GAN}(G, D_k)$ is the sum of adversarial loss
and $\mathcal{L}_{FM}(G, D_k)$ is the feature matching loss,
for all 3 discriminators of pix2pixHD \cite{wang_high-resolution_2018}.
The feature matching and rendering loss terms 
are controlled by $\lambda_{FM}$ and $\lambda_{R}$.

\section{Experiments}
\label{sec:experiments}
\subsection{Implementation Details}
\label{sec:experiments_impl}
The task of disentagling the diffuse and specular components,
from a given input image with baked illumination
can be formulated as an image-to-image translation problem. 
Nevertheless, as discussed previously: 
(a) our captured data are of very high-resolution (more than 4K) 
and thus cannot be used for training ``as-is',
due to hardware limitations 
(note not even on a 32GB GPU we can fit such high-resolution data in their original format), 
(b) pix2pixHD \cite{wang_high-resolution_2018} takes into account 
only the texture and optionally labels,
and thus geometric details,
in the form of the shape and shape normals cannot be exploited 
to improve the quality of the generated diffuse and specular components.

\subsubsection{Patch-Based Image-to-Image Translation}
\label{sec:experiments_impl_patch}
To alleviate the aforementioned shortcomings, we
split the original high-resolution data into smaller patches of $\hat{H}_p\times\hat{H}_p$ size. 
More specifically, using a stride of size $256$, 
we derive the partially overlapping patches 
by passing through each original UV horizontally as well as vertically.
For each translation task we utilize the shape or shape normals,
projected and interpolated in the same UV parameterisation as the textures.
This increases the accuracy and level of detail in 
the derived outputs as the geometry act as a ``guide'' to the network.
Finally, we downsample the whole input texture to the patch size and
include it as well, as we find it greatly improves the consistency of the predicted patches.
For the AvatarMe\textsuperscript{++} pipeline,
we concatenate them channel-wise with the texture input 
and thus feed to the network a $9$D tensor comprising of $\mathbf{\hat{T}}, \mathbf{N_O}, \hat{\mathbf{T}}^{(w)}$ 
and generate a $13$D tensor comprising of $\mathbf{A_D}, \mathbf{A_S}, \mathbf{N_D}, \mathbf{N_S}$ (Eq.~\ref{eq:method_renderme_net}) and $\hat{\mathbf{T}}^{(w)}$, which is discarded.
During inference, that patch size can be larger (e.g.~$1k\times1k$),
since the network is fully-convolutional.

\subsubsection{Training Setup}
\label{sec:experiments_impl_training}
To train RCAN \cite{zhang_image_2018}, we use the default hyper-parameters.
For the rest of the translation of models, 
we use a custom translation network as described earlier, 
which is based on pix2pixHD \cite{wang_high-resolution_2018}. 
More specifically, we use $9$ and $3$ residual blocks 
in the global and local generators, respectively. 
The learning rate we used is $0.0002$, 
whereas the Adam betas are $0.5$ for $\beta_1$ and $0.999$ for $\beta_2$. 
In our best model, we use a feature matching loss controller of $\lambda_{FM}=10.0$
and rendering loss controller of $\lambda_R=0.3$,
for which perform an ablation study in the following section.
Finally, we use a variable number of input and outputs as $N-$dimensional tensors,
based on the implementation 
(Sec.~\ref{sec:method_recon_avatarme_delight},
\ref{sec:method_recon_avatarme_specAlb}, or \ref{sec:method_recon_renderme}).
As mentioned earlier, this substantially improves the results by 
accentuating the details and enforcing patch consistency.

\subsubsection{Rendering Setup}
\label{sec:experiments_impl_rendering}
To implement the facial photorealistic differentiable rendering (Sec.~\ref{sec:method_diffrender}),
we extend the recently published PyTorch3D \cite{nikhila_ravi_pytorch3d_2020}
(version 0.3.0),
for its speed, easily modifiable modular design
and its compatibility with our image-translation network.
Specifically, we fully integrate it with the generator and discriminator networks
of our image-translation network,
implement objects for multiple reflectance textures
and implement a texture-space shader,
that uses our framework from Sec.~\ref{sec:method_diffrender}.
For the shader's parameters, i.e.~shininess exponent and translucency masks,
we use a single manually created UV map with spatially varying values.

For the self-occlusion prediction, we train an autoencoder 
and a linear regressor that maps light source features to the autoencoder's latent values.
The encoder and decoder consist of 5 convolutional blocks,
each block having 2 convolutional layers with ELU \cite{clevert2015fast},
batch normalization \cite{ioffe2015batch} and a down-~or up-sampling layer.
The hidden layer has 64 features and connects to the encoder and decoder
with a fully connected layer of 256 features.

\subsection{Evaluation}
\label{sec:experiments_eval}

\begin{figure*}[ht]
\centering
\newcommand\myheight{1.85cm}
\captionsetup[subfigure]{labelformat=empty}
    \subfloat{
        \includegraphics[height=\myheight{}]{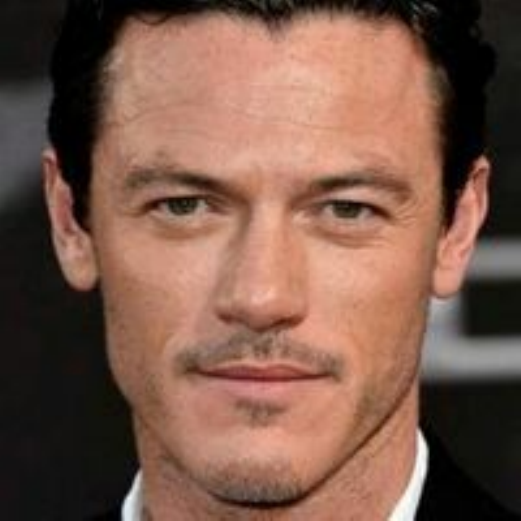}}
    \subfloat{
        \includegraphics[height=\myheight{}]{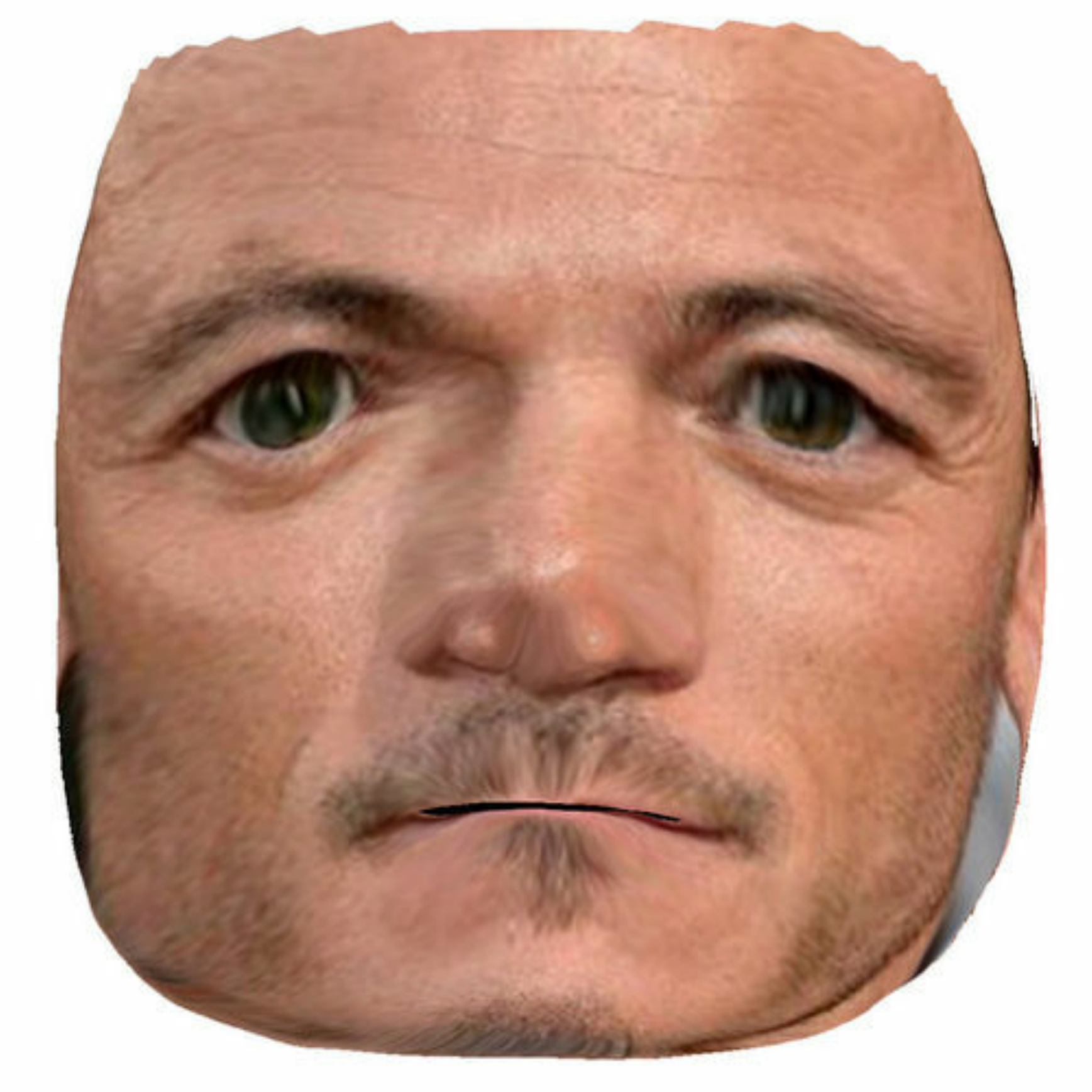}}
    \subfloat{
        \includegraphics[height=\myheight{}]{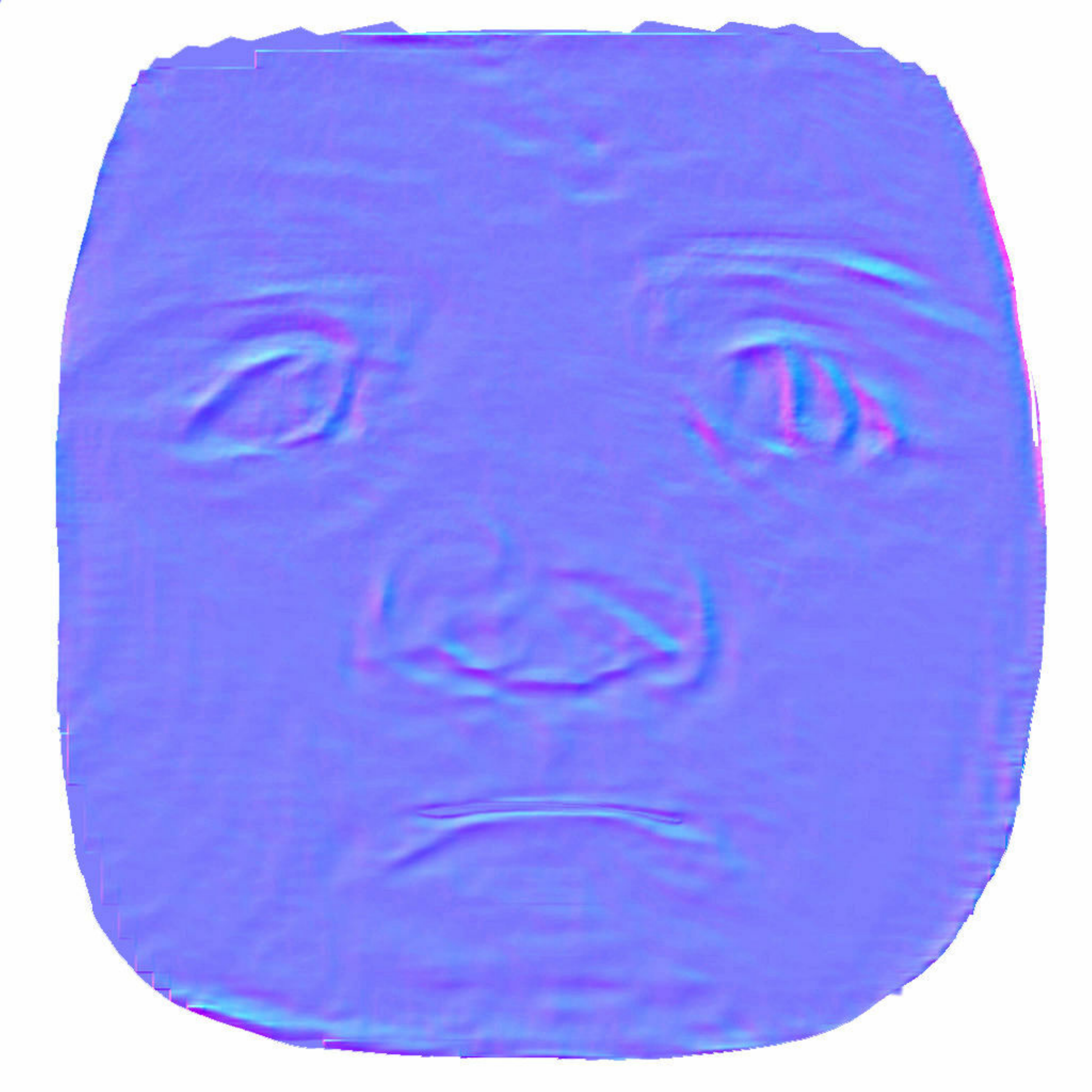}}
    \subfloat{
        \includegraphics[height=\myheight{}]{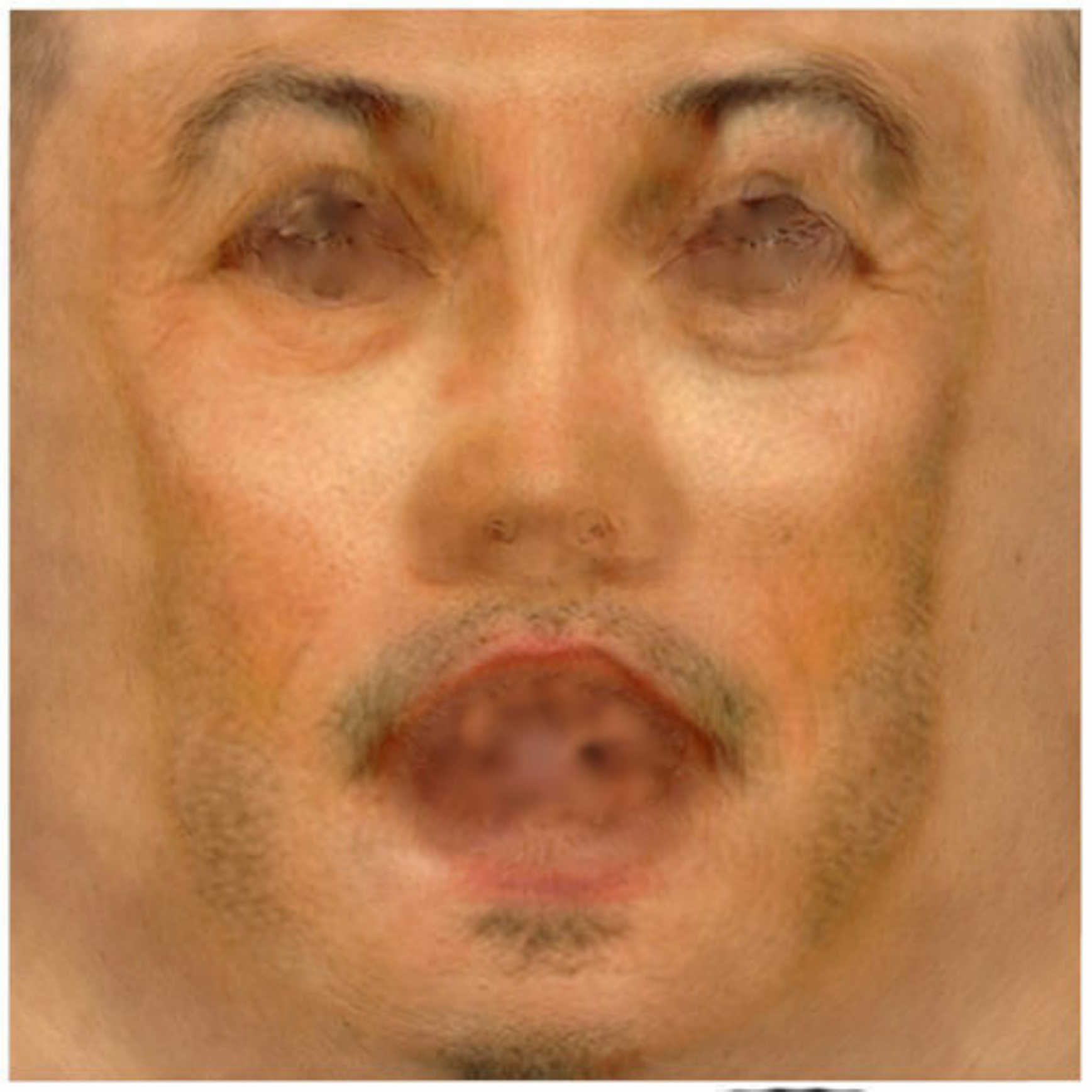}}
    \subfloat{
        \includegraphics[height=\myheight{}]{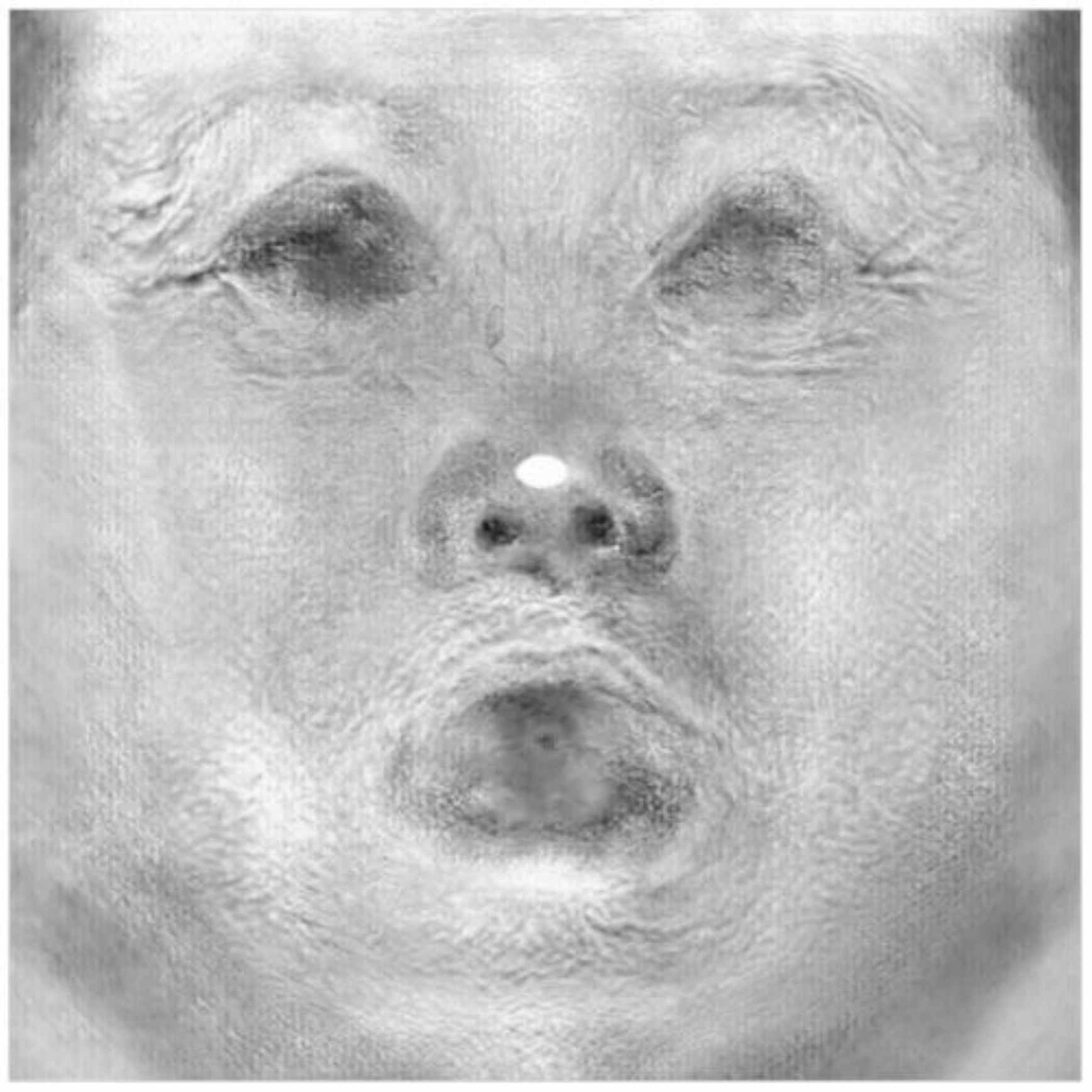}}
    \subfloat{
        \includegraphics[height=\myheight{}]{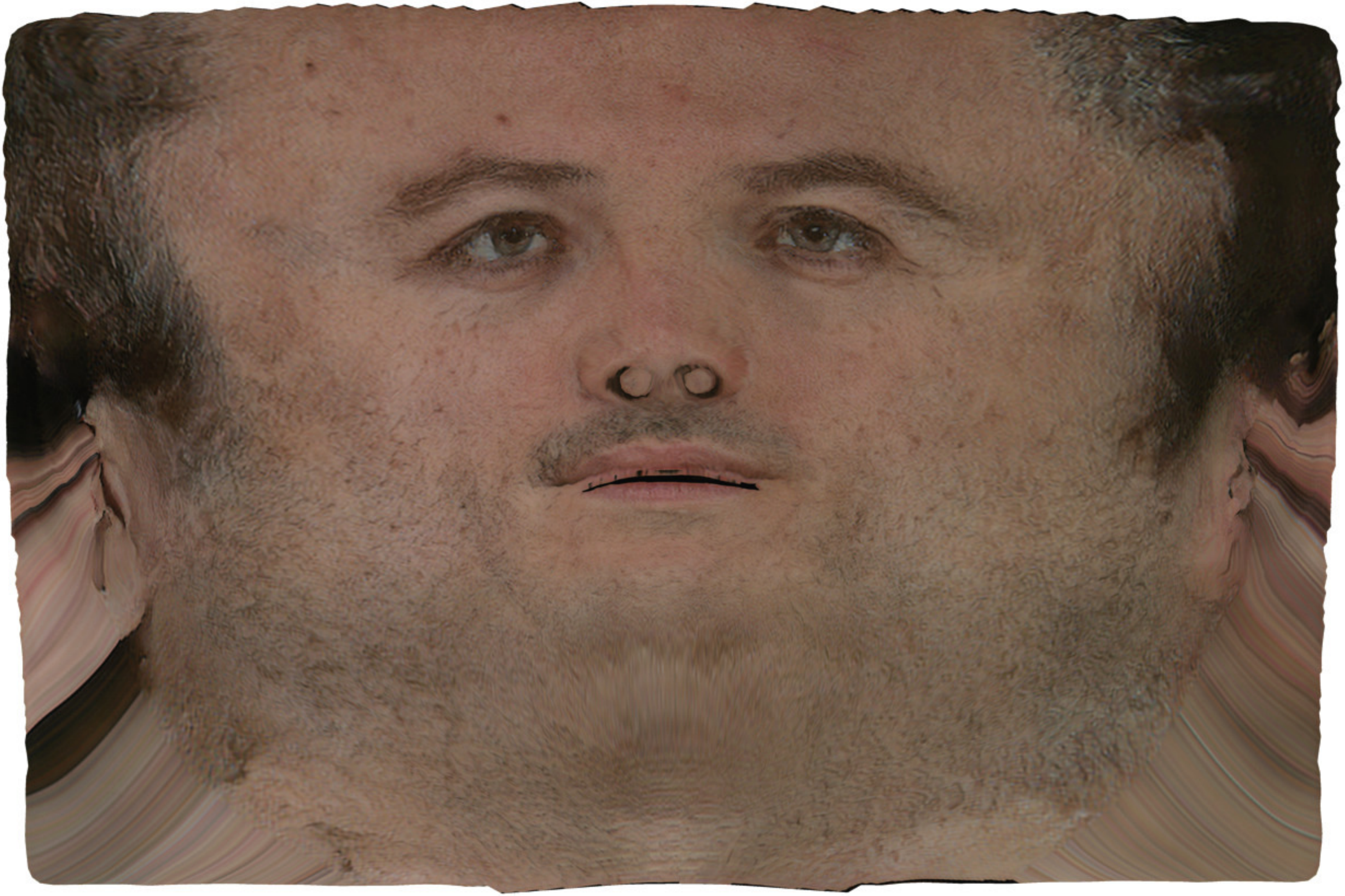}}
    \subfloat{
        \includegraphics[height=\myheight{}]{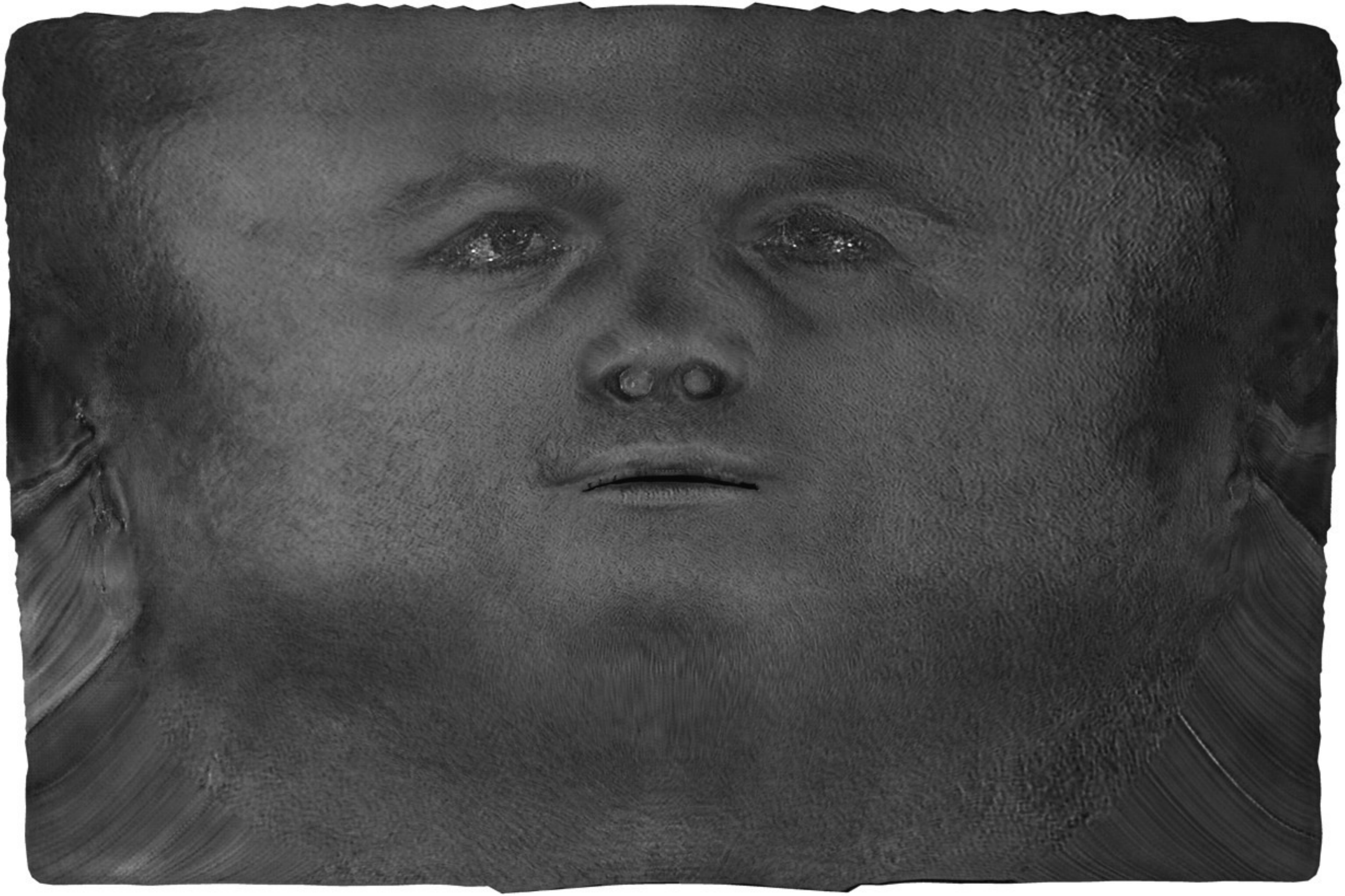}}
    \subfloat{
        \includegraphics[height=\myheight{}]{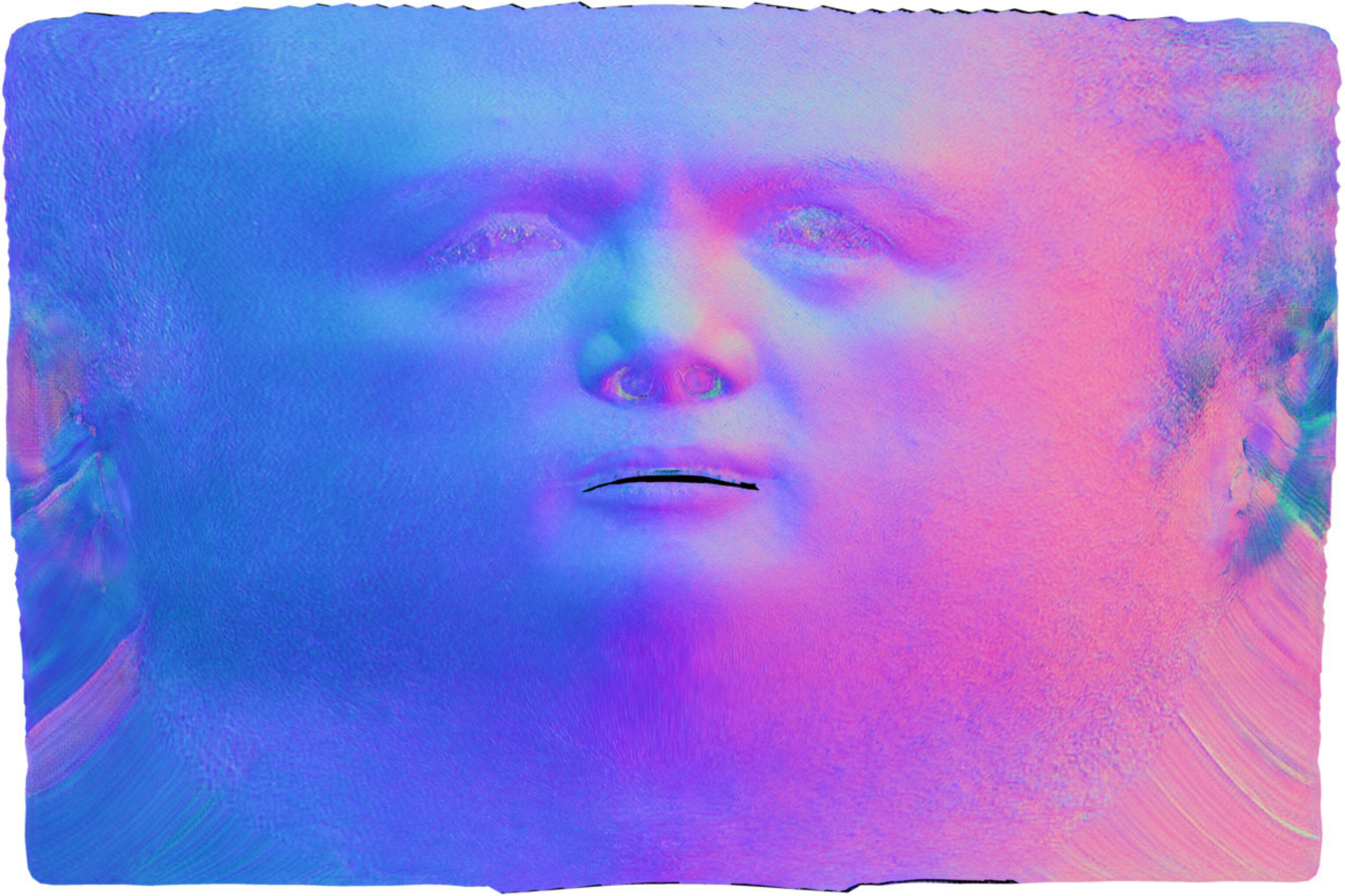}}\\
    \subfloat{
        \includegraphics[height=\myheight{}]{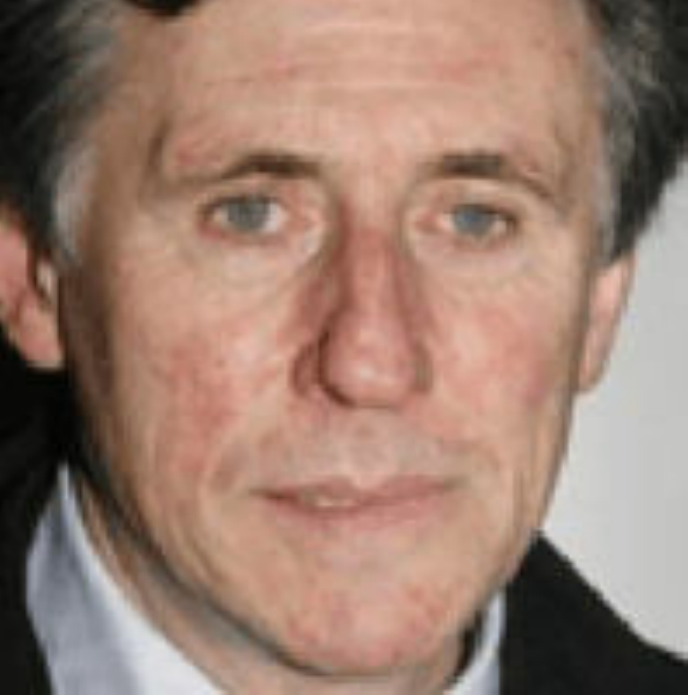}}
    \subfloat{
        \includegraphics[height=\myheight{}]{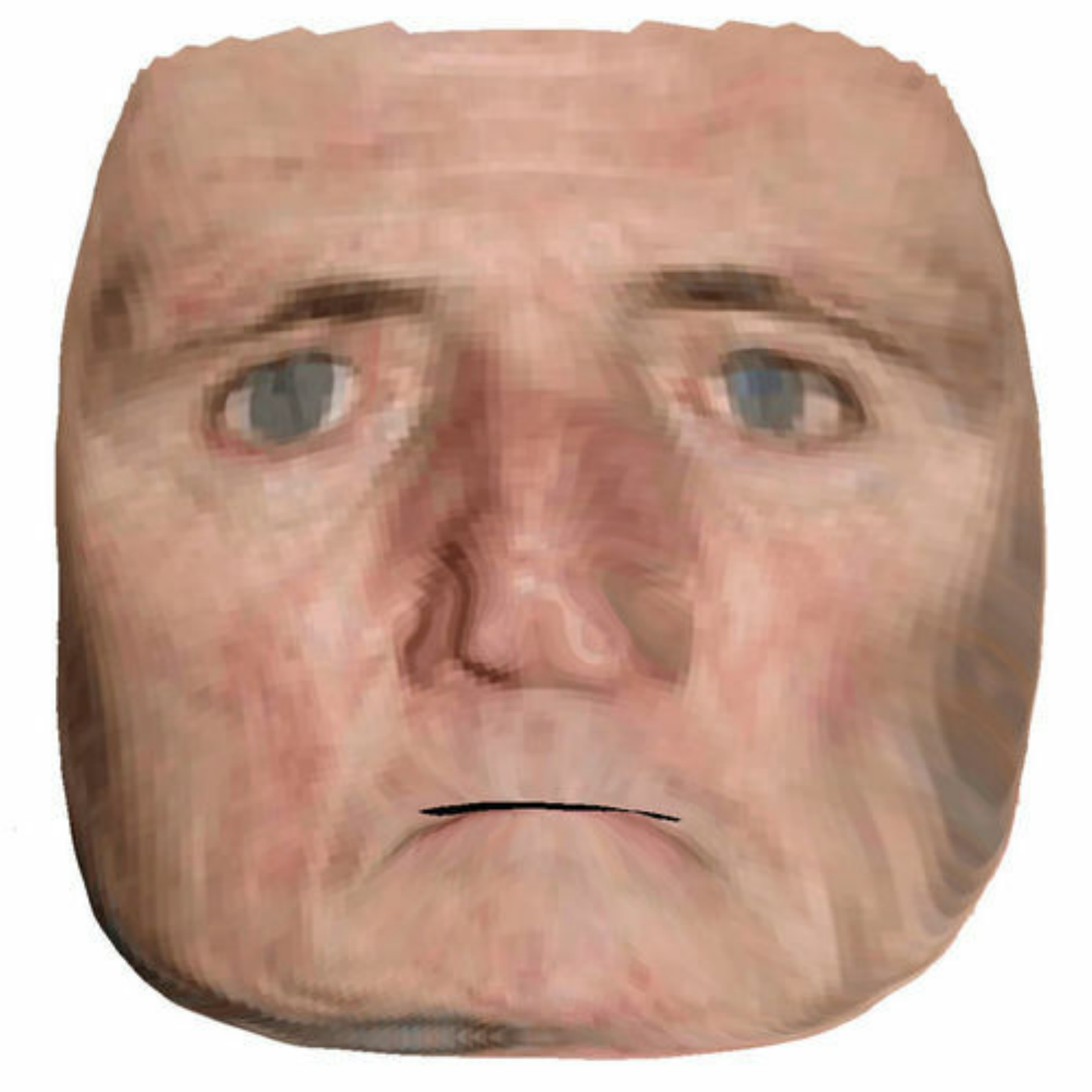}}
    \subfloat{
        \includegraphics[height=\myheight{}]{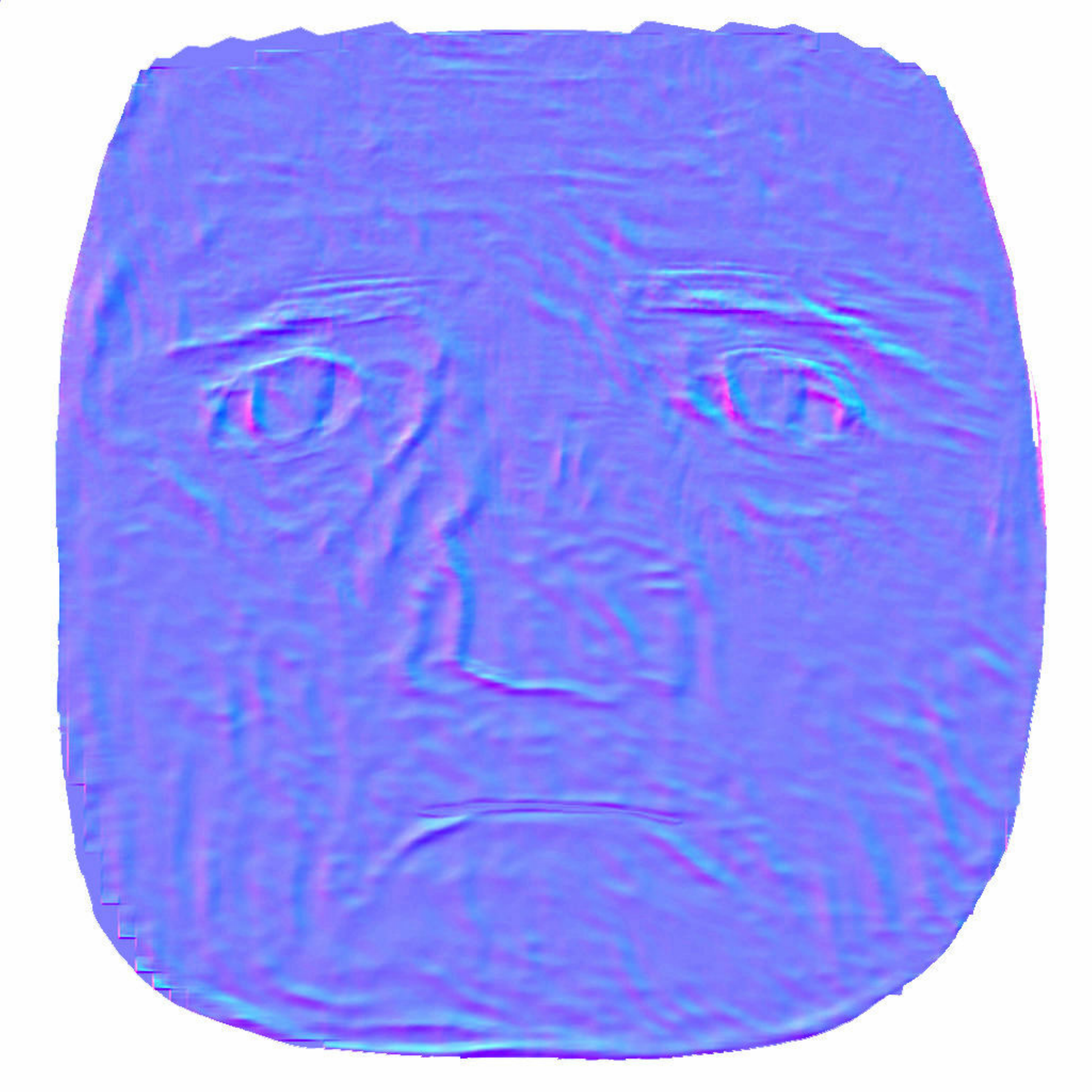}}
    \subfloat{
        \includegraphics[height=\myheight{}]{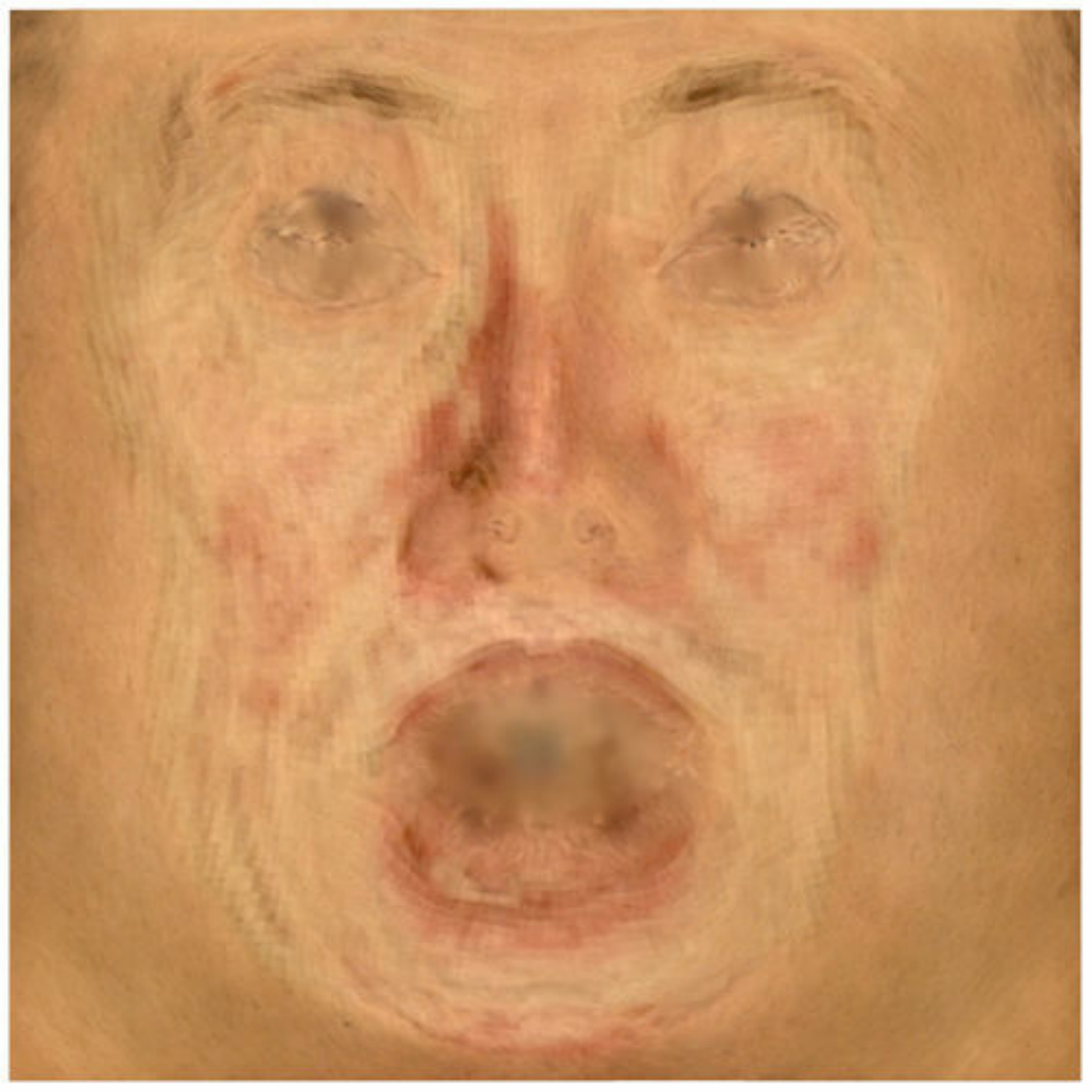}}
    \subfloat{
        \includegraphics[height=\myheight{}]{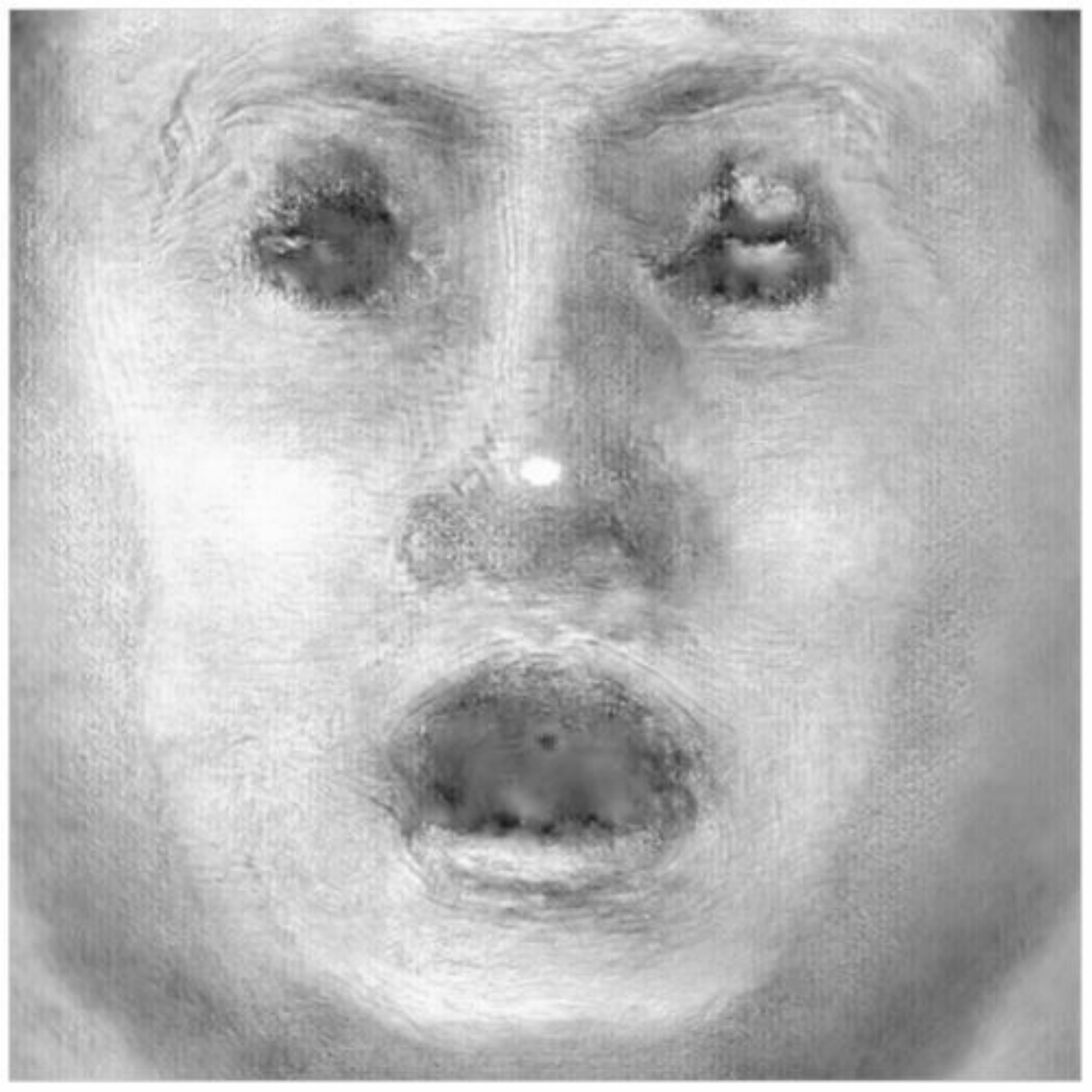}}
    \subfloat{
        \includegraphics[height=\myheight{}]{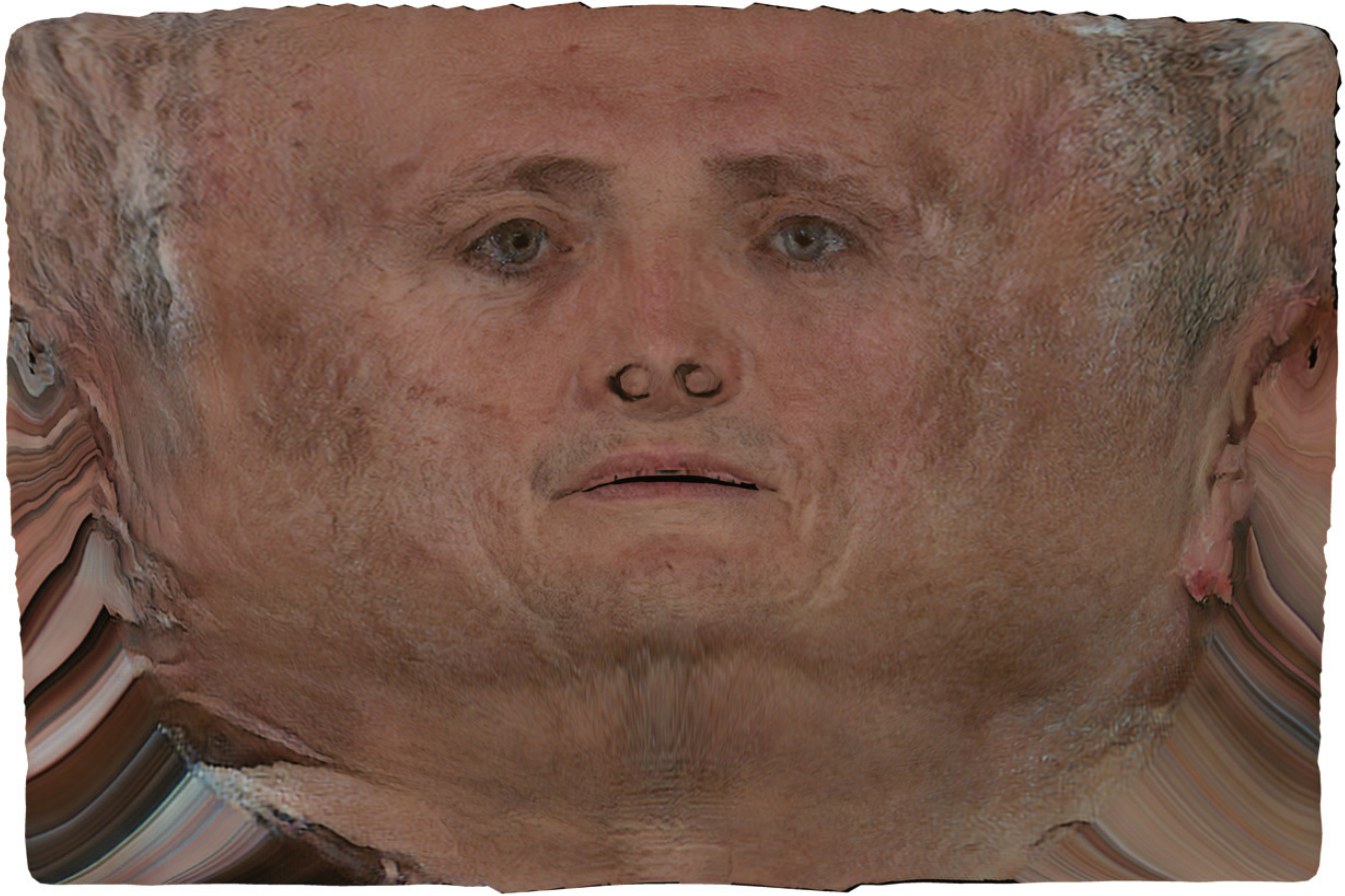}}
    \subfloat{
        \includegraphics[height=\myheight{}]{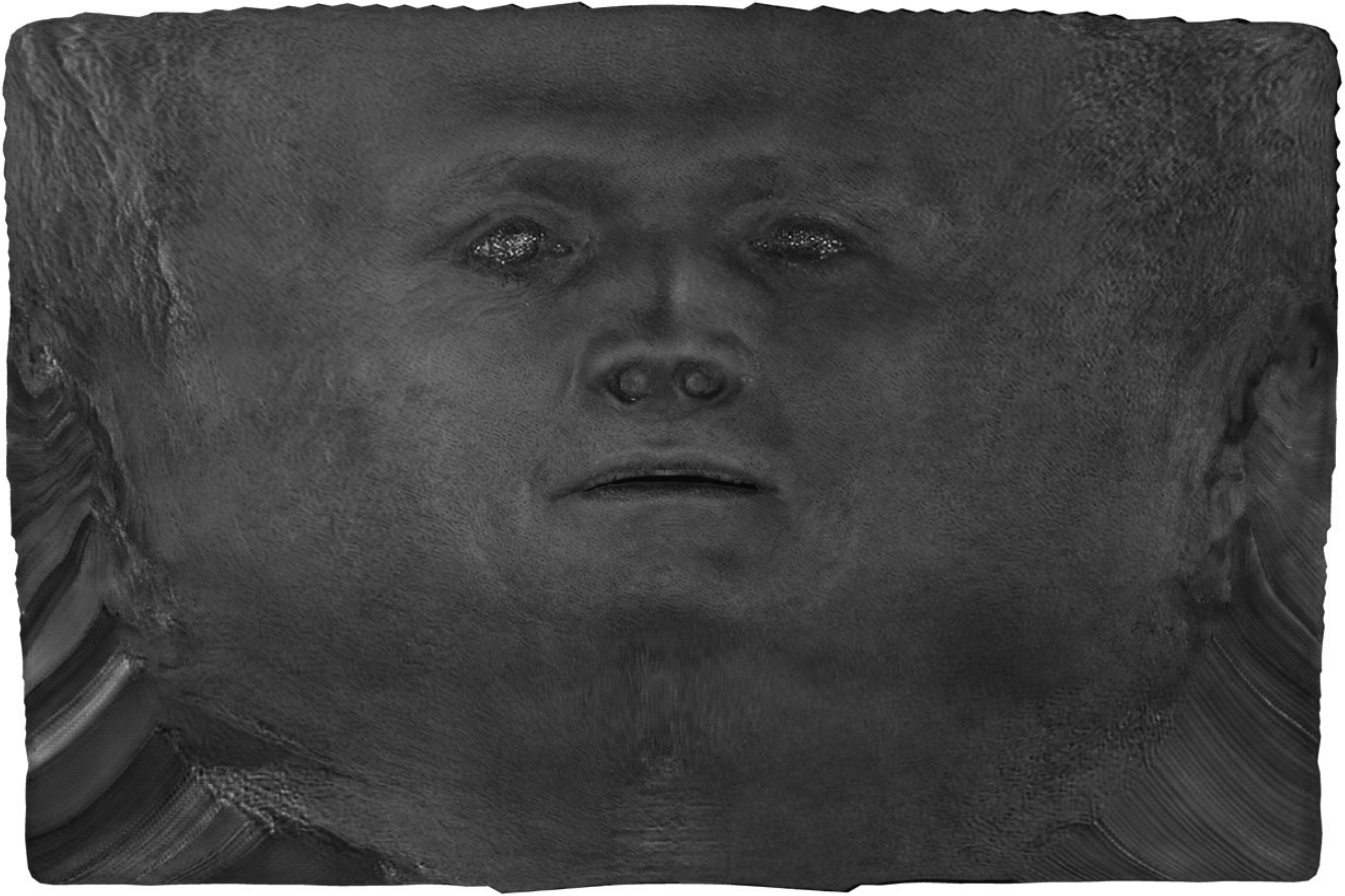}}
    \subfloat{
        \includegraphics[height=\myheight{}]{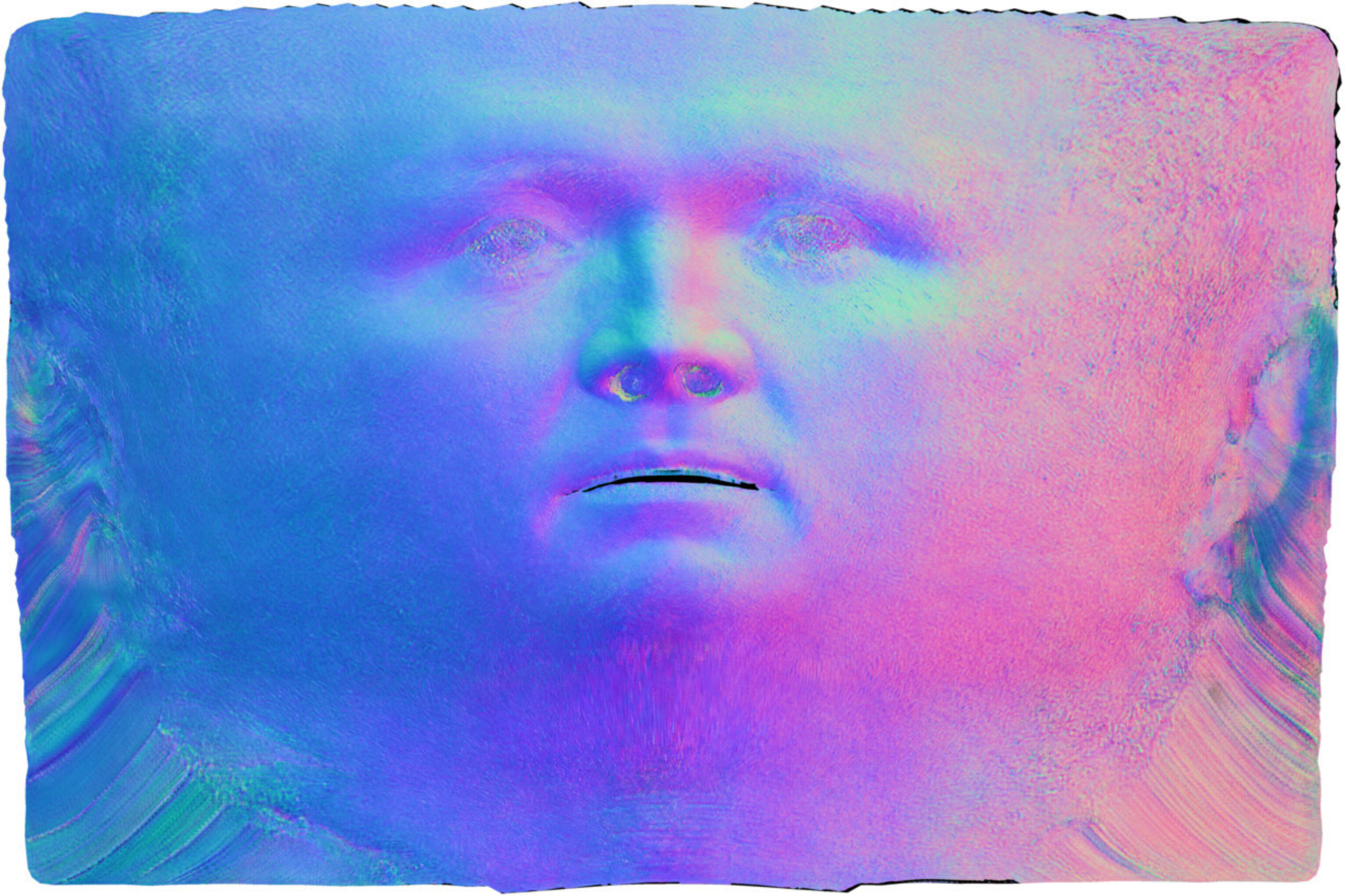}}\\
    \captionsetup[subfloat]{justification=centering}
   \subfloat[Input]{
        \includegraphics[height=\myheight{}]{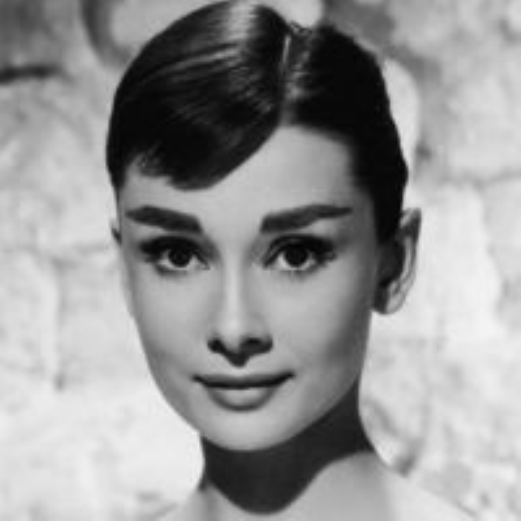}}
    \subfloat[Albedo.~\cite{chen_photo-realistic_2019}]{
        \includegraphics[height=\myheight{}]{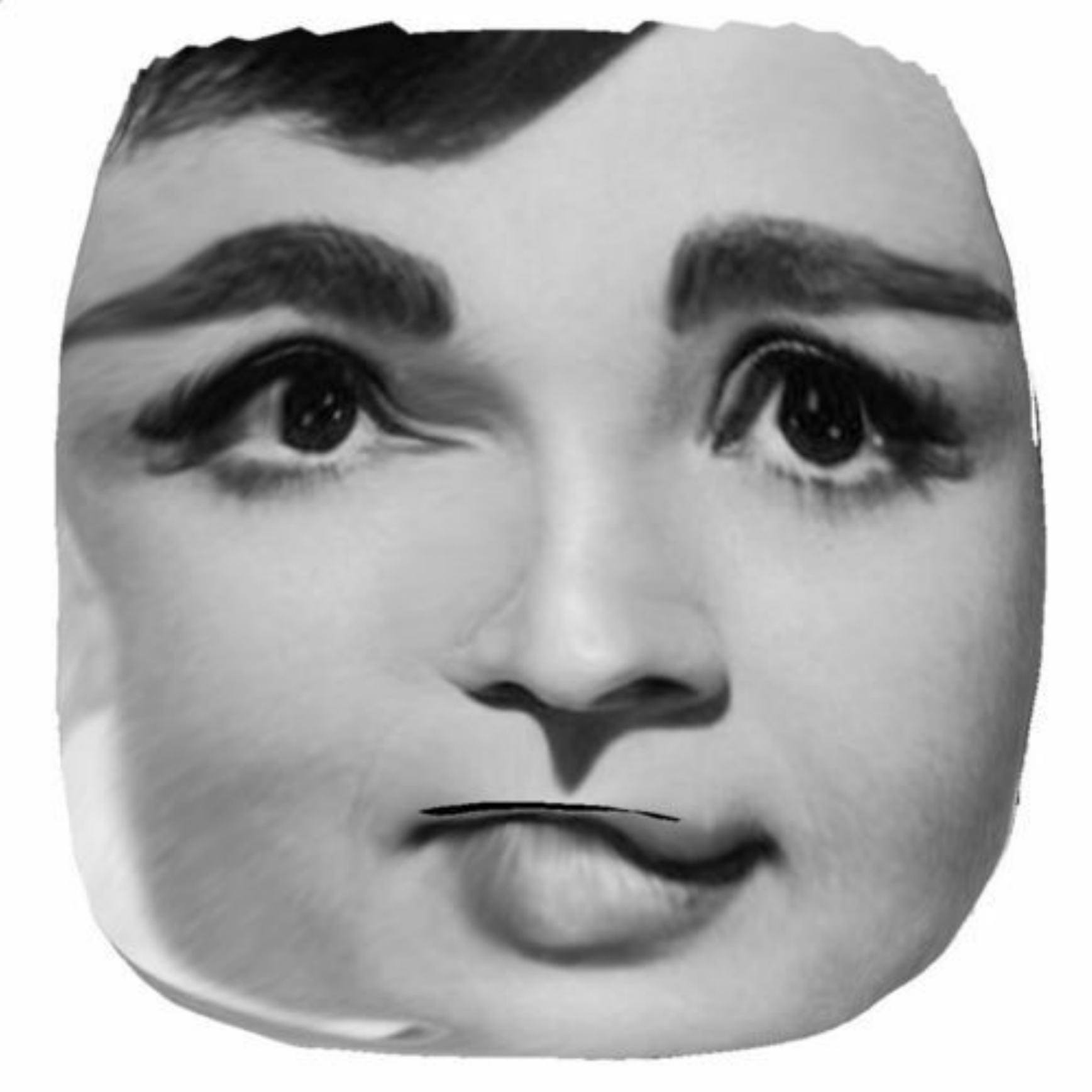}}
    \subfloat[Normals~\cite{chen_photo-realistic_2019}]{
        \includegraphics[height=\myheight{}]{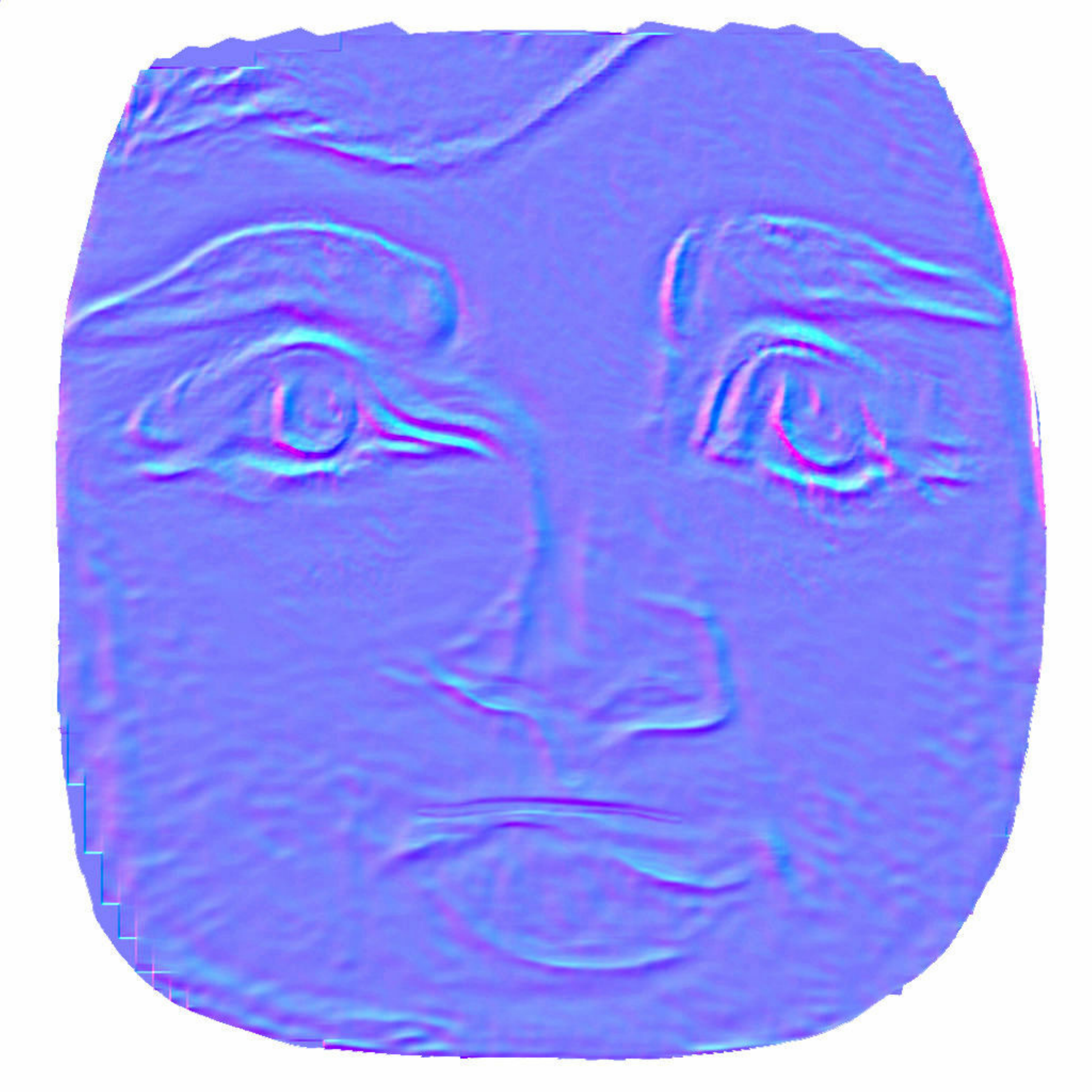}}
    \subfloat[Diff.~Alb.~\cite{yamaguchi_high-fidelity_2018}]{
        \includegraphics[height=\myheight{}]{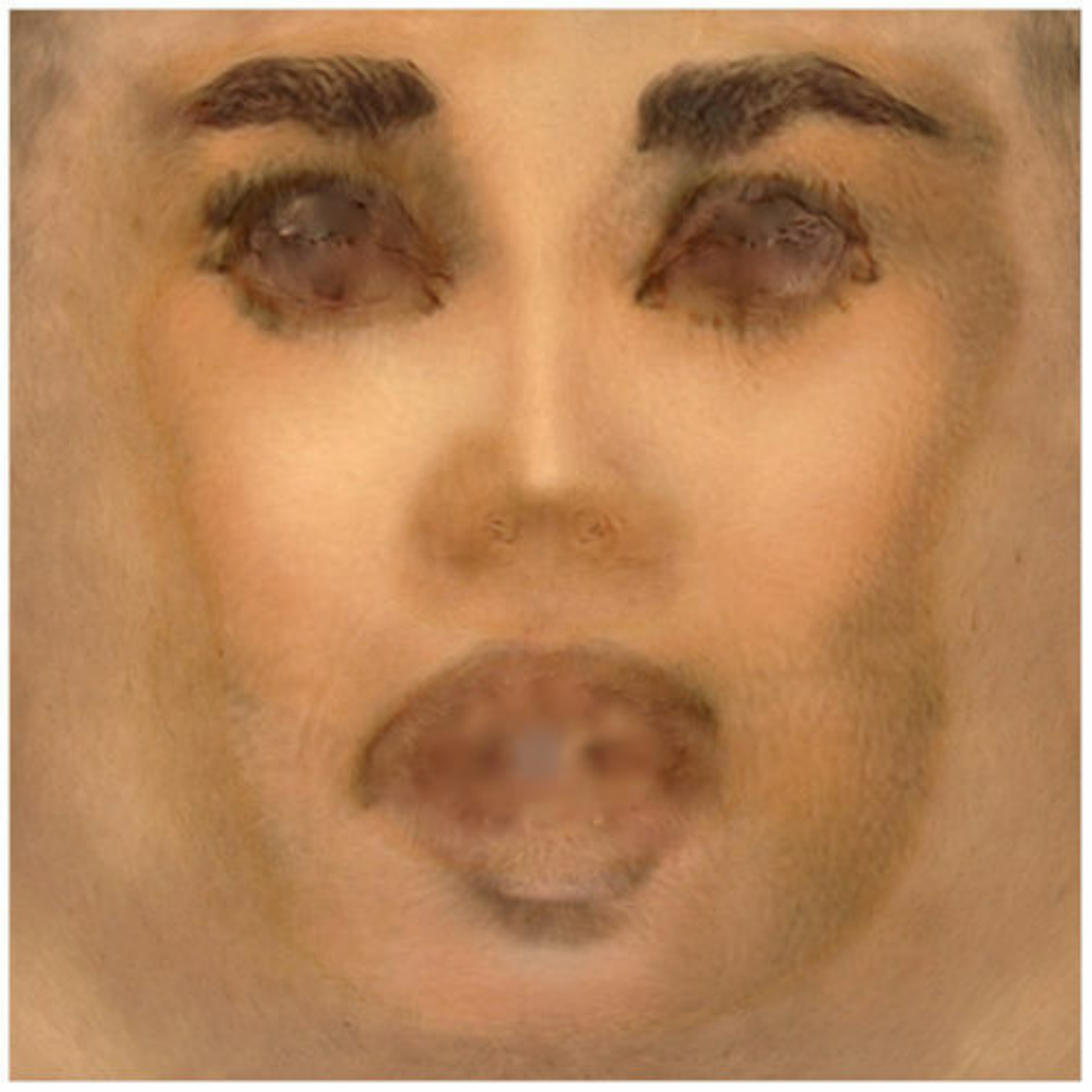}}
    \subfloat[Spec.~Alb.~\cite{yamaguchi_high-fidelity_2018}]{
        \includegraphics[height=\myheight{}]{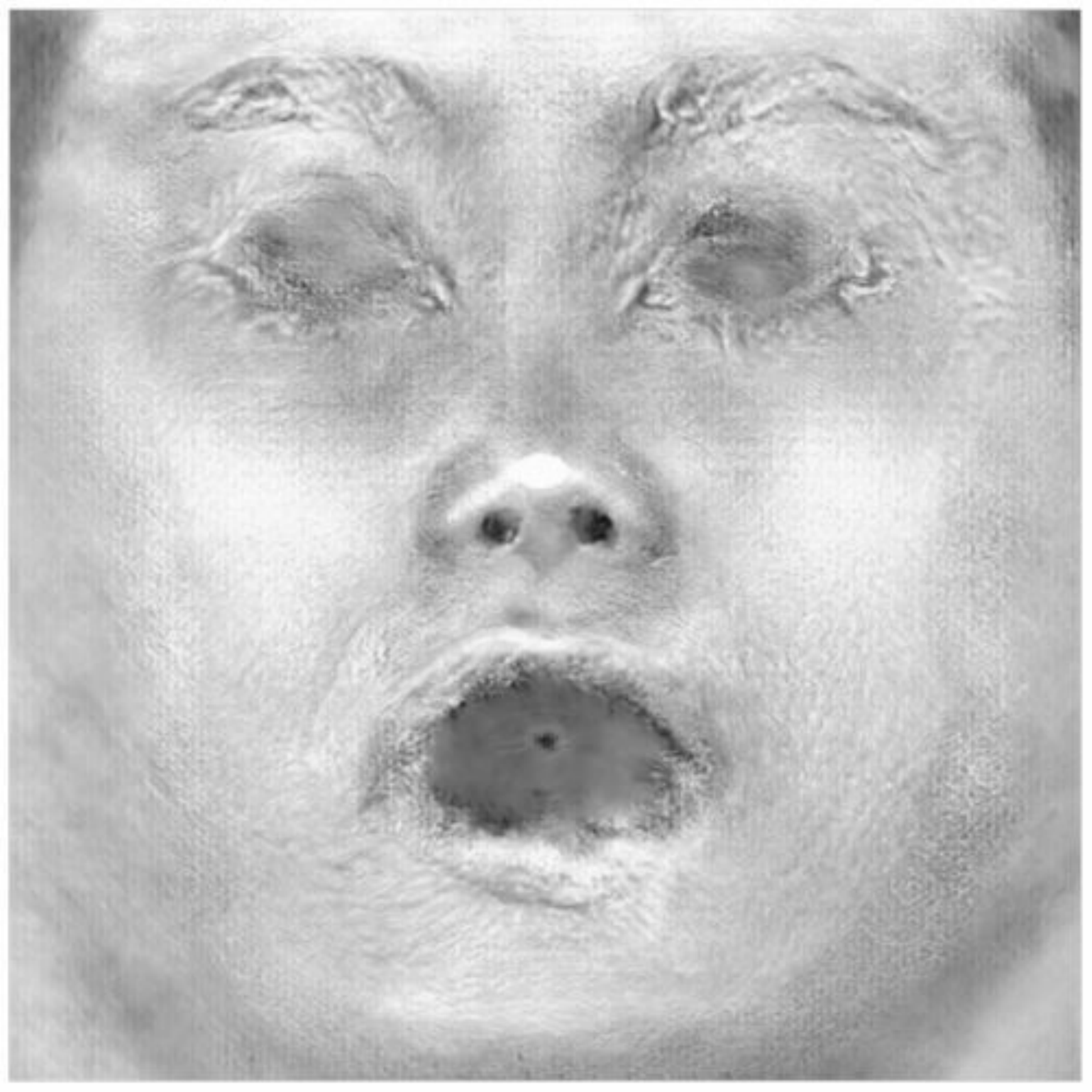}}
    \subfloat[Diffuse~Albedo~Ours]{
        \includegraphics[height=\myheight{}]{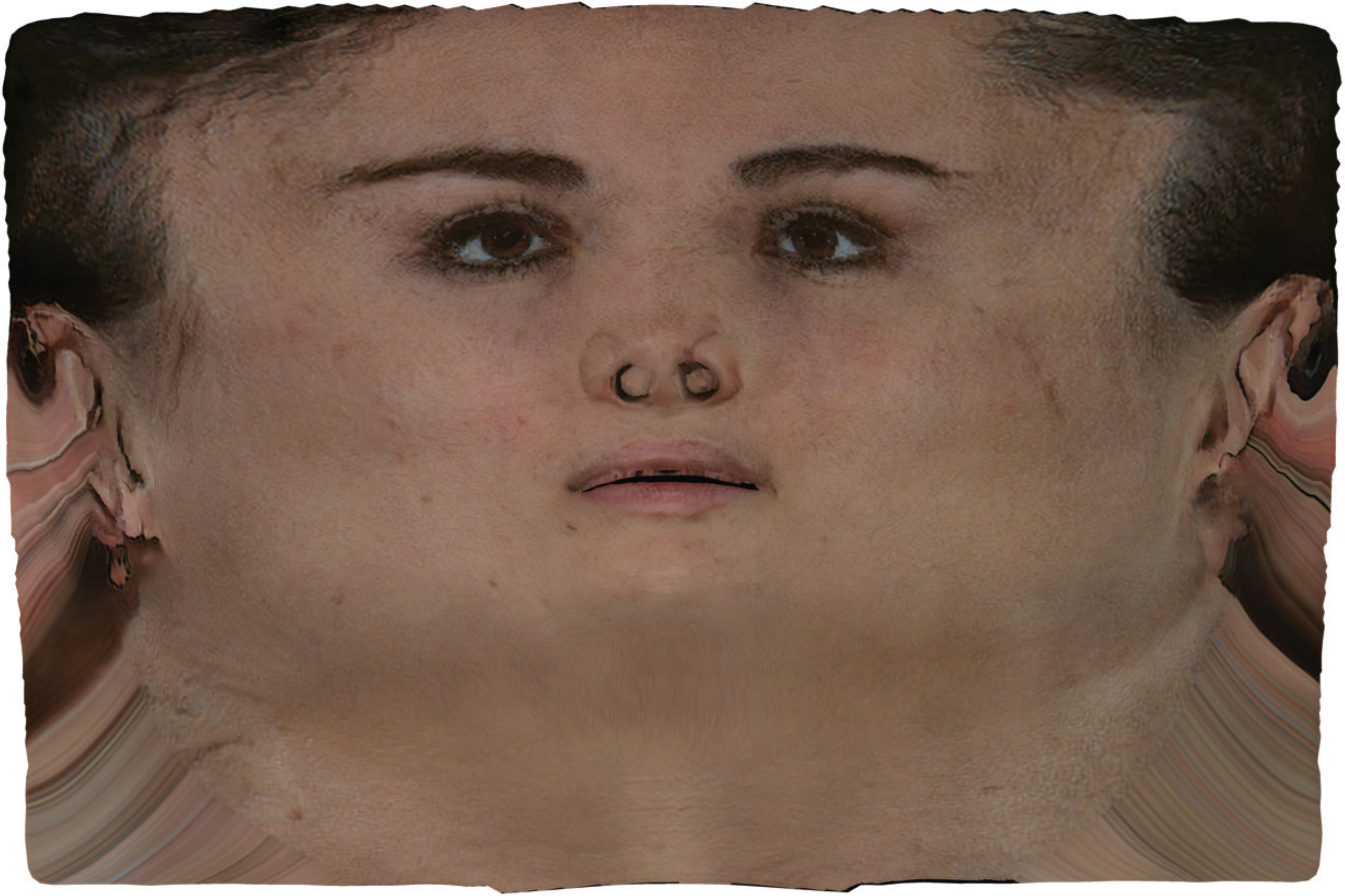}}
    \subfloat[Specular~Albedo~Ours]{
        \includegraphics[height=\myheight{}]{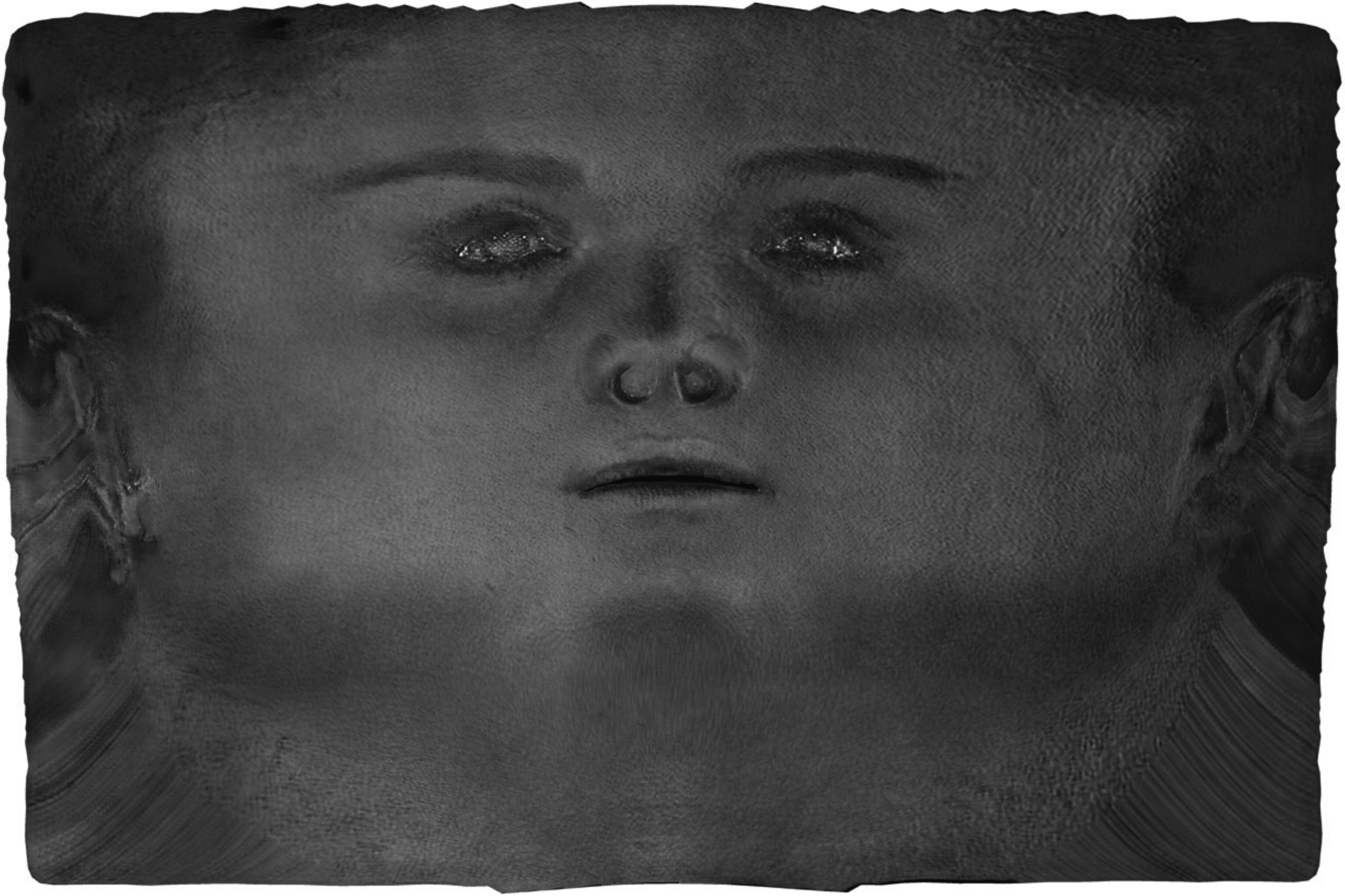}}
    \subfloat[Spec.~Normals~Ours]{
        \includegraphics[height=\myheight{}]{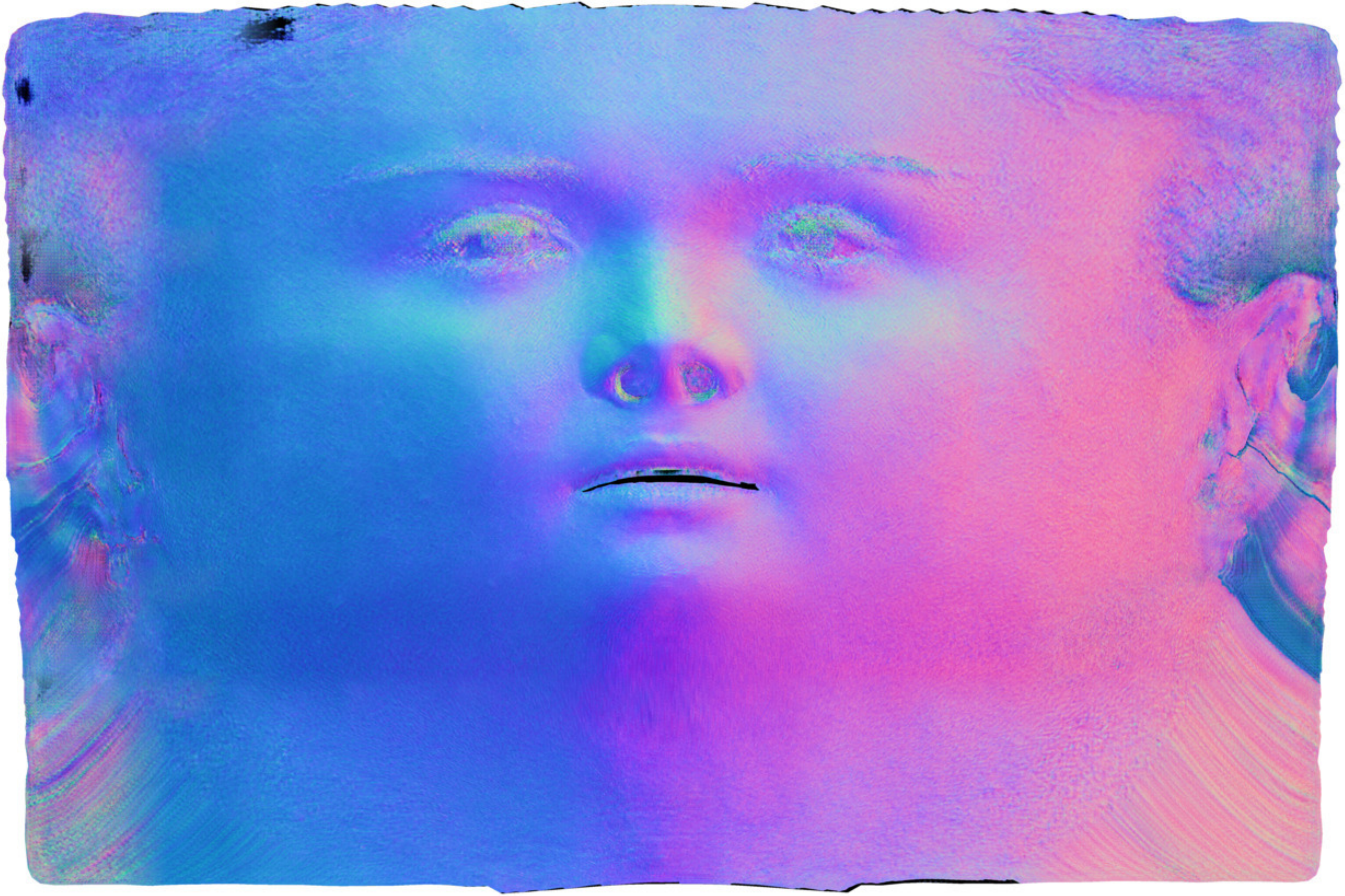}}
    \caption{
        Reflectance maps produced by our method \modelnameplus{}, against state-of-the-art methods. Reconstructions of \cite{yamaguchi_high-fidelity_2018} are provided by the authors and \cite{chen_photo-realistic_2019} are acquired using their open-sourced models.
    }
    \label{fig:results_comp_comparison}
\vspace{-0.5cm}
\end{figure*}

\begin{figure}[h!]
    \centering
    \captionsetup[subfigure]{labelformat=empty}
    \captionsetup[subfloat]{justification=centering}
    \subfloat{
        \includegraphics[width=0.24\linewidth]{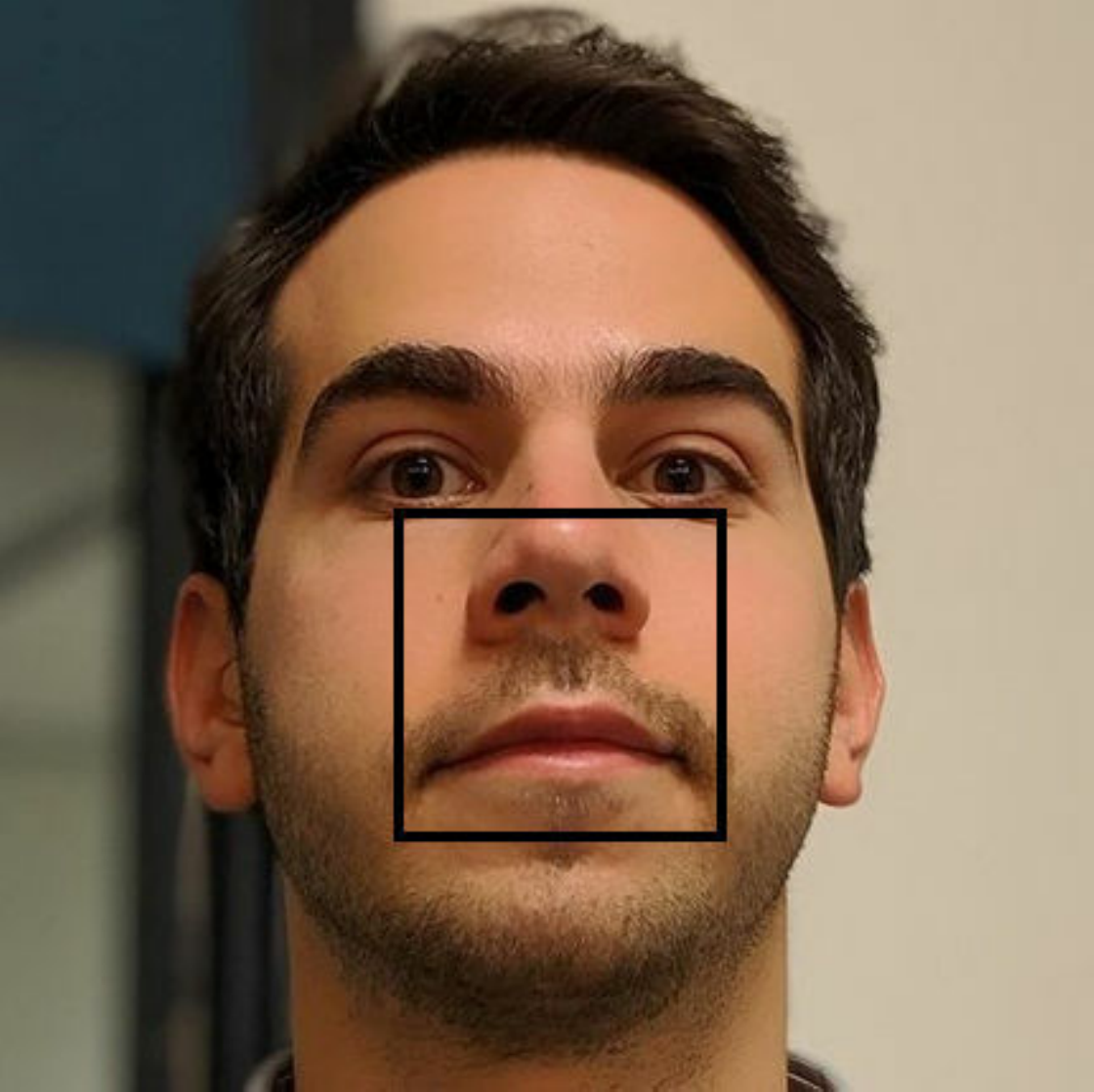}}
    \subfloat{
        \includegraphics[width=0.24\linewidth]{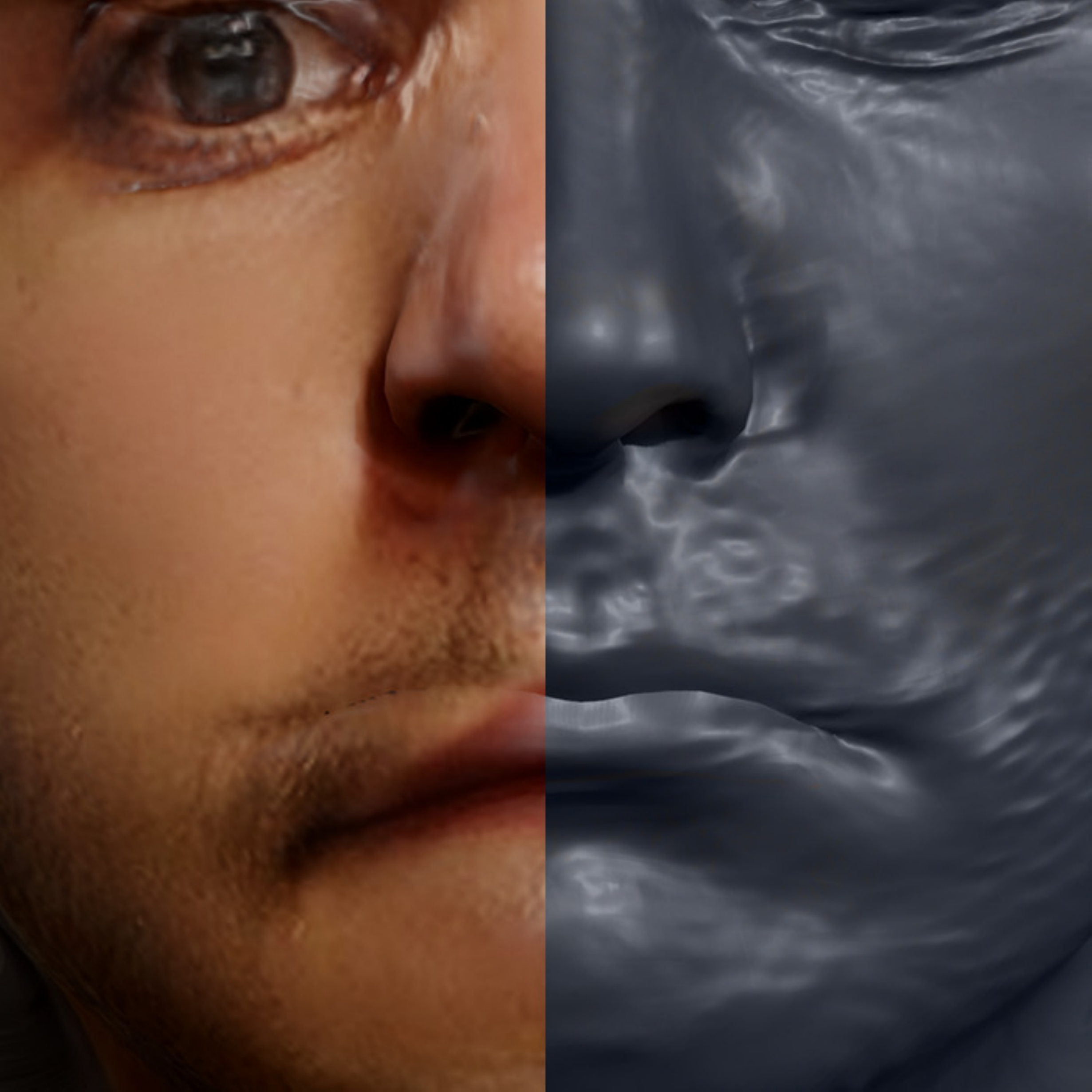}}
    \subfloat{
        \includegraphics[width=0.24\linewidth]{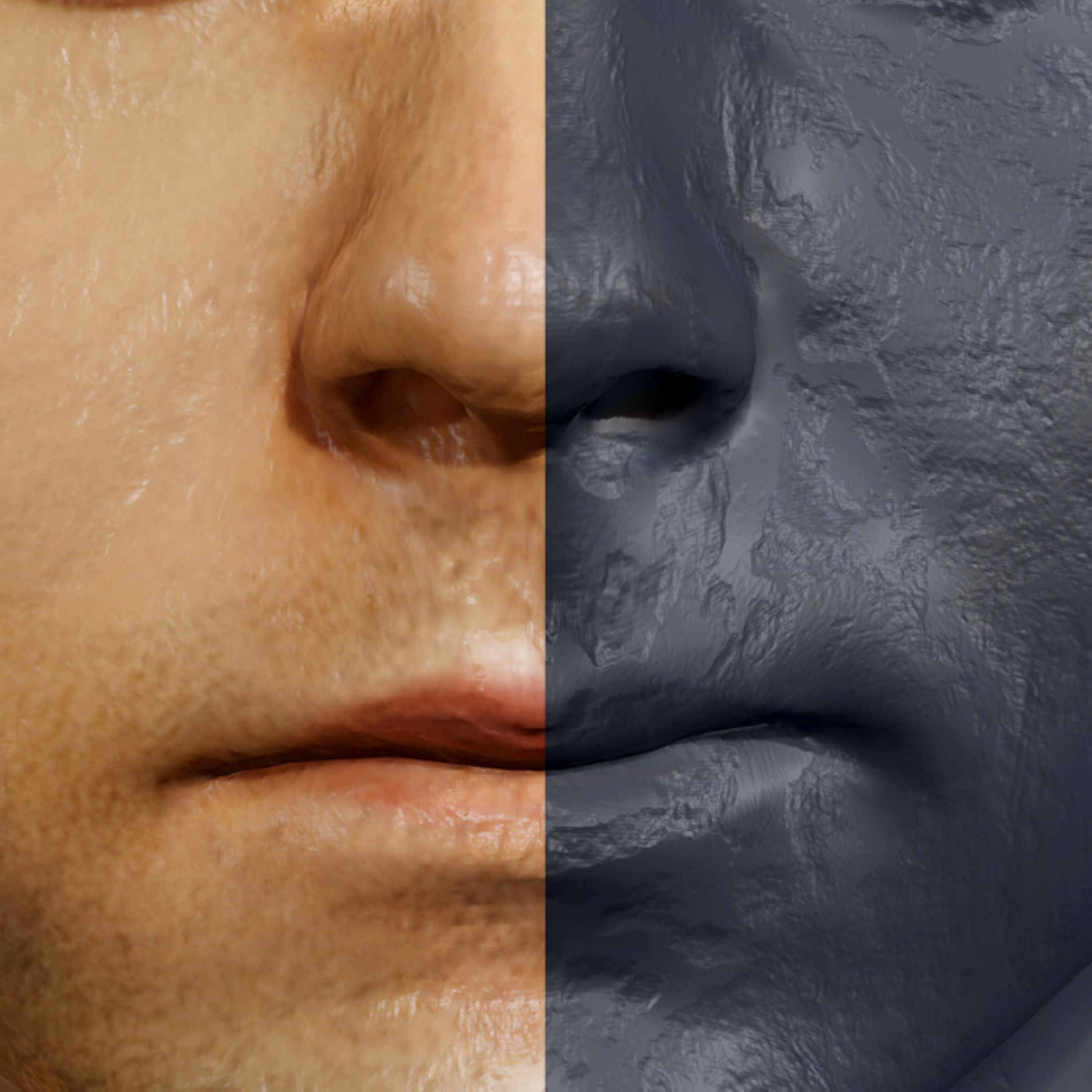}}
    \subfloat{
        \includegraphics[width=0.24\linewidth]{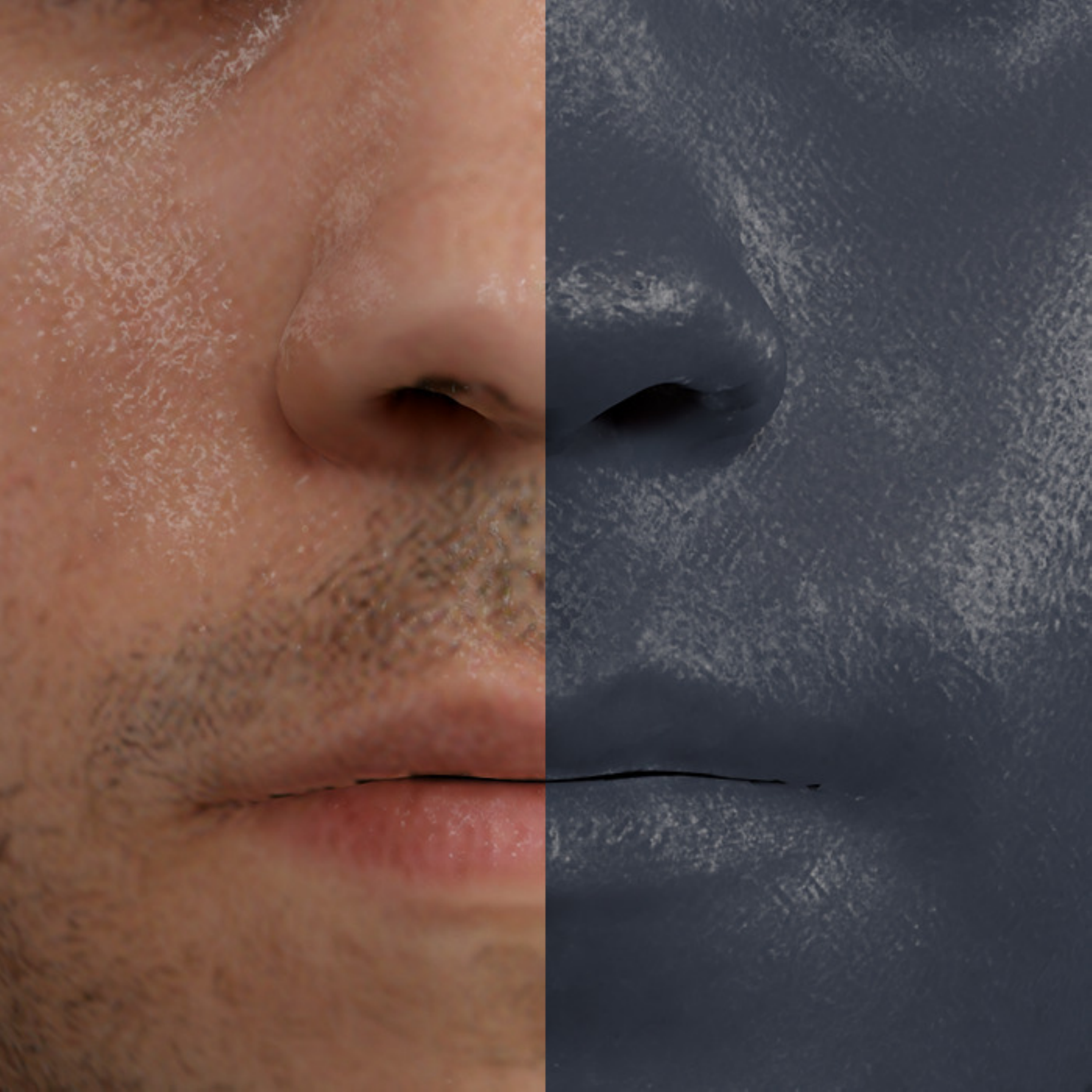}}\\
    \setcounter{subfigure}{0}
    \subfloat[
        Input]{
        \includegraphics[width=0.24\linewidth]{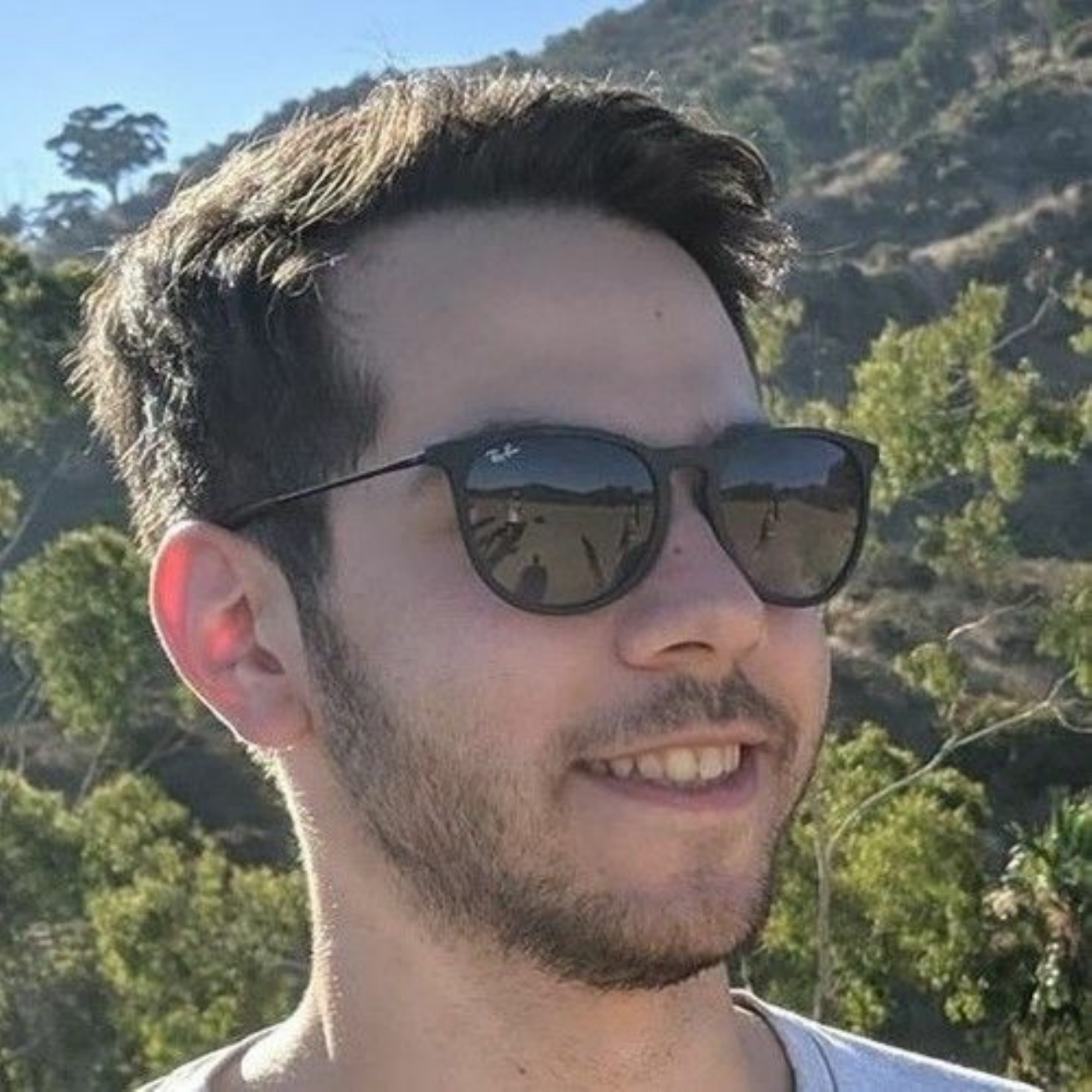}}
    \subfloat[
        Chen et al.~\cite{chen_photo-realistic_2019}]{
        \includegraphics[width=0.24\linewidth]{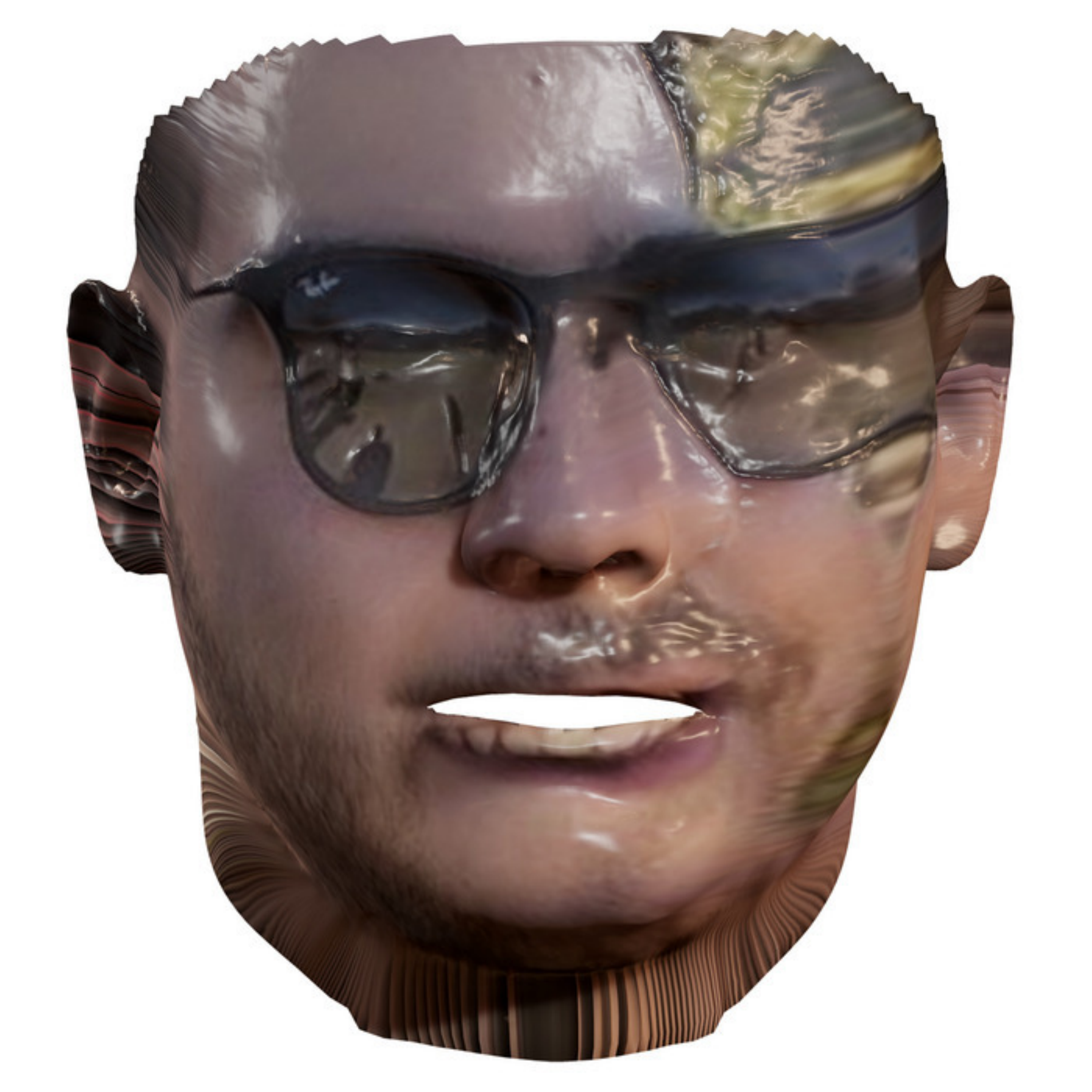}}
    \subfloat[
        Yamaguchi et al.~\cite{yamaguchi_high-fidelity_2018}]{
        \includegraphics[width=0.24\linewidth]{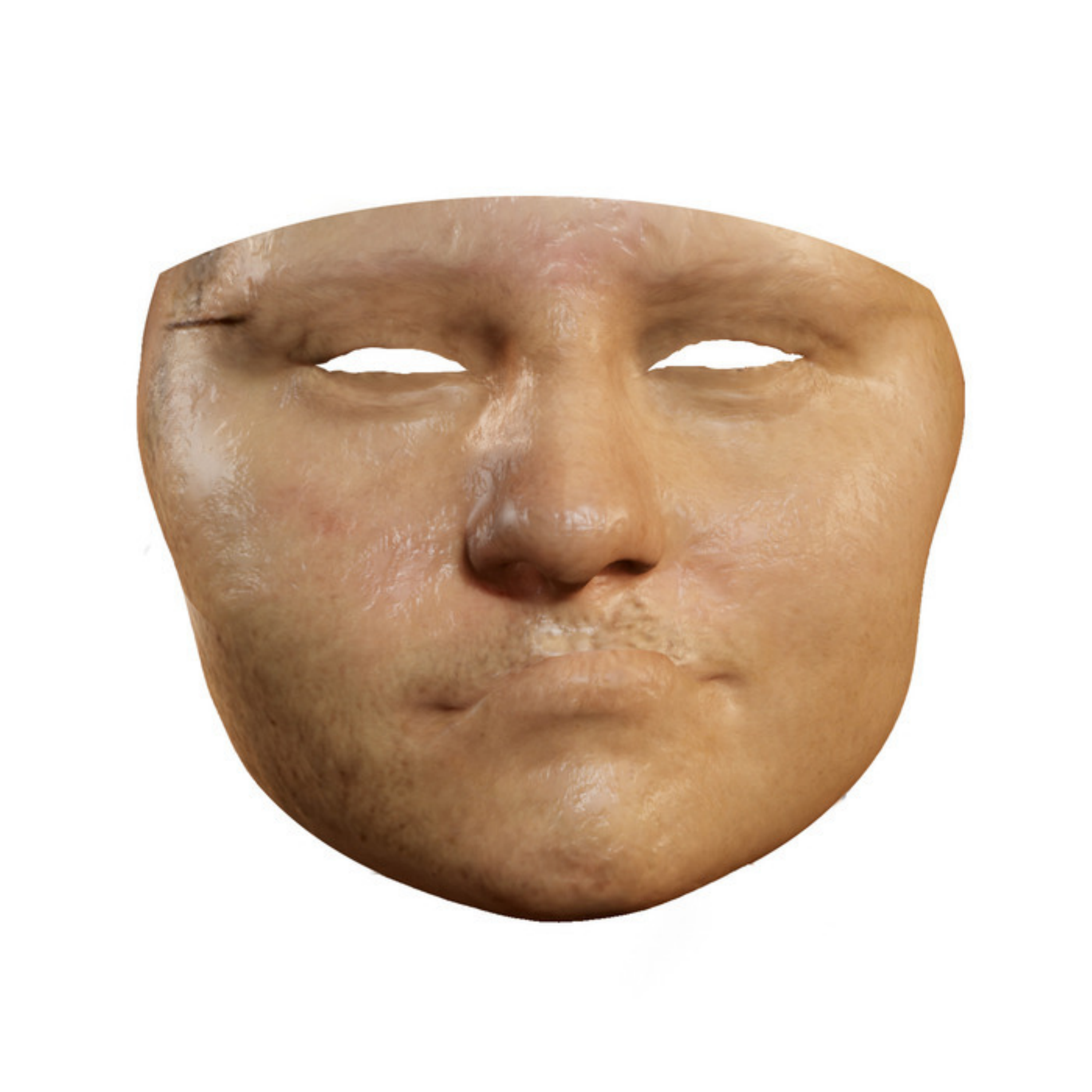}}
    \subfloat[
        Ours, AvatarMe\textsuperscript{++}]{
        \includegraphics[width=0.24\linewidth]{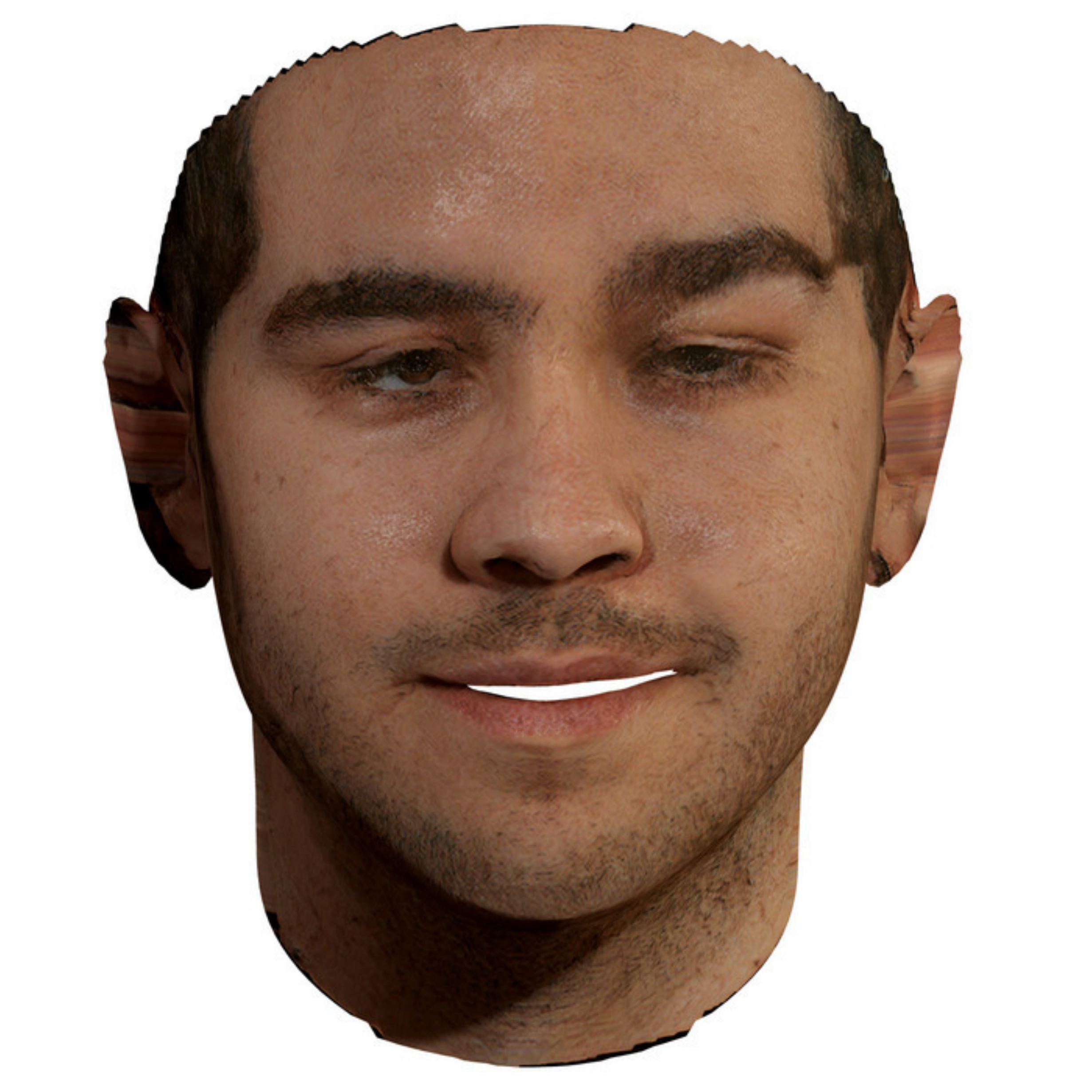}}
    \caption{
        Qualitative comparison of rendered reconstructions
        from a frontal and challenging side image.
        \cite{yamaguchi_high-fidelity_2018} results are provided by the
        authors and \cite{chen_photo-realistic_2019} results are acquired from their open-sourced models.
    }
    \label{fig:comp_qualitative_comp}
\end{figure}

To evaluate our reconstruction pipeline,
we compare reconstructed relfectance maps and renderings acquired with \modelnameplus{},
with ground truth data captured in a similar manner as our dataset RealFaceDB,
the digital Emily Project \cite{alexander_digital_2010} and current state-of-the-art.
We use the Mean Squared Error (MSE) and Peak Signal-to-Noise Ratio (PSNR) \cite{hore2010image}.
In Table~\ref{table:comparisons}, we conduct quantitative comparisons
against the state-of-the-art \cite{chen_learning_2019, yamaguchi_high-fidelity_2018}. 
As can be seen, our method outperforms \cite{chen_photo-realistic_2019} and \cite{yamaguchi_high-fidelity_2018} by a significant margin.
All meshes were manually registered to the same topology and UV parameterization.
Moreover using a state-of-the-art face recognition algorithm \cite{deng_arcface_2019},
we also find the highest match of facial identity compared to the input images when using our method. 
The input images were compared against renderings of the faces 
with reconstructed geometry and reflectance, 
including eyes added manually to \cite{yamaguchi_high-fidelity_2018}.
We also present qualitative comparisons in Fig.~\ref{fig:results_comp_comparison} and Fig.~\ref{fig:comp_qualitative_comp}.

\begin{table}[h!]
\centering
\captionsetup[table]{singlelinecheck=false}
\caption{
    Quantitative comparisons with state-of-the-art, between 6 reconstructions
    of the same subject, from different ``in-the-wild`` images, and ground truth using \cite{lattas_multi-view_2019}.
    We transform \cite{chen_photo-realistic_2019, yamaguchi_high-fidelity_2018}
    results to our UV topology and compute only for a $2K\times2K$ centered crop,
    as they only produce the frontal part of the face
    and manually add eyes to \cite{yamaguchi_high-fidelity_2018}.
}\label{table:comparisons}
\begin{tabular}{@{}lrrrr@{}}
\toprule
Algorithm        & \cite{yamaguchi_high-fidelity_2018} & \cite{chen_photo-realistic_2019}
                                                        & \modelname{}      & \modelnameplus{}\\ 
\midrule
PSNR (Albedo)                               & 11.225    & 14.374    & 24.05 & \textbf{26.18}  \\
PSNR (Normals)                              & 21.889    & 17.321    & 26.97 & \textbf{27.12}  \\
MSE (Albedo)                                & 0.0225    & 0.0140    & 0.0049& \textbf{0.0038}  \\
MSE (Normals)                               & 0.0047    & 0.0049    & 0.0031& \textbf{0.0025}  \\
Rendered ID Score \cite{deng_arcface_2019}  & 0.629     & 0.632     & 0.873 & \textbf{0.881}  \\
\bottomrule
\end{tabular}
\end{table}

\begin{figure}[h]
\centering
\captionsetup[subfigure]{labelformat=empty}
\renewcommand{\tabcolsep}{0pt}
    \begin{tabular}{@{}lllll@{}}
    
    \multirow{2}{*}{
        \subfloat[Input]{
            \includegraphics[width=0.18\linewidth]{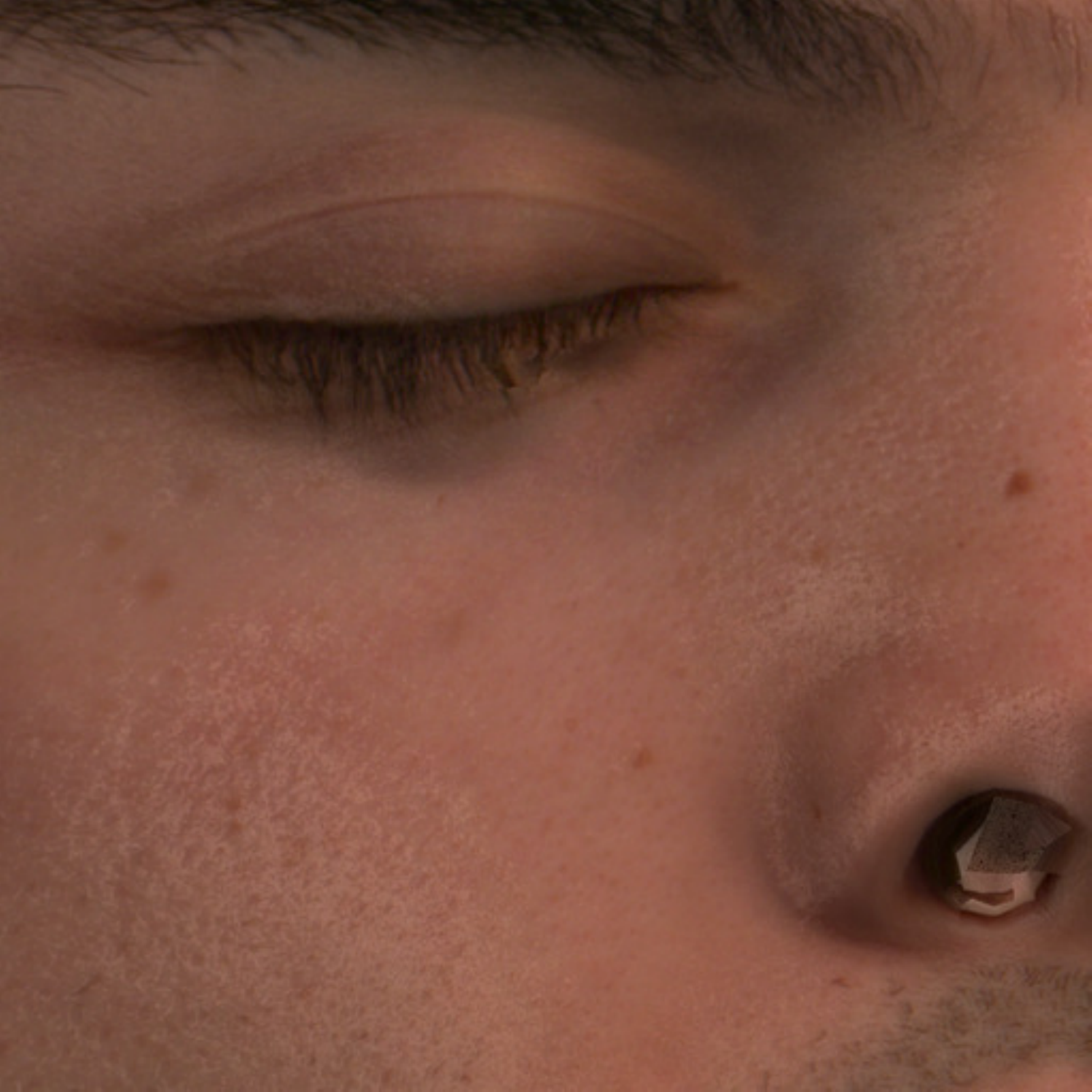}
        }
    }&
    \subfloat{
        \includegraphics[width=0.18\linewidth]{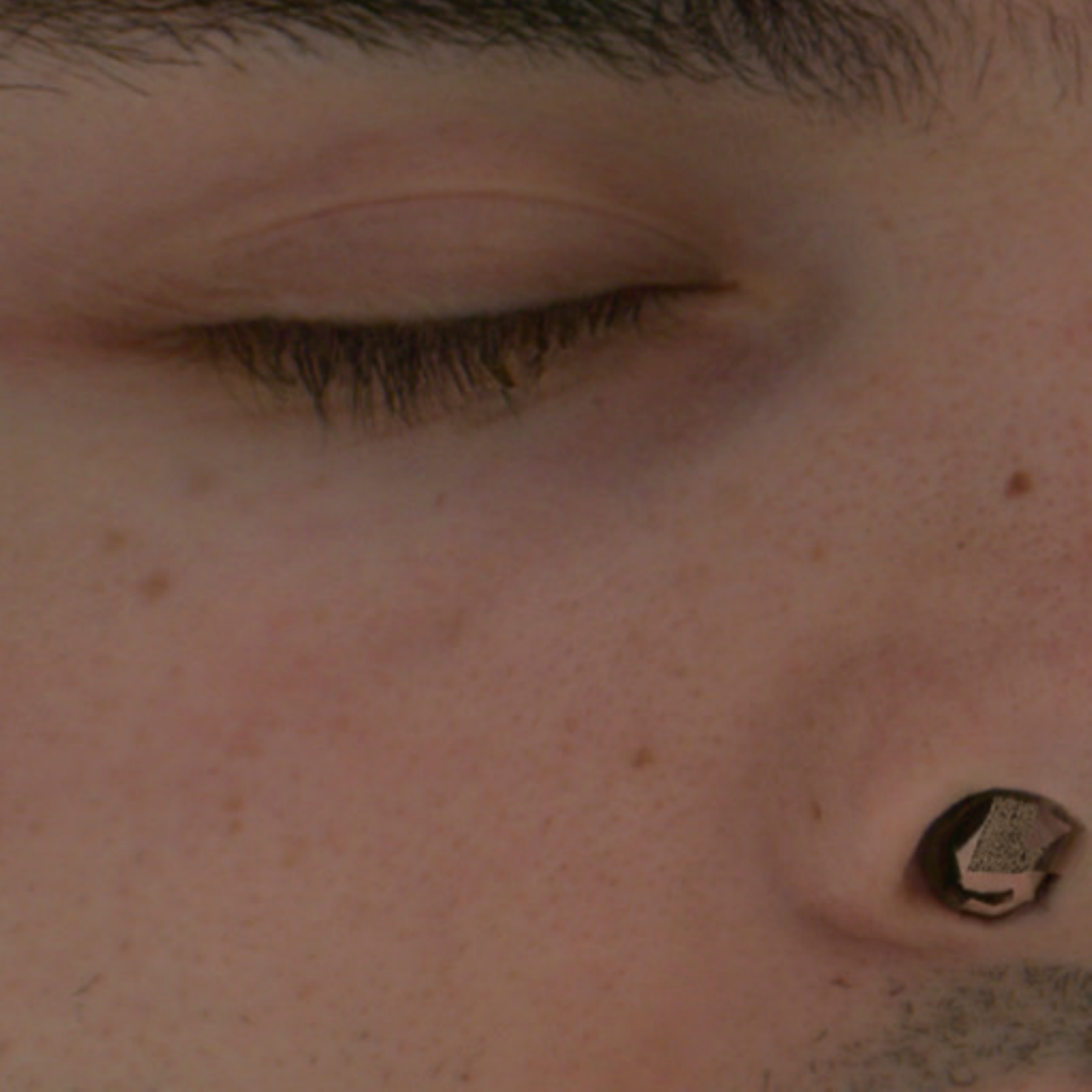}
    }&
    \subfloat{
        \includegraphics[width=0.18\linewidth]{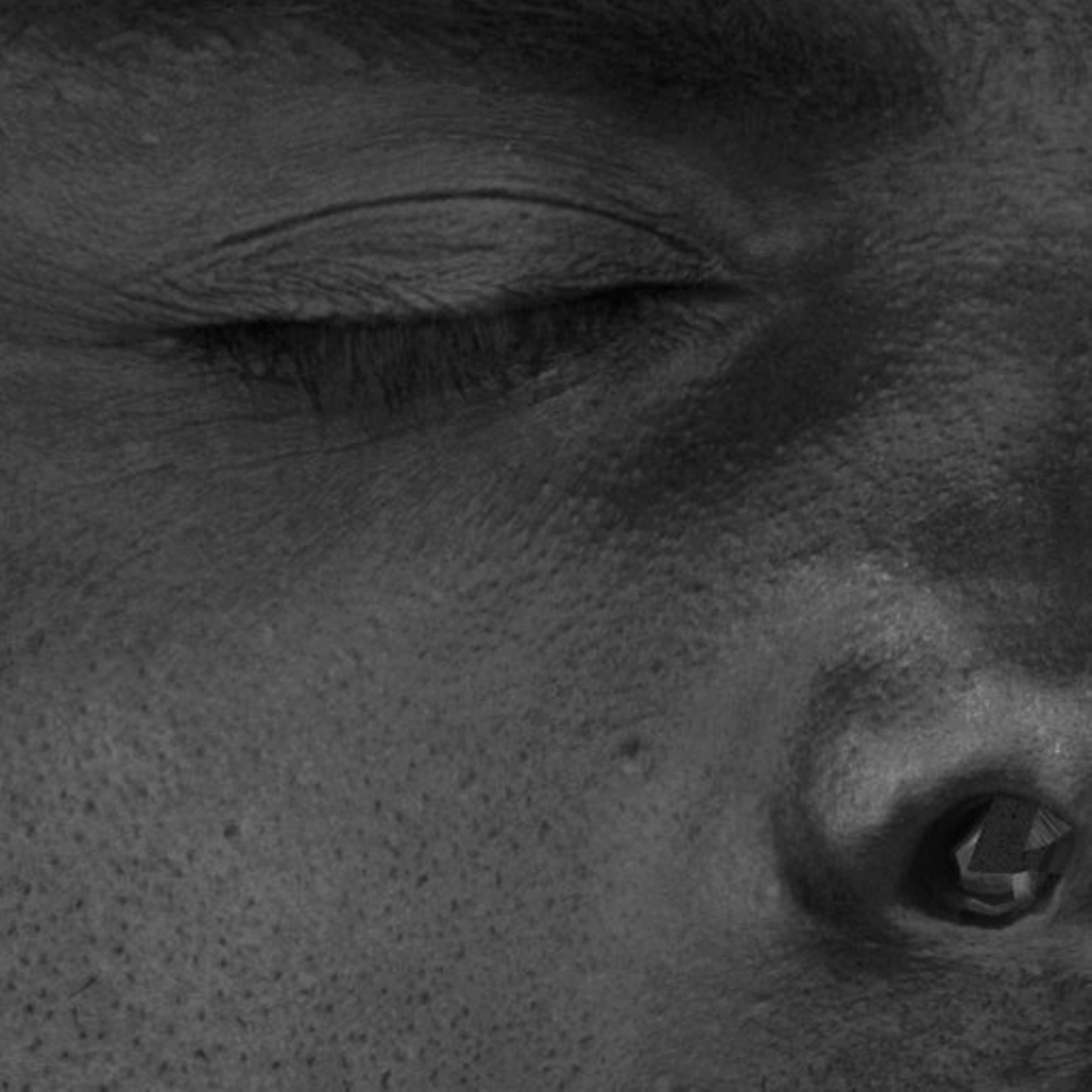}
    }&
    \subfloat{
        \includegraphics[width=0.18\linewidth]{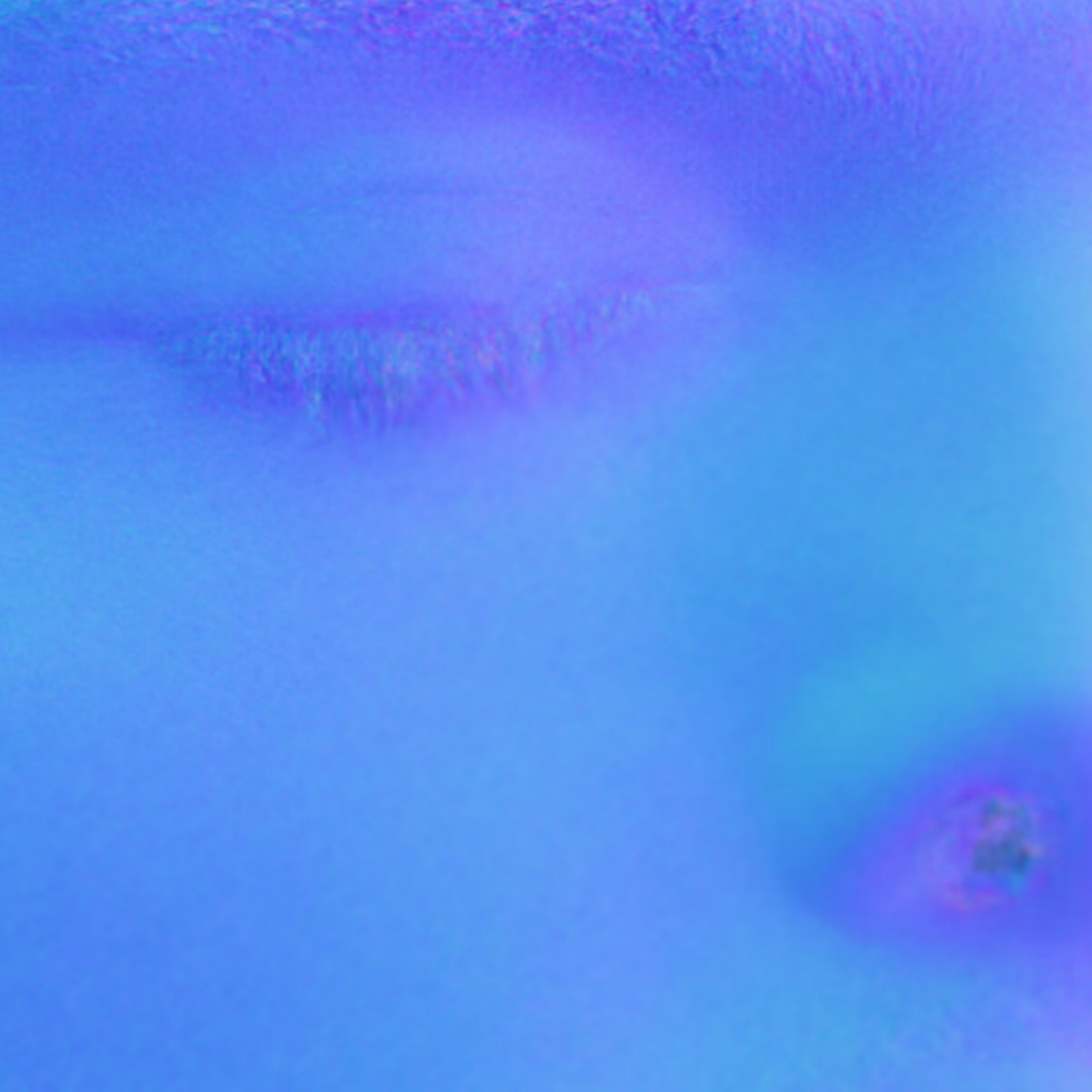}
    }&
    \subfloat{
        \includegraphics[width=0.18\linewidth]{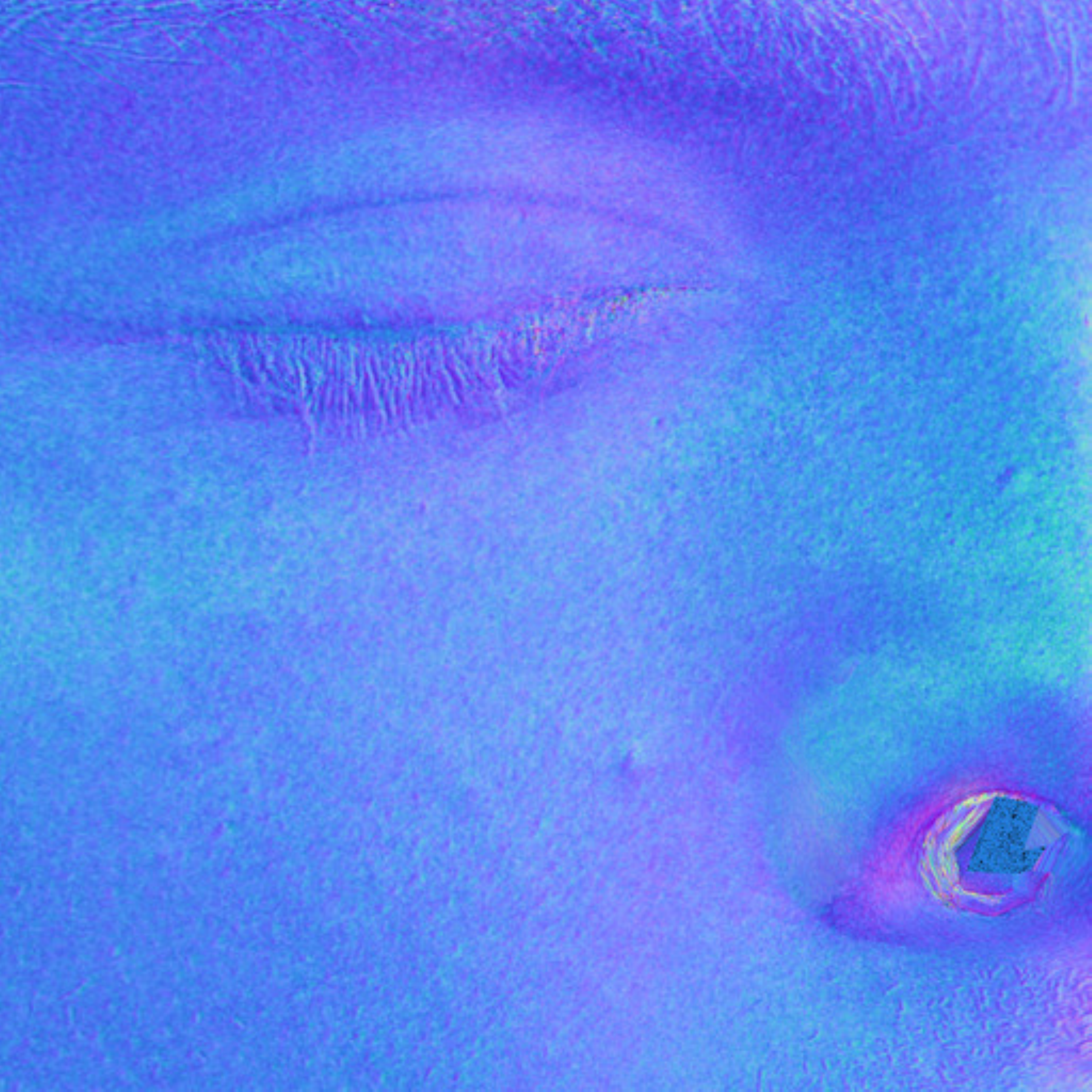}
    }\\
    \setcounter{subfigure}{1}
    &
    \subfloat[Diff.Alb.]{
        \includegraphics[width=0.18\linewidth]{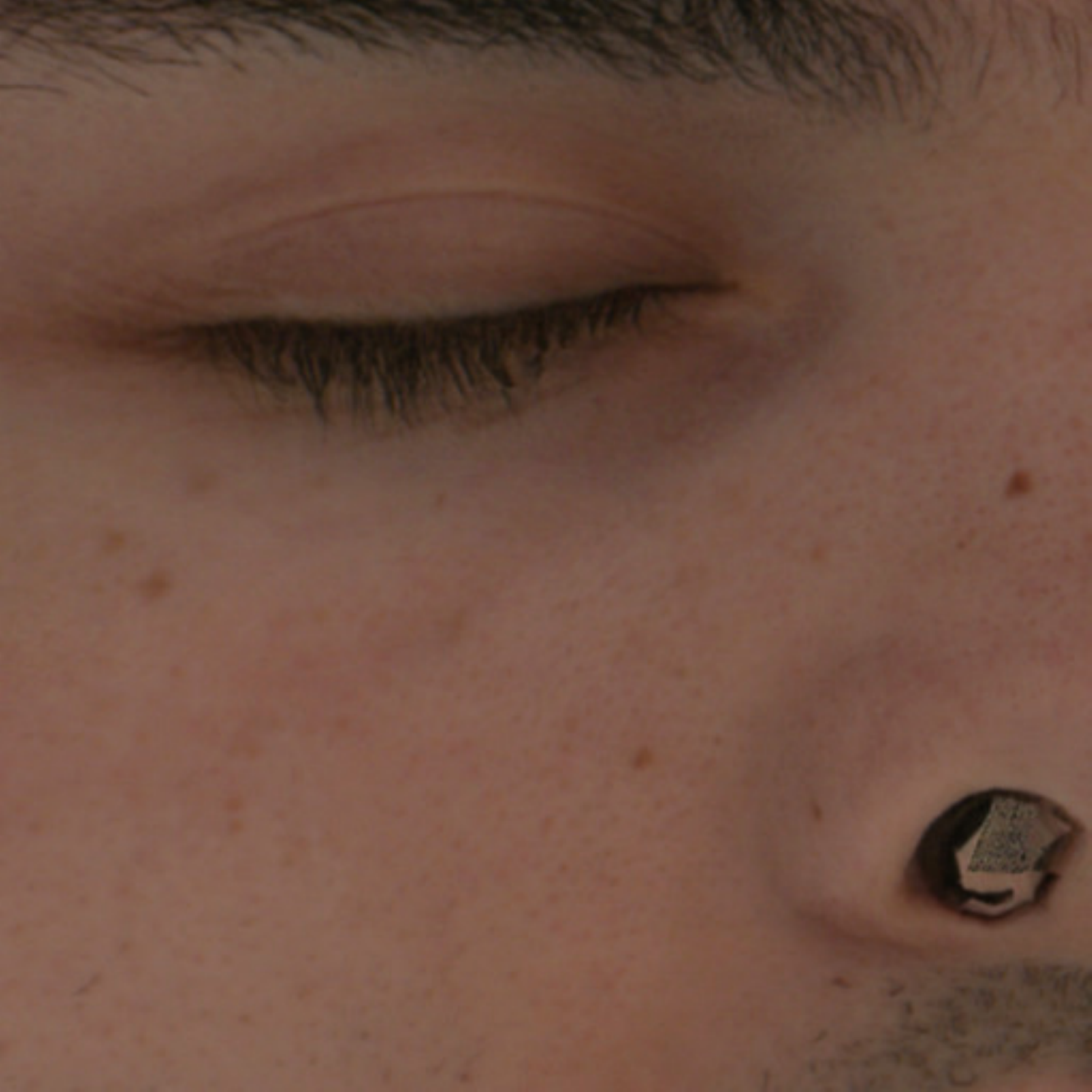}
    }&
    \subfloat[Spec.Alb.]{
        \includegraphics[width=0.18\linewidth]{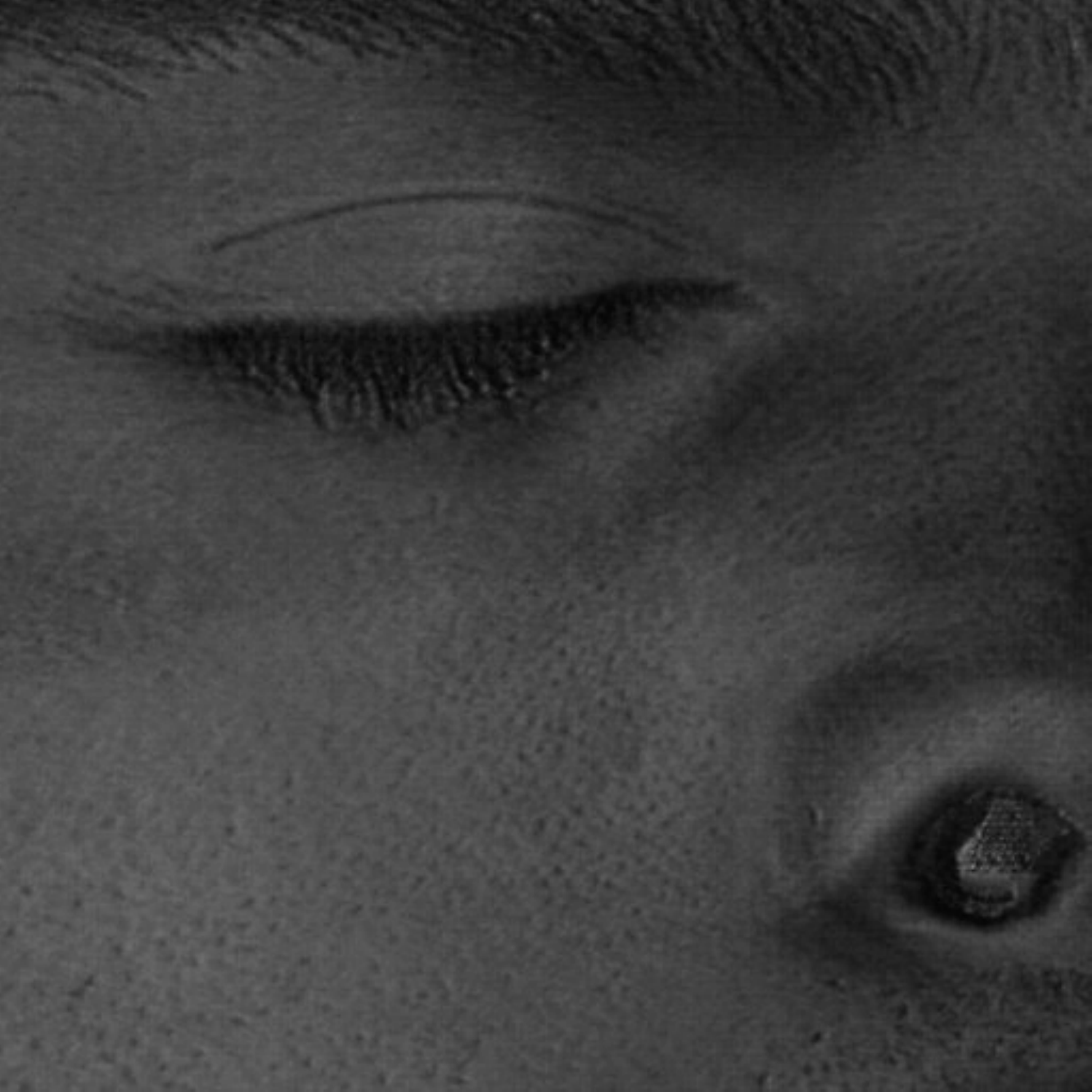}
    }&
    \subfloat[Diff.Nor.]{
        \includegraphics[width=0.18\linewidth]{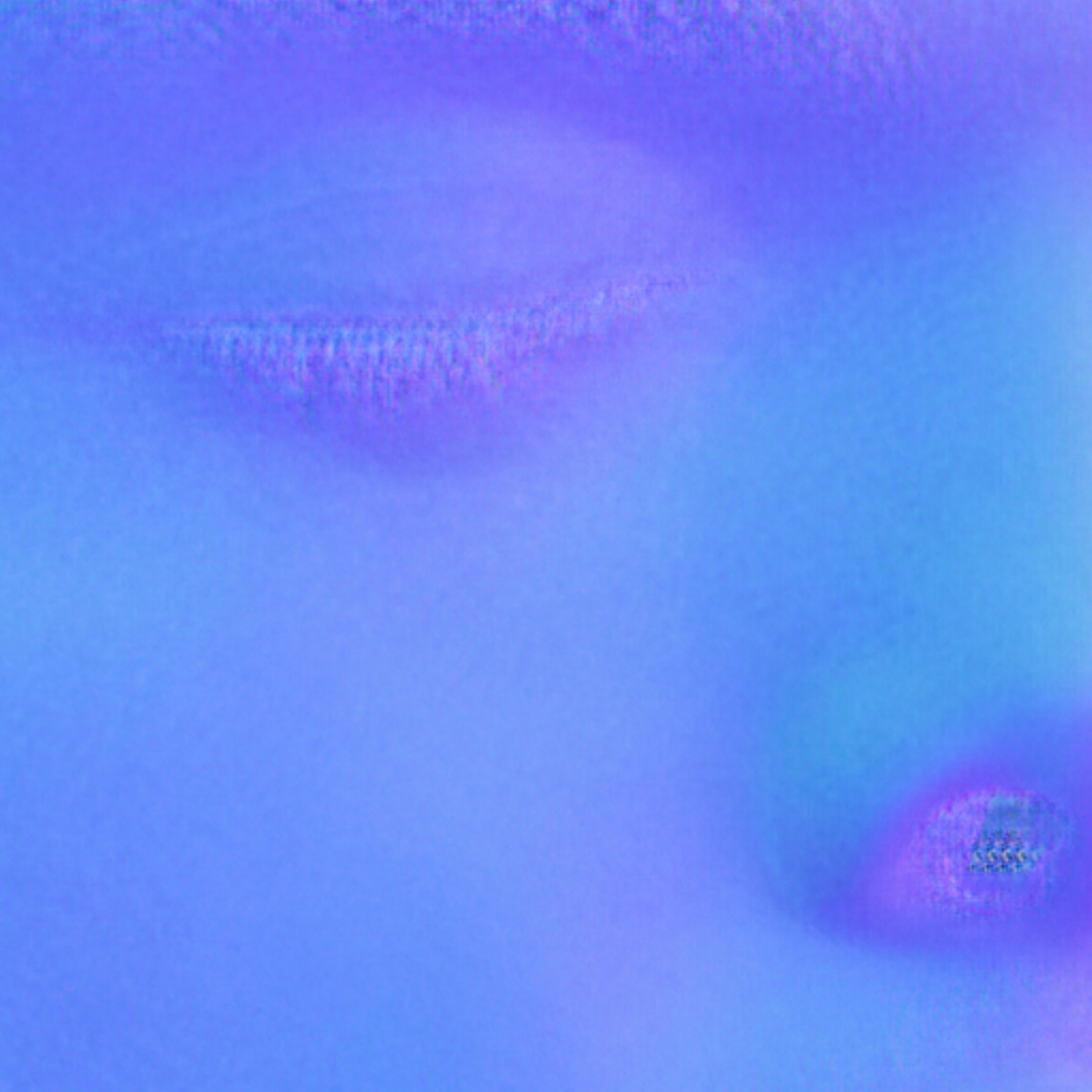}
    }&
    \subfloat[Spec.Nor.]{
        \includegraphics[width=0.18\linewidth]{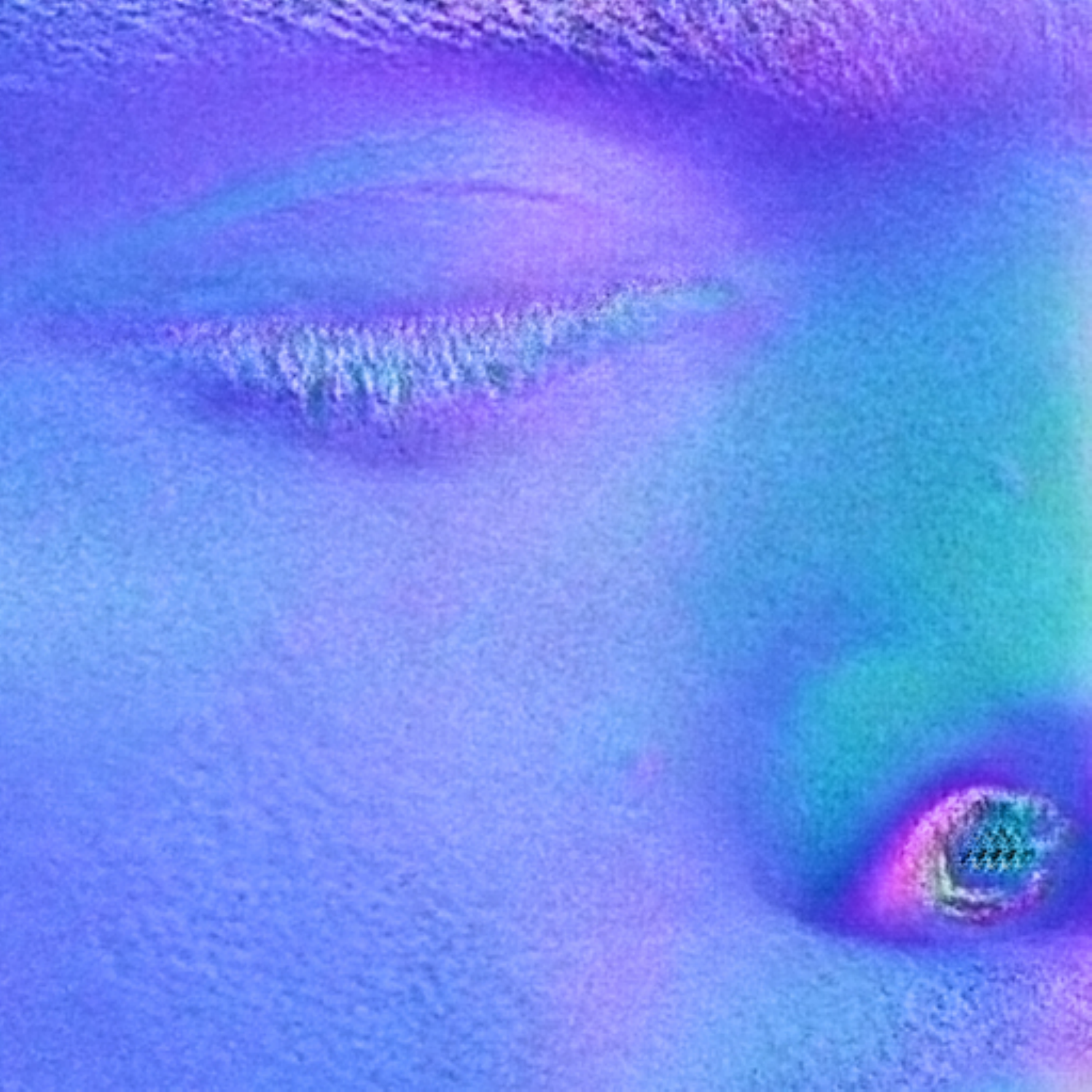}
    }
    \end{tabular}
    \caption{
        Comparison of AvatarMe\textsuperscript{++} predicted reflectance with ground truth.
        Left: Patch of rendered test subject in target domain,
        Top row: predicted reflectance with our method,
        Bottom row: ground truth.
    }
    \label{fig:results_gt_comparison}
\vspace{-0.3cm}
\end{figure}

\begin{figure}
    \centering
    
    \subfloat{
        \includegraphics[width=0.24\linewidth]{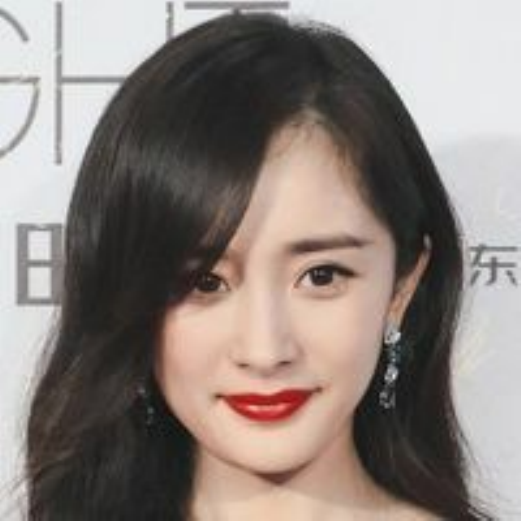}}
    \subfloat{
        \includegraphics[width=0.24\linewidth]{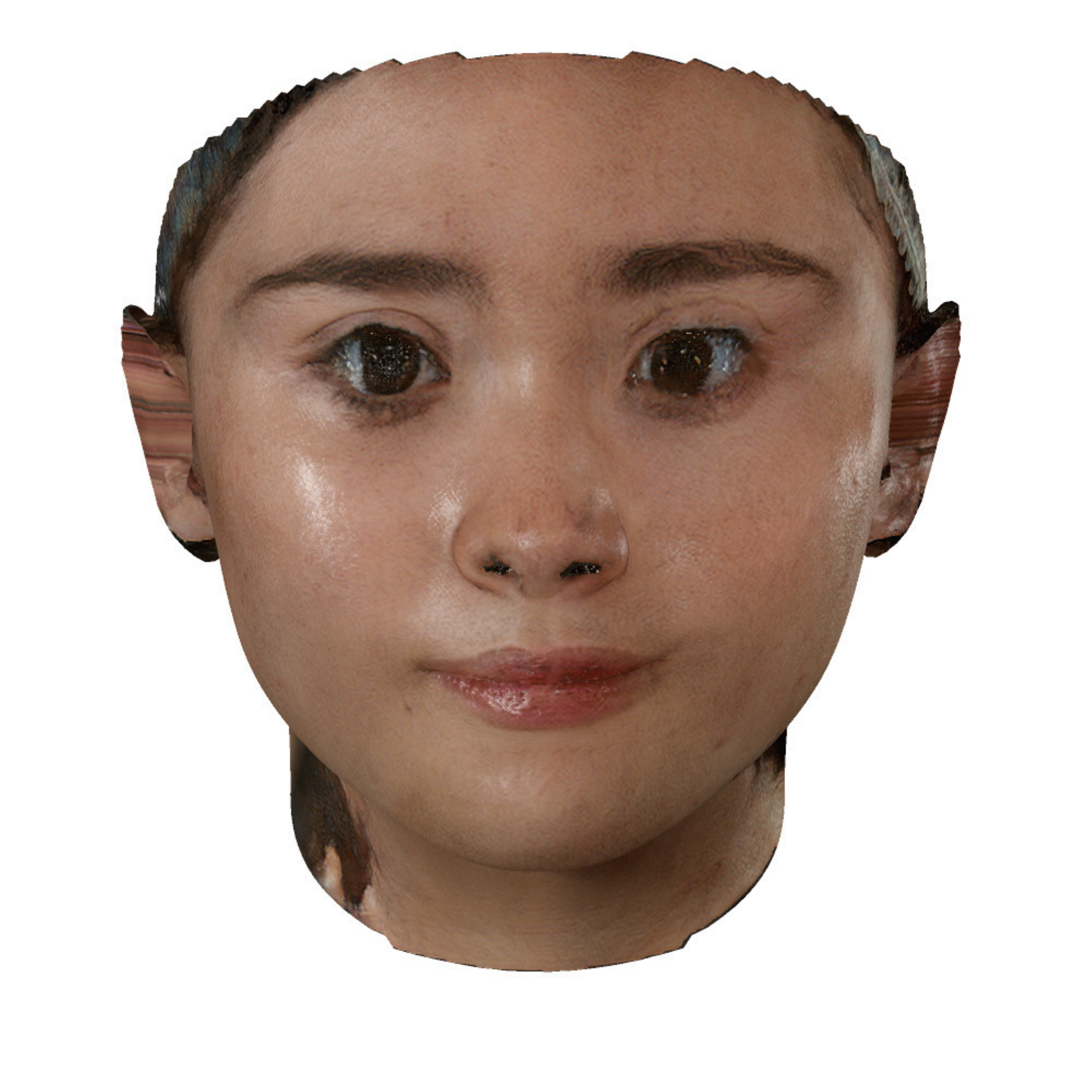}}
    \subfloat{
        \includegraphics[width=0.24\linewidth]{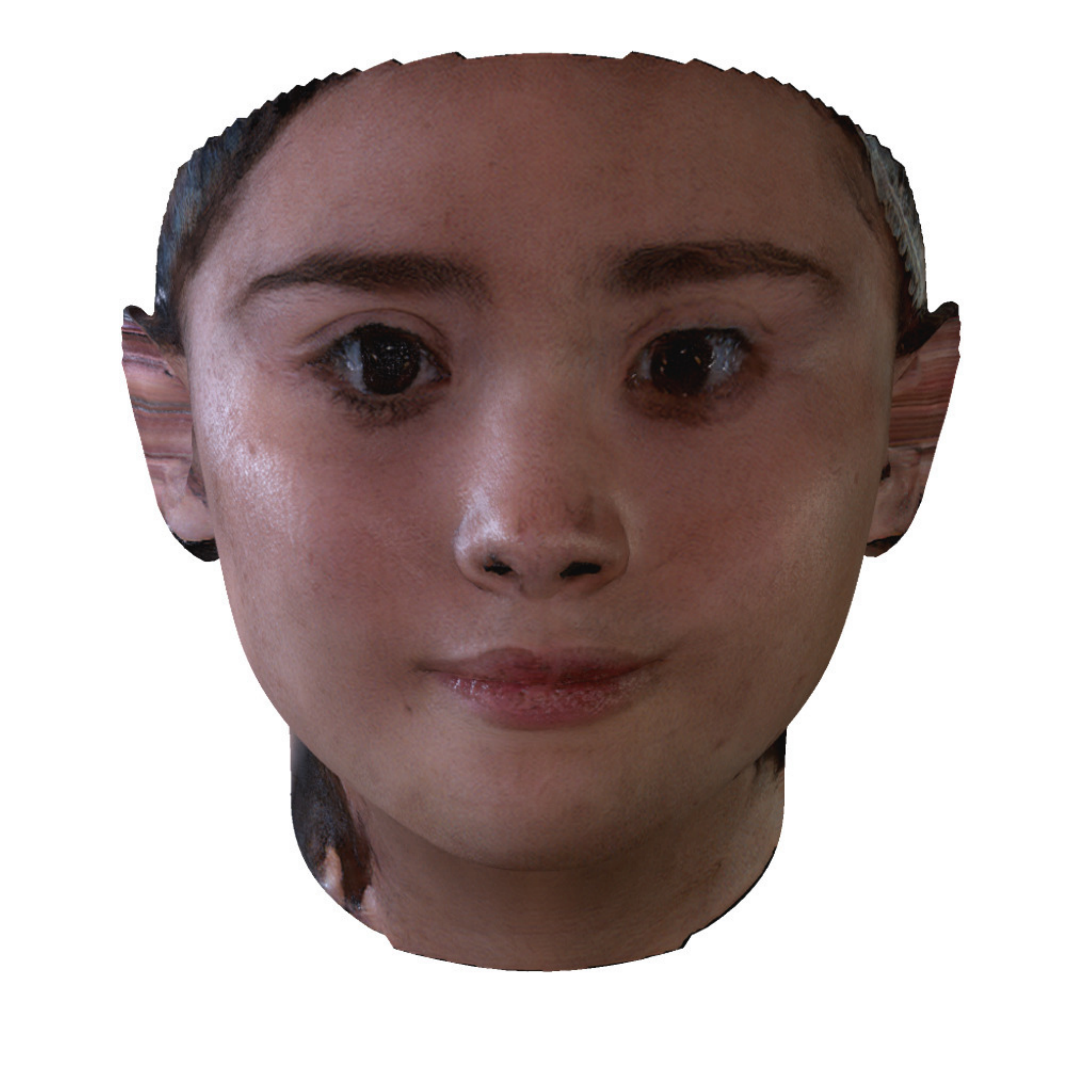}}
    \subfloat{
        \includegraphics[width=0.24\linewidth]{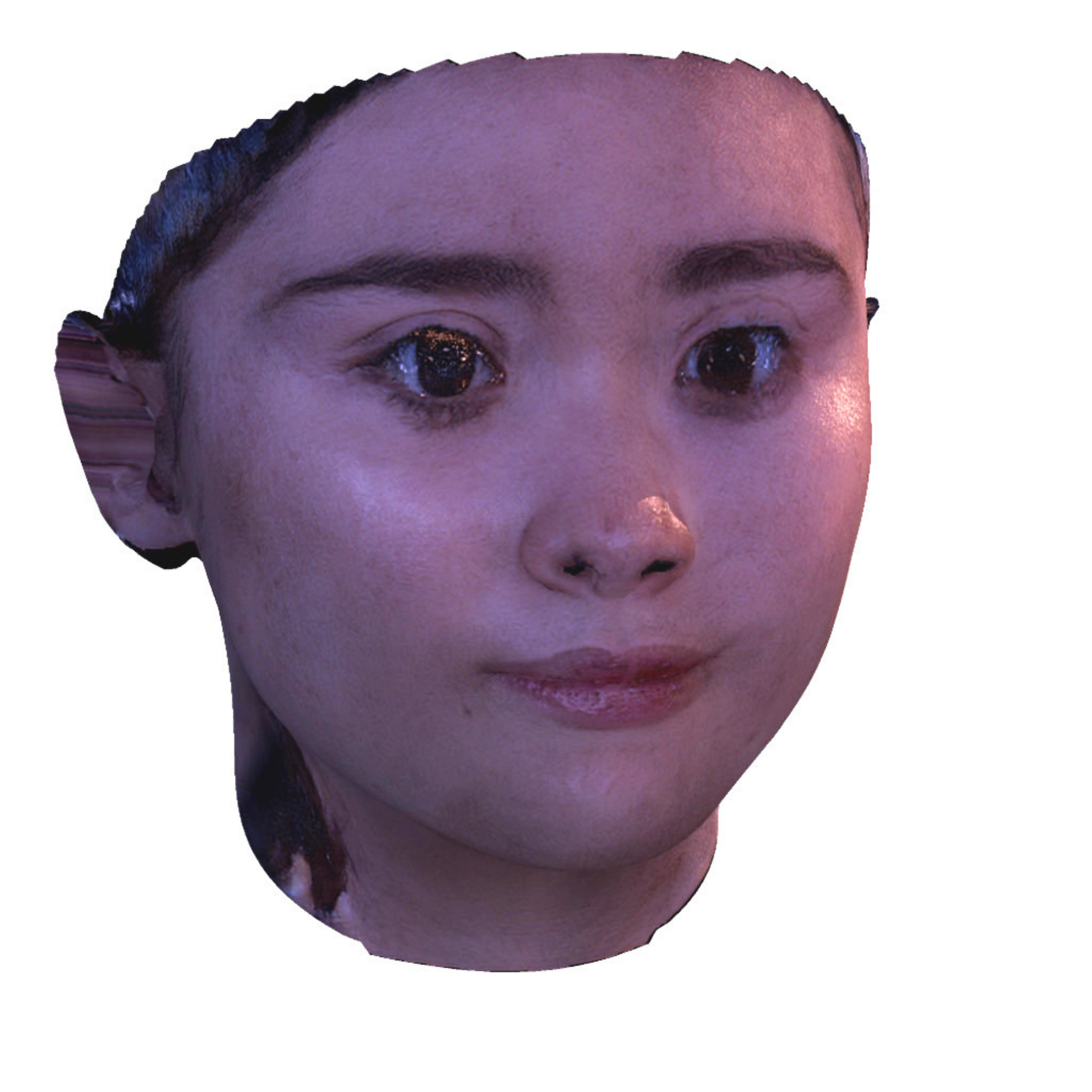}} \\
        
    \subfloat{
        \includegraphics[width=0.24\linewidth]{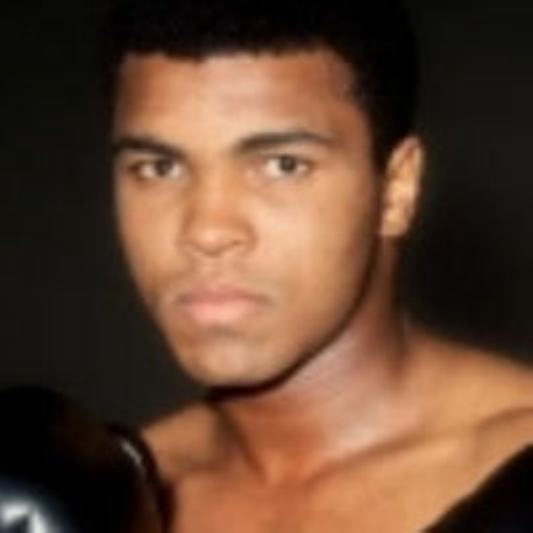}}
    \subfloat{
        \includegraphics[width=0.24\linewidth]{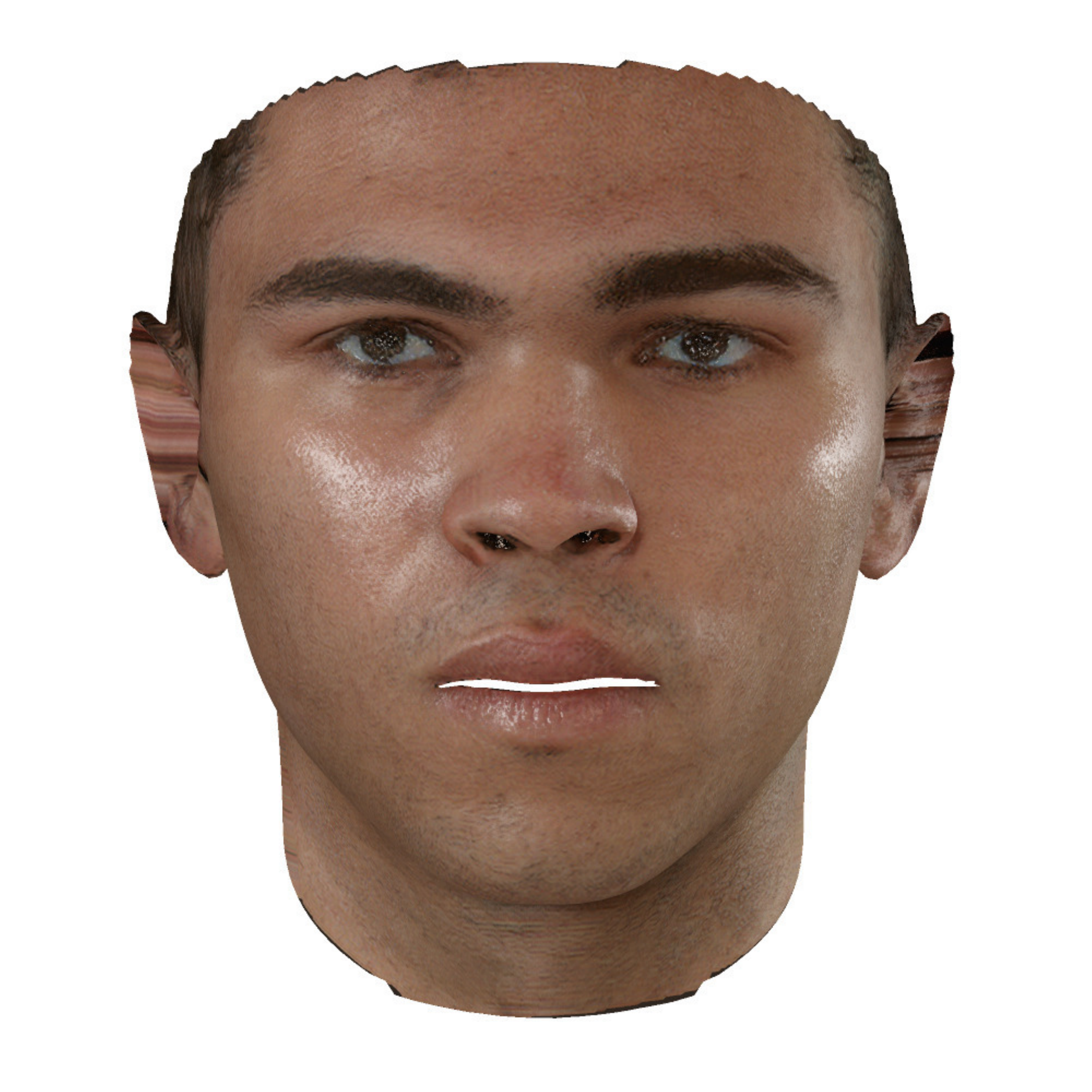}}
    \subfloat{
        \includegraphics[width=0.24\linewidth]{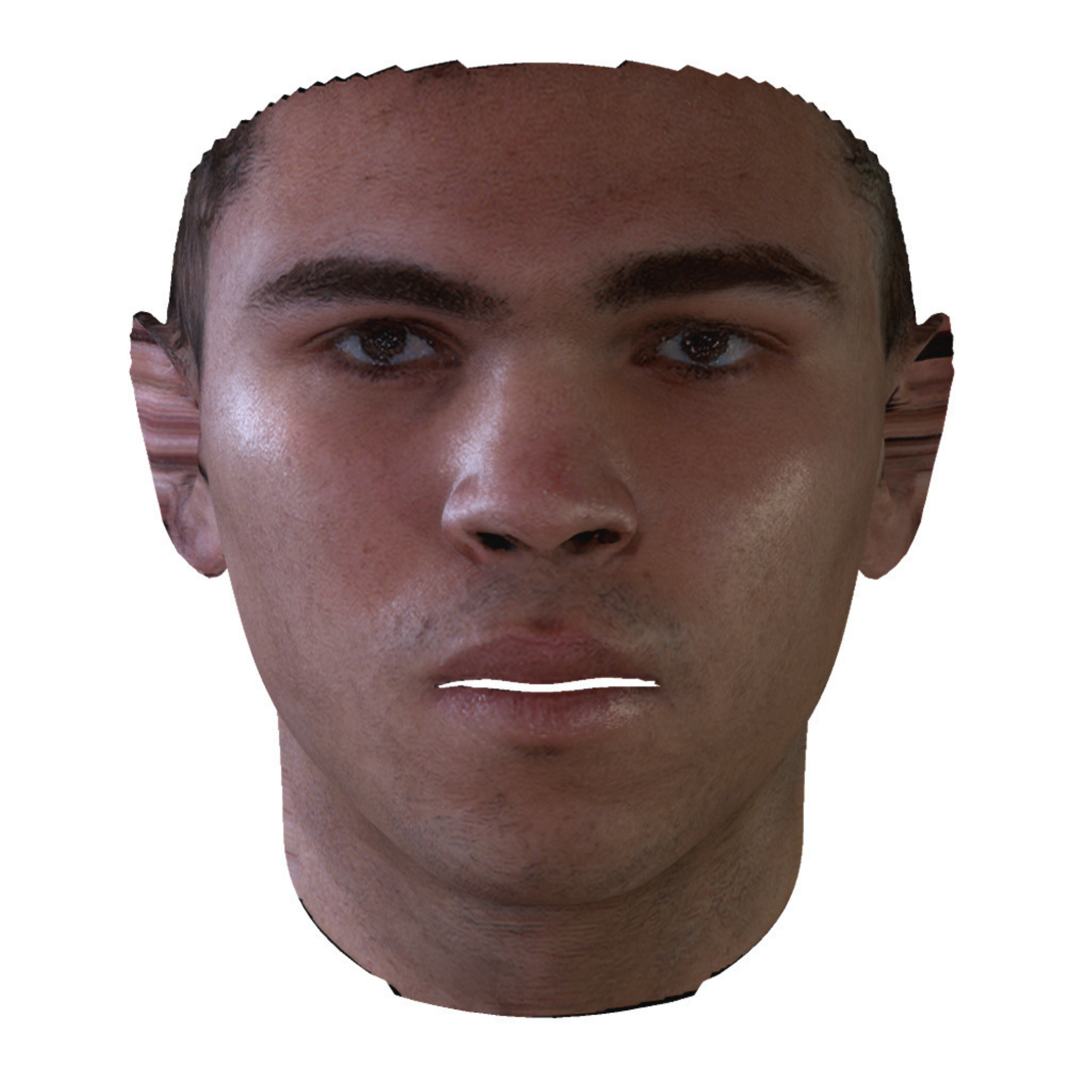}}
    \subfloat{
        \includegraphics[width=0.24\linewidth]{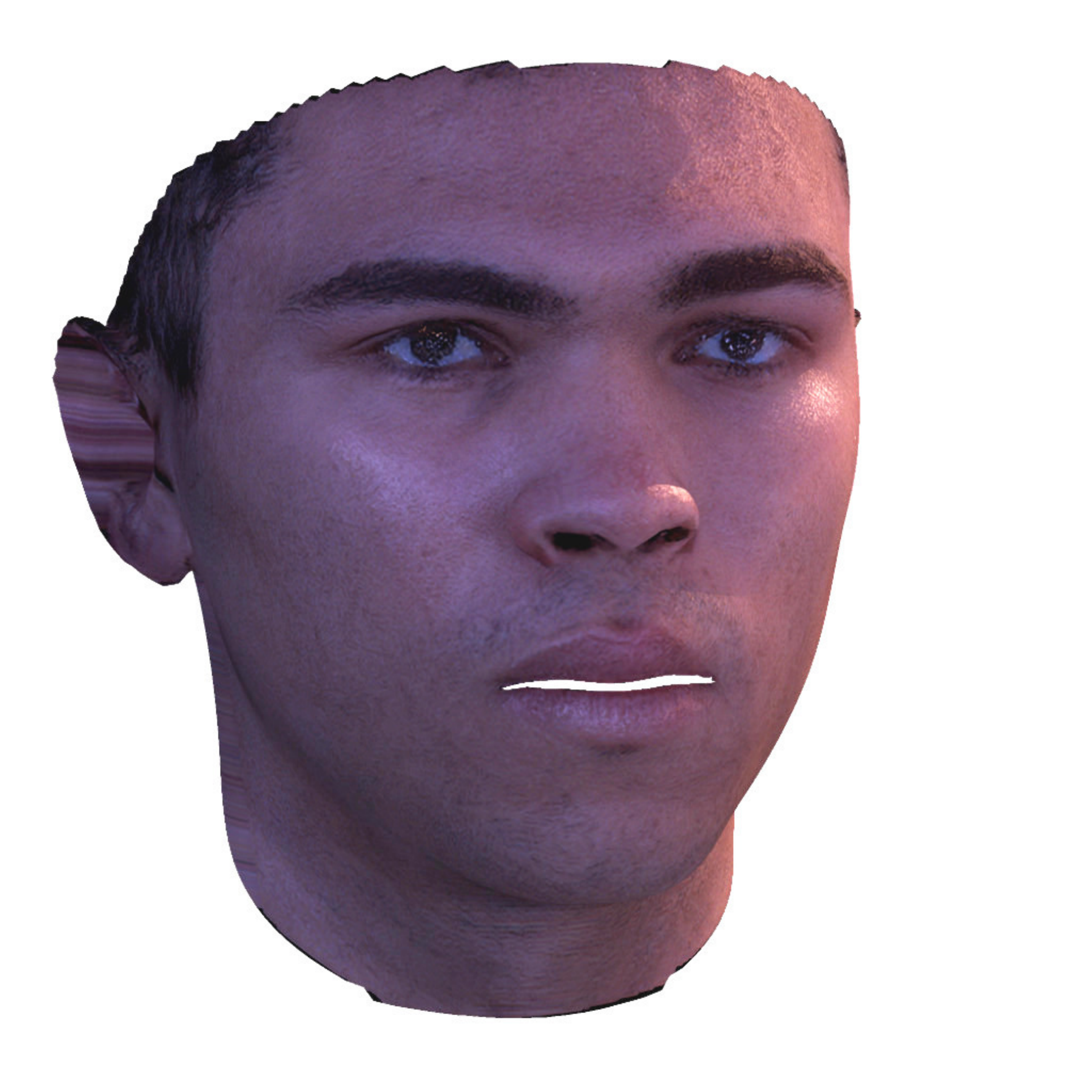}} \\
        
    \subfloat{
        \includegraphics[width=0.24\linewidth]{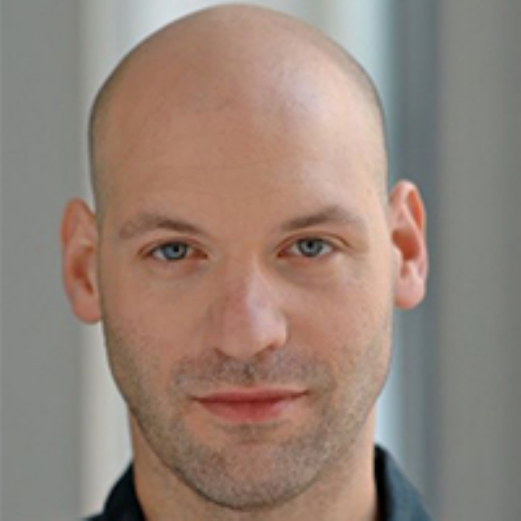}}
    \subfloat{
        \includegraphics[width=0.24\linewidth]{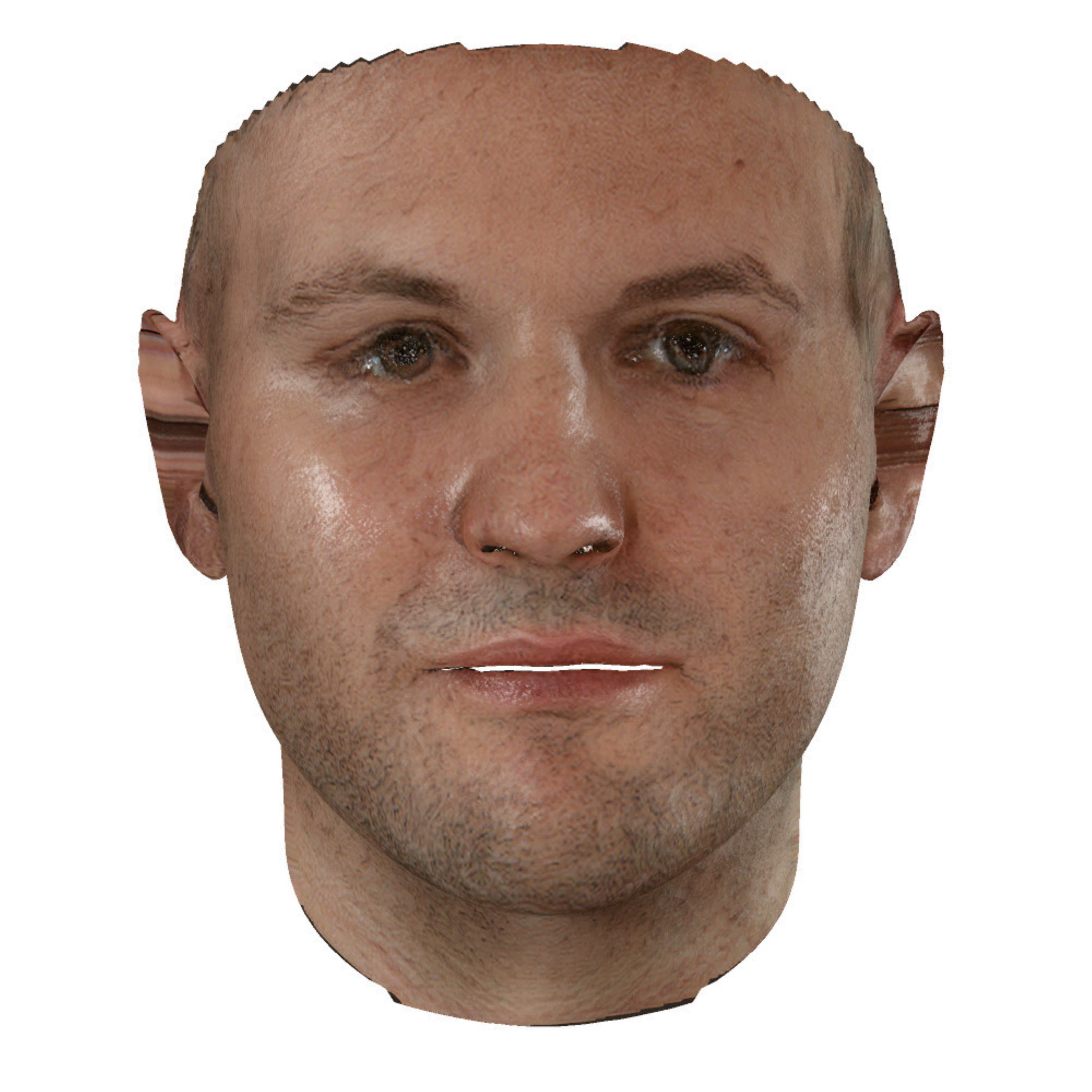}}
    \subfloat{
        \includegraphics[width=0.24\linewidth]{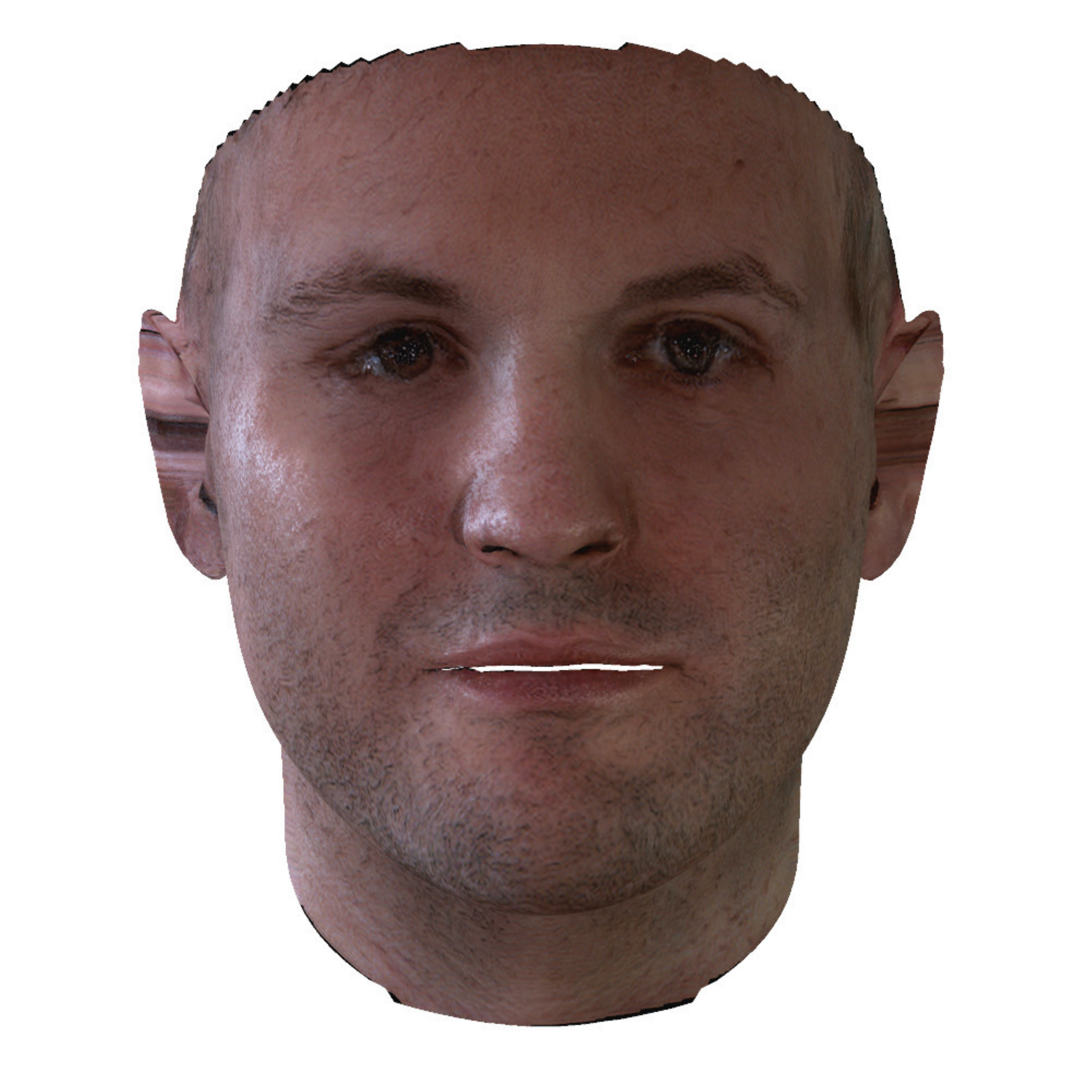}}
    \subfloat{
        \includegraphics[width=0.24\linewidth]{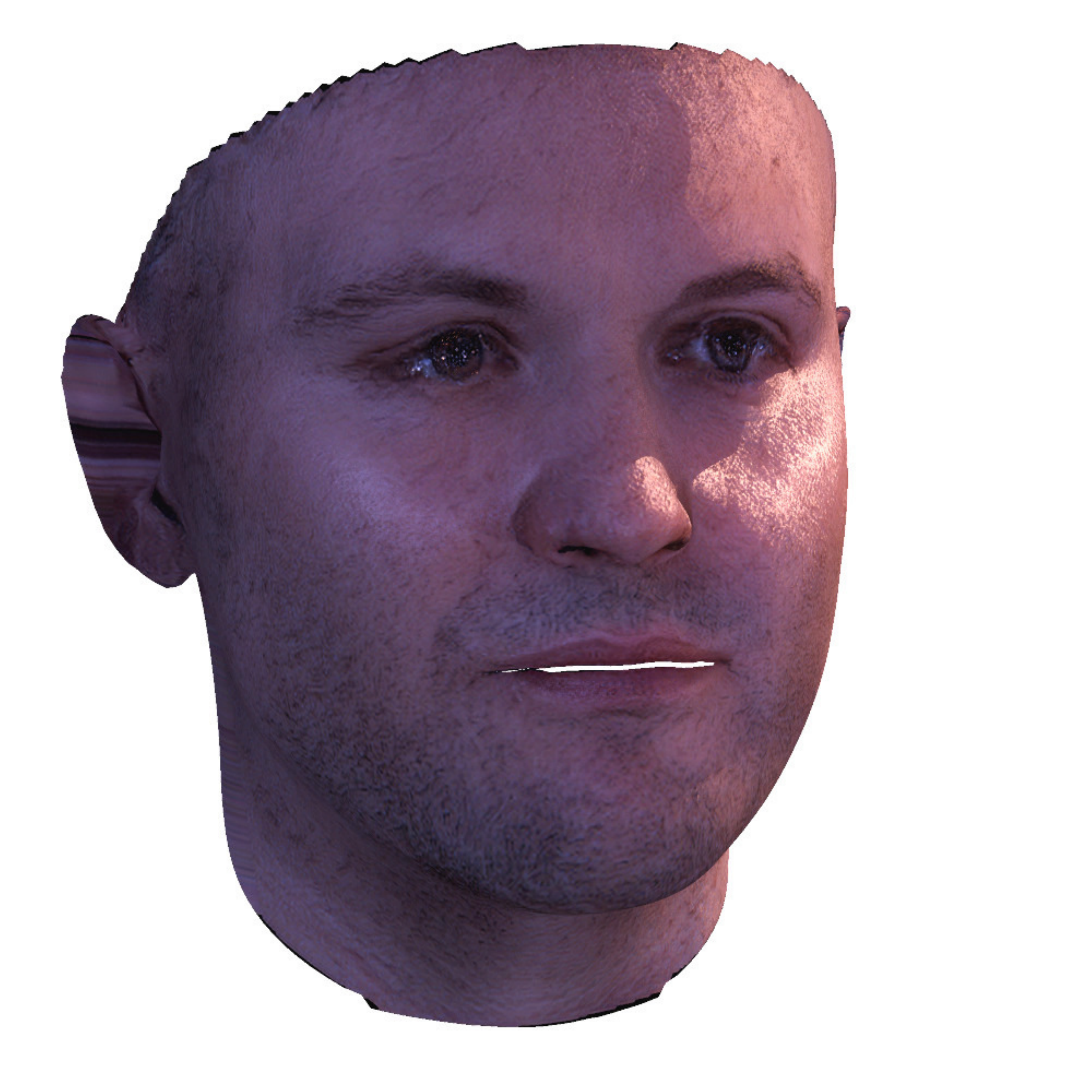}} \\
        
    \subfloat{
        \includegraphics[width=0.24\linewidth]{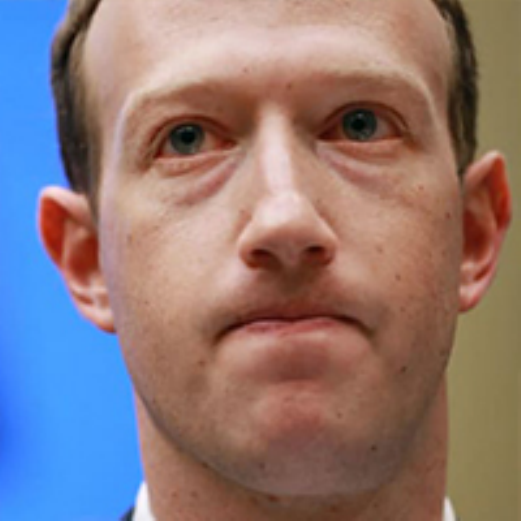}}
    \subfloat{
        \includegraphics[width=0.24\linewidth]{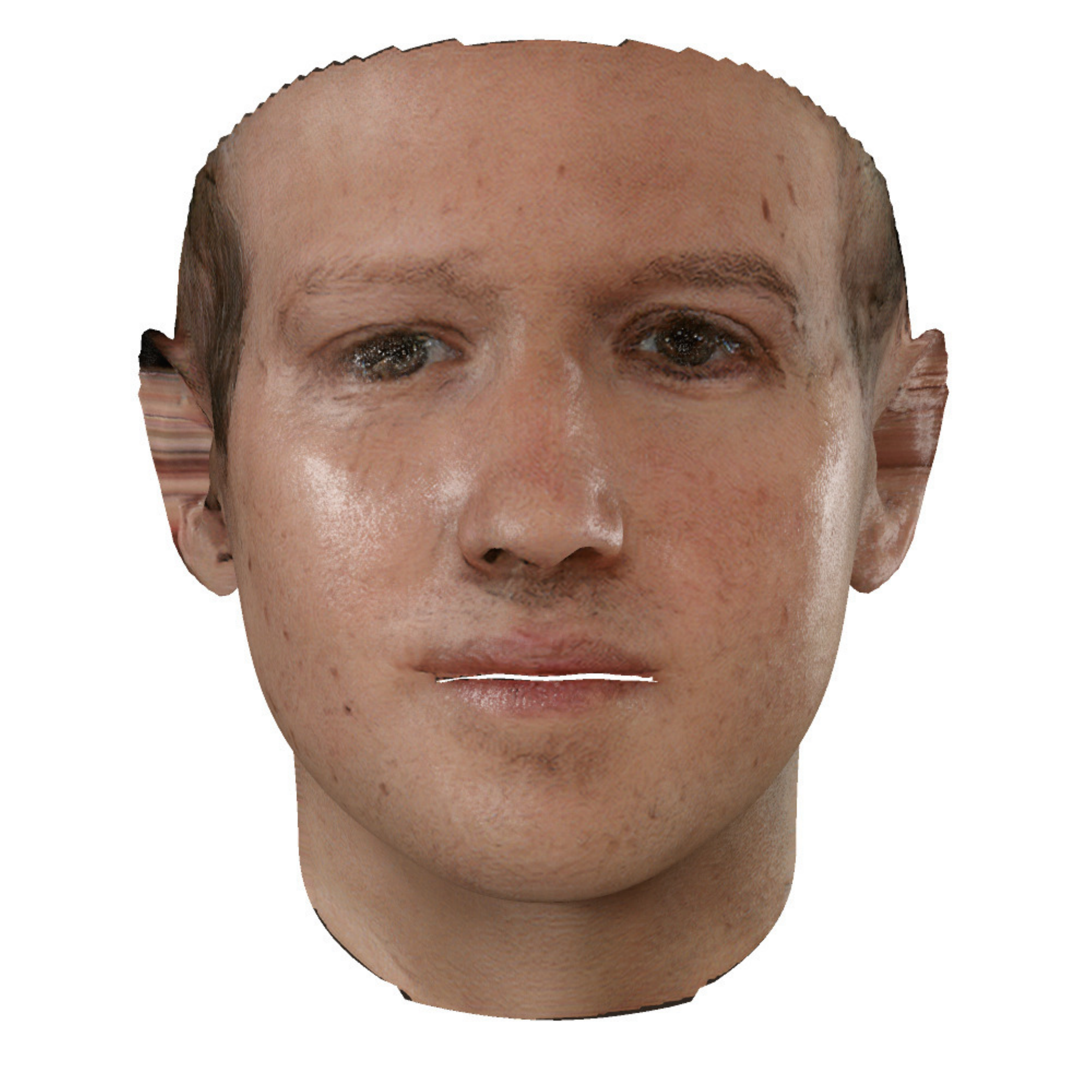}}
    \subfloat{
        \includegraphics[width=0.24\linewidth]{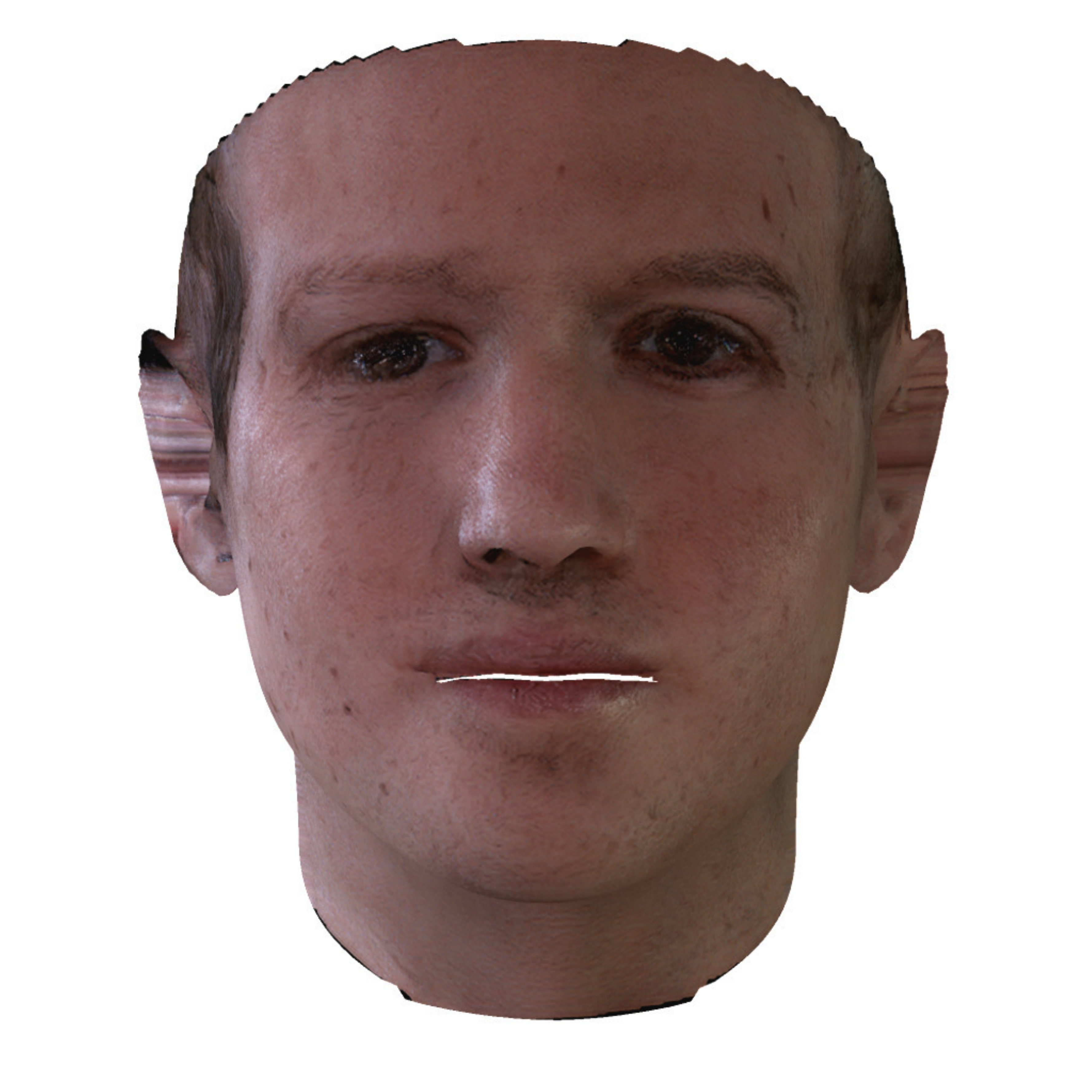}}
    \subfloat{
        \includegraphics[width=0.24\linewidth]{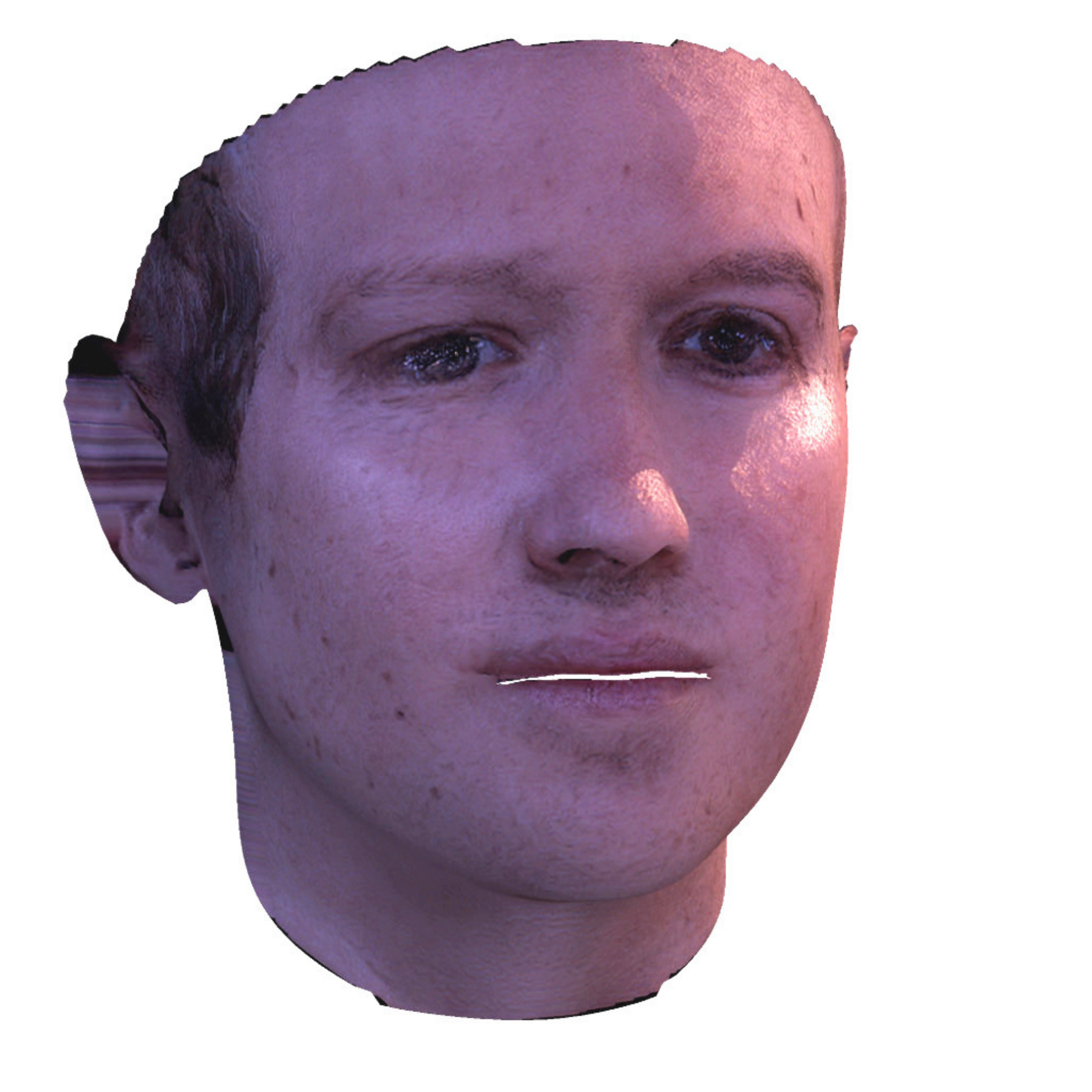}} \\

    \subfloat{
        \includegraphics[width=0.24\linewidth]{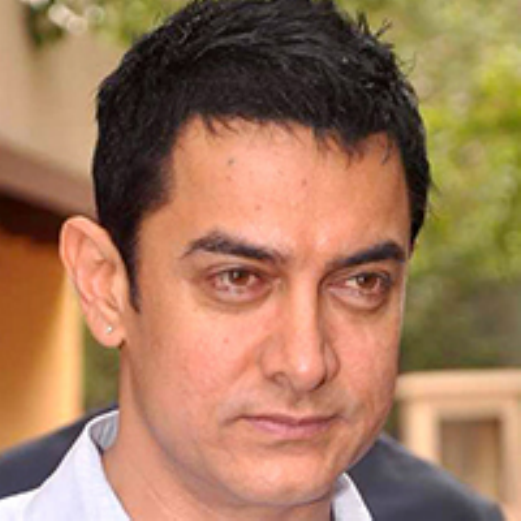}}
    \subfloat{
        \includegraphics[width=0.24\linewidth]{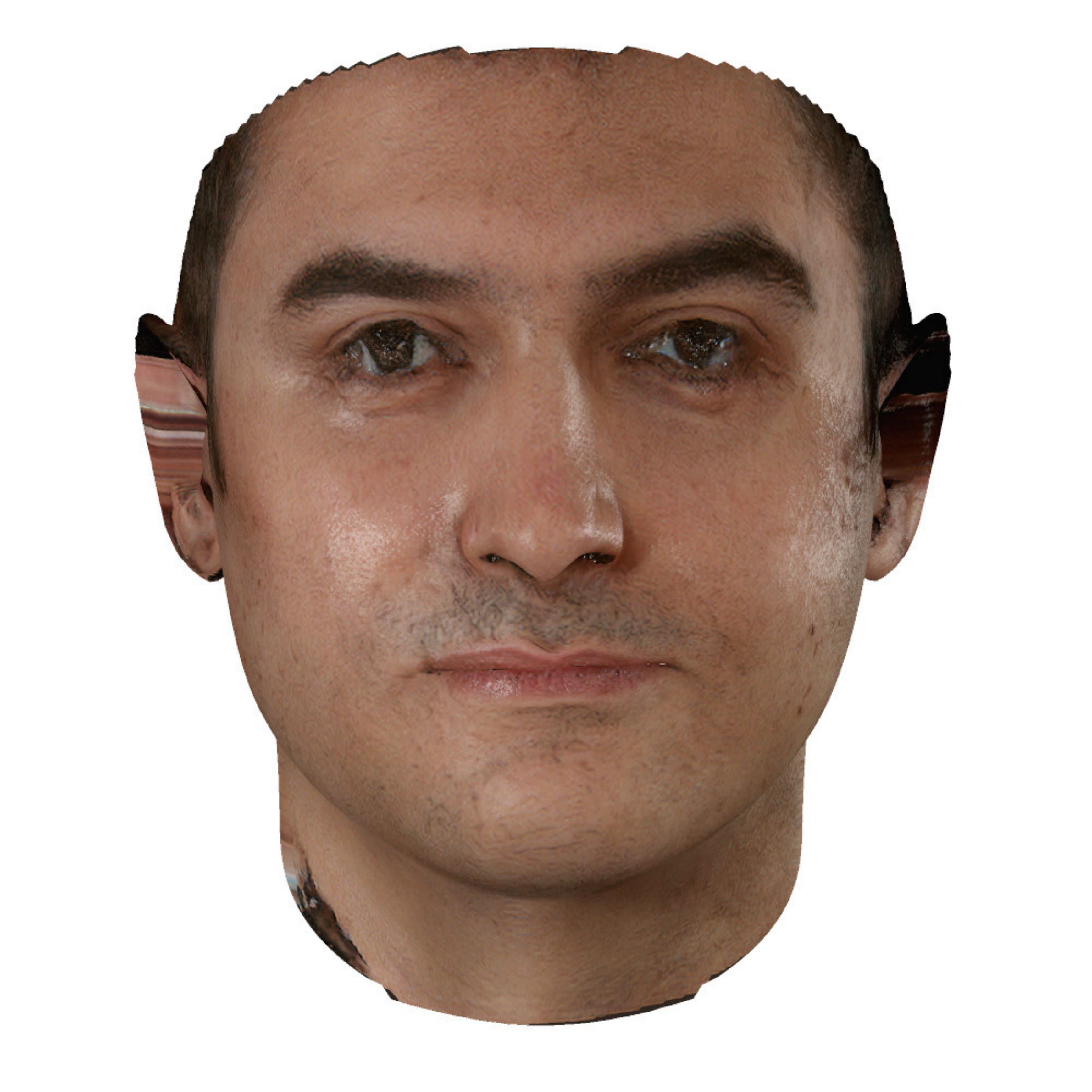}}
    \subfloat{
        \includegraphics[width=0.24\linewidth]{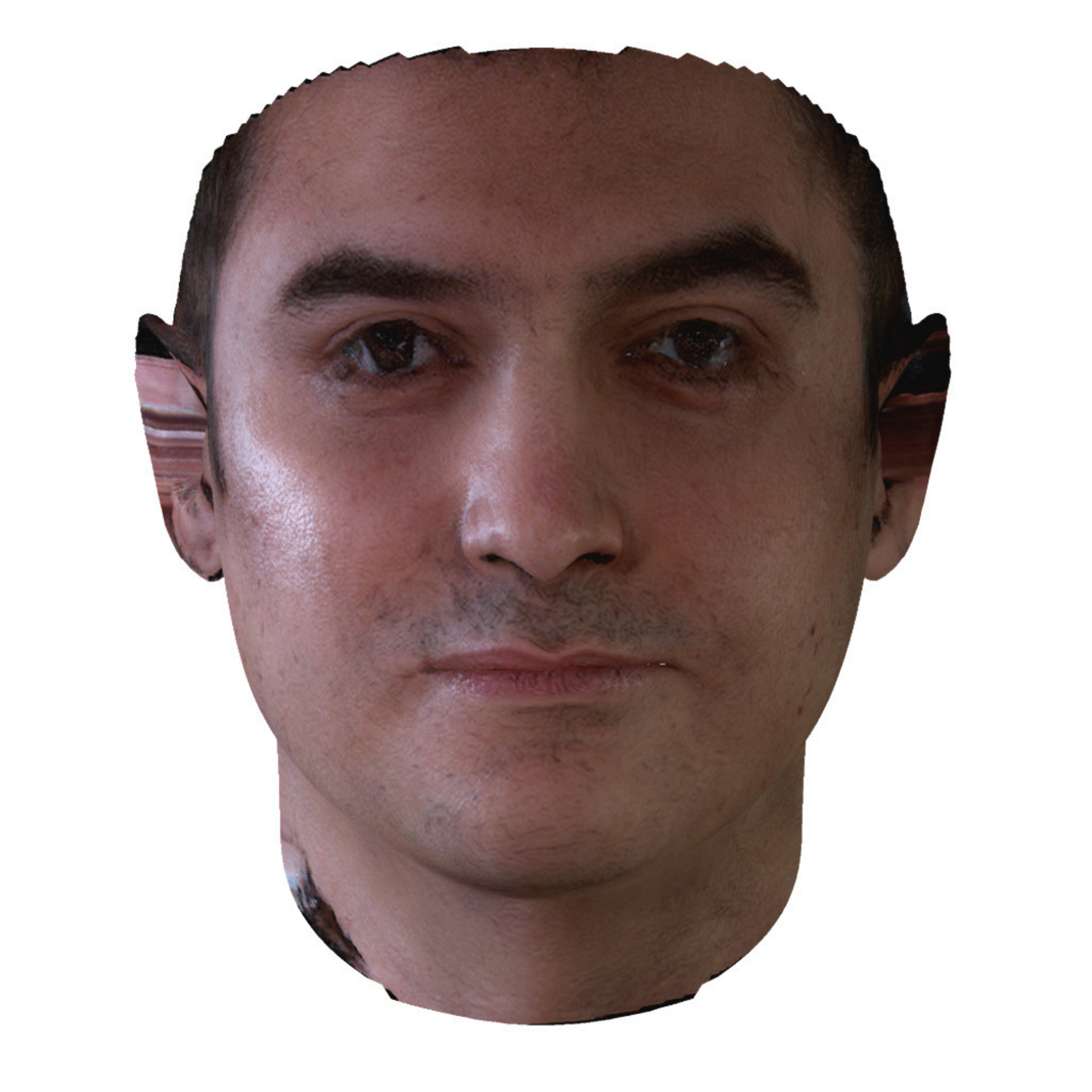}}
    \subfloat{
        \includegraphics[width=0.24\linewidth]{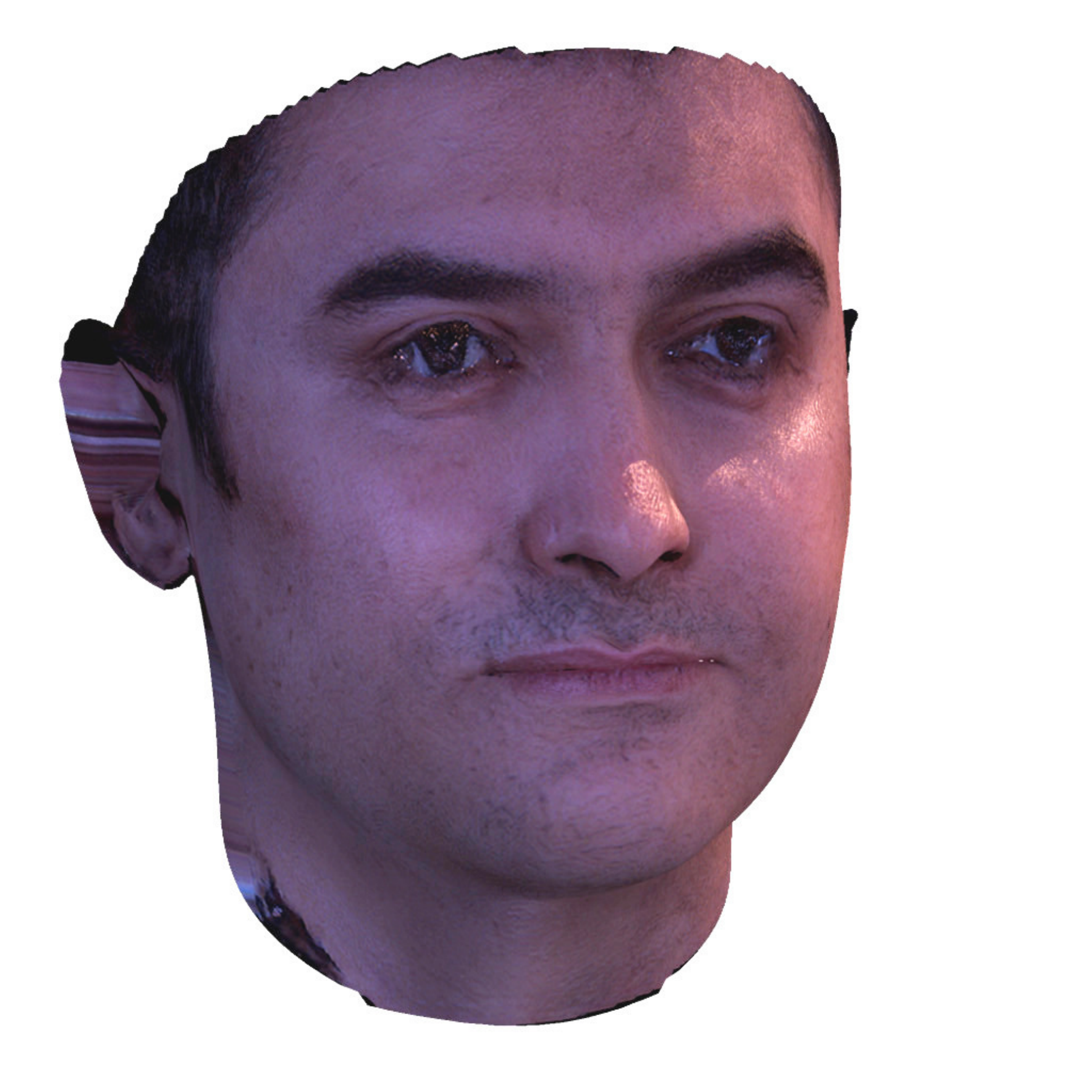}}\\

    \setcounter{subfigure}{0}

    \captionsetup[subfloat]{justification=centering}
    \subfloat[
        Input]{
        \includegraphics[width=0.24\linewidth]{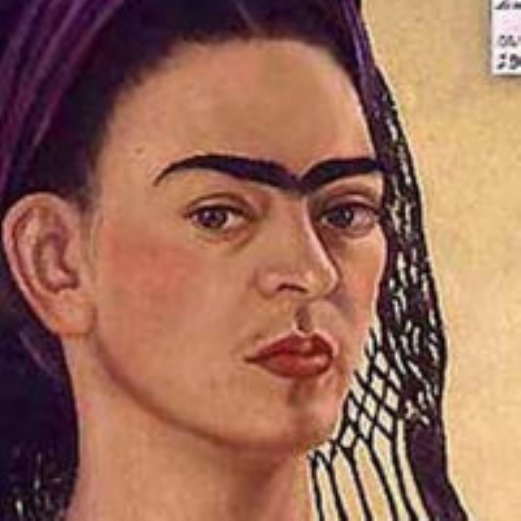}}
    \subfloat[
        Cathedral]{
        \includegraphics[width=0.24\linewidth]{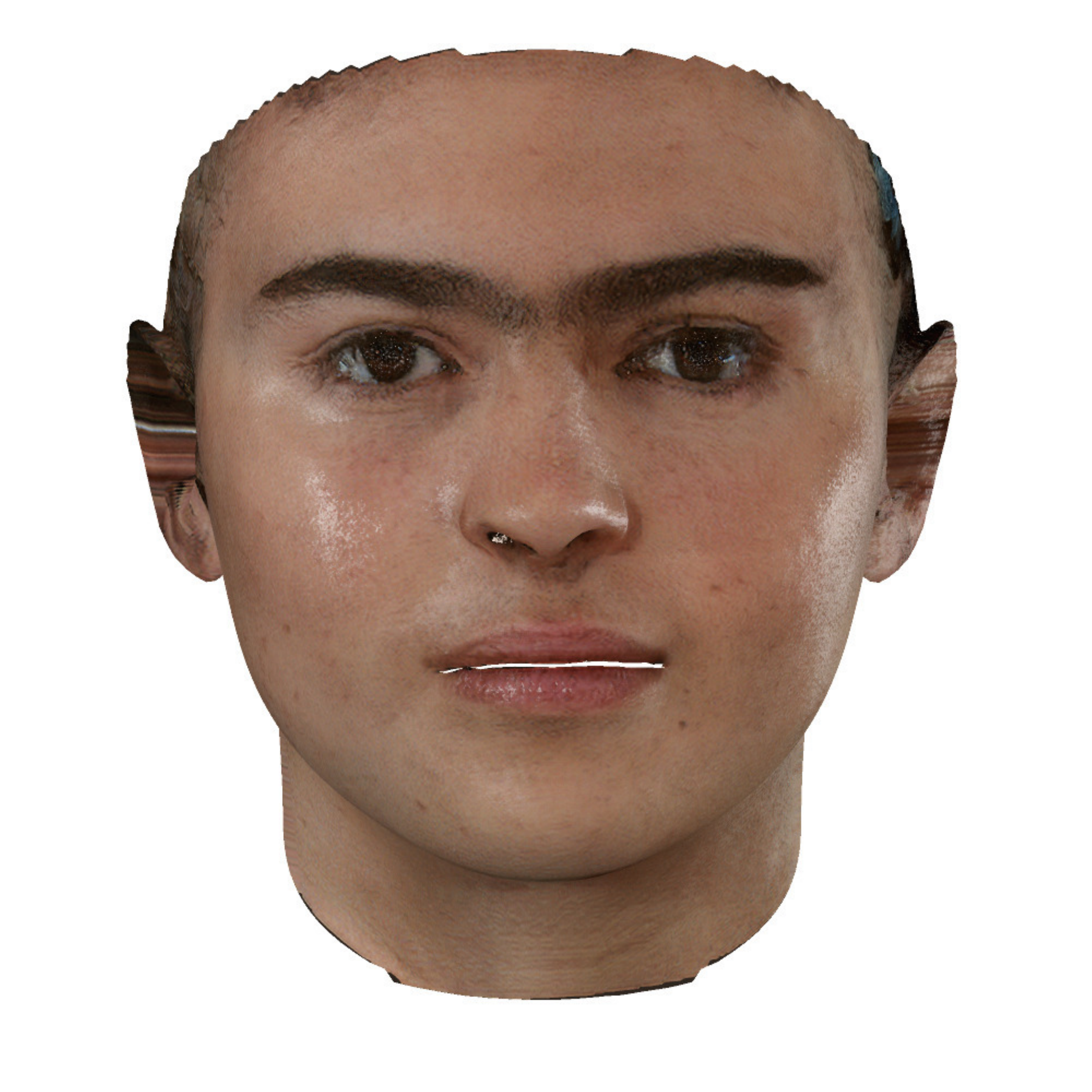}}
    \subfloat[
        Night Street]{
        \includegraphics[width=0.24\linewidth]{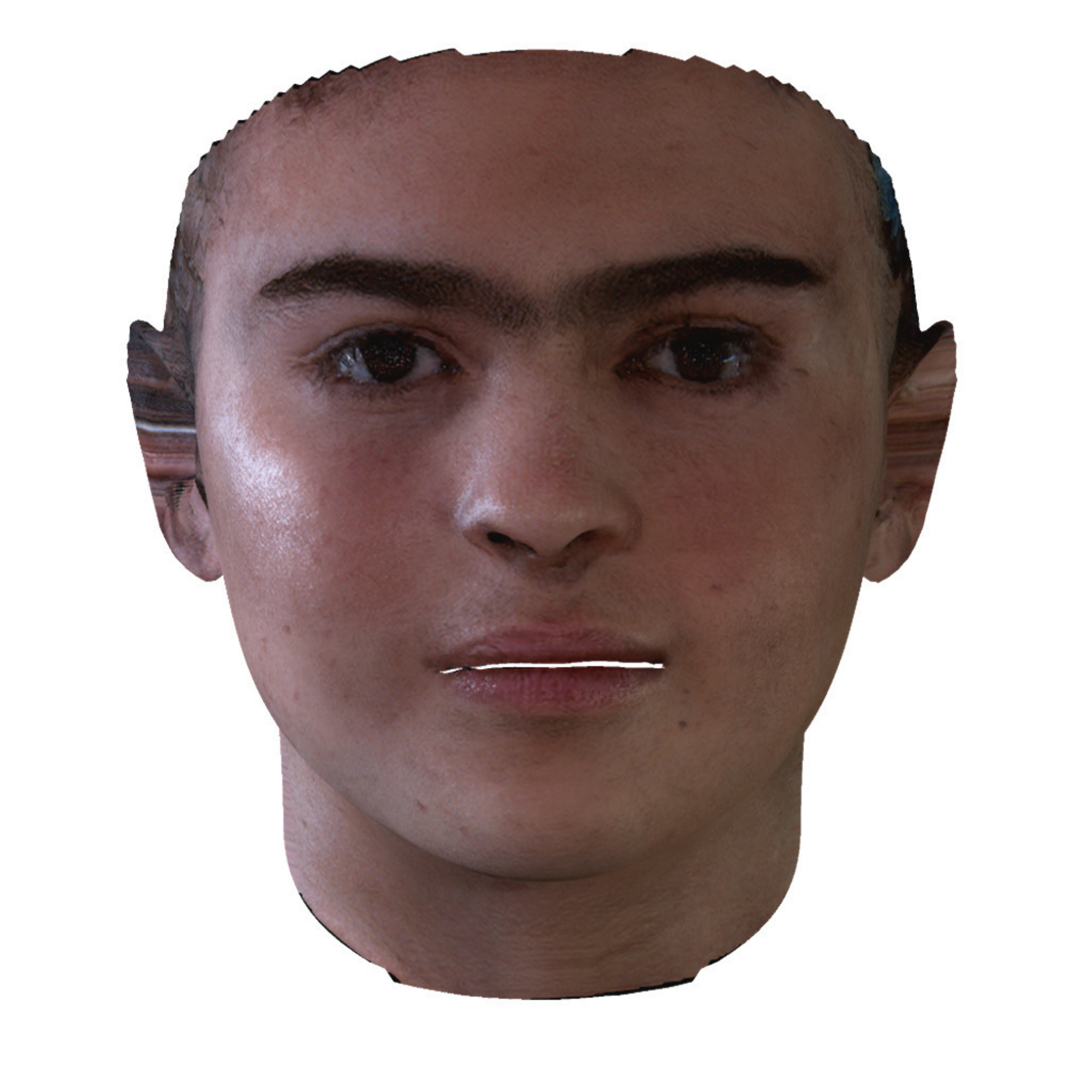}}
    \subfloat[
        Sunset]{
        \includegraphics[width=0.24\linewidth]{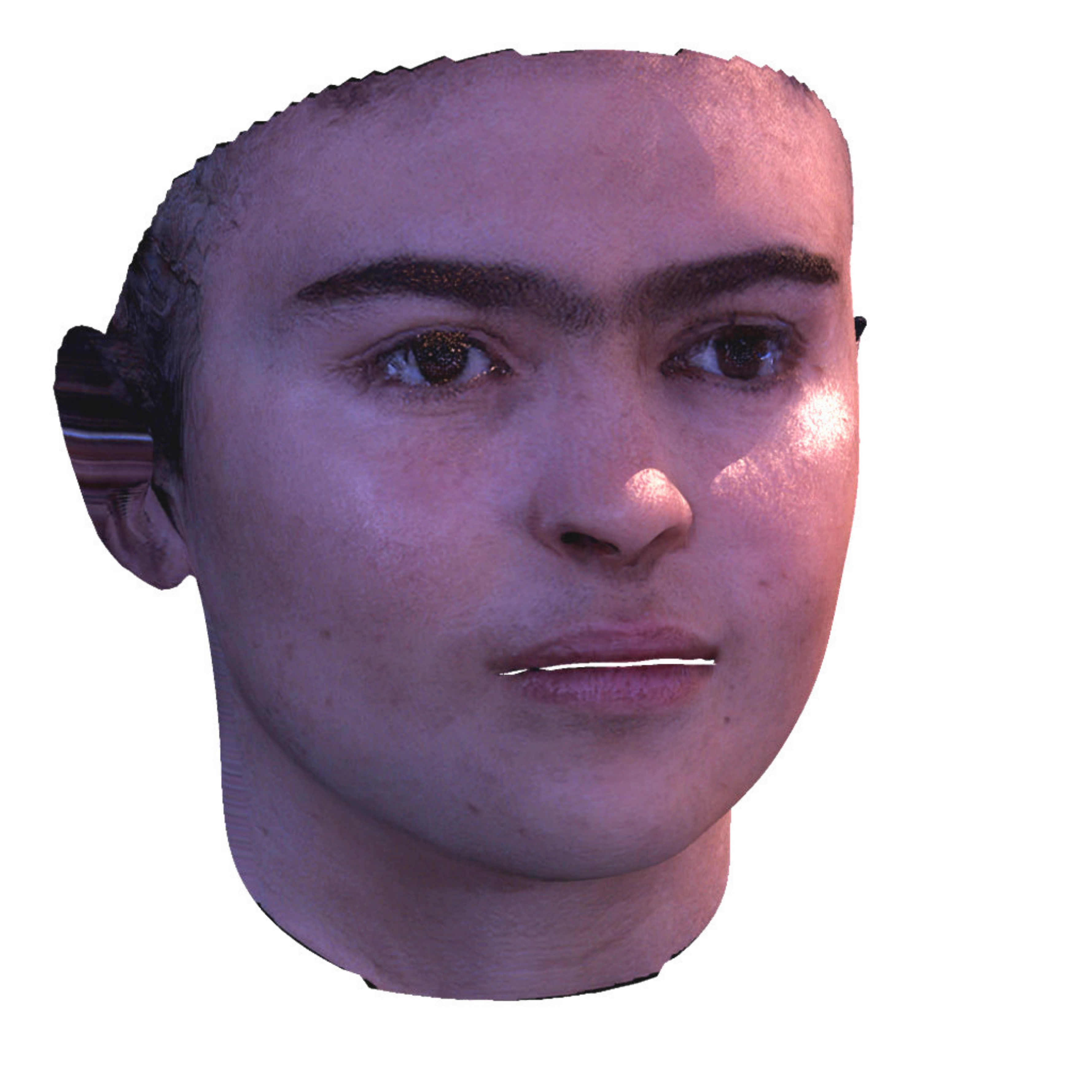}}
    
    \caption{
        AvatarMe\textsuperscript{++} results rendered
        in different environments.
    }
    \label{fig:results_many_results}
    \vspace{-0.35cm}
\end{figure}

\begin{figure}[h]
    \centering
    \vspace{-0.3cm}
    \captionsetup[subfloat]{justification=centering}
    \subfloat[
        Input]{
        \includegraphics[width=0.19\linewidth]{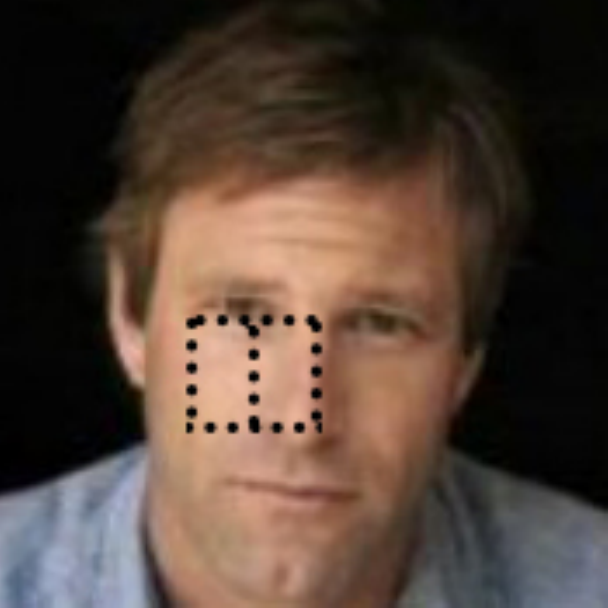}}
    \subfloat[
        3DMM Recon \cite{gecer_ganfit_2019}]{
        \includegraphics[width=0.19\linewidth]{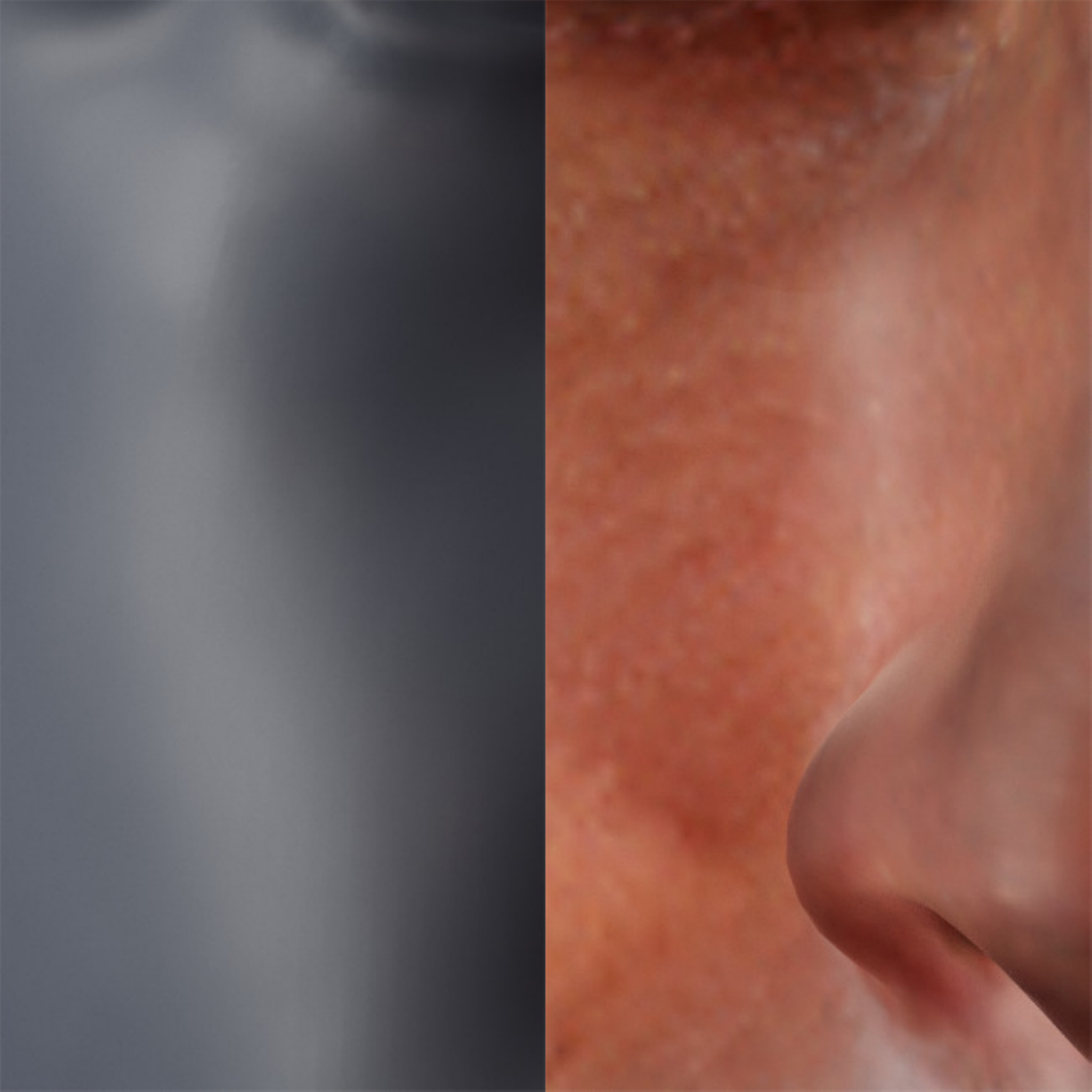}}
    \subfloat[
        Ours, 3DMM Super Res.]{
        \includegraphics[width=0.19\linewidth]{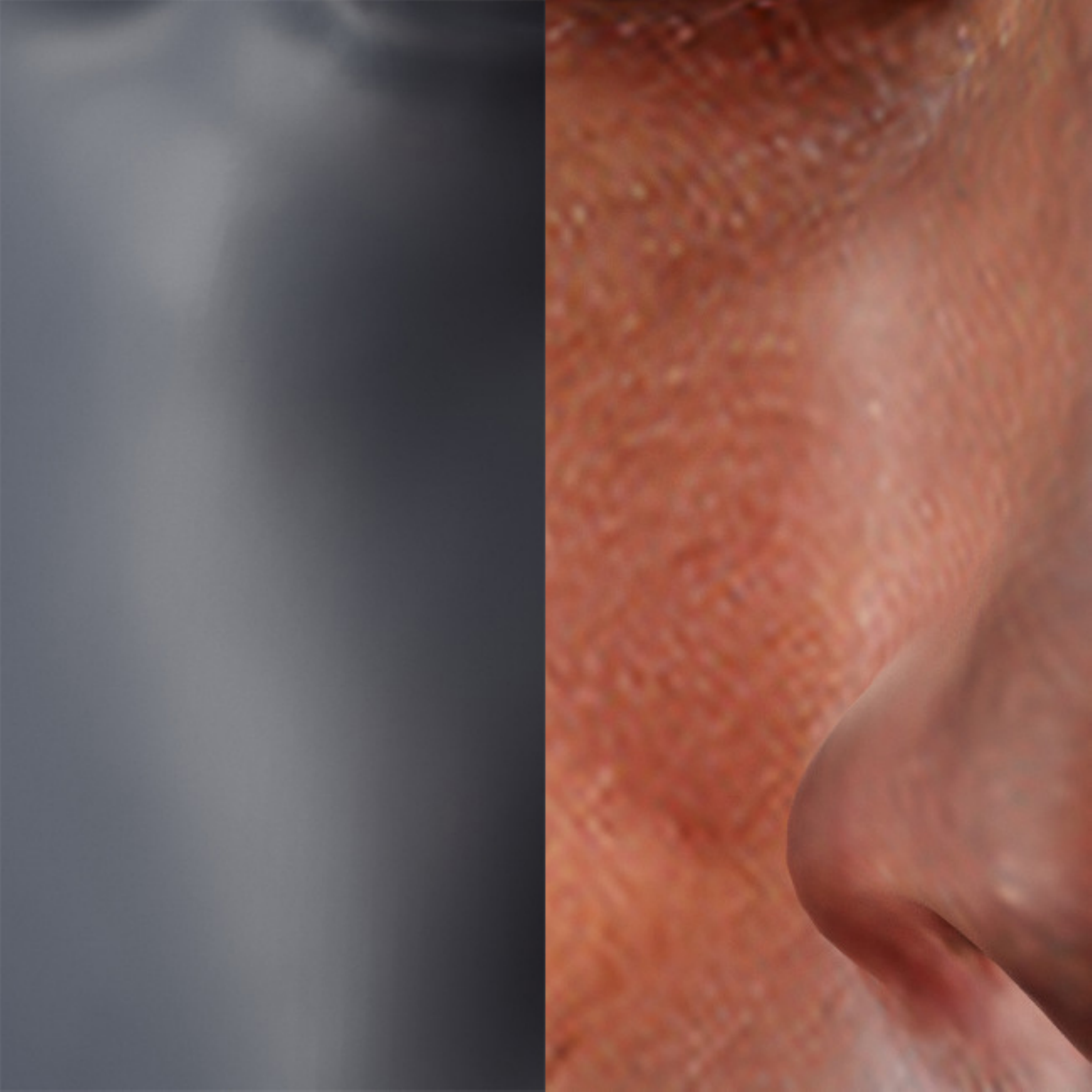}}
    \subfloat[
        Ours, AvatarMe]{
        \includegraphics[width=0.19\linewidth]{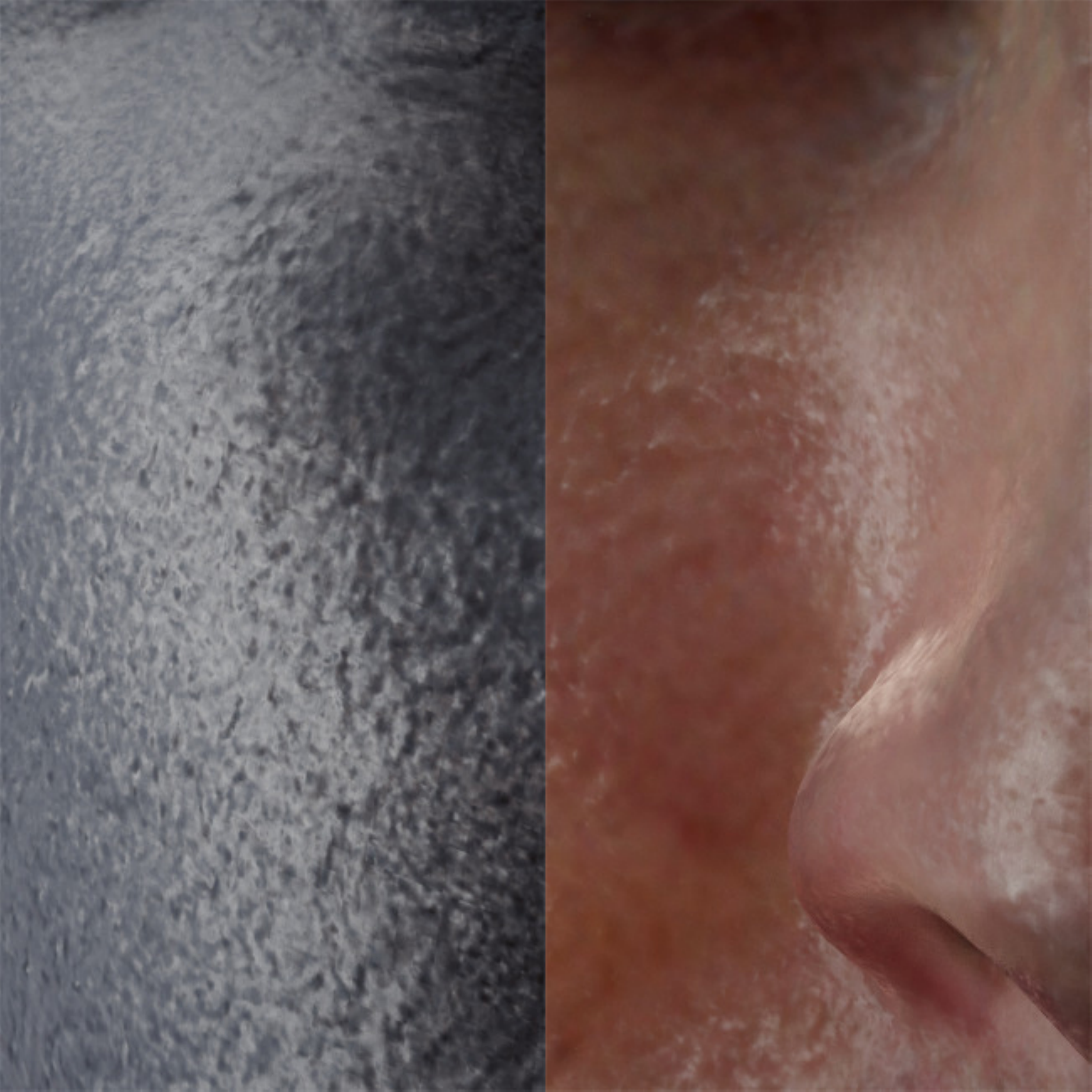}}
    \subfloat[
        Ours, AvatarMe\textsuperscript{++}]{
        \includegraphics[width=0.19\linewidth]{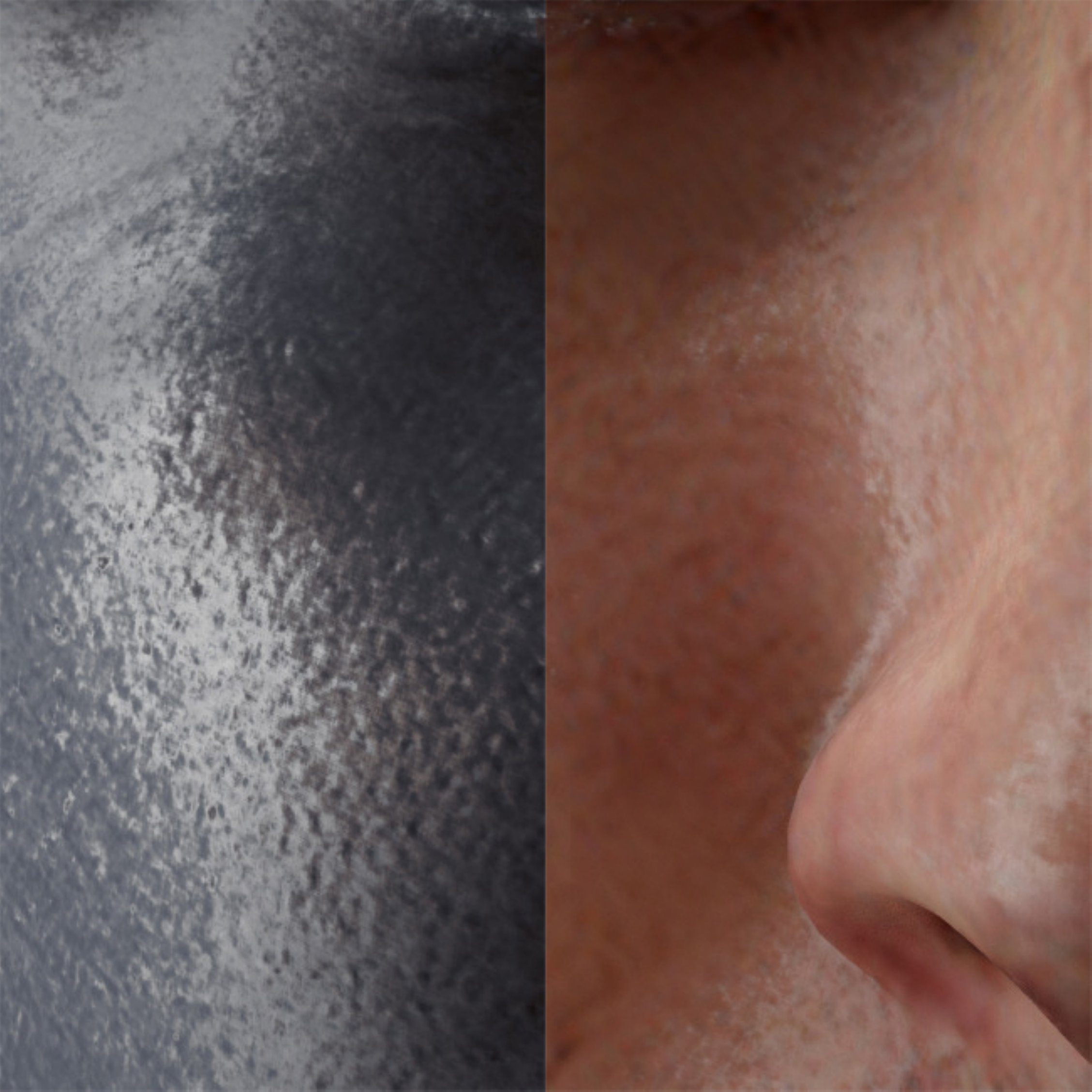}}
    
    \caption{
        Comparison of our shape reconstruction (left) and rendering (right)
        between the 3DMM fitting with GANFIT \cite{gecer_ganfit_2019},
        upsampled texture (Sec.~\ref{sec:method_recon_super_res}),
        AvatarMe (Sec.~\ref{sec:method_recon_avatarme})
        and AvatarMe\textsuperscript{++} (Sec.~\ref{sec:method_recon_renderme}).
    }
    \label{fig:results_ablation_rendering_steps}
    \vspace{-0.2cm}
\end{figure}

For the qualitative comparisons, 
we perform 3D reconstructions of arbitrary images. 
As shown in Figs.~\ref{fig:results_comp_comparison} and \ref{fig:comp_qualitative_comp},
our method does not produce any artifacts in the final renderings 
and successfully handles extreme poses and occlusions such as sunglasses. 
We infer the texture maps in a patch-based manner from high-resolution input, 
which produces higher-quality details than \cite{chen_photo-realistic_2019} and \cite{yamaguchi_high-fidelity_2018}, 
who train on high-quality scans but infer the maps for the whole face, 
in lower resolution. 
This is also apparent in Fig.~\ref{fig:results_gt_comparison} and Fig.~\ref{fig:results_ablation_rendering_steps}, 
which shows our reconstruction after each step of our process.
Moreover, we can successfully acquire each component from
black-and-white images (Fig.~\ref{fig:results_comp_comparison})
and even paintings (Fig.~\ref{fig:results_many_results}).

\begin{figure}[h]
\vspace{-0.3cm}
    \centering
    \captionsetup[subfigure]{labelformat=empty}
    \subfloat{
        \includegraphics[height=1.31cm]{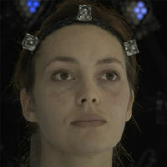}}
    \subfloat{
        \includegraphics[height=1.31cm]{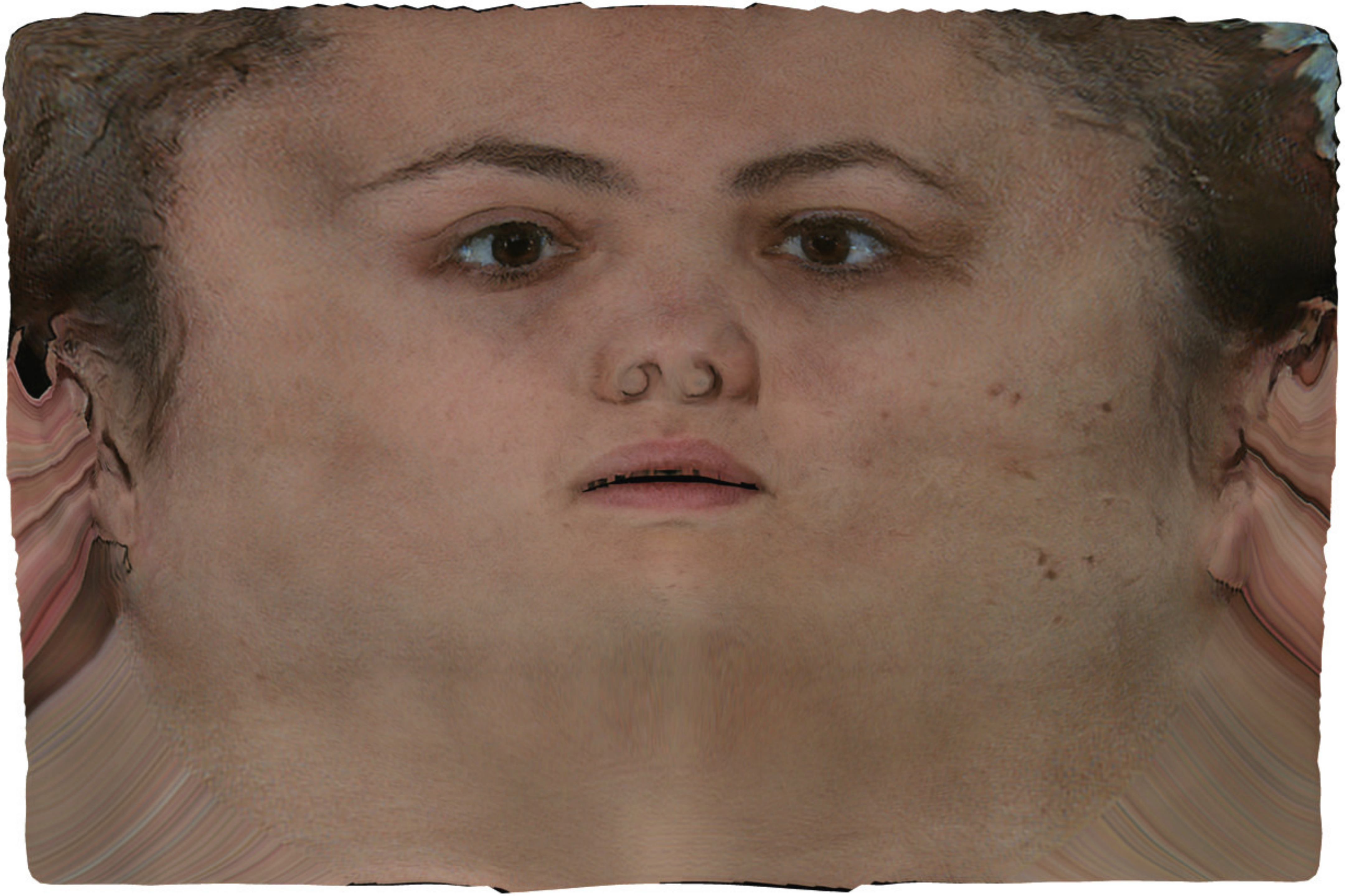}}
    \subfloat{
        \includegraphics[height=1.31cm]{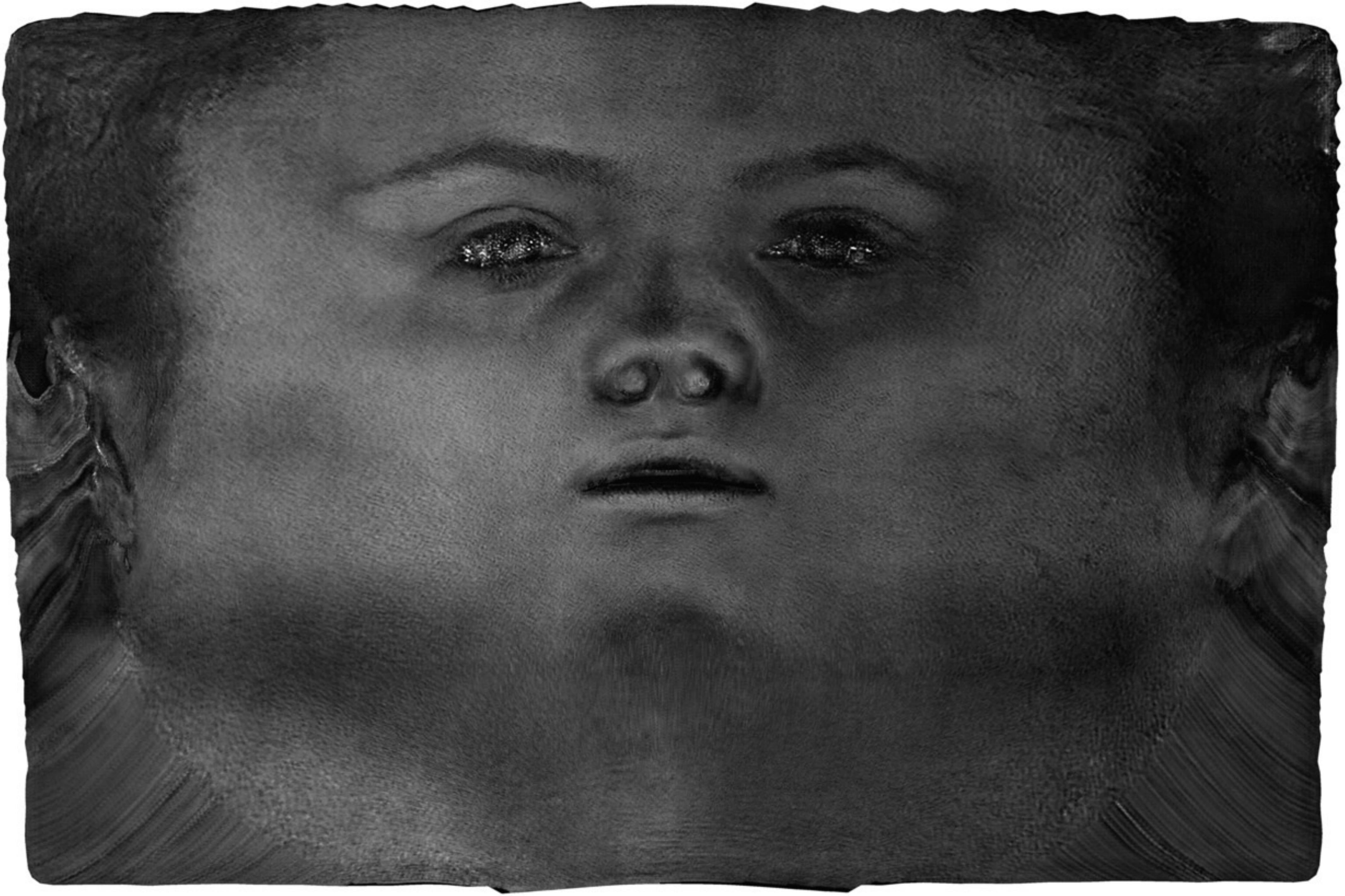}}
    \subfloat{
        \includegraphics[height=1.31cm]{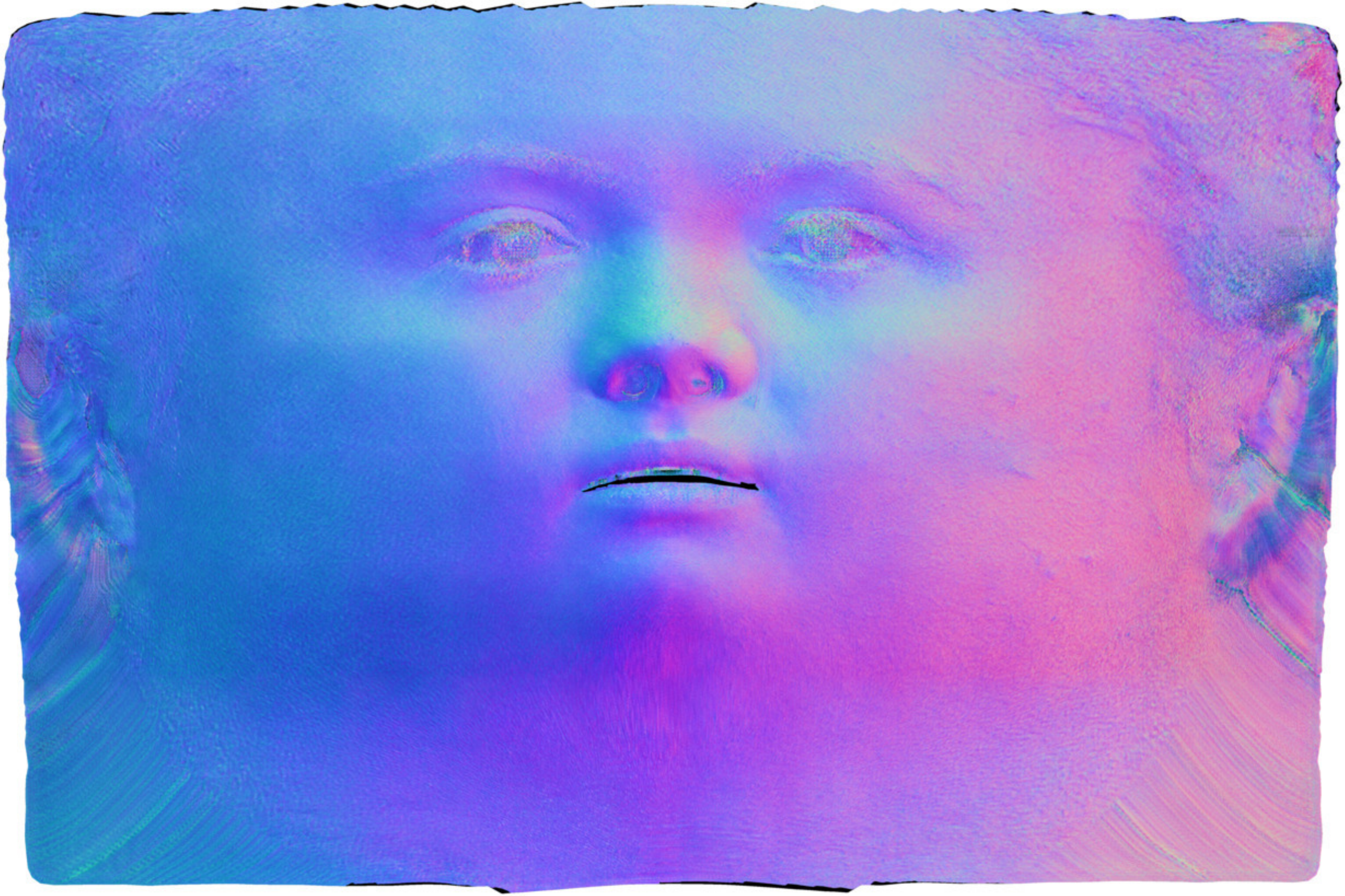}}
    \subfloat{
        \includegraphics[height=1.31cm]{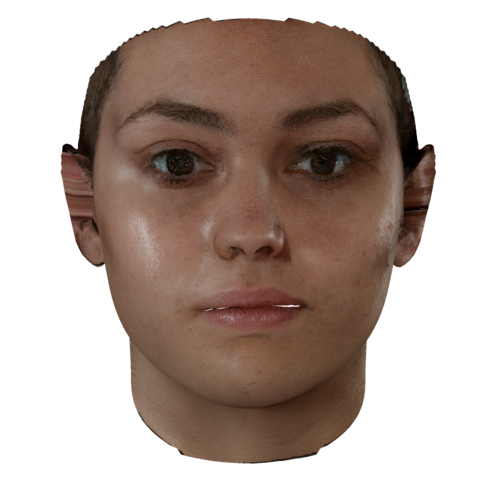}}\\
    \setcounter{subfigure}{0}
    \captionsetup[subfloat]{justification=centering}
    \subfloat[
        Input]{
        \includegraphics[height=1.31cm]{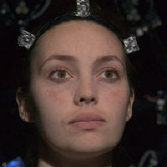}}
    \subfloat[
        Diff.Alb.]{
        \includegraphics[height=1.31cm]{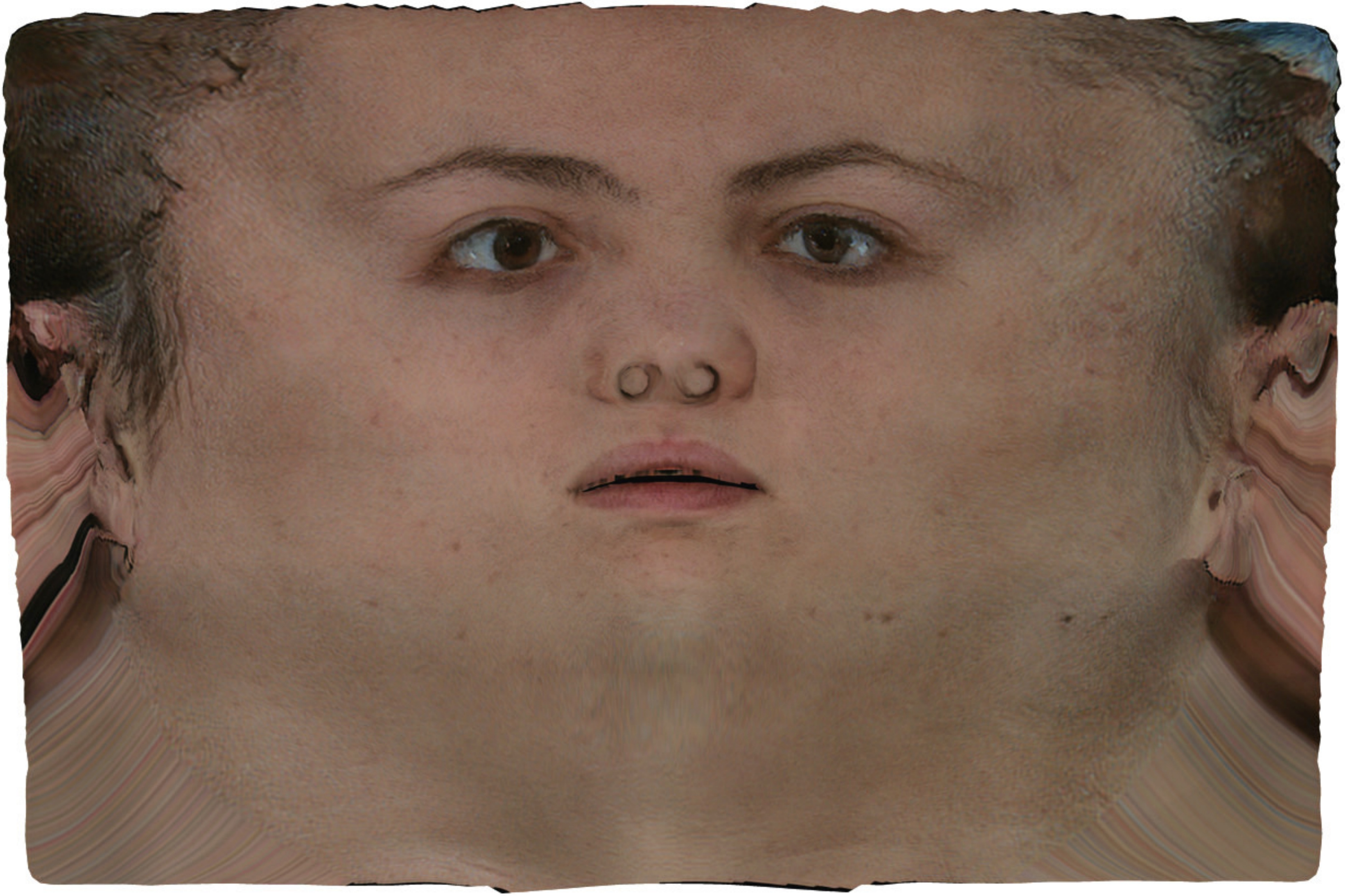}}
    \subfloat[
        Sp.Alb.]{
        \includegraphics[height=1.31cm]{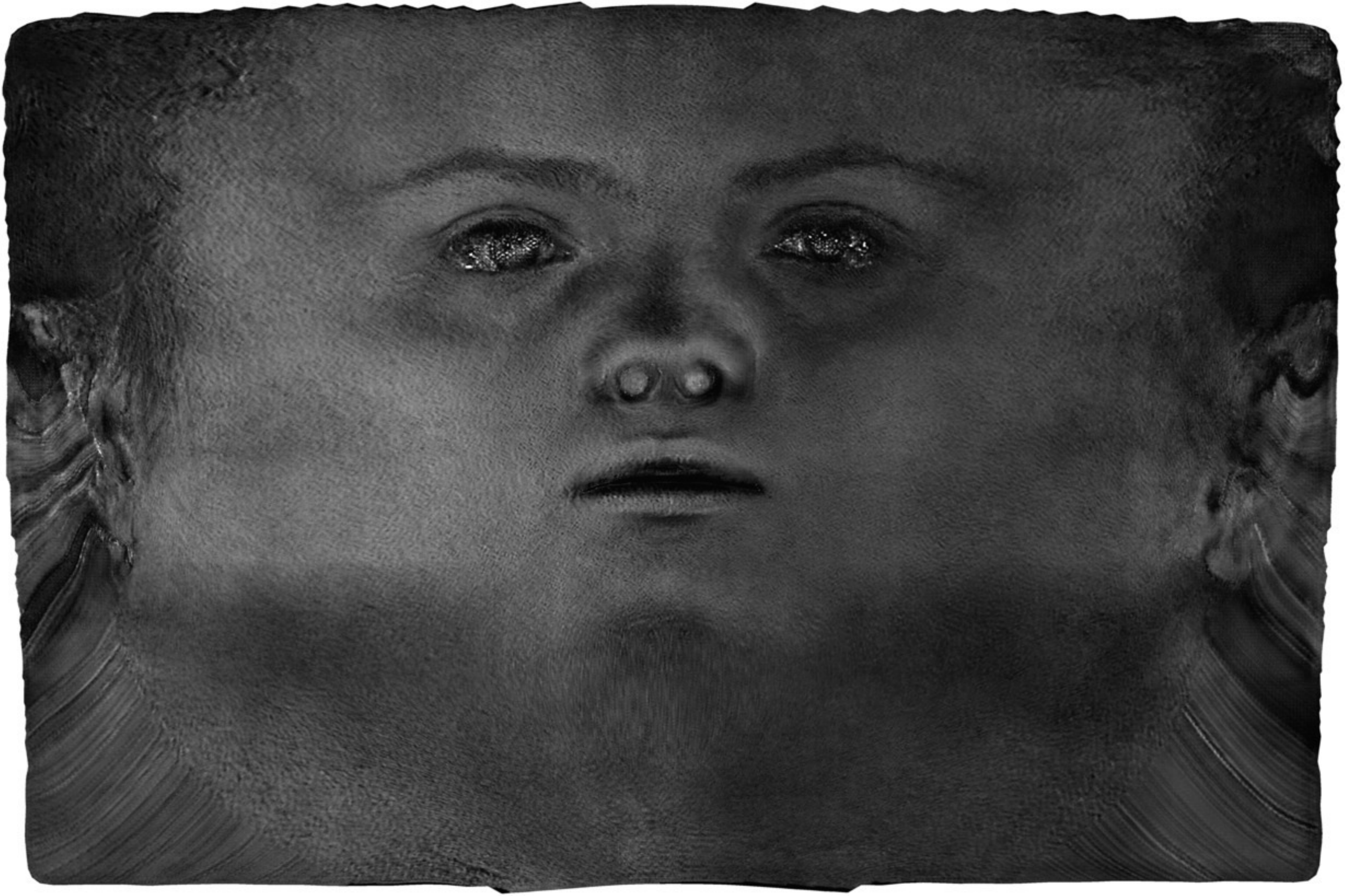}}
    \subfloat[
        Sp.Norm]{
        \includegraphics[height=1.31cm]{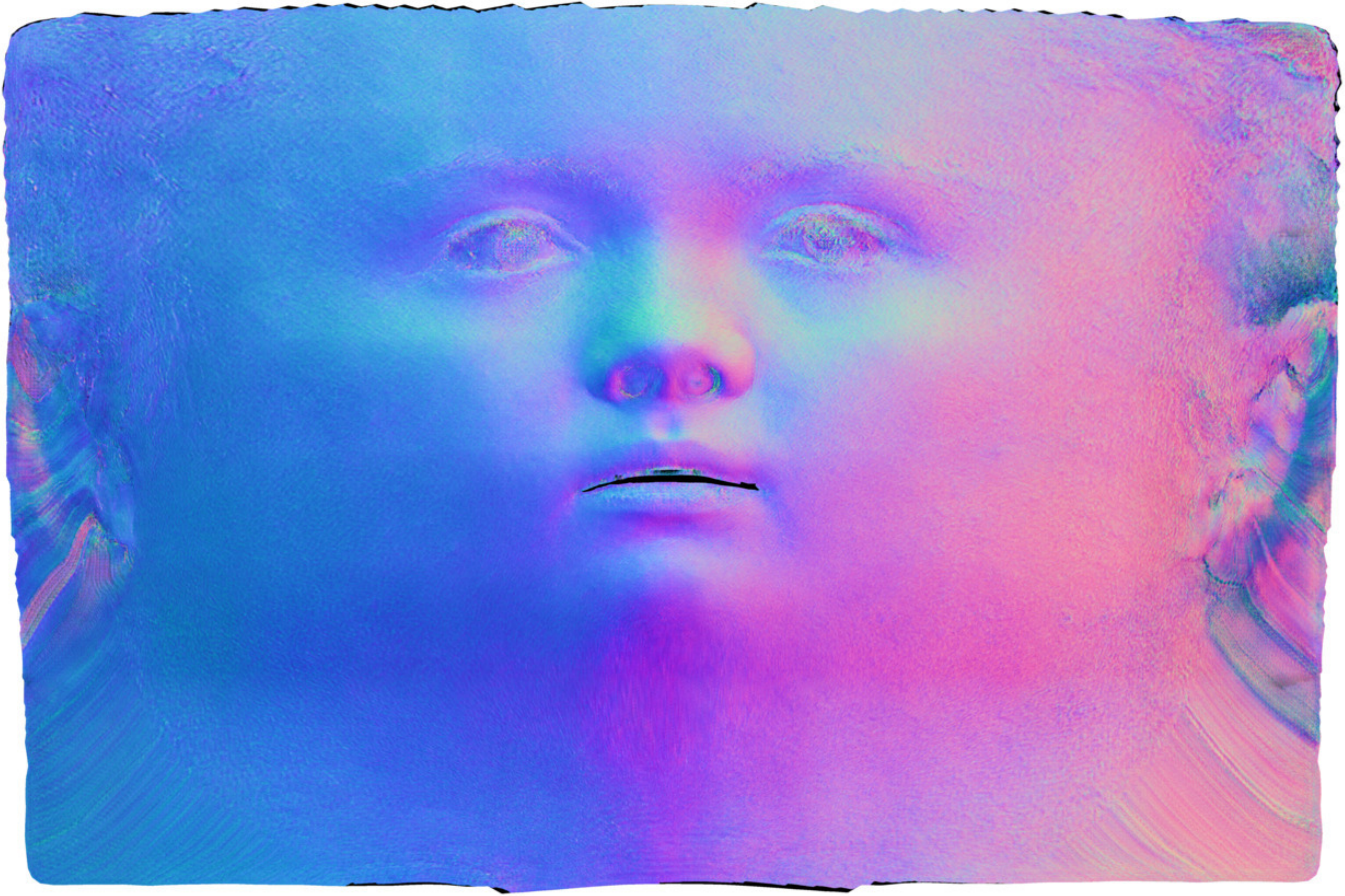}}
    \subfloat[
        Render]{
        \includegraphics[height=1.31cm]{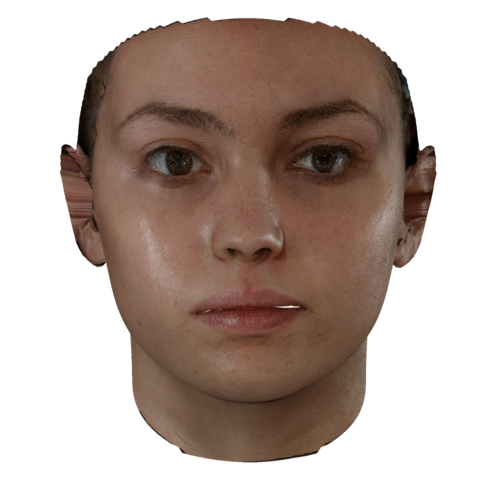}}
    
    \caption{
       Consistency of \modelnameplus{} on varying
        conditions, from the Digital Emily Project \cite{alexander_digital_2010}.
        We calculate on average $30.94$ PSNR and $0.0007$ MSE between our results.
        Compared to the ground truth from \cite{alexander_digital_2010},
        we achieve on average $0.0083$ MSE and $20.13$ PSNR on albedo and
        and $0.011$ MSE and $24.02$ PSNR on normals.
    }
    \label{fig:results_emily}
    \vspace{-0.3cm}
\end{figure}

Furthermore, we experiment with different environment conditions,
in the input images and while rendering.
As presented in Fig.~\ref{fig:results_emily}, 
the extracted normals, diffuse and specular albedos are consistent, 
regardless of the illumination on the original input images. 
Moreover, Fig.~\ref{fig:results_many_results} 
shows different subjects rendered under different environments.
We can realistically illuminate each subject in each scene and
accurately reconstruct the environment reflectance,
including detailed specular reflections and subsurface scattering.

\begin{figure}[ht]
    \centering
    \captionsetup[subfigure]{labelformat=empty}
    \subfloat{
        \includegraphics[width=0.24\linewidth]{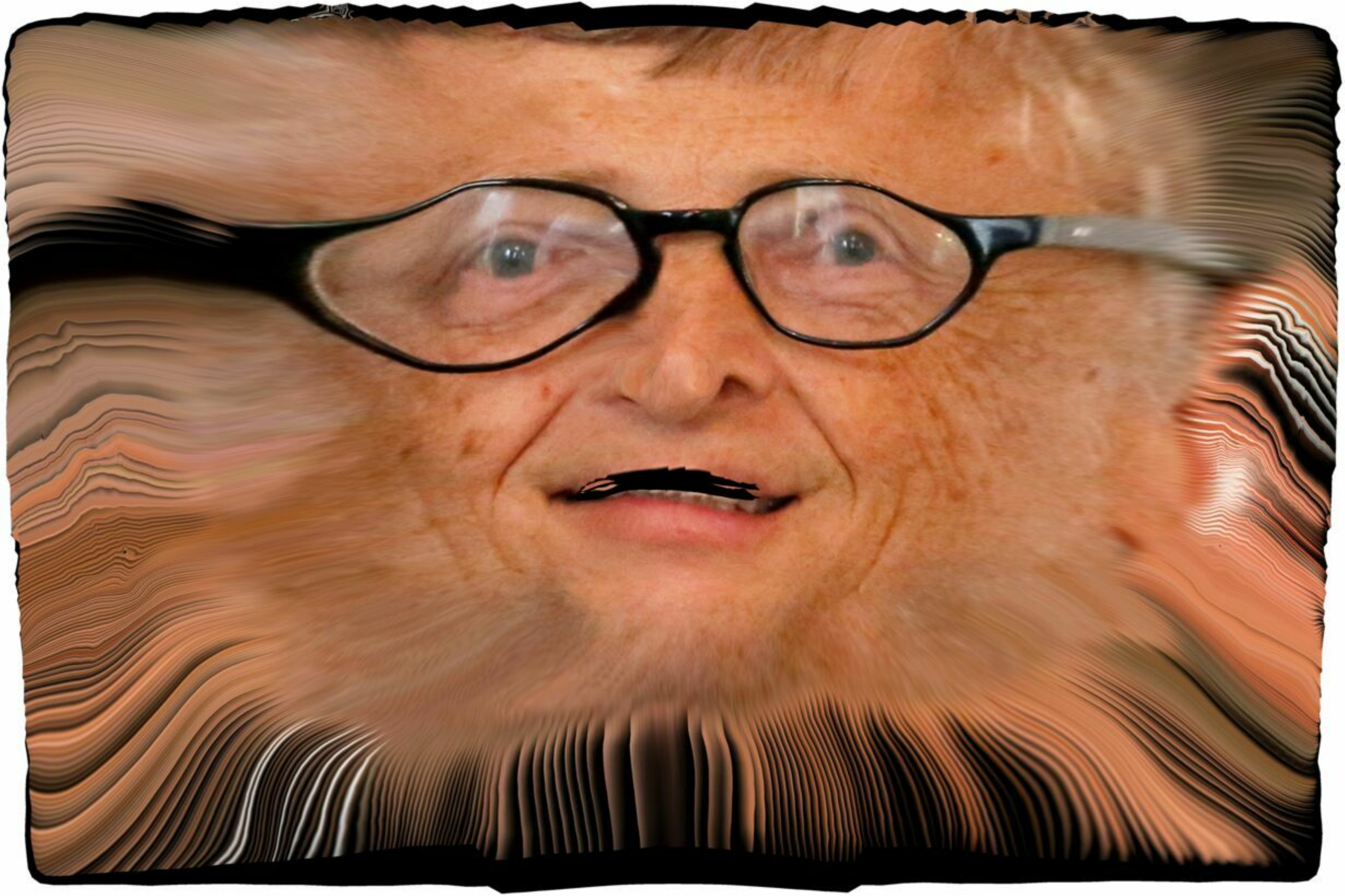}}
    \subfloat{
        \includegraphics[width=0.24\linewidth]{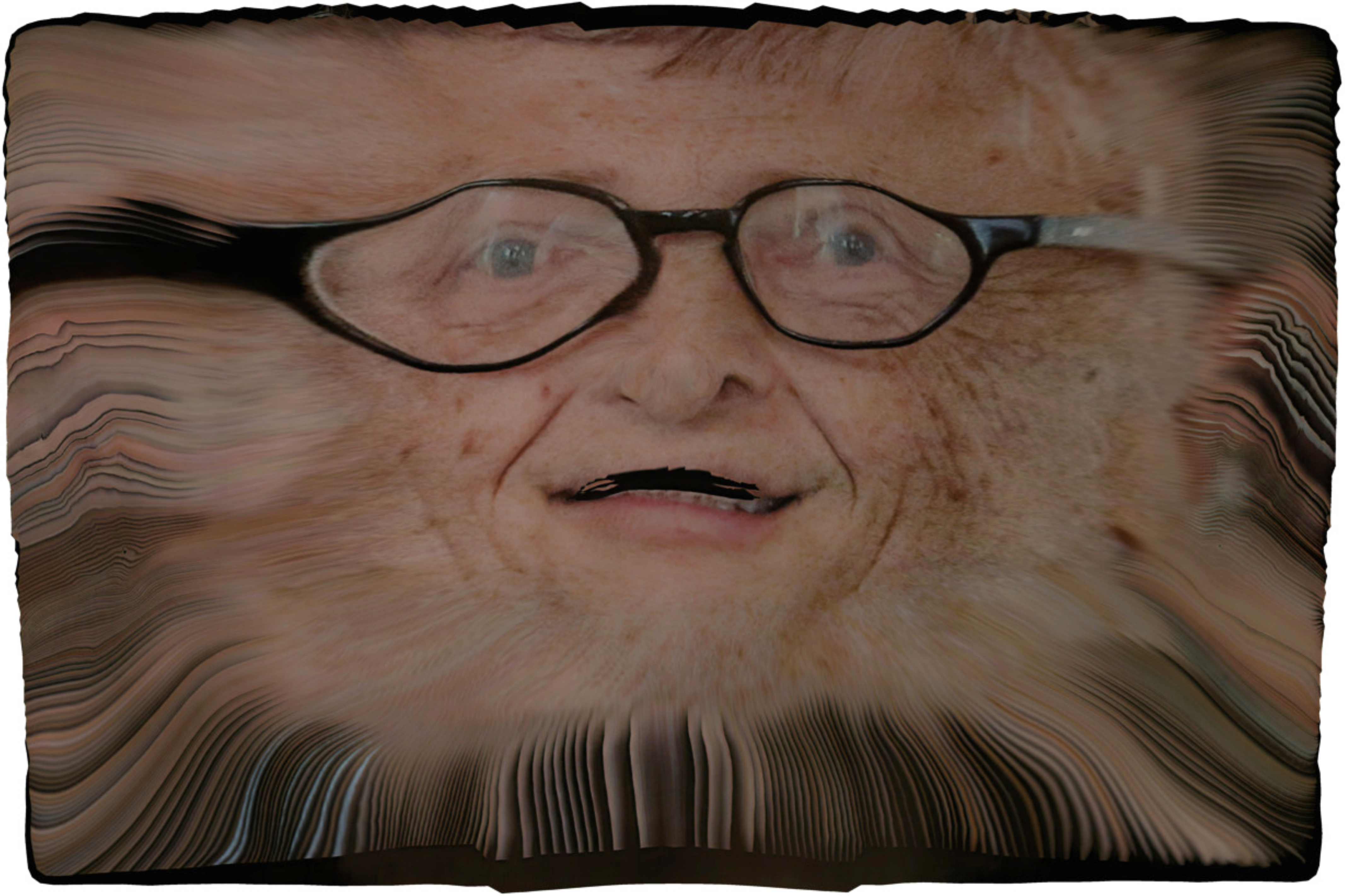}}
    \subfloat{
        \includegraphics[width=0.24\linewidth]{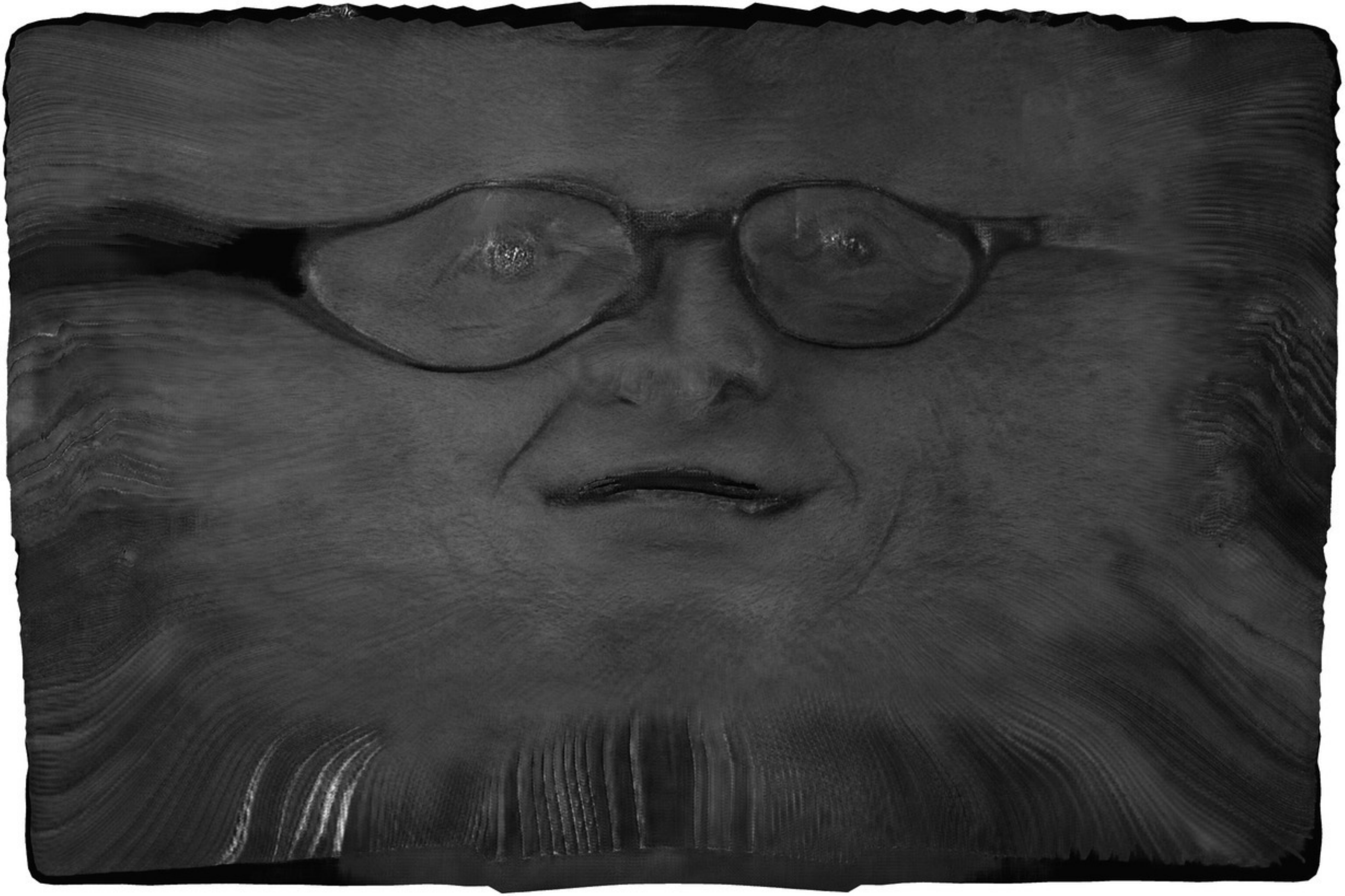}}
    \subfloat{
        \includegraphics[width=0.24\linewidth]{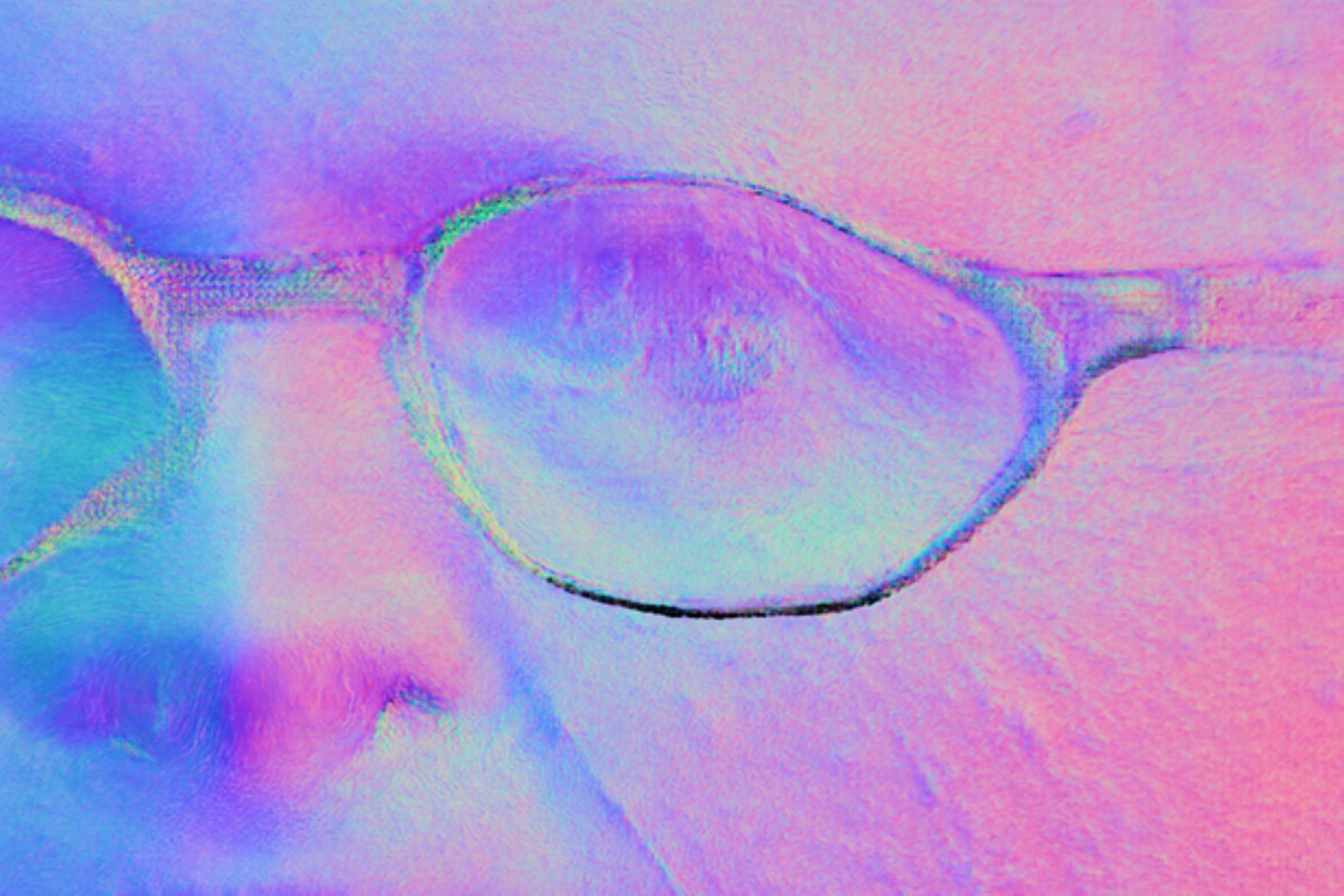}}
        
    \subfloat{
        \includegraphics[width=0.24\linewidth]{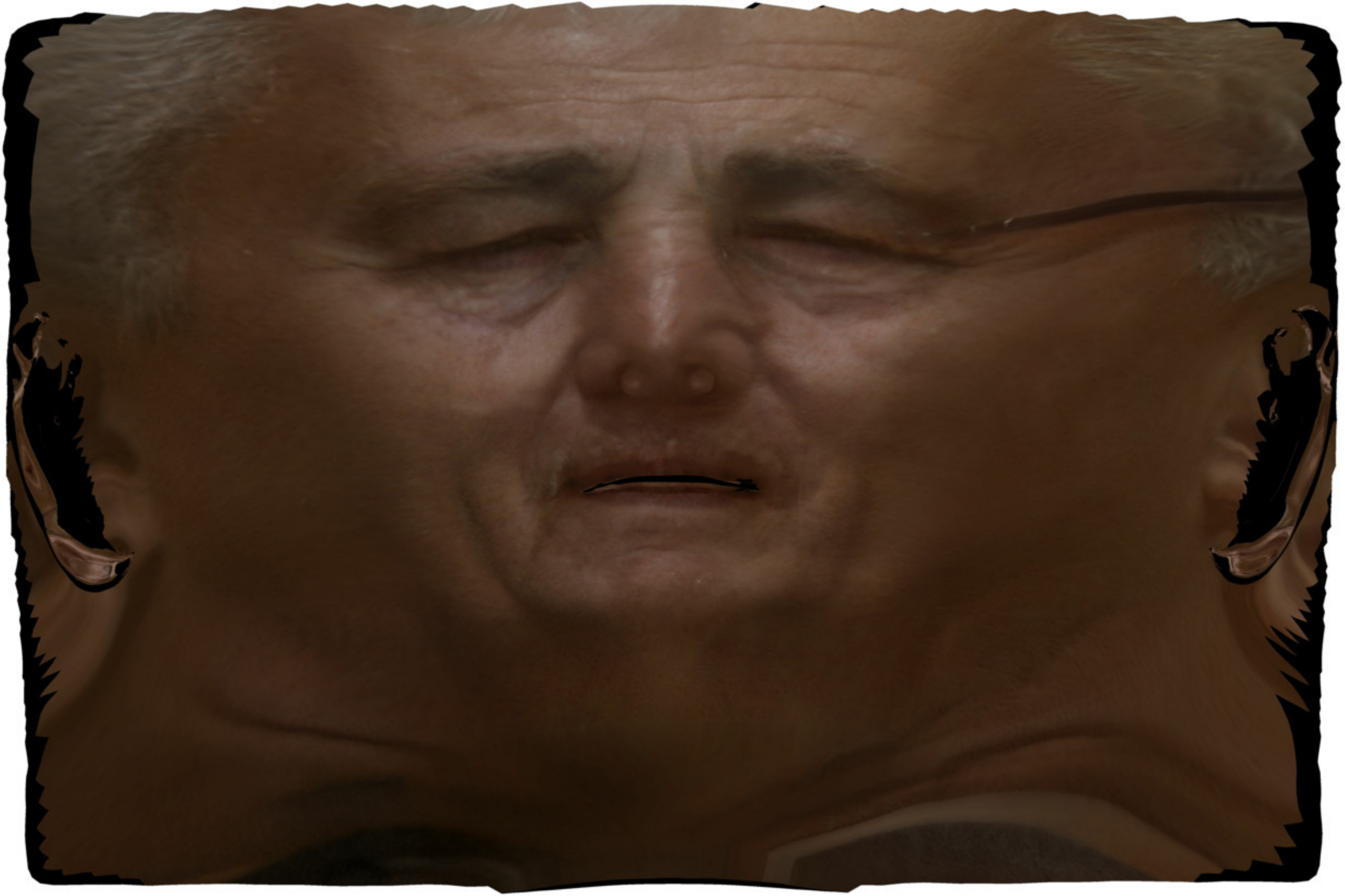}}
    \subfloat{
        \includegraphics[width=0.24\linewidth]{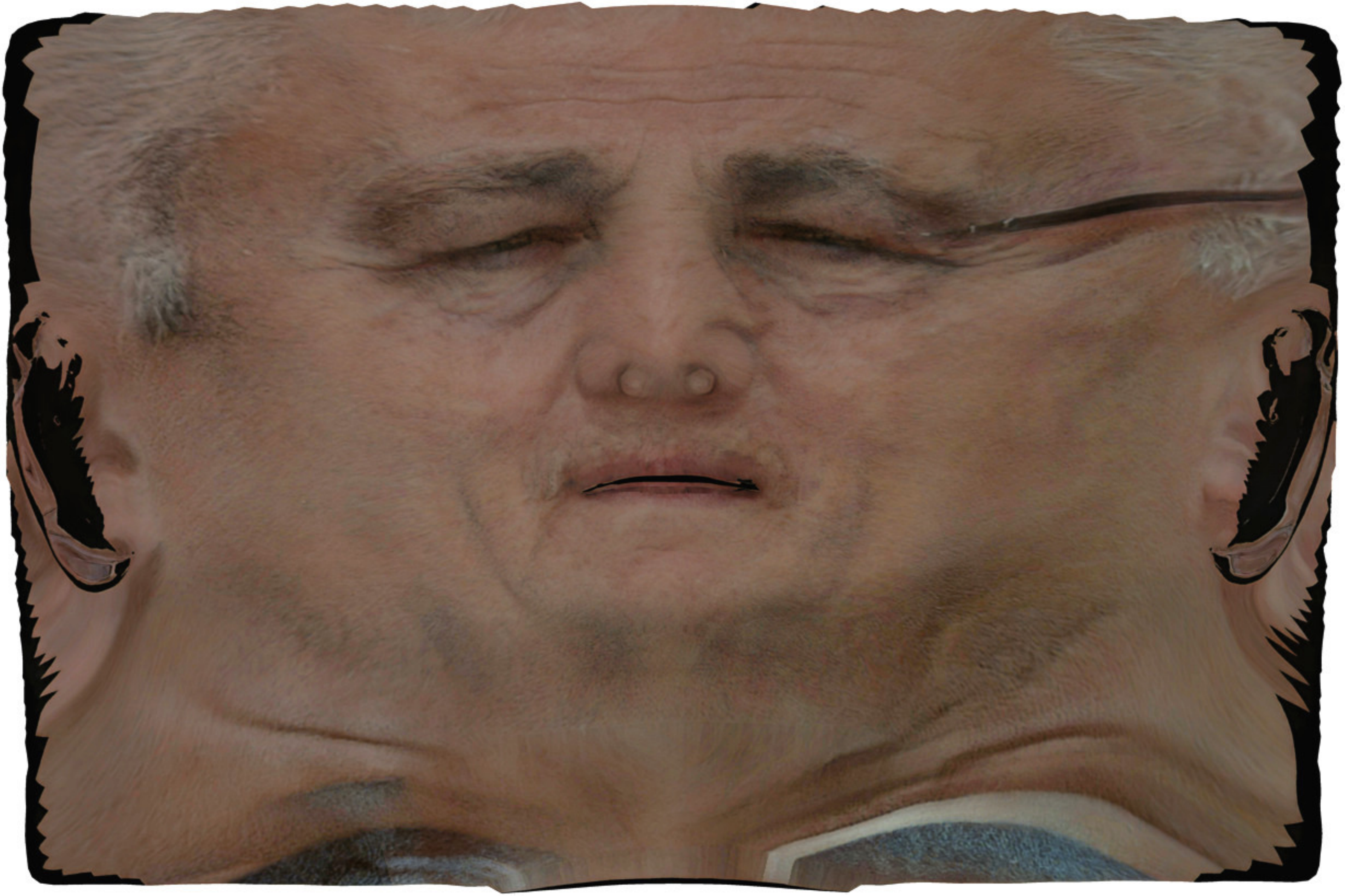}}
    \subfloat{
        \includegraphics[width=0.24\linewidth]{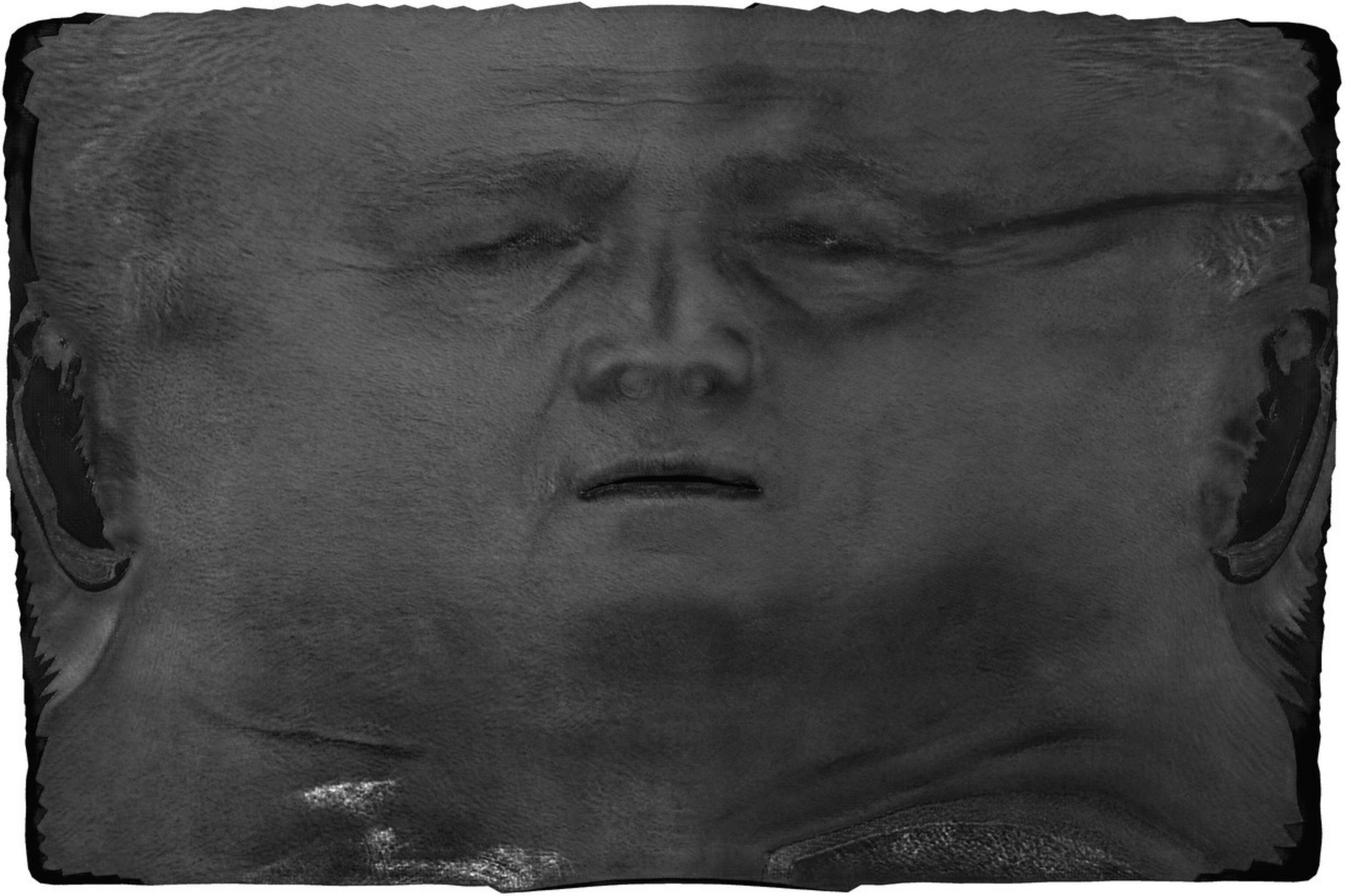}}
    \subfloat{
        \includegraphics[width=0.24\linewidth]{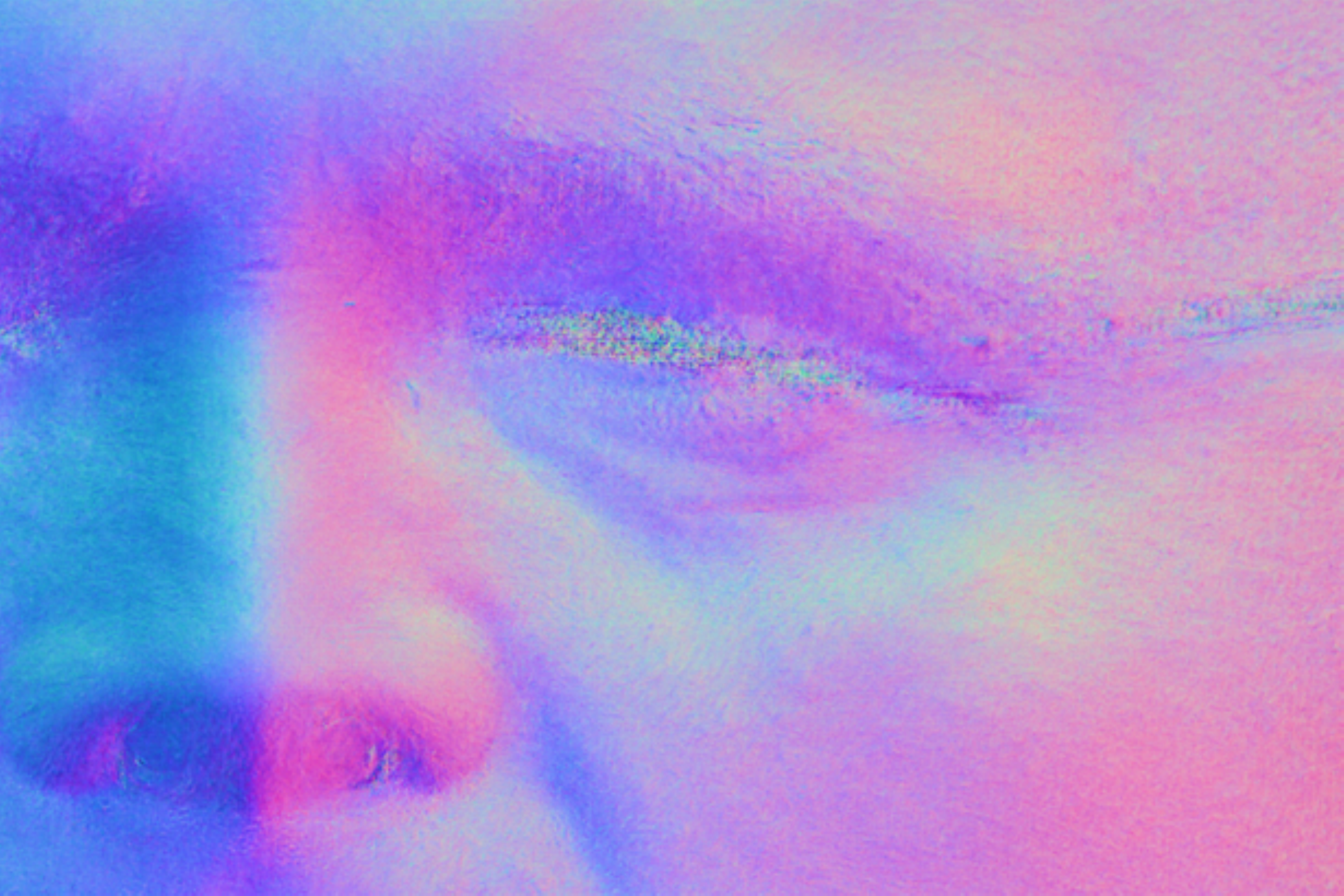}}
        
    \subfloat{
        \includegraphics[width=0.24\linewidth]{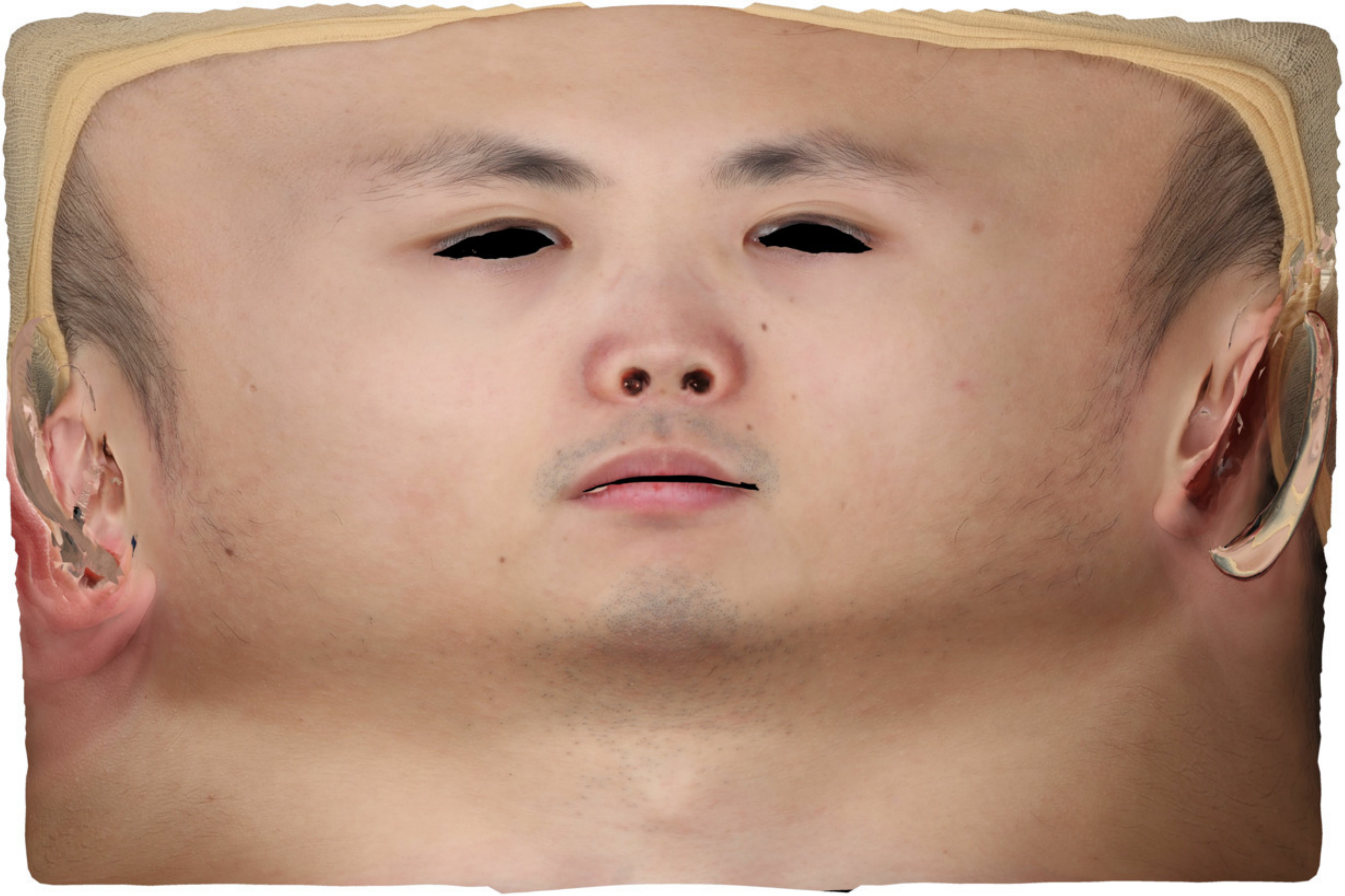}}
    \subfloat{
        \includegraphics[width=0.24\linewidth]{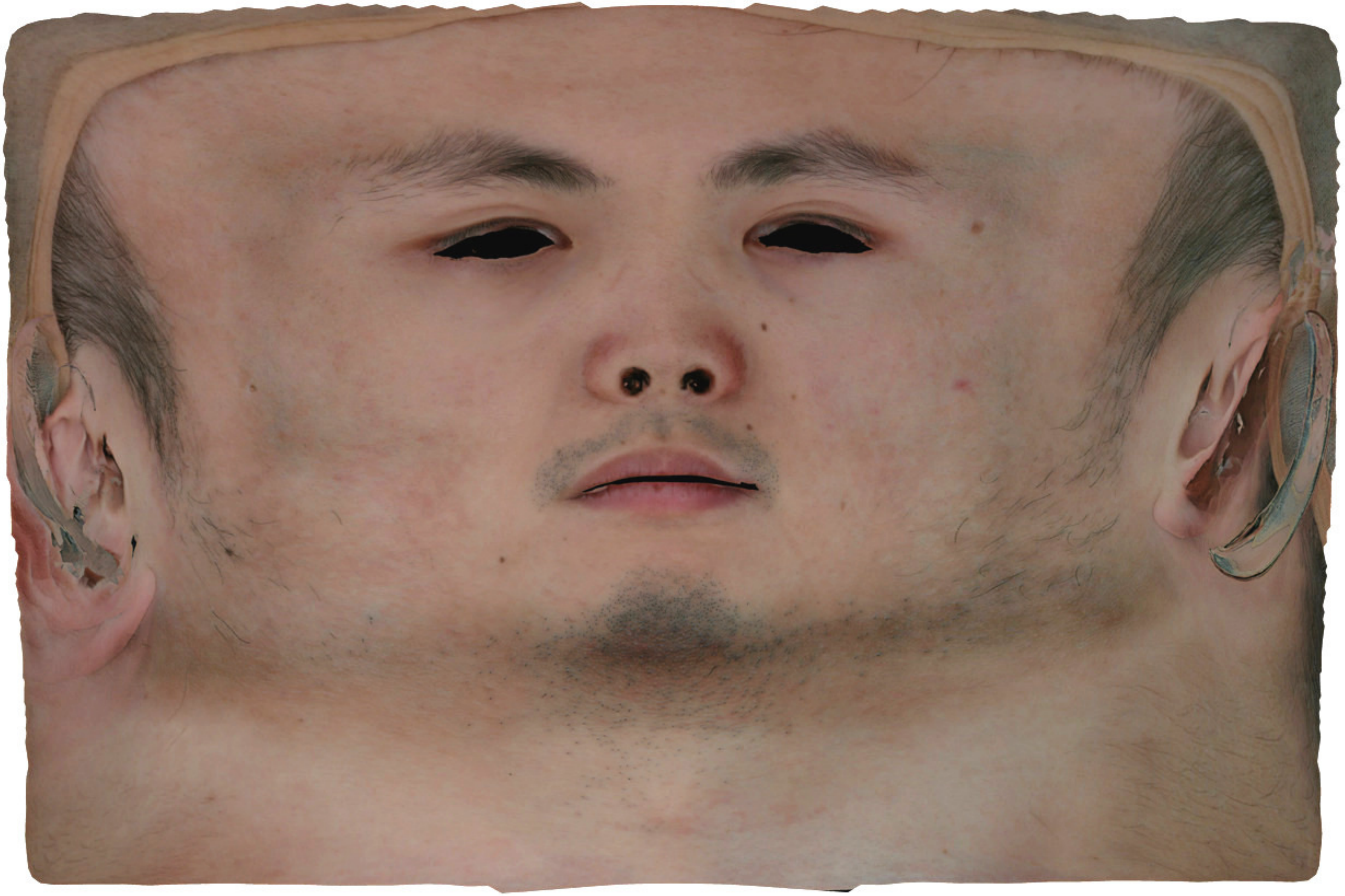}}
    \subfloat{
        \includegraphics[width=0.24\linewidth]{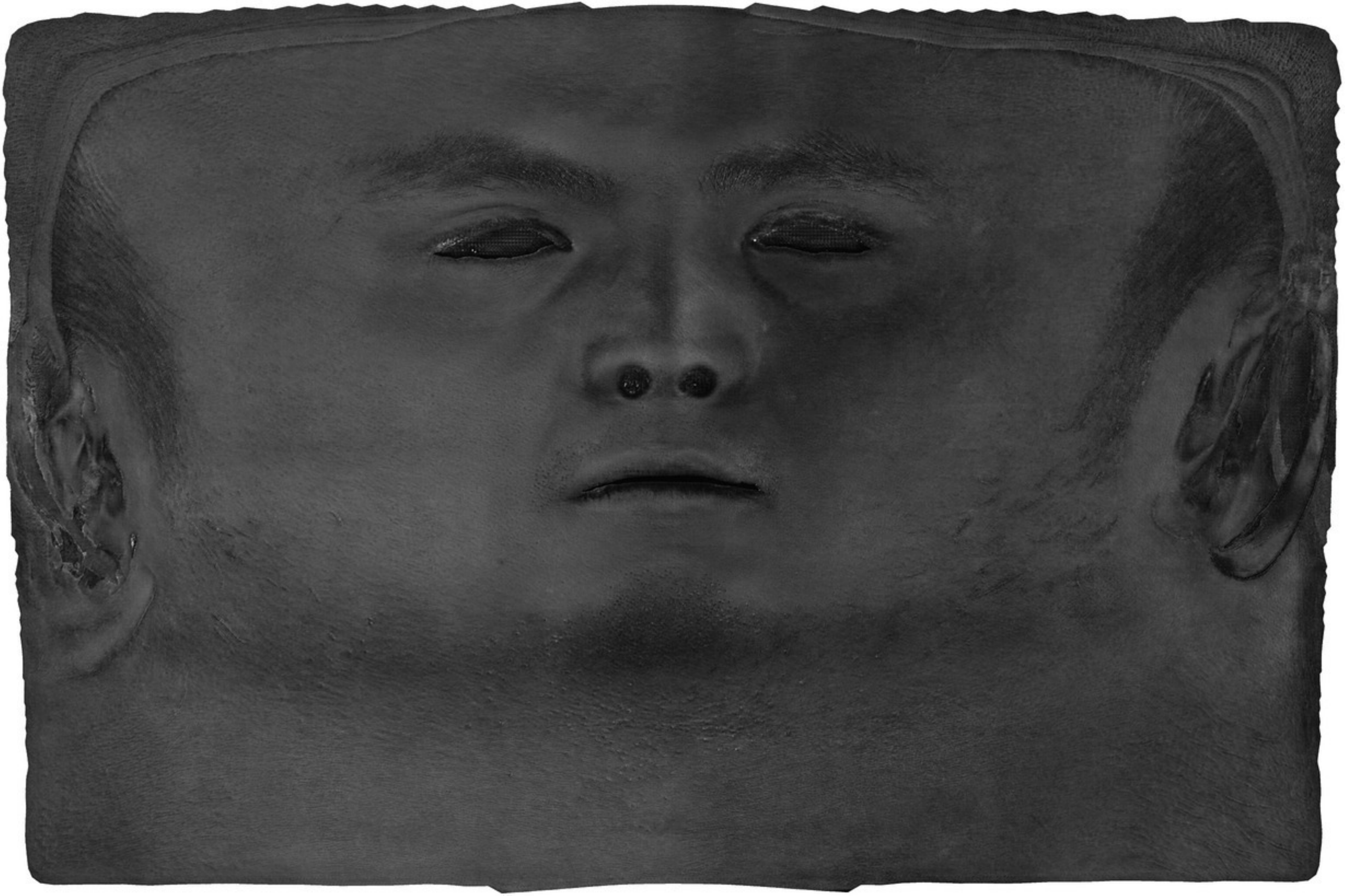}}
    \subfloat{
        \includegraphics[width=0.24\linewidth]{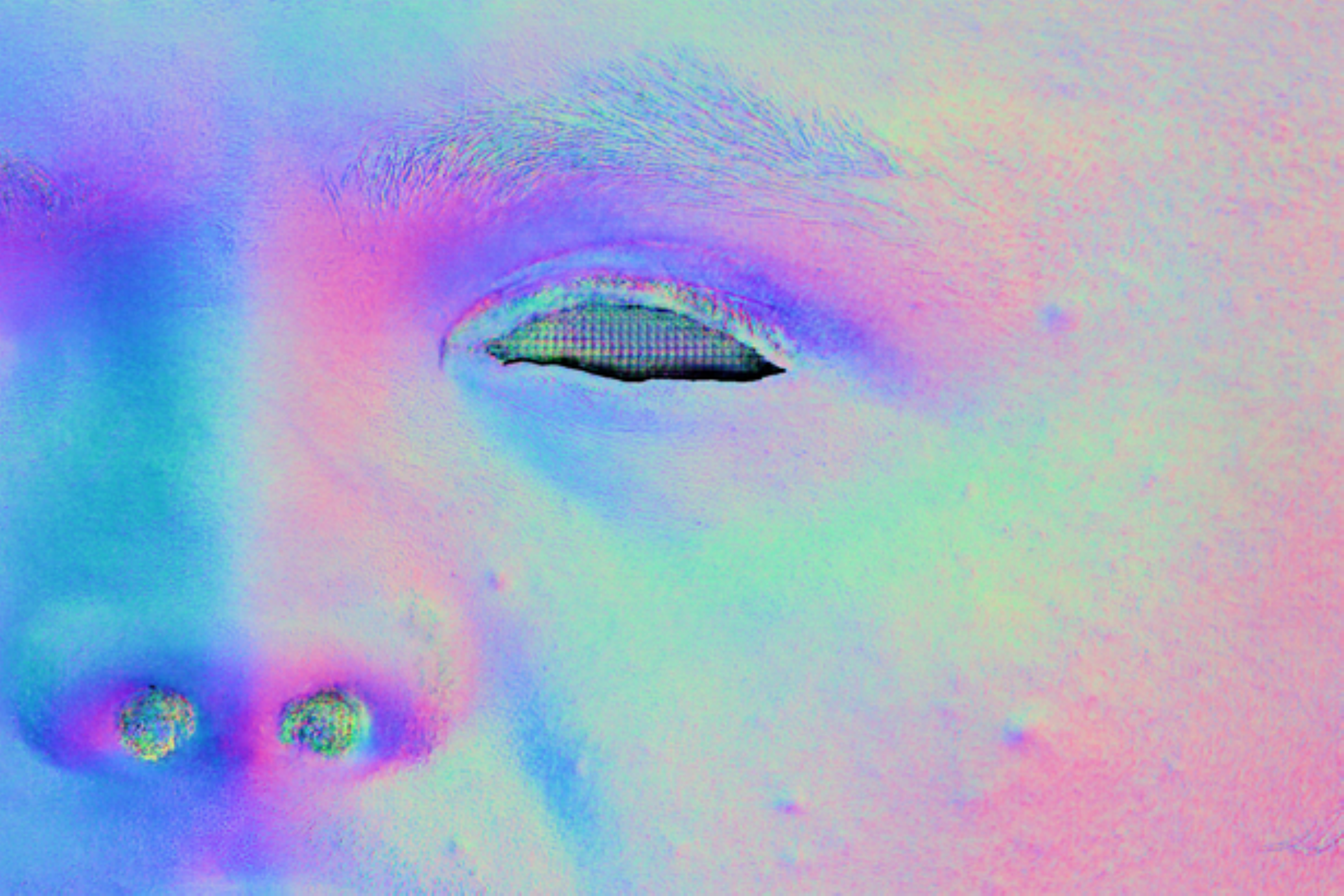}}   
        
    \subfloat{
        \includegraphics[width=0.24\linewidth]{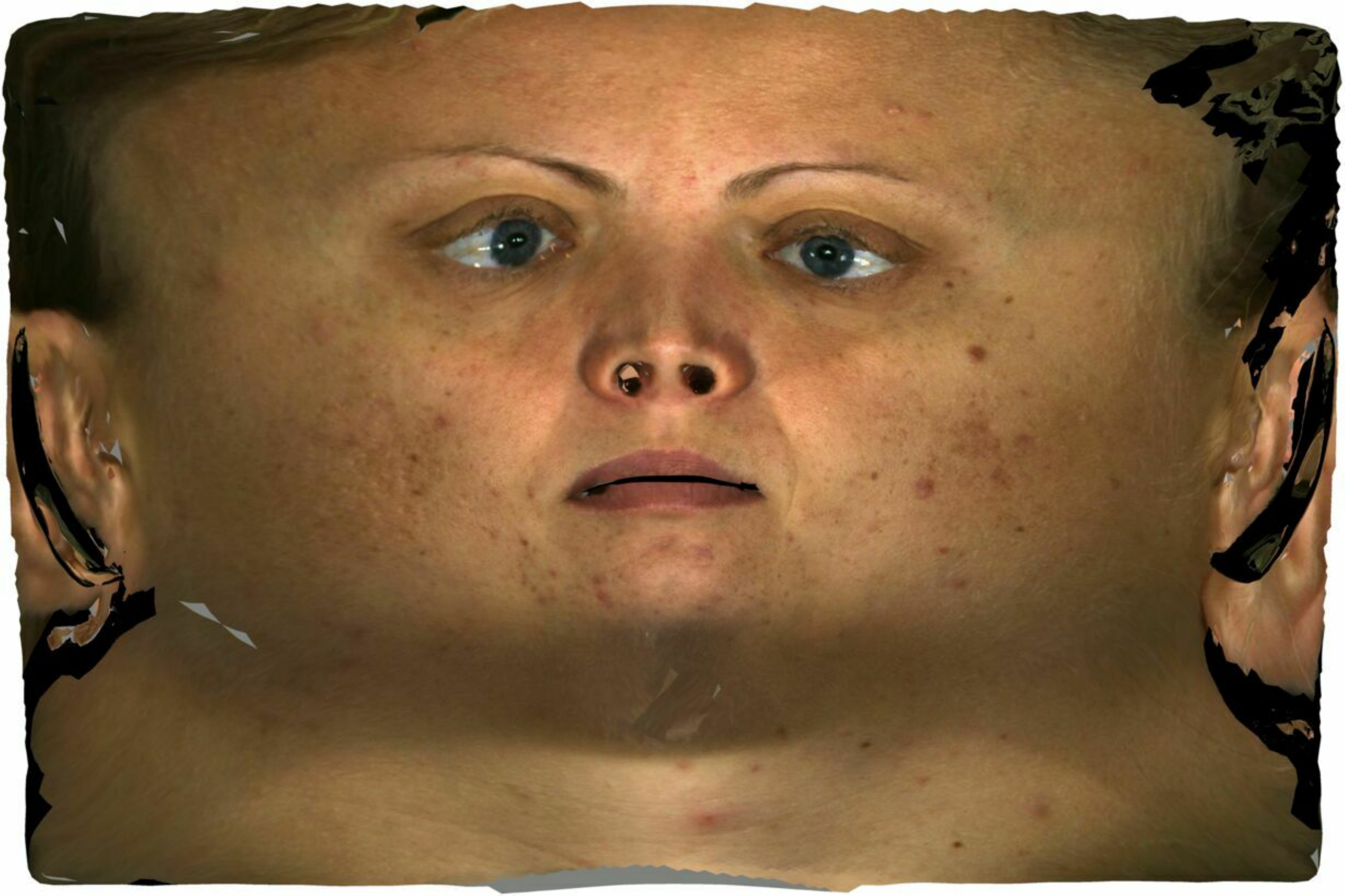}}
    \subfloat{
        \includegraphics[width=0.24\linewidth]{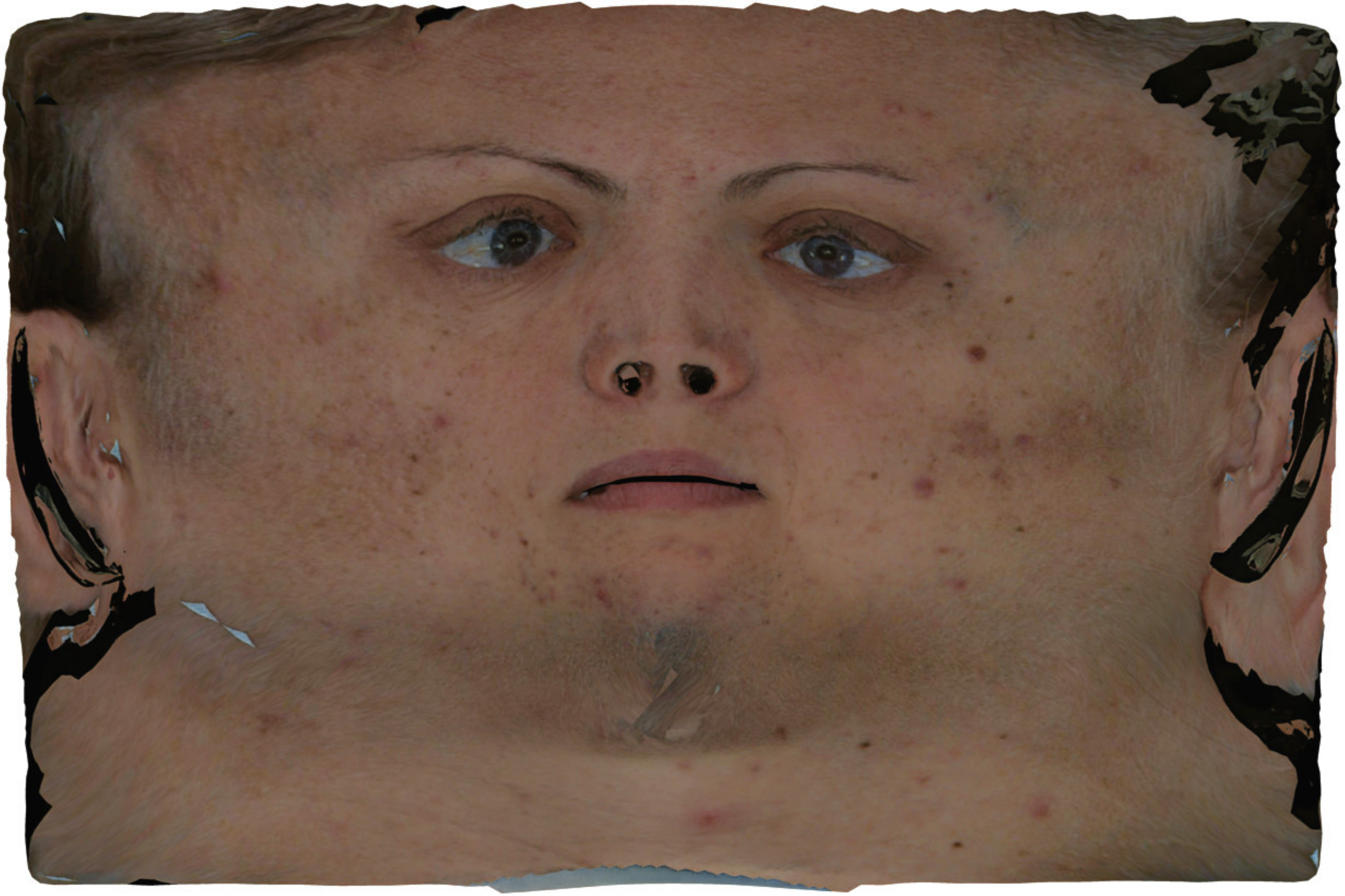}}
    \subfloat{
        \includegraphics[width=0.24\linewidth]{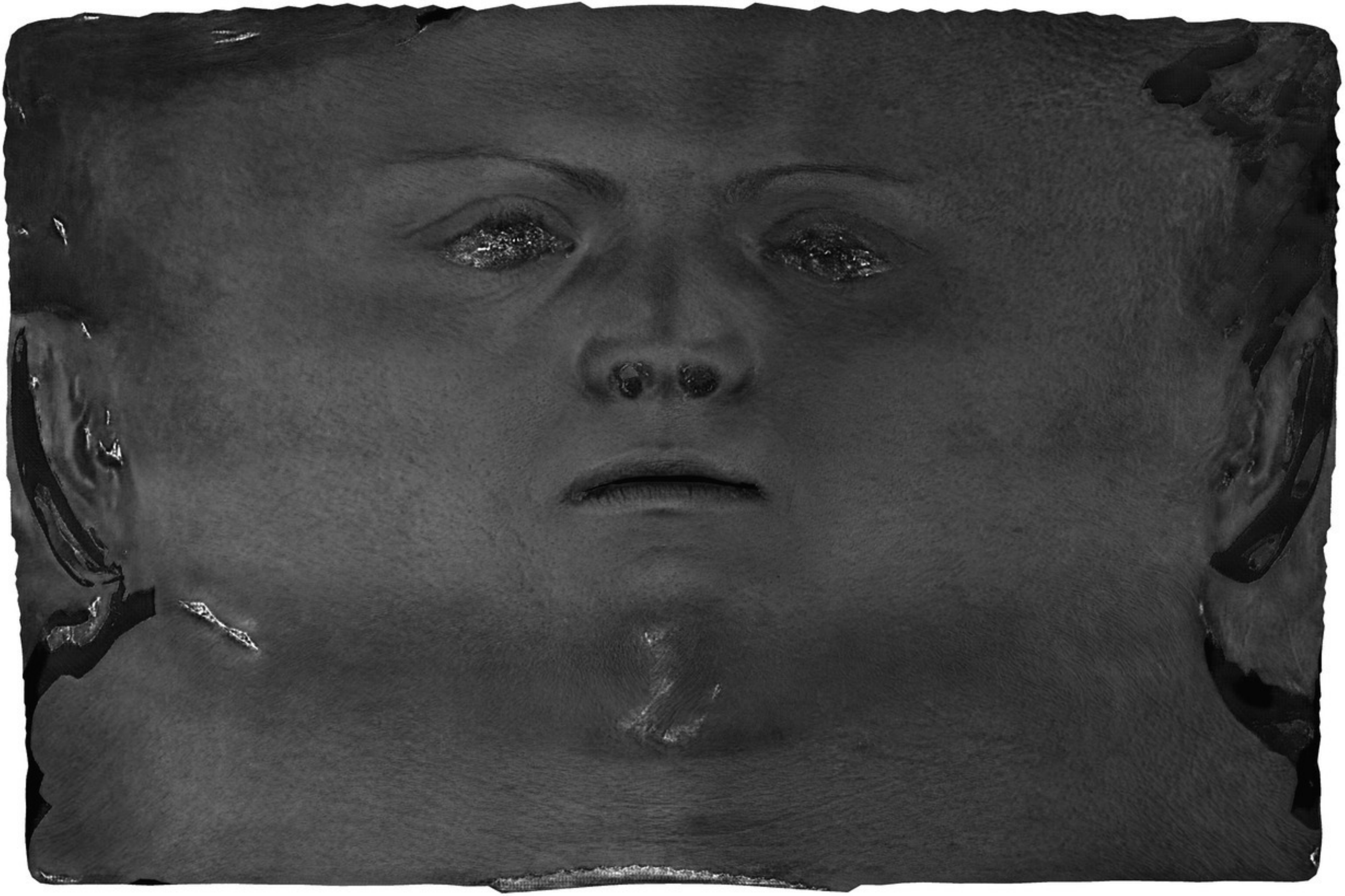}}
    \subfloat{
        \includegraphics[width=0.24\linewidth]{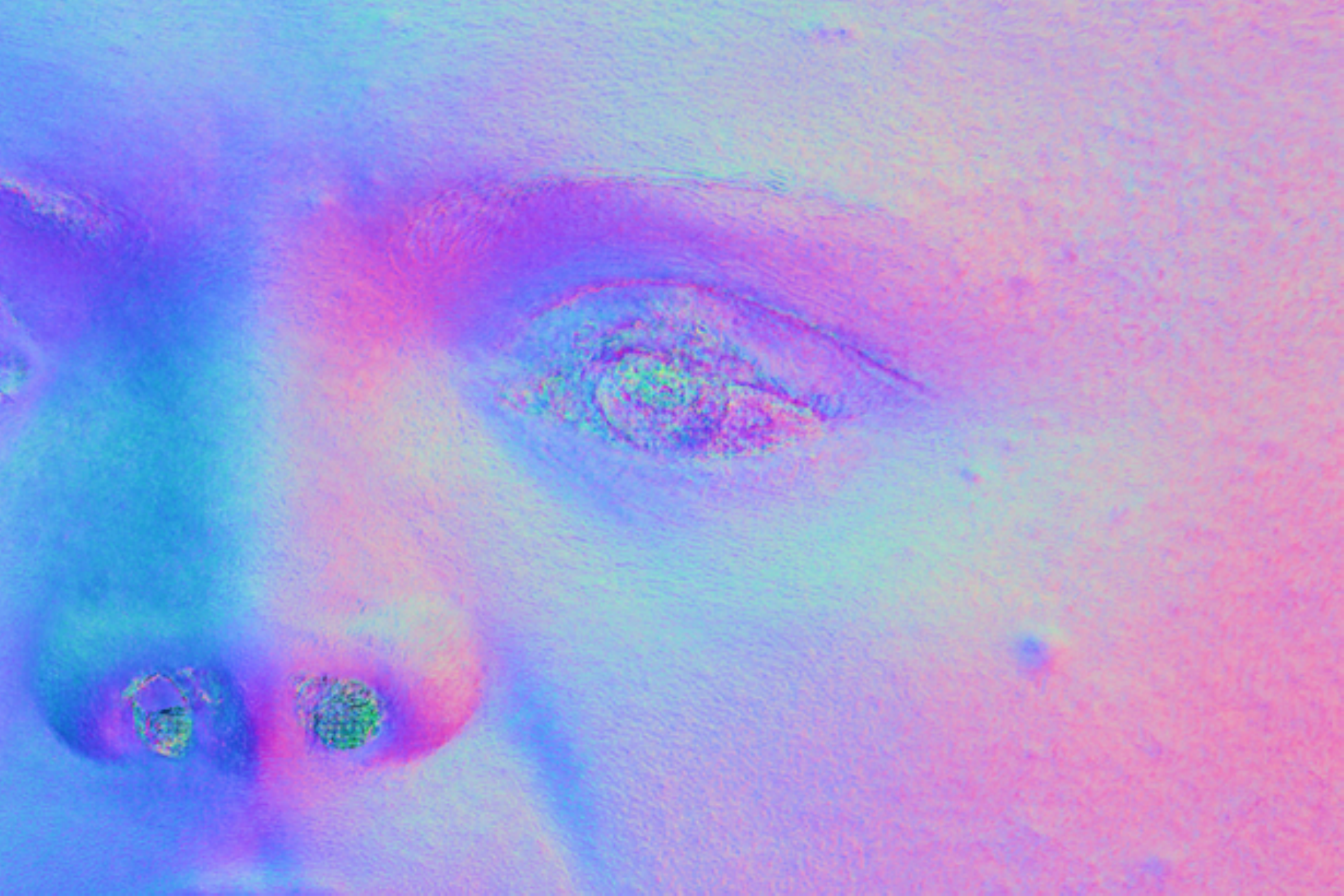}}
        
    \captionsetup[subfloat]{justification=centering}
    \subfloat[
        Input]{
        \includegraphics[width=0.24\linewidth]{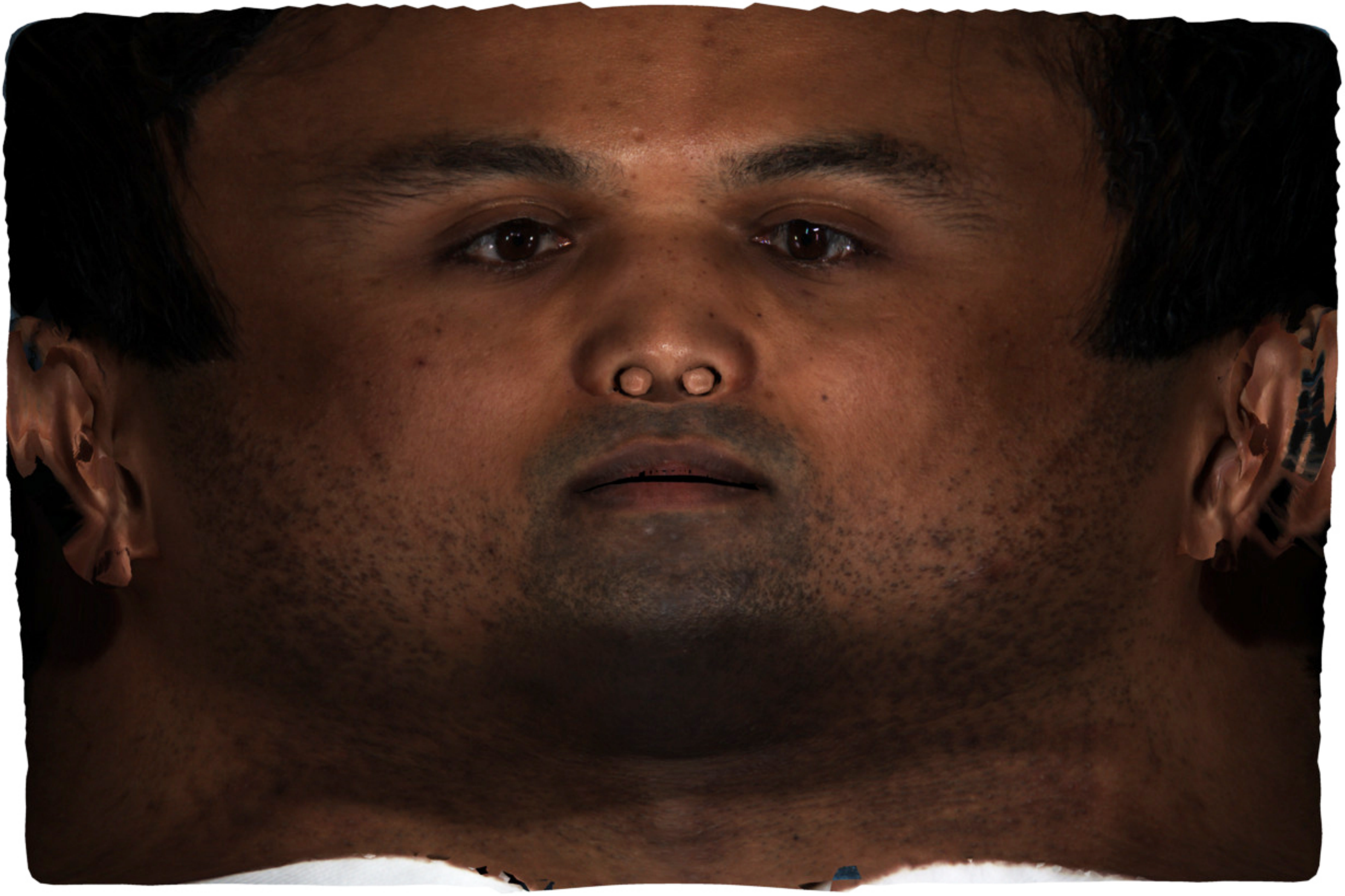}}
    \subfloat[
        Diffuse Albedo]{
        \includegraphics[width=0.24\linewidth]{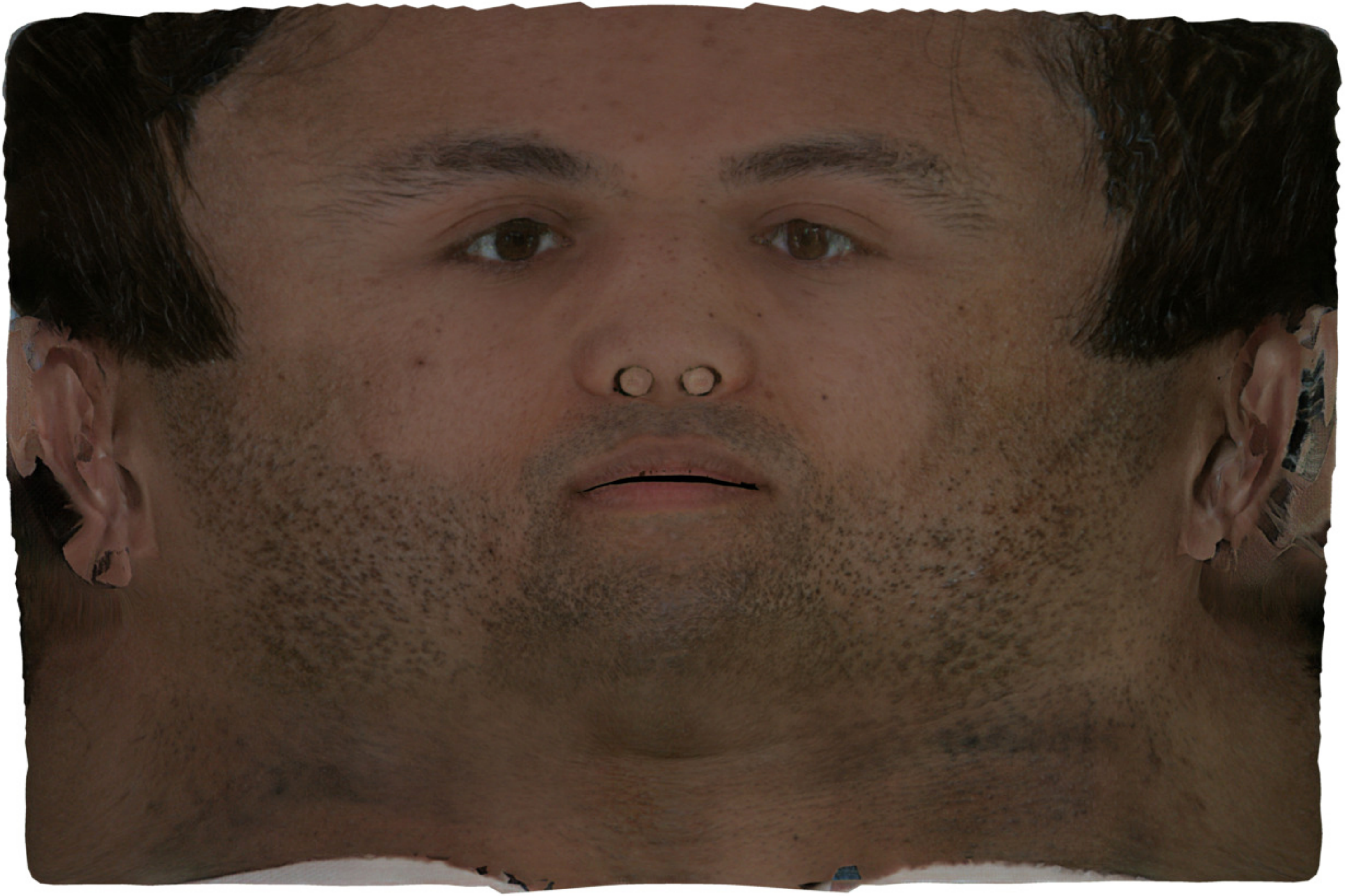}}
    \subfloat[
        Specular Albedo]{
        \includegraphics[width=0.24\linewidth]{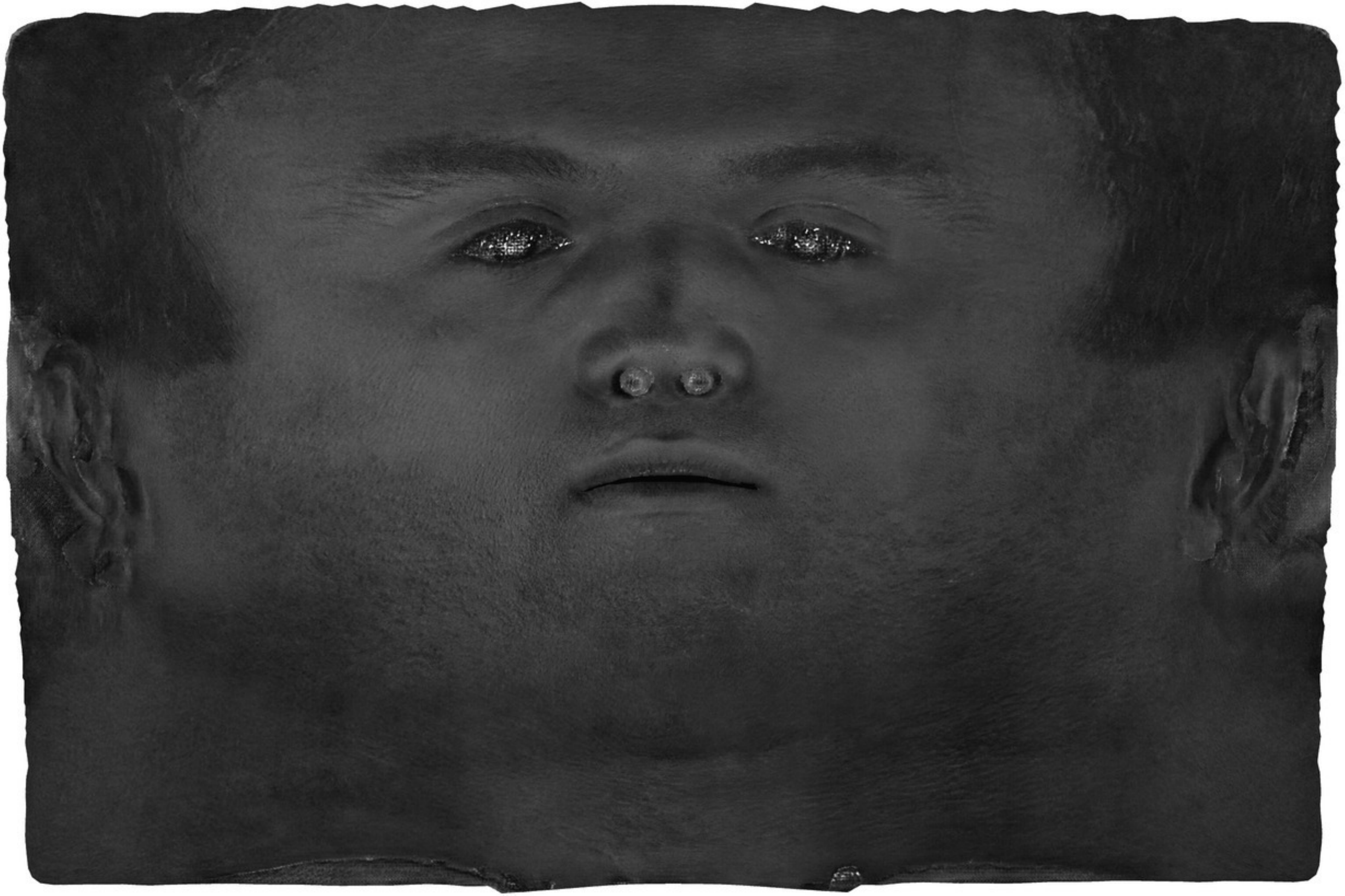}}
    \subfloat[
        Spec.~Normals]{
        \includegraphics[width=0.24\linewidth]{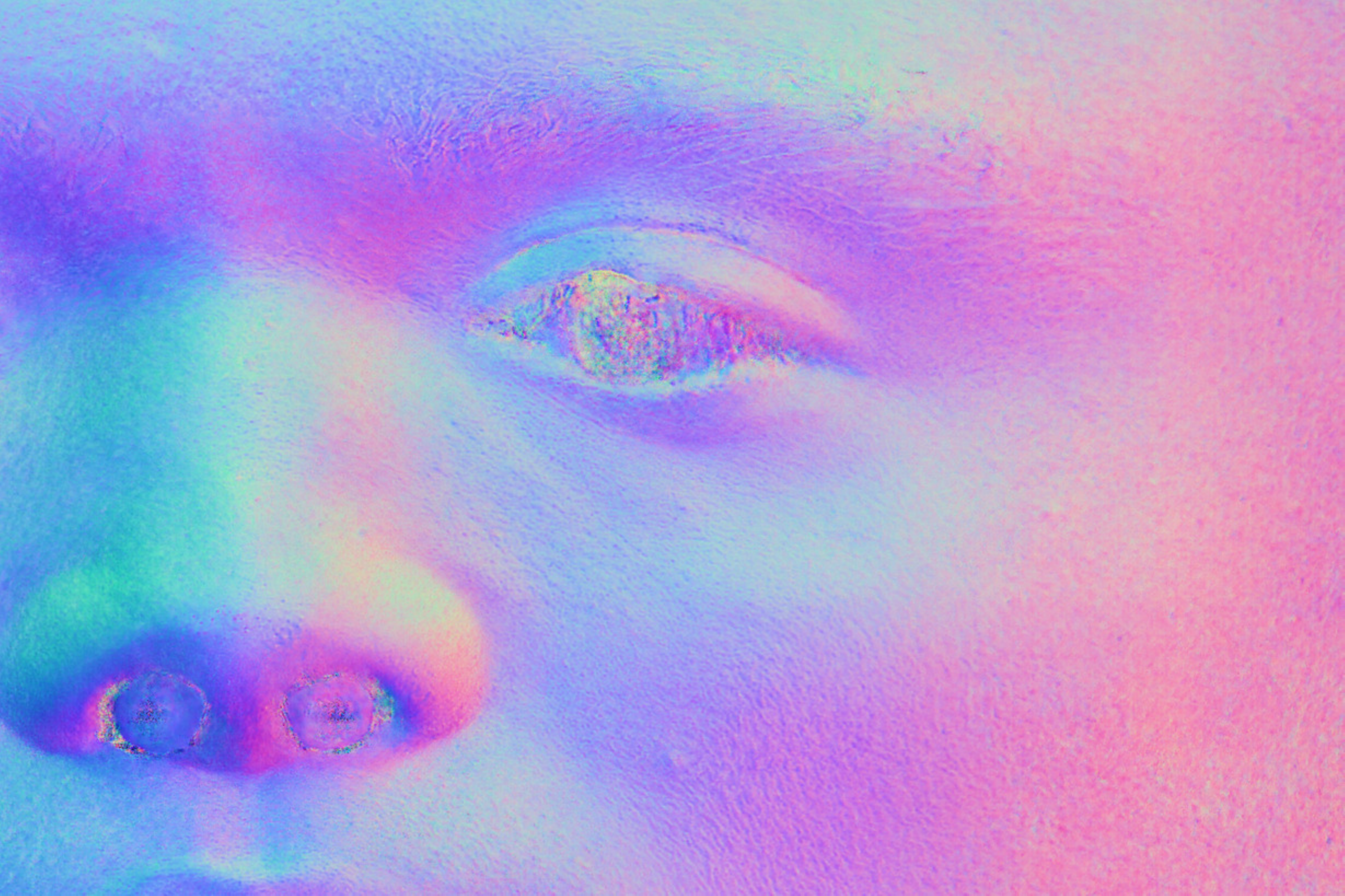}}
    
    \caption{
        Generalization of \modelnameplus{}:
        Training \modelnameplus{} with stochastically varied rendering scene parameters and our rendering loss (Sec.~\ref{sec:method_recon_renderme_renderloss}),
        makes the network domain-agnostic to an extend.
        Here we show results of a single \modelnameplus{} network,
        on reconstructions and captured data, registered to our topology. From top to bottom:
        a) Reconstructed subject, with Facial Details Synthesis proxy and texture \cite{chen_photo-realistic_2019},
        b) Reconstructed subject with OSTeC fitting and texture completion \cite{gecer2020ostec},
        c) Captured subject from FaceScape \cite{yang_facescape_2020},
        d) Captured subject from Superface \cite{DBLP:conf/eccv/BerrettiBP12},
        e) Captured subject with 3dMDface (3dmd.com).
    }
    \label{fig:results_others}
\end{figure}

\subsection{Ablation}
\label{sec:experiments_abblation}
\subsubsection{Method Components and Variants}
\label{sec:method_and_test_set}

\begin{figure*}[ht]
    \centering
    \includegraphics[width=\linewidth]{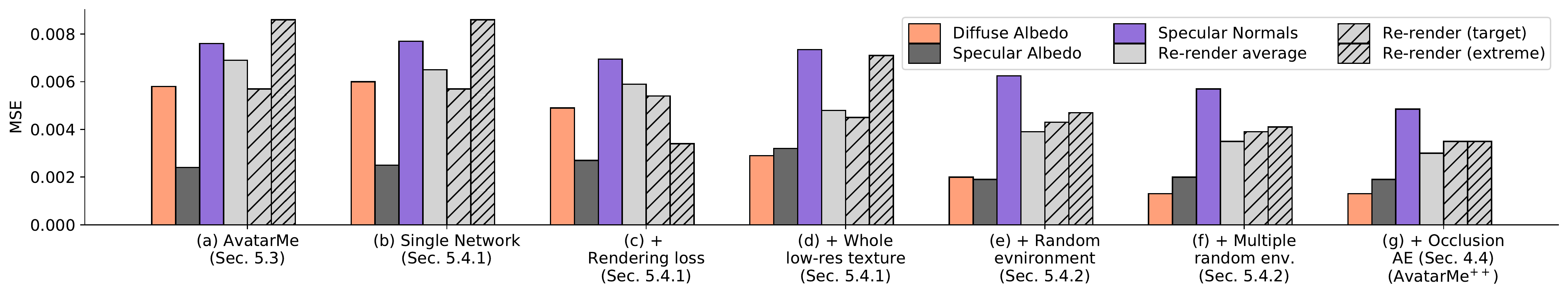}
    \vspace{-0.8cm}
    \caption{
        Quantitative ablation results for the different variations of our method.
        From left to right, we start with the base AvatarMe and introduce each component of AvatarMe\textsuperscript{++}.
        Each variation is applied on test set $\mathcal{T}$ (Sec.~\ref{sec:method_and_test_set}).
        We measure the mean squared error (MSE)
        between ground truth and prediction's relfectance, and re-rendering
        in the target environment and an extreme side illumination environment
        used in $\mathcal{T}$.
    }
    \label{fig:results_ablation_barplot}
    \subfloat{
        \includegraphics[width=0.1055\linewidth]{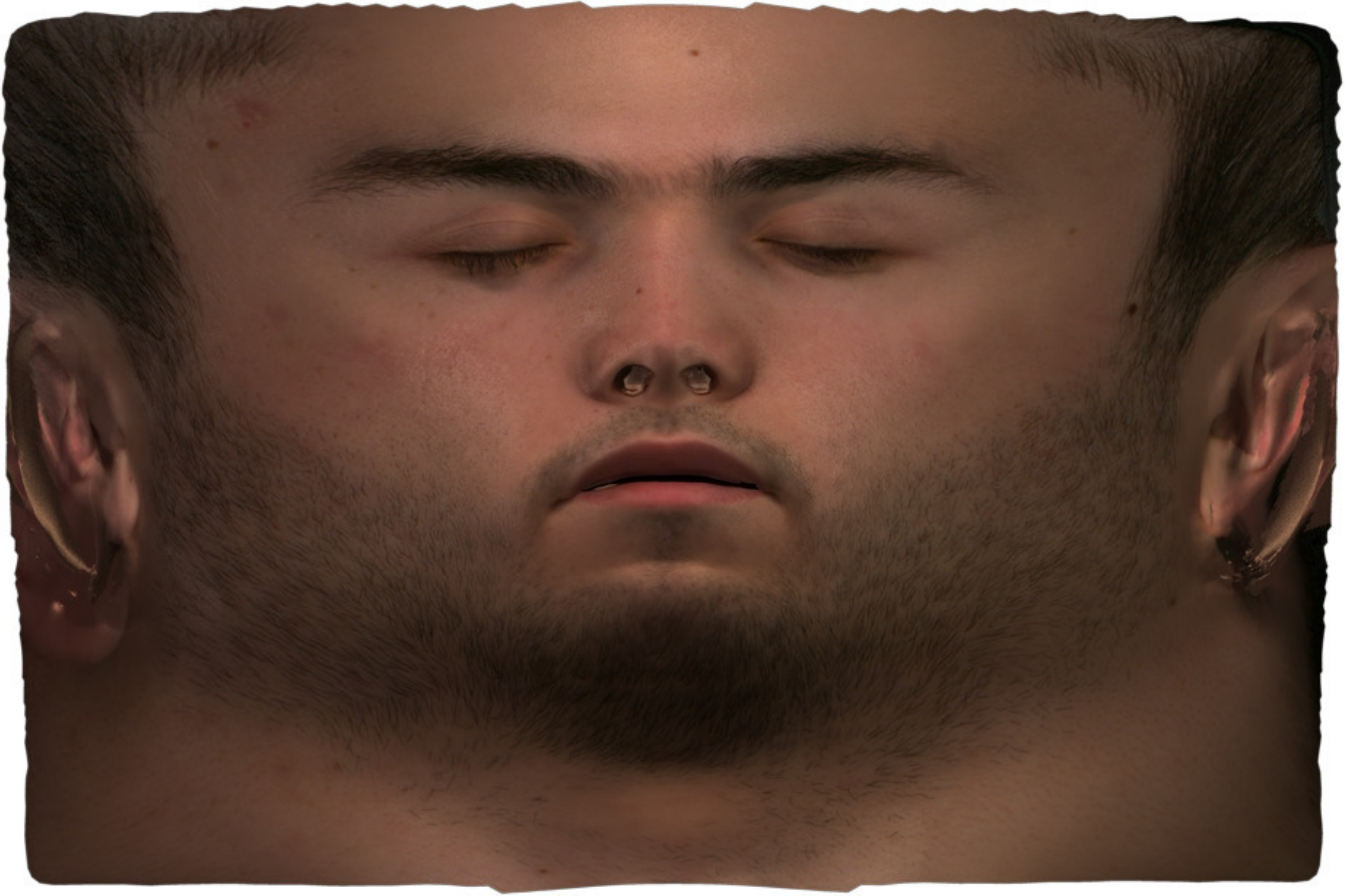}}
    \subfloat{
        \includegraphics[width=0.1055\linewidth]{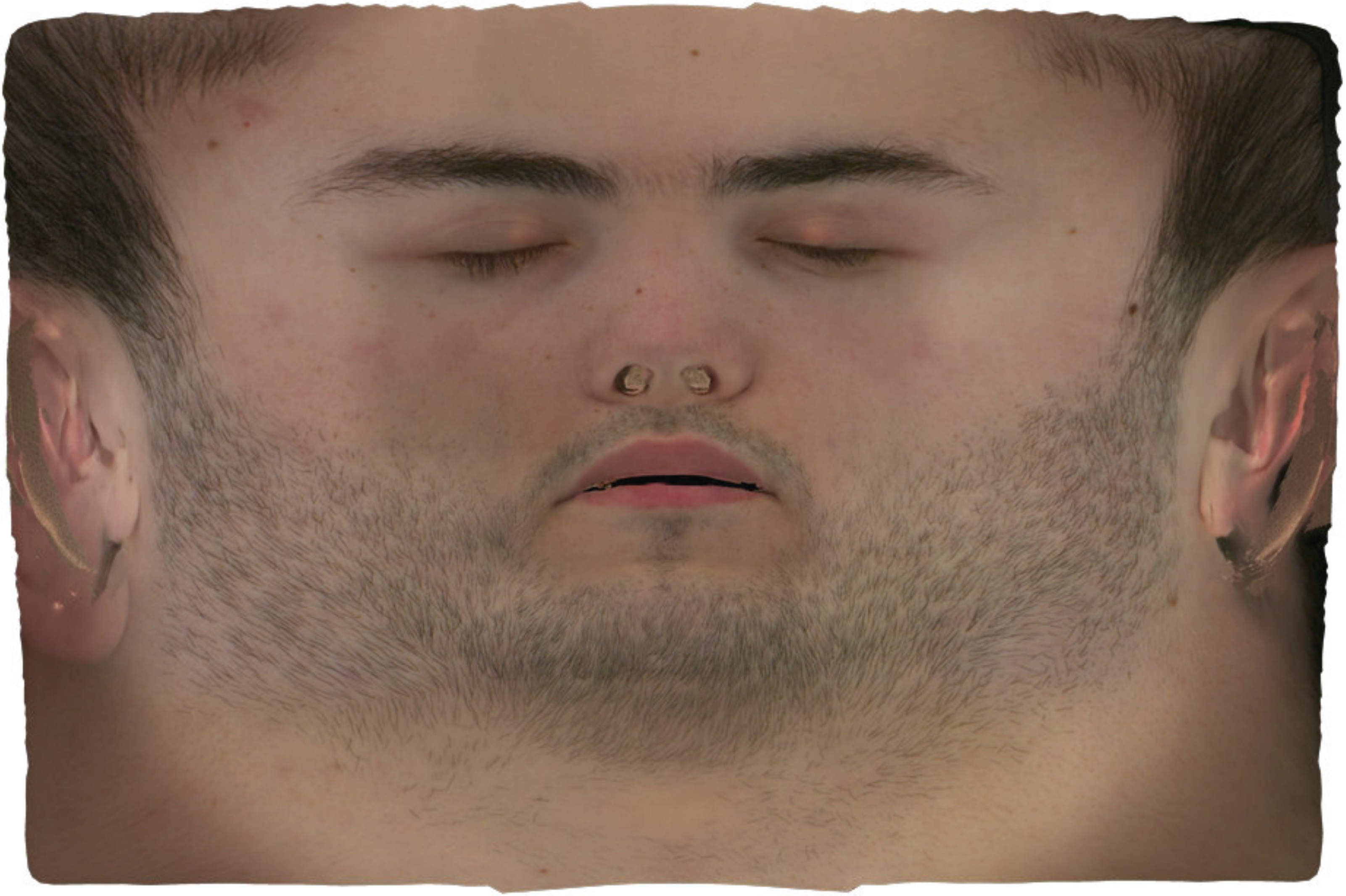}}
    \subfloat{
        \includegraphics[width=0.1055\linewidth]{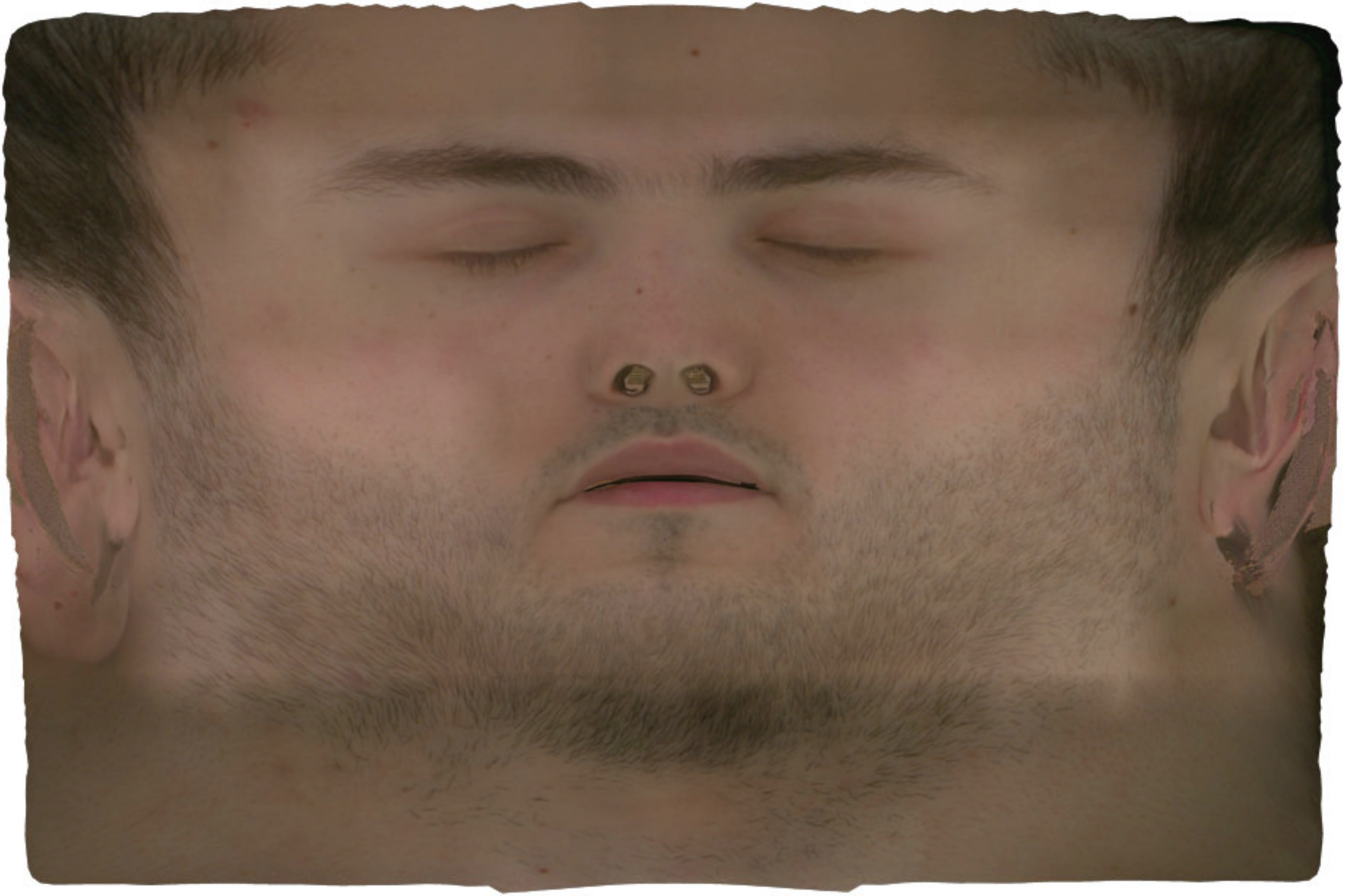}}
    \subfloat{
        \includegraphics[width=0.1055\linewidth]{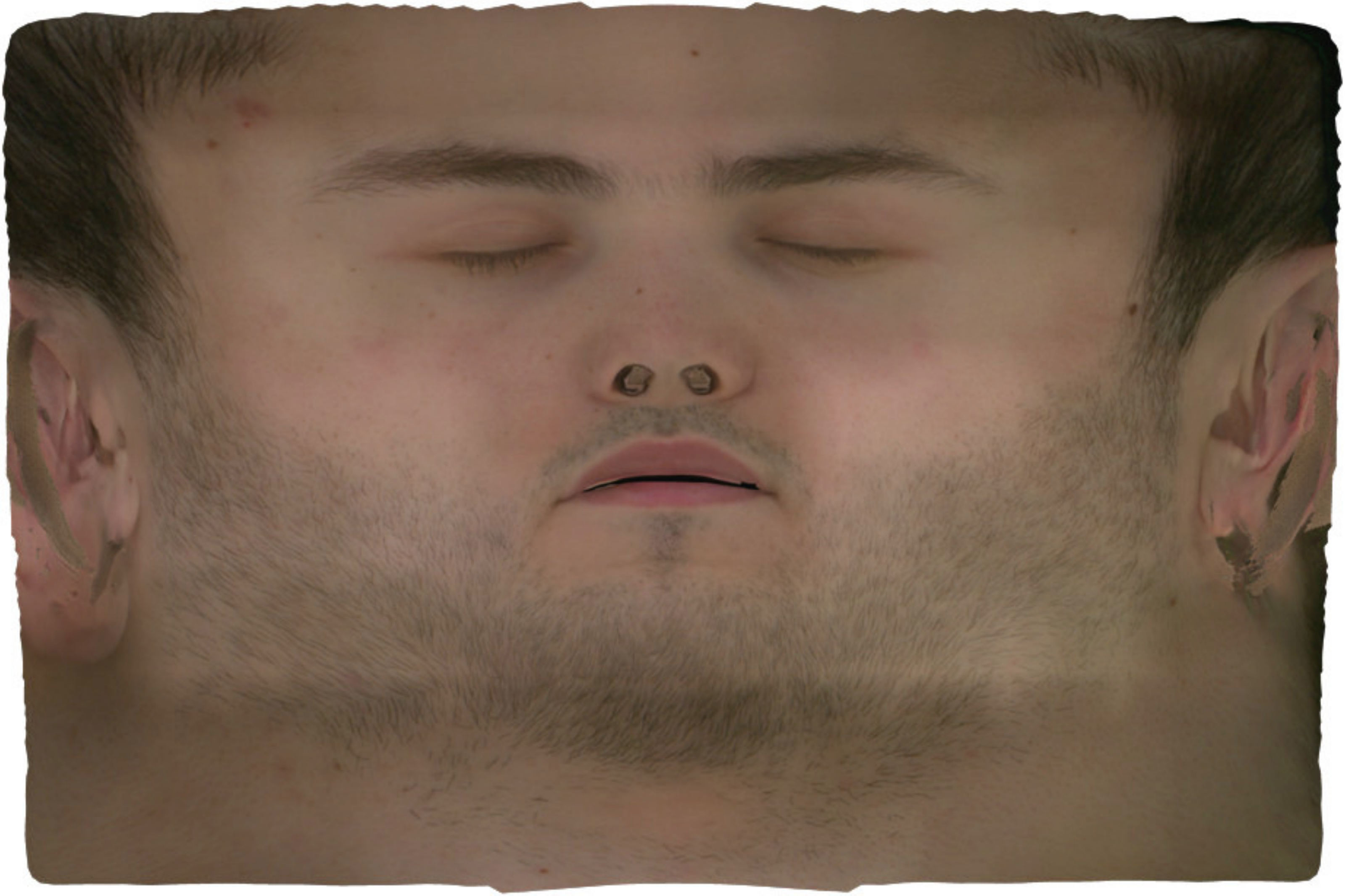}}
    \subfloat{
        \includegraphics[width=0.1055\linewidth]{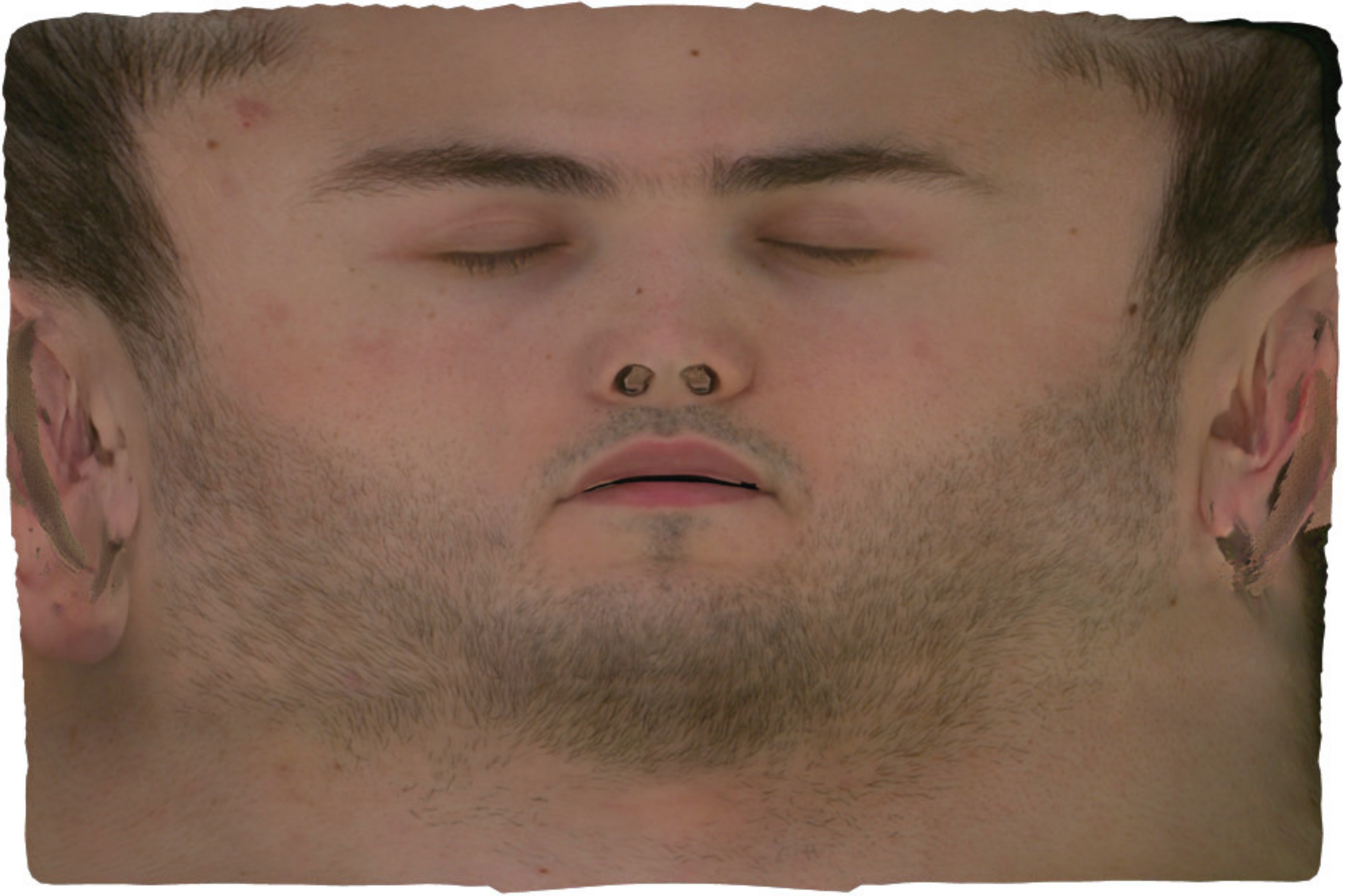}}
    \subfloat{
        \includegraphics[width=0.1055\linewidth]{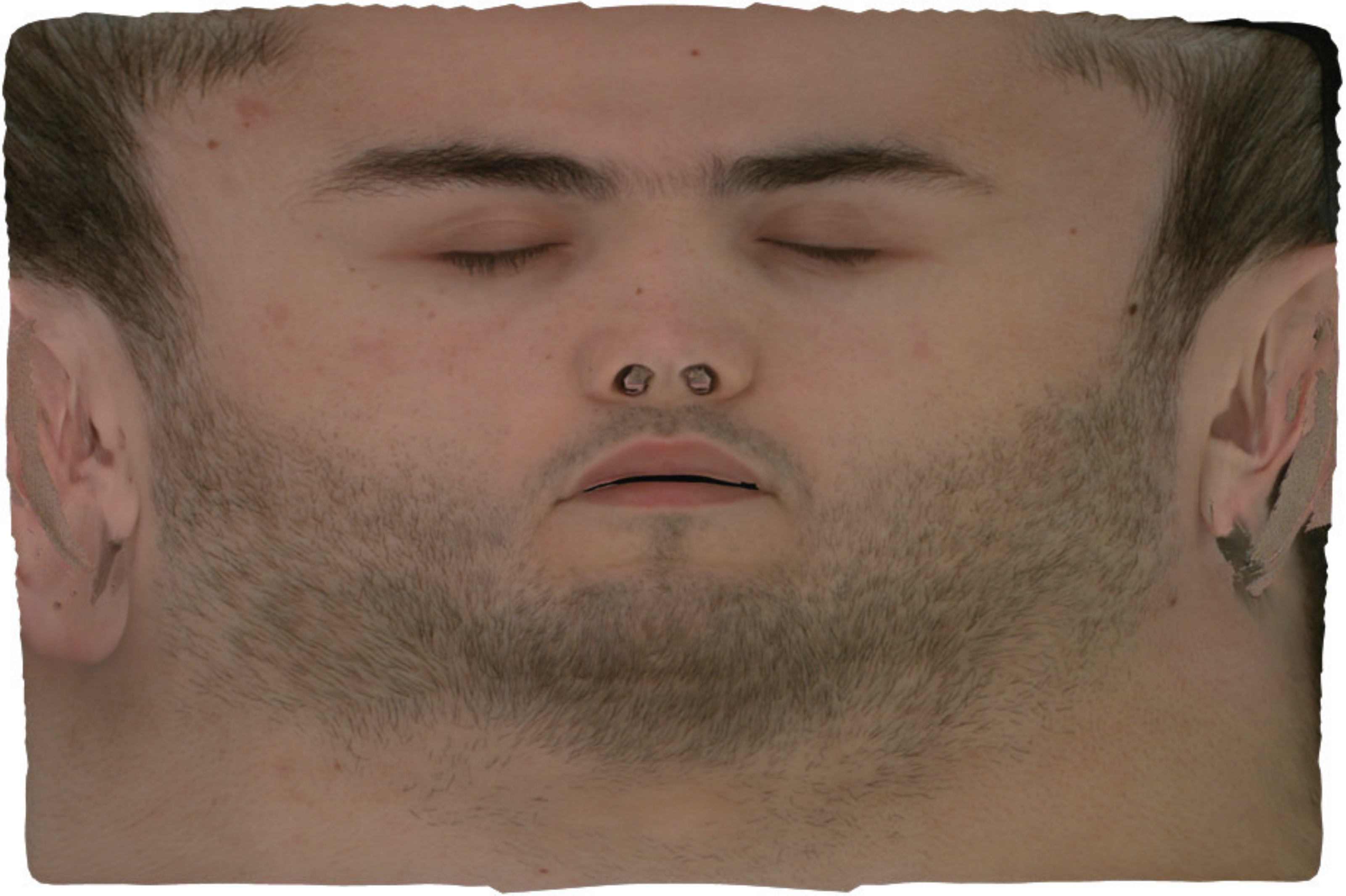}}
    \subfloat{
        \includegraphics[width=0.1055\linewidth]{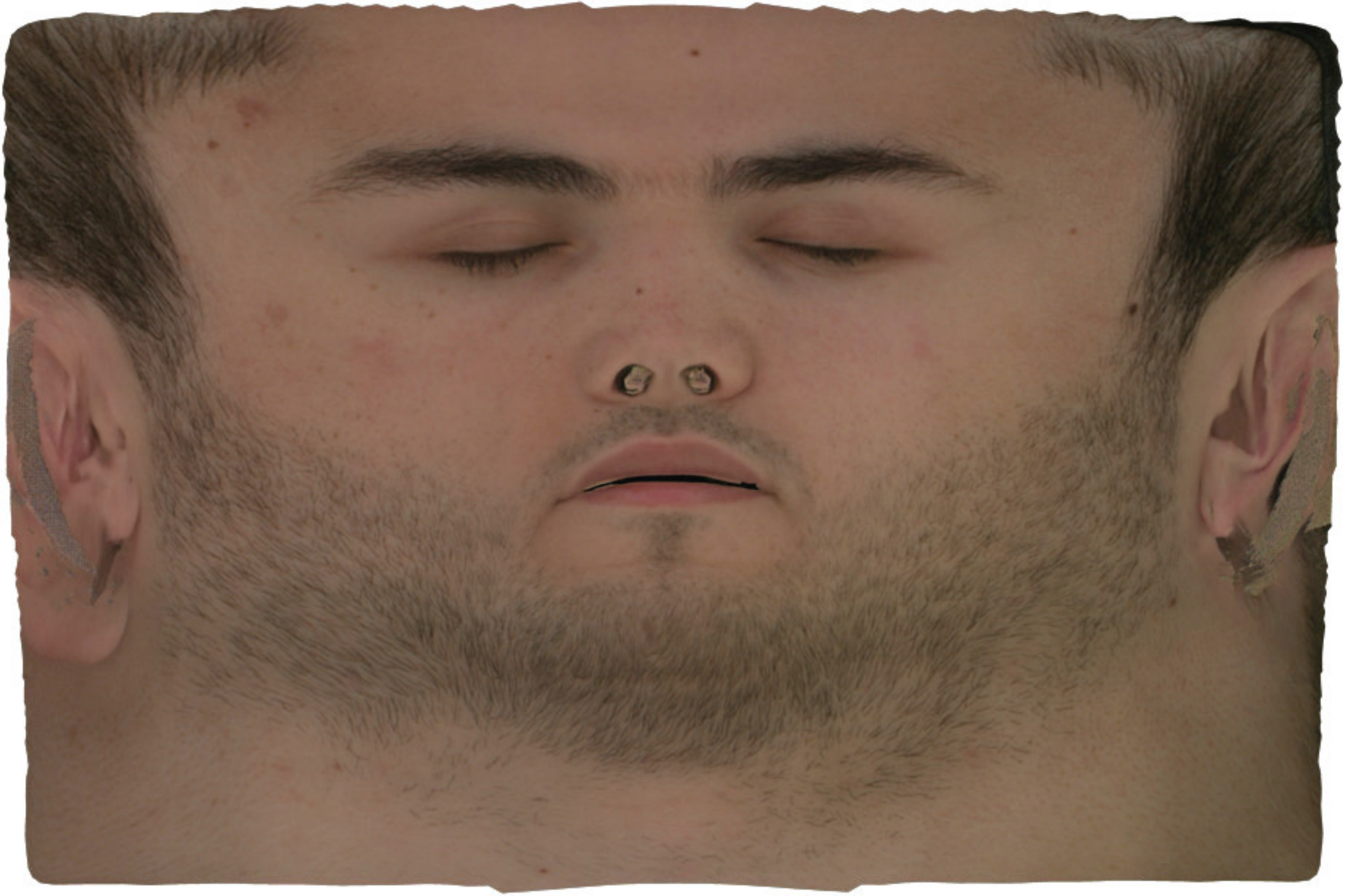}}
    \subfloat{
        \includegraphics[width=0.1055\linewidth]{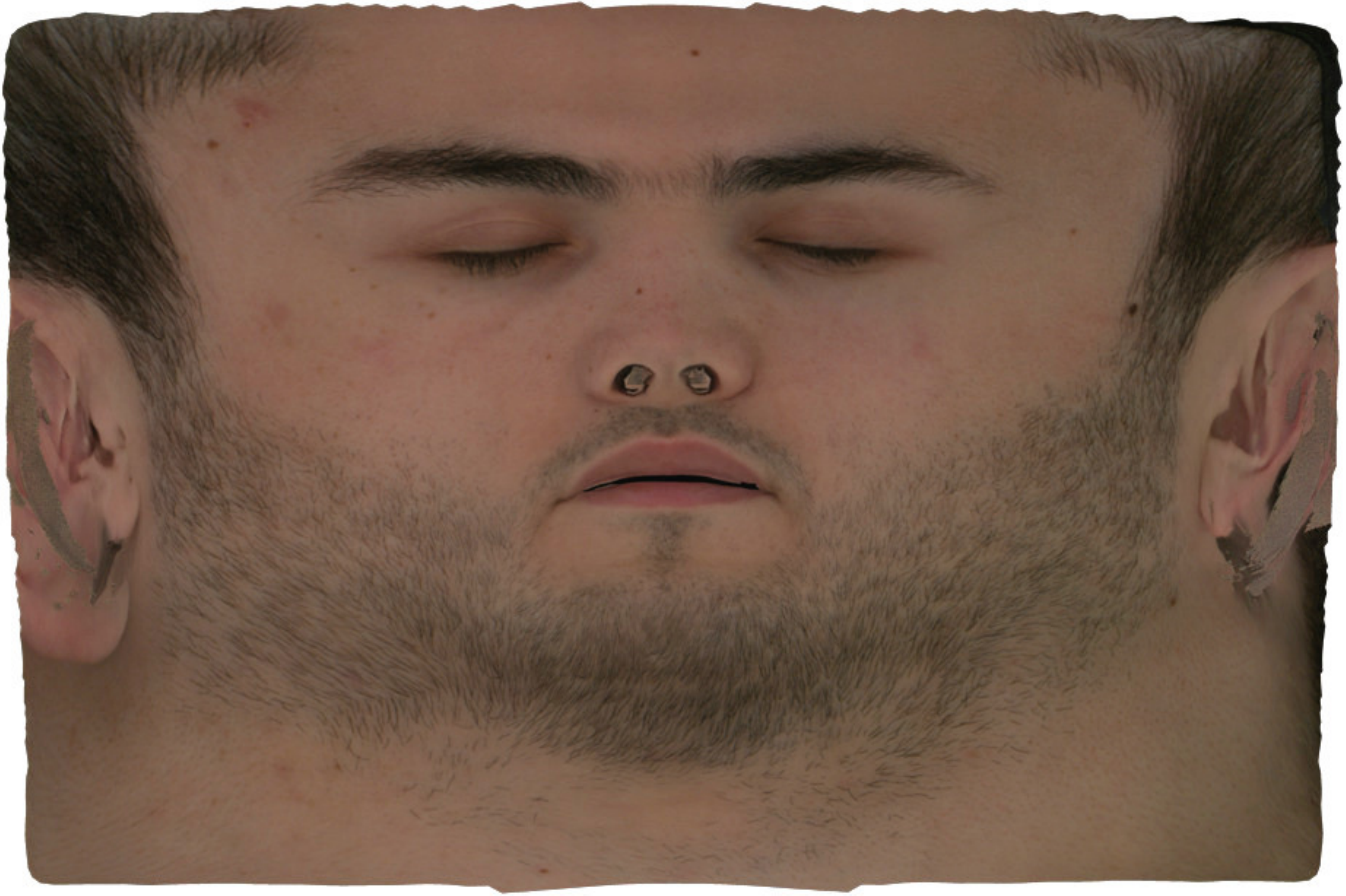}}
    \subfloat{
        \includegraphics[width=0.1055\linewidth]{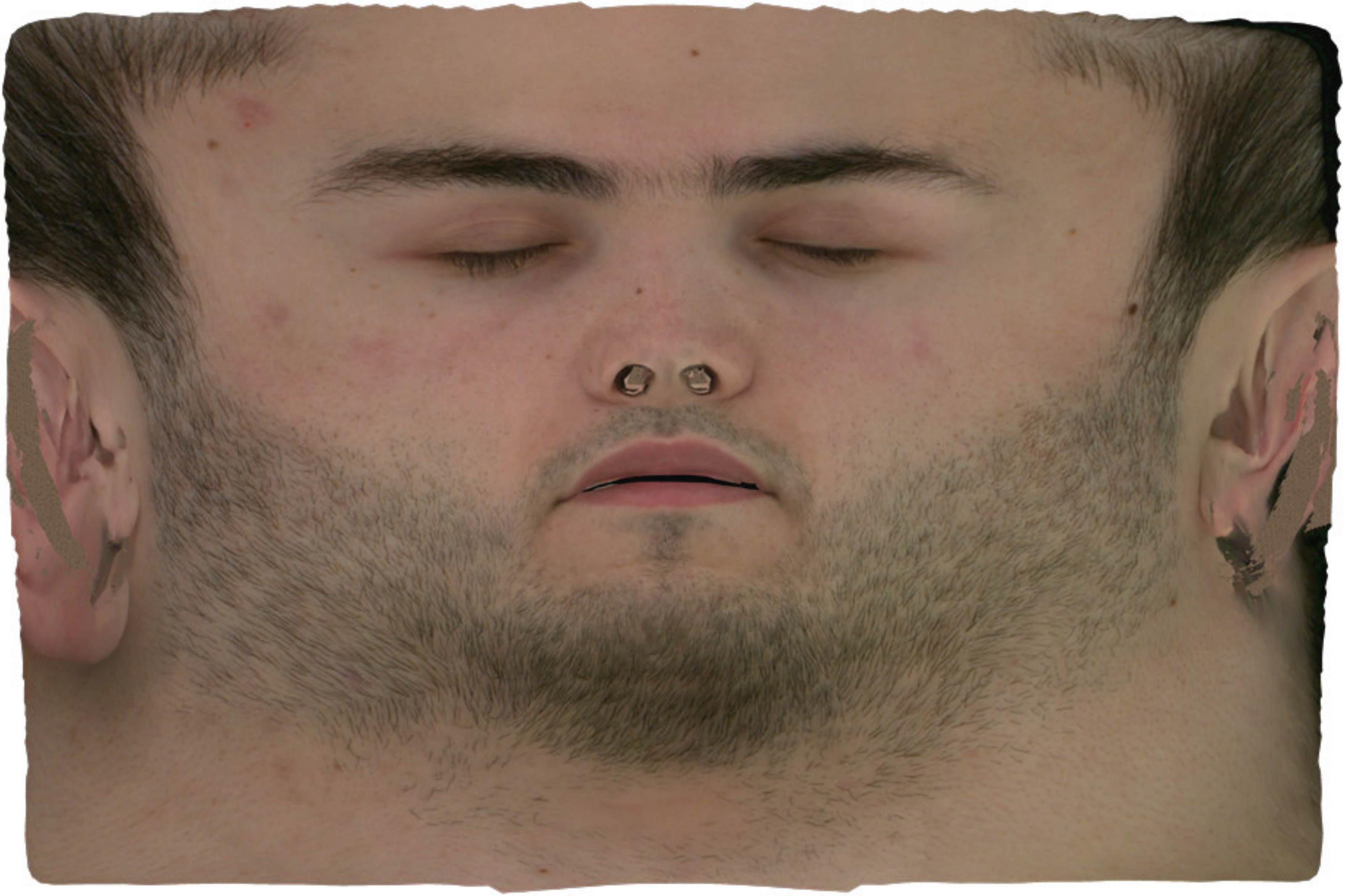}} \\
    
    \setcounter{subfigure}{0}
    \captionsetup[subfloat]{justification=centering}
    \subfloat[Input]{
        \includegraphics[width=0.1055\linewidth]{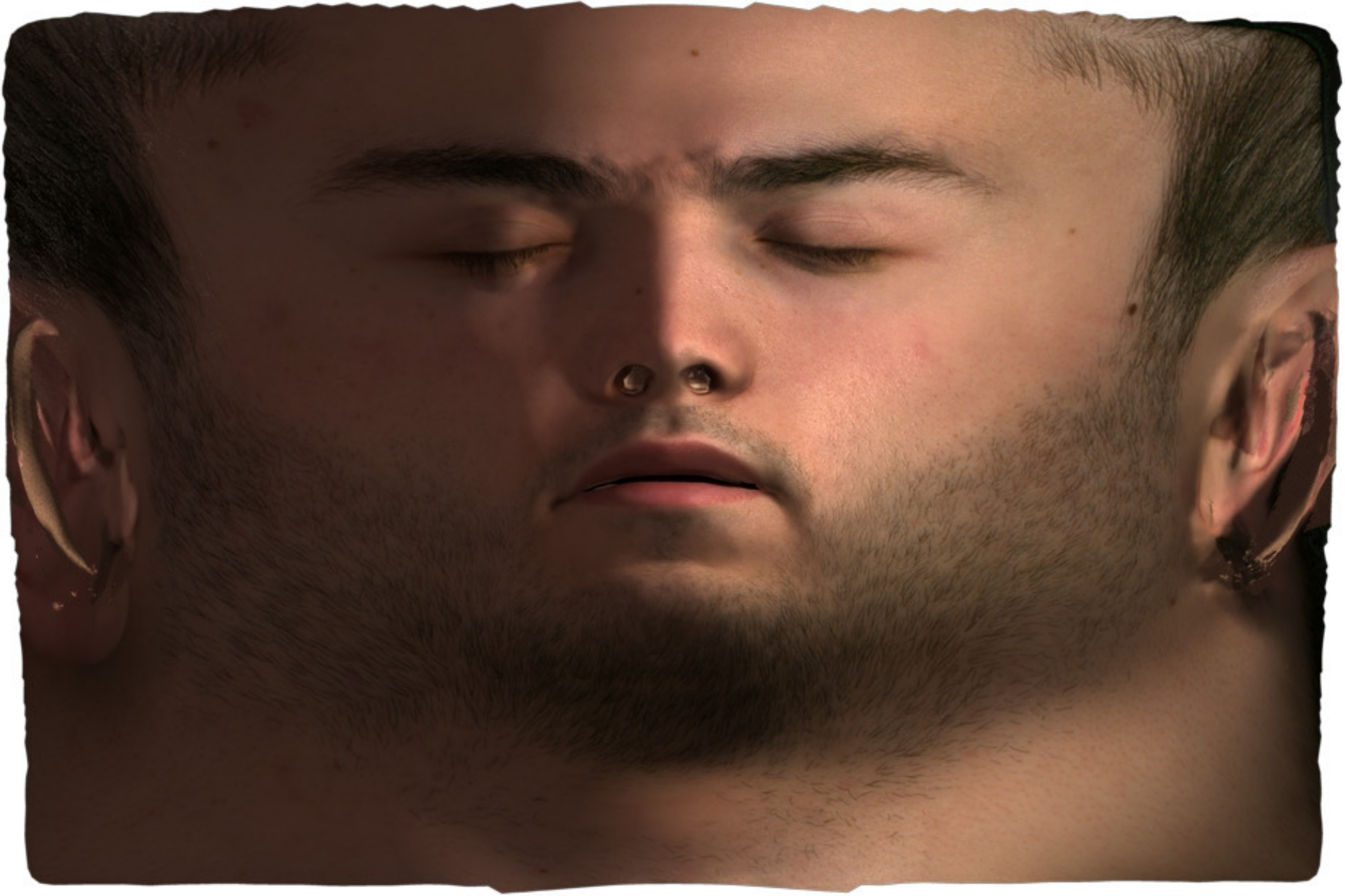}}
    \subfloat[AvatarMe (Sec.~\ref{sec:method_recon_avatarme})]{
        \includegraphics[width=0.1055\linewidth]{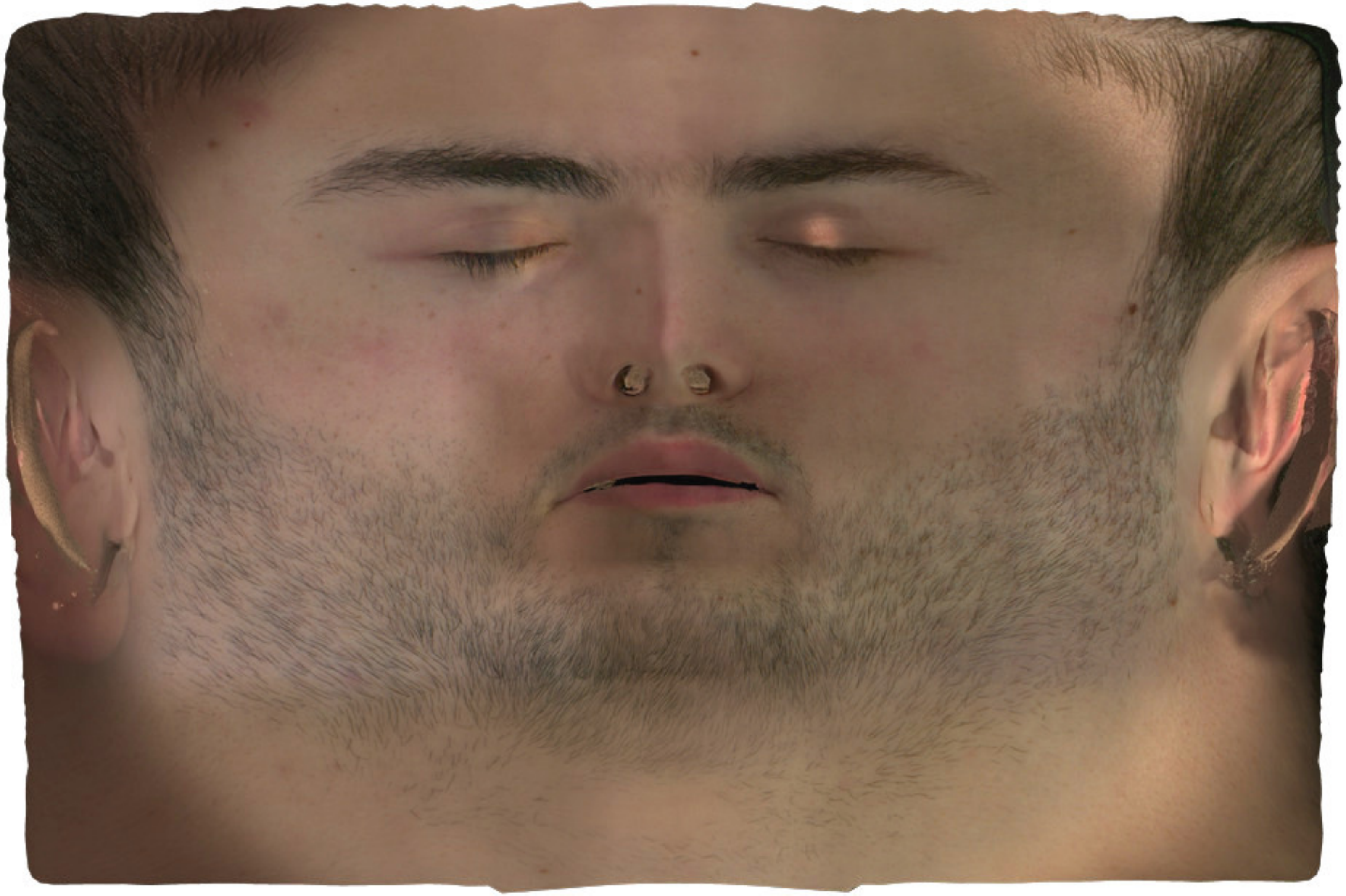}}
    \subfloat[Single Network (Sec.~\ref{sec:method_recon_renderme_combined})]{
        \includegraphics[width=0.1055\linewidth]{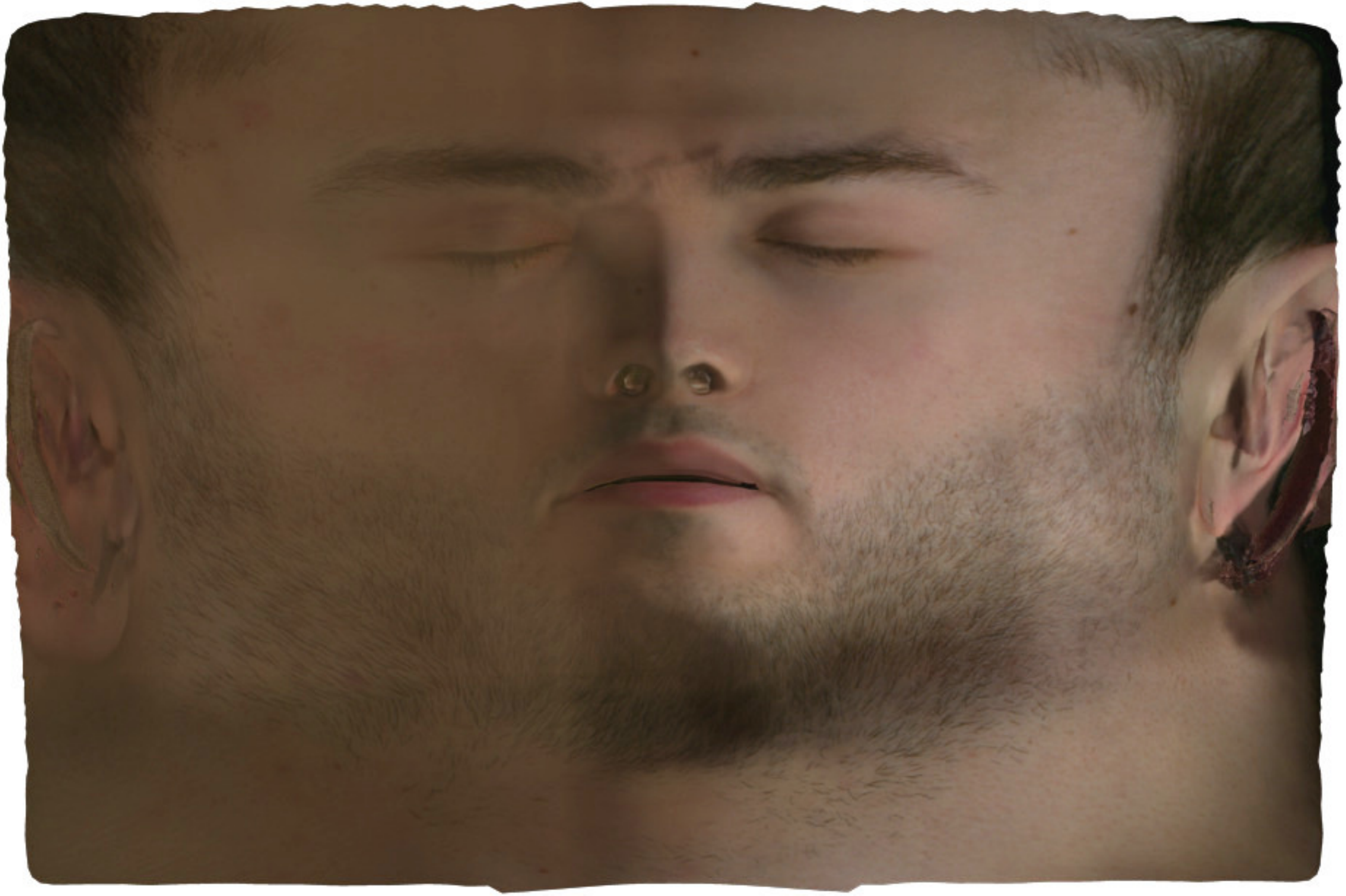}}
    \subfloat[+ Rendering loss (Sec.~\ref{sec:method_recon_renderme_renderloss})]{
        \includegraphics[width=0.1055\linewidth]{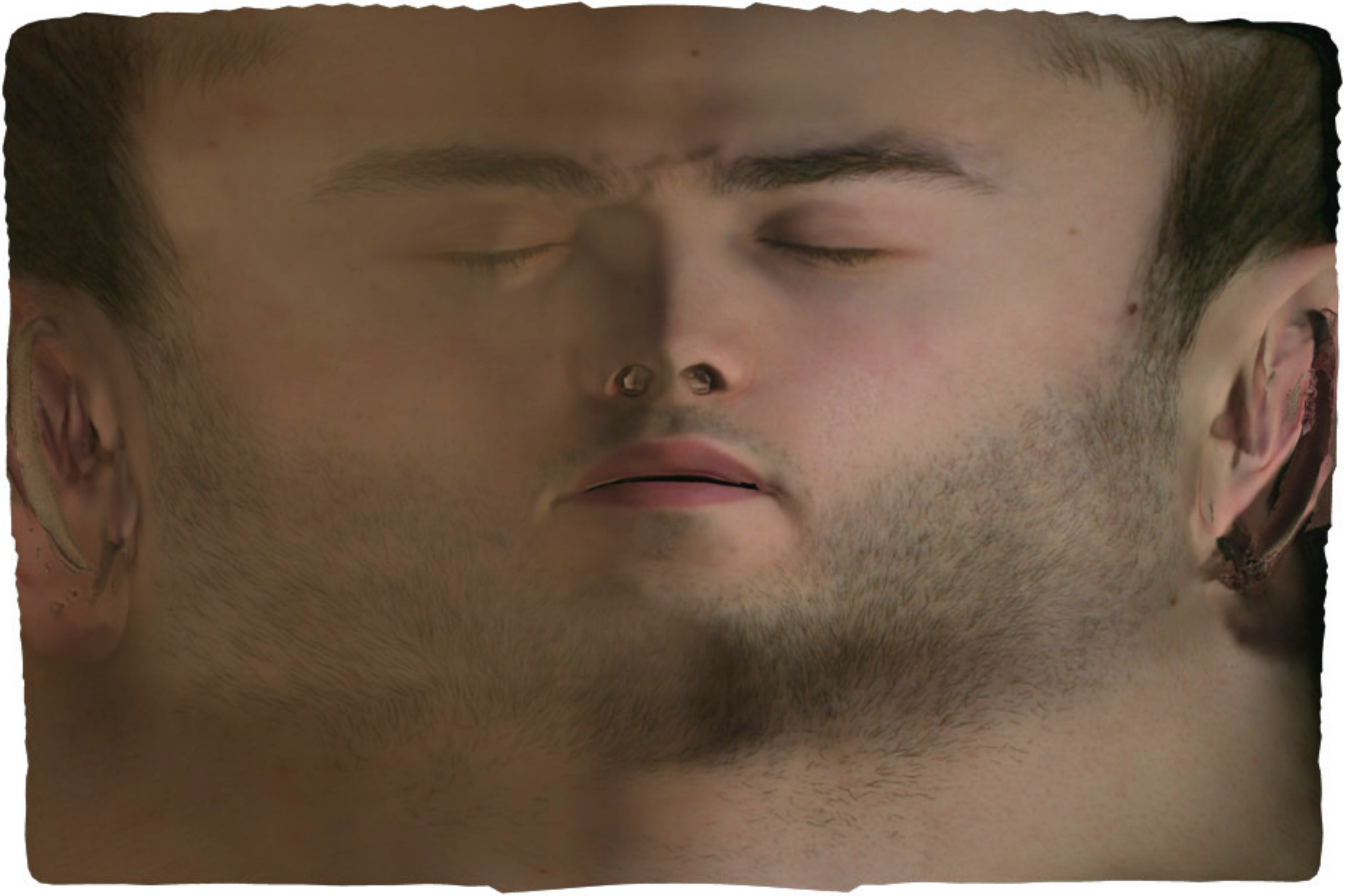}}
    \subfloat[+~Whole low-res texture (Sec.~\ref{sec:method_recon_renderme_combined})]{
        \includegraphics[width=0.1055\linewidth]{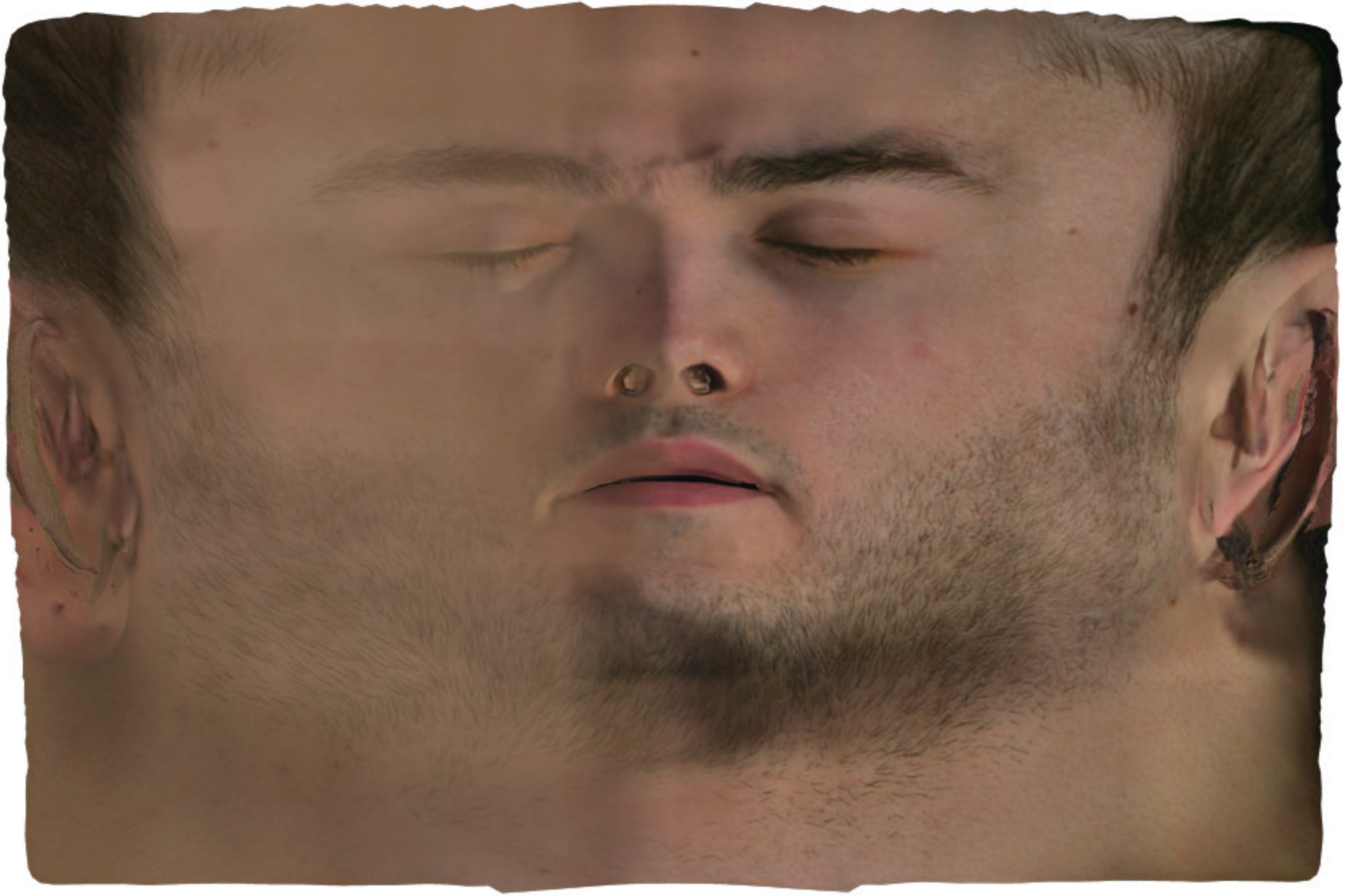}}
    \subfloat[+~Random environment (Sec.~\ref{sec:method_recon_renderme_renderloss})]{
        \includegraphics[width=0.1055\linewidth]{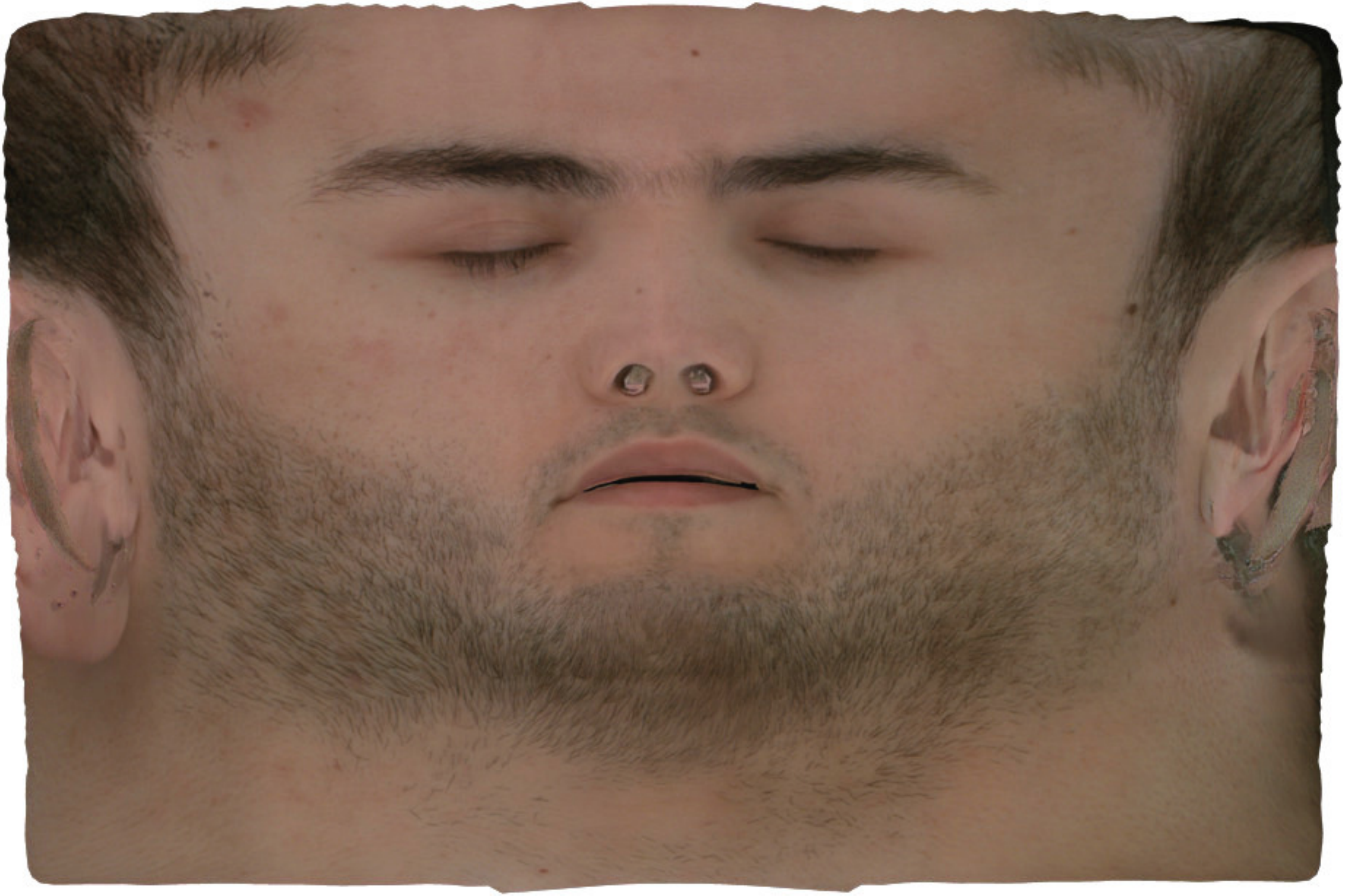}}
    \subfloat[+~Multiple random env. (Sec.~\ref{sec:method_recon_renderme_renderloss})]{
        \includegraphics[width=0.1055\linewidth]{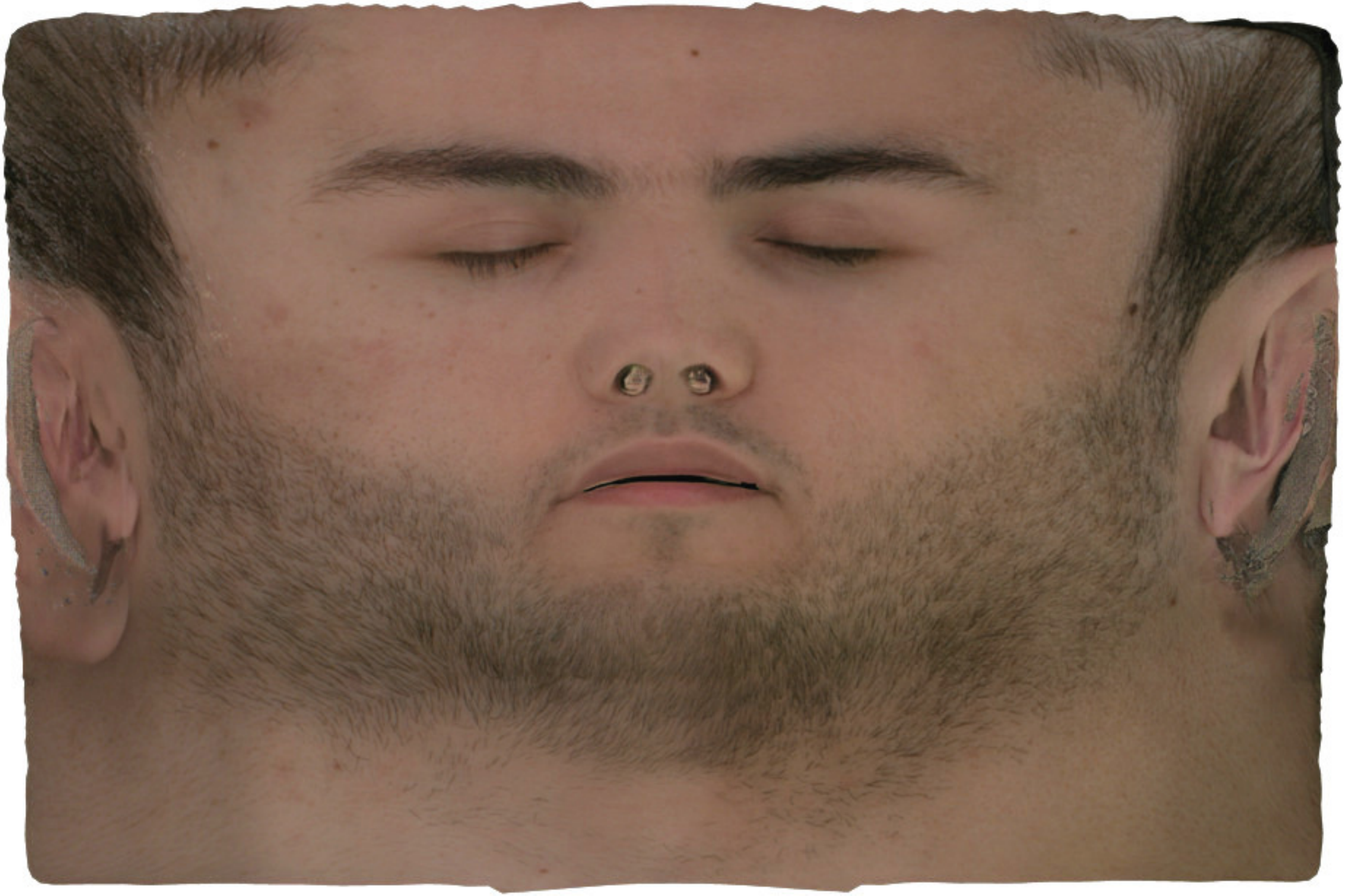}}
    \subfloat[+~Occlusion AE (Sec.~\ref{sec:method_diffrender_shadows}) (AvatarMe\textsuperscript{++})]{
        \includegraphics[width=0.1055\linewidth]{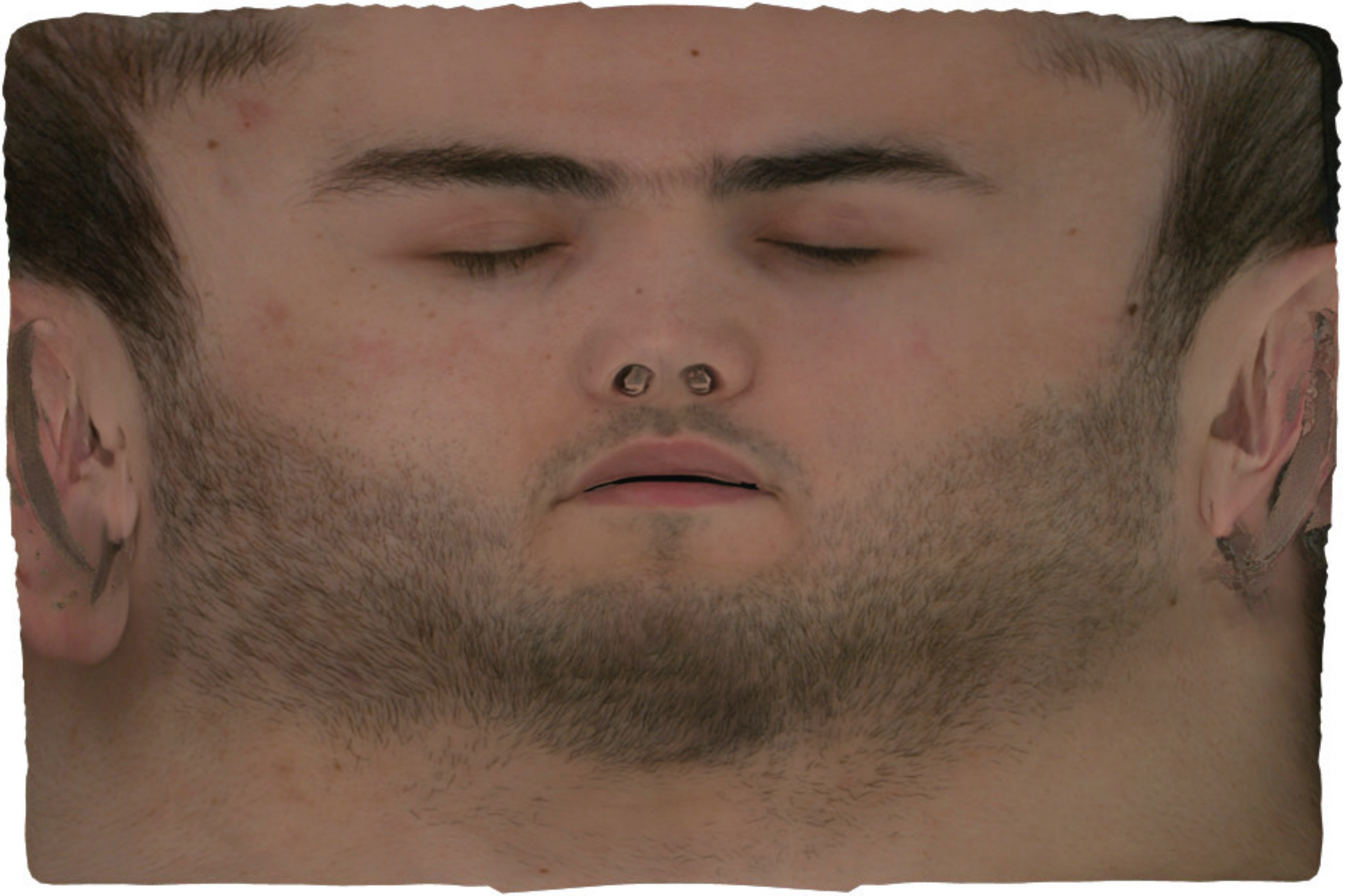}}
    \subfloat[Ground Truth]{
        \includegraphics[width=0.1055\linewidth]{fig/ablation/ground_truth.pdf}}
    \caption{
        From left to right, rendered test image, predicted diffuse albedo $\mathbf{A_D}$
        from each of the main method variations and the ground truth
        diffuse albedo $\mathbf{A_D^*}$.
        Top row input is rendered in our target environment,
        bottom row in an extreme side illumination environment from $\mathcal{T}$.
    }
    \label{fig:results_ablation_images}
\vspace{-0.2cm}
\end{figure*}
\begin{figure}[h]
\vspace{-0.3cm}
    \centering
    \captionsetup[subfigure]{labelformat=empty}
    \subfloat[
        Input]{
        \includegraphics[width=0.24\linewidth]{fig/ablation/gf_input.pdf}}
    \subfloat[
        AvatarMe $\mathbf{N_D}$]{
        \includegraphics[width=0.24\linewidth]{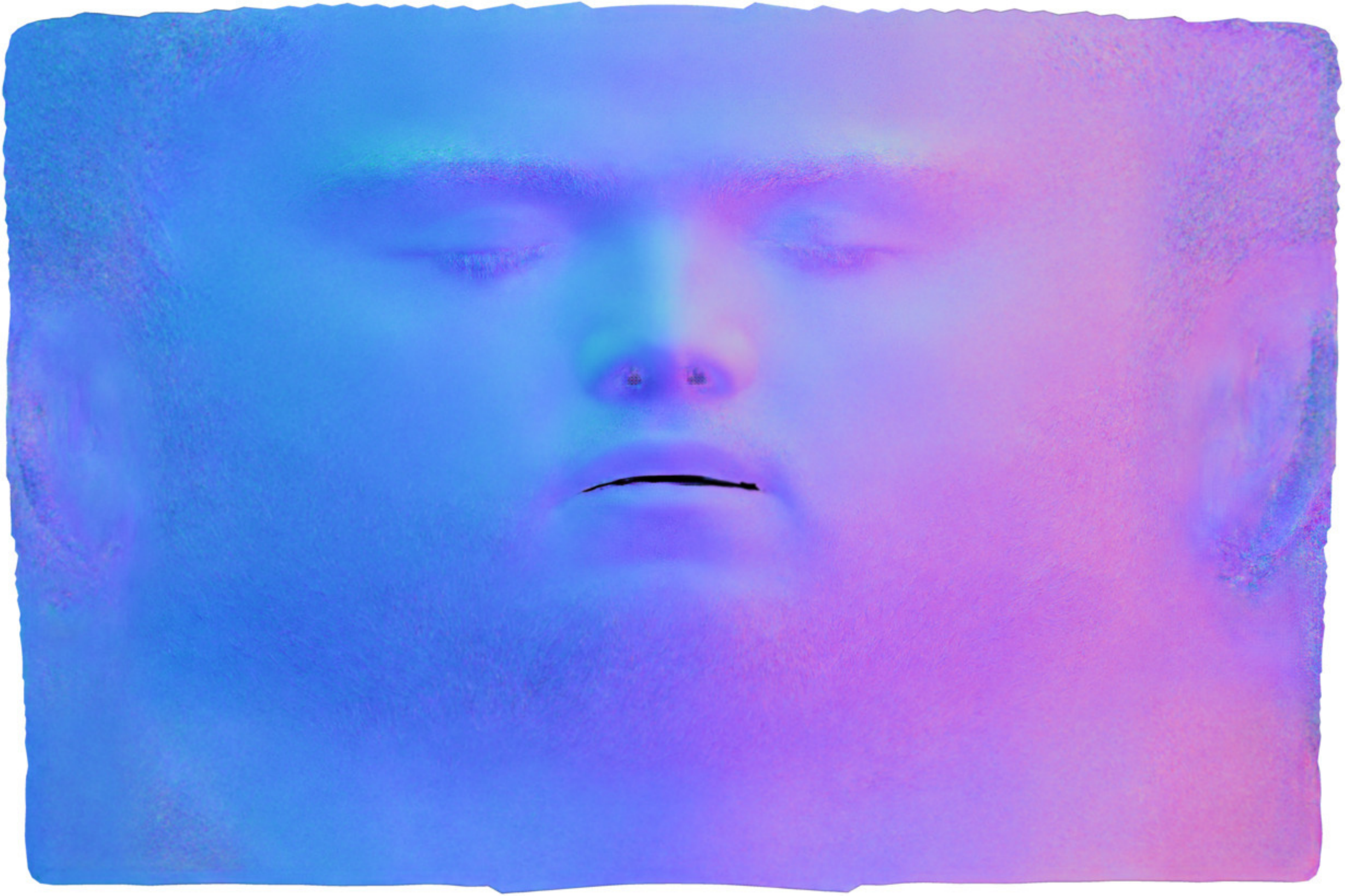}}
    \subfloat[
        AvatarMe $\mathbf{A_S}$]{
        \includegraphics[width=0.24\linewidth]{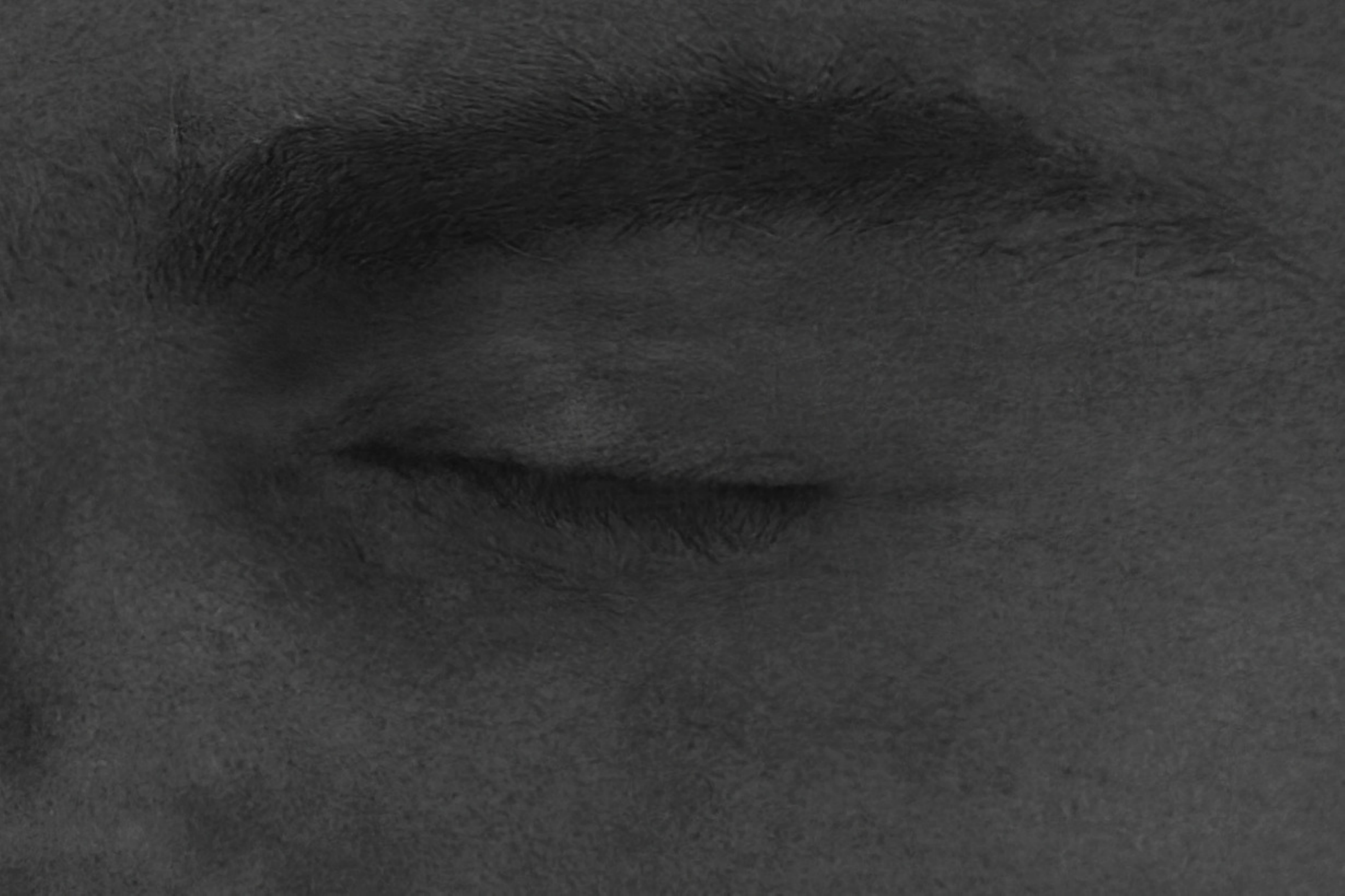}}
    \subfloat[AvatarMe $\mathbf{N_S}$]{
        \includegraphics[width=0.24\linewidth]{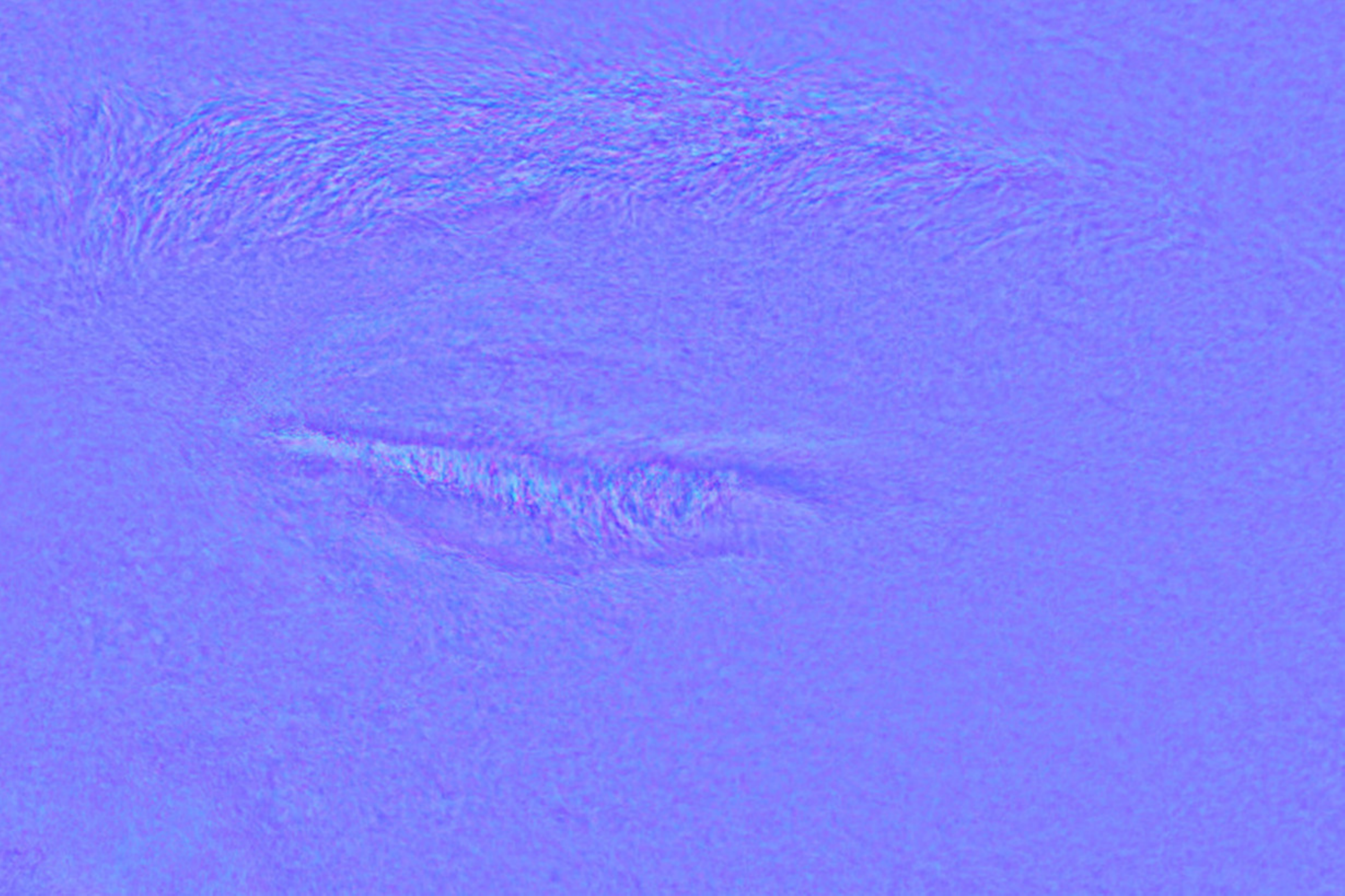}}
    \\
    \subfloat[
        Input]{
        \includegraphics[width=0.24\linewidth]{fig/ablation/gf_input.pdf}}
    \subfloat[
        AvatarMe\textsuperscript{++} $\mathbf{N_D}$]{
        \includegraphics[width=0.24\linewidth]{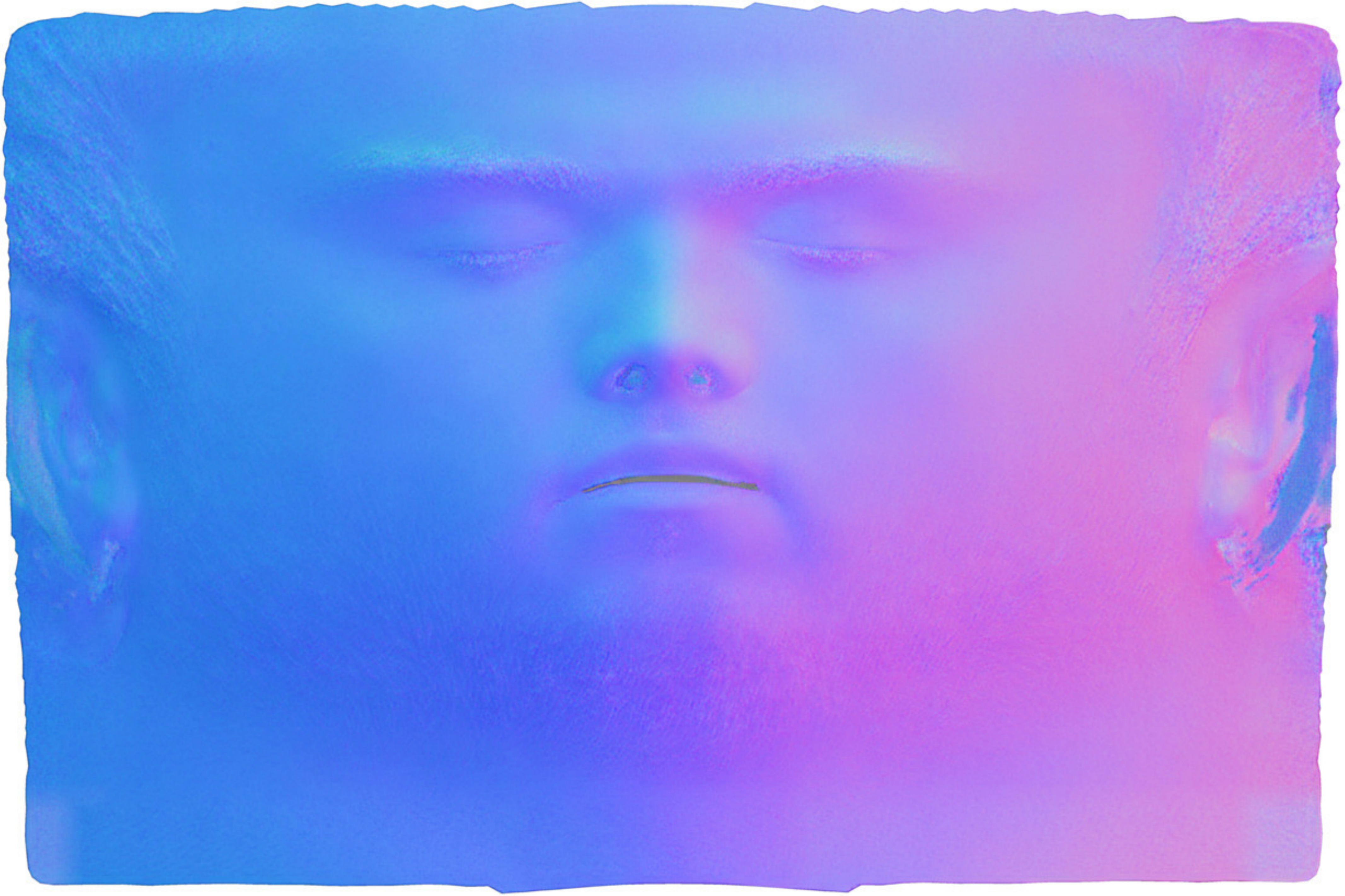}}
    \subfloat[
        AvatarMe\textsuperscript{++} $\mathbf{A_S}$]{
        \includegraphics[width=0.24\linewidth]{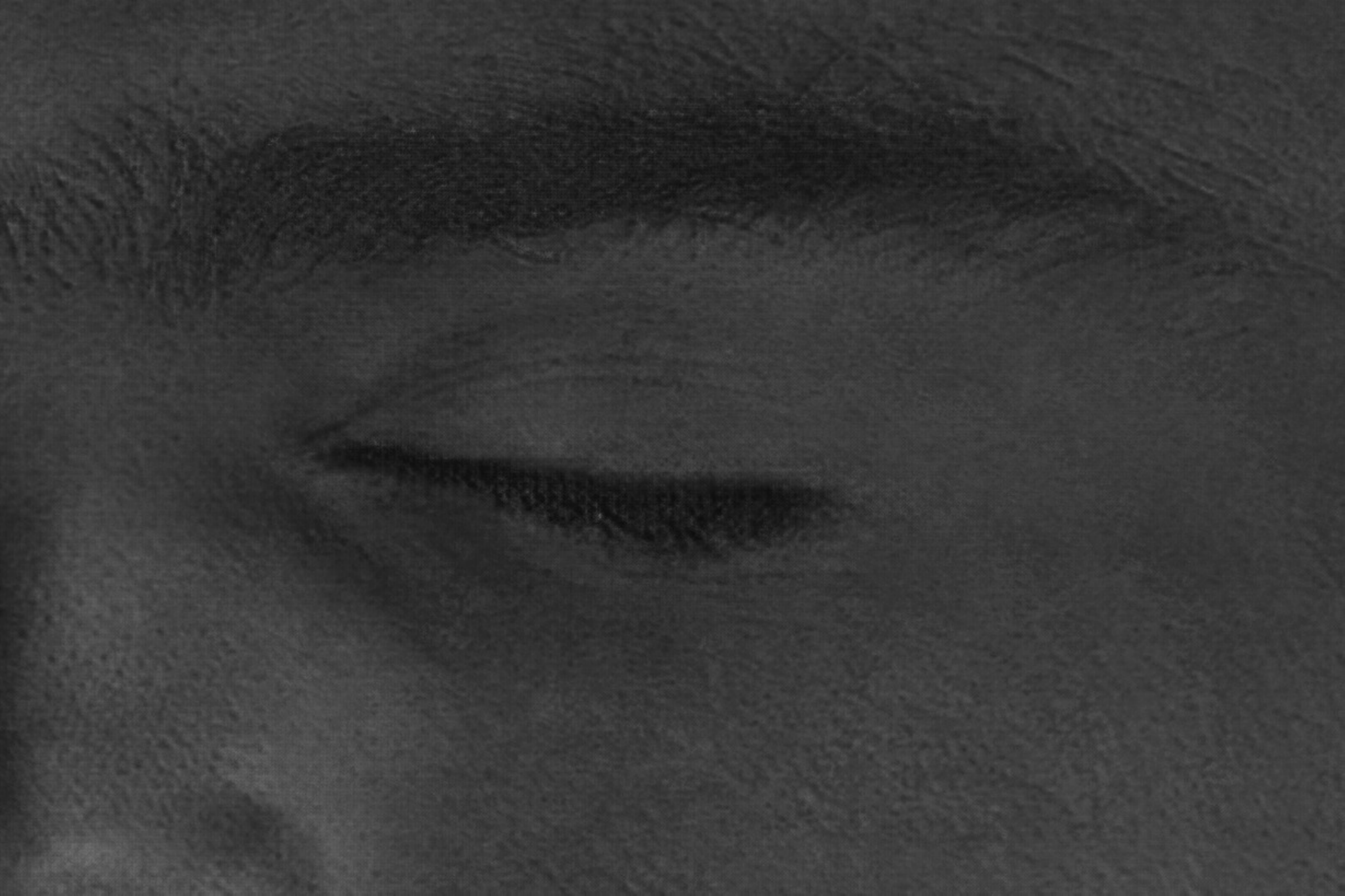}}
    \subfloat[AvatarMe\textsuperscript{++} $\mathbf{N_S}$]{
        \includegraphics[width=0.24\linewidth]{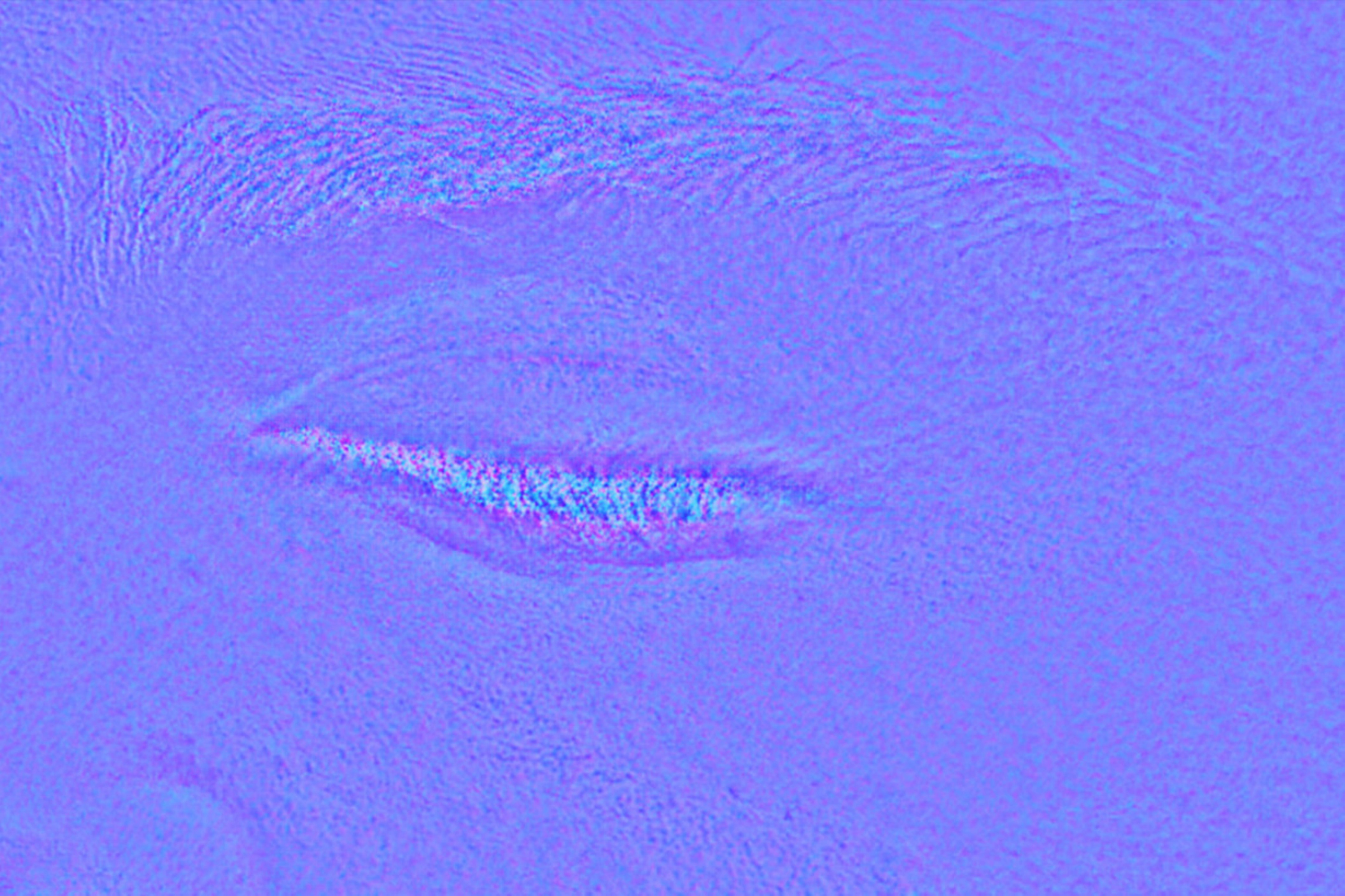}}
    
    \caption{
        Predicted diffuse normals $\mathbf{N_D}$,
        specular albedo $\mathbf{A_D}$, and specular normals details $\mathbf{N_S}$ in tangent space,
        between AvatarMe (Sec.\ref{sec:method_recon_avatarme})
        and AvatarMe\textsuperscript{++}.
        Diffuse albedo $\mathbf{A_D}$ comparison included in Fig.~\ref{fig:results_ablation_images}.
    }
    \label{fig:results_avatarme_avatarme++}
\end{figure}

\begin{table}[h]
\centering
\caption{
    Method components ablation.   
    Training time is given per training iteration,
    testing time is given for the inference of whole $\hat{W}\times\hat{H}$ textures.
    AvatarMe time includes the training time for all 4 of its networks.
    We measure the mean squared error (MSE) between ground truth and inferred reflectance maps, for the test set $\mathcal{T}$ (Sec \ref{sec:method_and_test_set}).
}
\label{tab:results_ablation_time}
\begin{tabular}{@{}lrrr@{}}
\toprule
\textbf{Method} & \textbf{Train time} & \textbf{Test time} & \textbf{MSE} \\
\midrule
AvatarMe (Sec.~\ref{sec:method_recon_avatarme})
    & 22.8  & 15.98 & 0.0079 \\
Single Network (Sec.~\ref{sec:method_recon_renderme_combined})
    & 8.0   & 6.35  & 0.0080 \\
+ Rendering loss (Sec.~\ref{sec:method_recon_renderme_combined})
    & 14.4  & 6.48  & 0.0075 \\
+ Whole low-res texture (Sec.~\ref{sec:method_recon_renderme_combined})
    & 14.8  & 9.31  & 0.0066 \\
+ Random environment (Sec.~\ref{sec:method_recon_renderme_renderloss})
    & 14.8  & 9.35  & 0.0055 \\
+ Multiple random env. (Sec.~\ref{sec:method_recon_renderme_renderloss})
    & 24.0  & 9.35  & 0.0048 \\
+ Occl. AE (Sec.~\ref{sec:method_diffrender_shadows}) (AvatarMe\textsuperscript{++})
    & 28.8  & 9.35  & 0.0043 \\
\bottomrule
\end{tabular}
\end{table}

We investigate the importance of the various components of our method,
when added on the base network.
We create a test set 
$\mathcal{T} = \{\mathcal{R}(\mathbf{R}_i, \mathbf{L}_j)\}$,
by rendering 5 test subjects from our captured dataset, RealFaceDB,
with reflectance $\mathbf{R}_j$,
in 10 total illumination environments $\mathbf{L}_j$ including:
a) the 3DMM-target environment (Sec.~\ref{sec:method_data_env_estimation}),
b) $-/+30\%$ light source variation in position,
c) $-/+30\%$ in intensity from a), and 
d) directional front and side illumination.
We compare the mean squared error,
between the ground truth and the model's predicted reflectance
and the rendering of the prediction in the source testing environment,
as well as in the target environment.
Testing in various illumination environments evaluates the generalization
abilities of our networks, outside the target environment.
We plot both inference and re-rendering error in Fig.~\ref{fig:results_ablation_barplot}
and show the reconstructed diffuse albedo $\mathbf{A_D}$ in 
Fig.~\ref{fig:results_ablation_images}. Moreover, 
although using a single network reduces the training and inference time
by about $75\%$, 
the additional rendering adds a significant overhead in training time,
but does not significantly increase testing time.
We show this trade-off between reconstruction quality 
and training and testing time
Table~\ref{tab:results_ablation_time}.

The introduction of the stochastic rendering loss (Sec.~\ref{sec:method_recon_renderme_renderloss}) 
enables AvatarMe\textsuperscript{++} to generalize to facial textures with different environments,
while AvatarMe is trained only on the target 3DMM environment.
Fig.~\ref{fig:results_ablation_barplot} shows the quantitative improvement of AvatarMe\textsuperscript{++} on the test set $\mathcal{T}$ with 10 different environments,
two of which are shown in Fig.~\ref{fig:results_ablation_images}.
Finally, Fig.~\ref{fig:results_others} shows the network's ability
to generalize to textures obtained from different datasets (i.e.~\cite{DBLP:conf/eccv/BerrettiBP12, yang_facescape_2020})
and acquisitions methods (i.e.~\cite{gecer2020ostec, wang_high-resolution_2018}).

\subsubsection{Network and Rendering Hyper-parameters}
\label{sec:experiments_abblation_hyperparams}

\begin{figure}[h]
    \centering
    \includegraphics[width=0.9\linewidth]{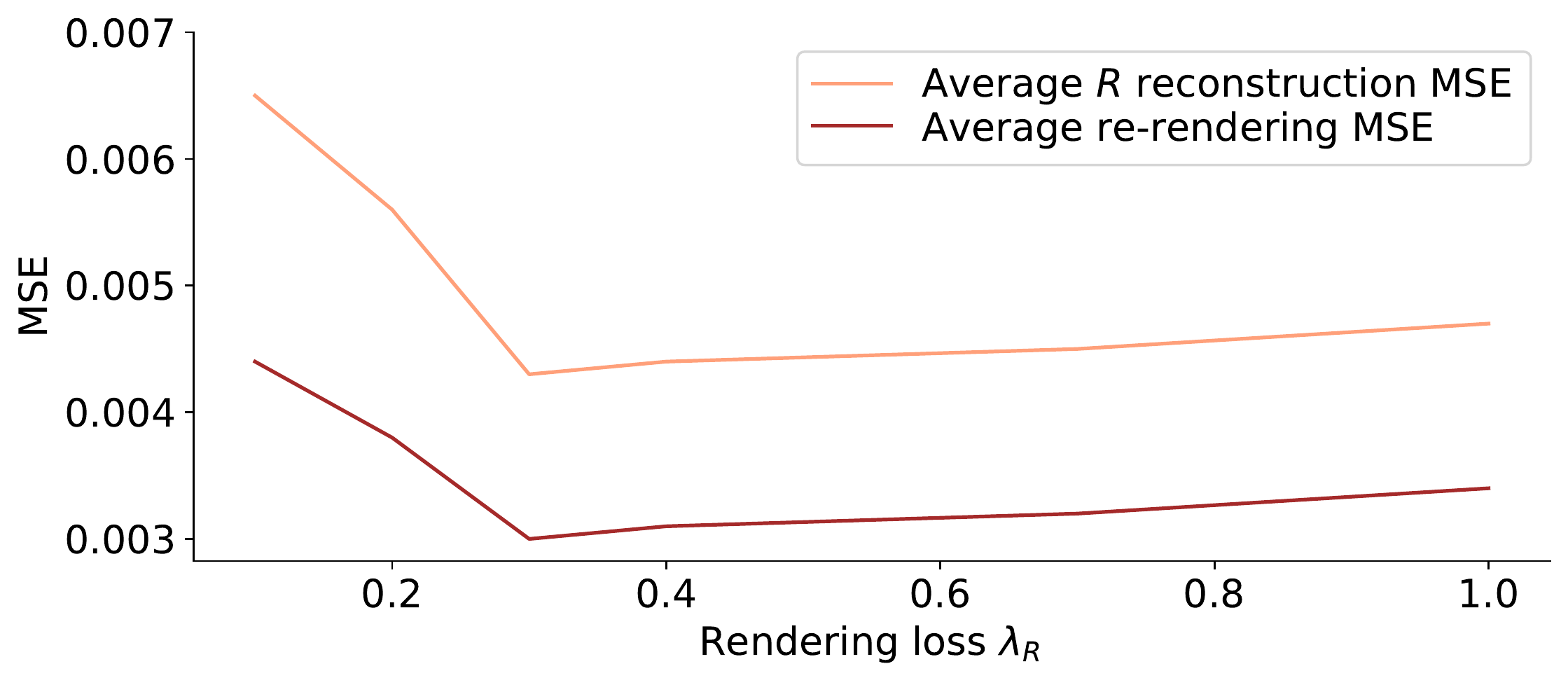}
    \caption{
        The effect of different values of the rendering loss controller
        $\lambda_R$ (Eq.~\ref{eq:recon_renderme_loss})
        in the average reflectance $\mathbf{R}$ reconstruction
        for 5 subjects rendered in 10 different environments
        and the average re-rendering mean squared truth
        with the ground-truth.
    }
    \label{fig:results_ablation_lambda}
\end{figure}

Training the network on 4 Tesla V100 GPUs
takes about one day for a base single network (Sec.~\ref{eq:recon_avatarme_diffalbedo} or Sec.~\ref{sec:method_recon_renderme_combined})
and up to three days for the complete method, 
with multiple rendering loss computations
and self-occlusion in rendering (Sec.~\ref{sec:method_recon_renderme_renderloss}). 
Given these restrictions,
we perform a study on the effect of the important hyper parameters.
We train the complete method with different seeds and record an average
$3.14\%$ standard deviation in the error of reflectance reconstruction and
$5.37\%$ standard deviation in the error of re-rendered textures.
All the other results are reported using the same seed for all stochastic operations.
The impact of the rendering loss controller 
$\lambda_R$ (Eq.~\ref{eq:recon_renderme_loss})
is shown in Fig.~\ref{fig:results_ablation_lambda}.
We find $\lambda_R = 0.3$ to yield the best results.

Table~\ref{tab:results_ablation_scene_variation}
shows the effect of the scene variation hyper-parameters when training with 
randomized light source $\mathbf{l}$, camera direction $\mathbf{v}$
and light source intensity $\mathbf{c_l}$.
We find the best results at up to
$50\%$ variation in light source position, and up to
$25\%$ in camera position and light source intensity.
Finally, Table~\ref{tab:results_ablation_scene_variation}
shows the effect of the number of random scenes $n_L$
used to evaluate the rendering loss (Eq.~\ref{eq:recon_renderme_renderloss}).
We find a significant improvement in using 2 scenes instead of 1,
and a smaller improvement when adding additional scenes.
However each additional rendering loss evaluation
introduces a trade-off with computational time.

\begin{table}[h]
\centering
\caption{
    Ablation study for rendering loss hyper-parameters.
    Comparison of reflectance prediction and re-rendering error
    for variation in scene parameters
    (light source position $\mathbf{l}$,
    camera view direction $\mathbf{v}$
    light source intensity $\mathbf{c_l}$)
    around the target environment
    (top table), and number of random scenes evaluated $n_L$ in the rendering loss (bottom table) (Sec.~\ref{sec:method_recon_renderme_renderloss}).
    Average error reported for test set $\mathcal{T}$ (Sec.~\ref{sec:method_and_test_set}).
}
\label{tab:results_ablation_scene_variation}
\begin{tabular}{@{}lll|rr@{}}
\toprule
\multicolumn{3}{c}{\textbf{Scene Variation}} & \multicolumn{2}{c}{\textbf{Results}}     \\
\cmidrule(lr){1-3} \cmidrule(lr){4-5}
$\mathbf{l}$ & $\mathbf{v}$ & $\mathbf{c_l}$
                                        & \textbf{Recon. MSE}   & \textbf{Render MSE}   \\
\midrule
$0\%$       & $0\%$     & $0\%$         & 0.0055                & 0.0059                \\
$50\%$      & $25\%$    & $25\%$        & \textbf{0.0043}       & \textbf{0.0030}                \\
$75\%$      & $50\%$    & $50\%$        & 0.0051                & 0.0036                \\
\bottomrule
\smallskip
\end{tabular}
\begin{tabular}{@{}r|rrr@{}}
\toprule
\textbf{Evaluations $n_L$} & \textbf{Train Time} & \textbf{Recon. MSE} 
                                                        & \textbf{Render MSE} \\
\midrule
1       & \textbf{15.6} & 0.0054             & 0.0037           \\
2       & 22.4          & 0.0045             & 0.0032           \\
3       & 28.8          & \textbf{0.0043}    & \textbf{0.0030}  \\
\bottomrule
\end{tabular}
\end{table}

\subsection{Limitations}
\label{sec:experiments_limitations}

While our dataset contains a relatively large number of subjects,
it does not contain sufficient examples of subjects from certain ethnicities
(Sec.~\ref{sec:method_data_collection}).
Despite our effective training data augmentation,
we find that the predicted diffuse albedo
for subjects of the lightest or darkest skin types,
produces patches of inconsistent intensity
and the specular normals of older subjects are reconstructed sub-optimally.
Moreover, the accuracy of facial reconstruction is not completely independent
of the quality of the input photograph, 
and well-lit, higher resolution photographs produce more accurate results,
depending on the 3DMM method used.
Additionally, we show that our method can generalize to various reconstruction and capturing methods, however, the model expects their light sources to be in front of the subject.
Finally, our renderer models self-occlusion but not occlusion from foreign objects.
This could be modeled by augmenting our dataset with randomized occlusions.

\section{Conclusion}
\label{sec:conclusion}
In this paper, 
we propose the first methodology that produces high-quality 
rendering-ready face reconstructions from arbitrary ``in-the-wild'' images. 
We build upon recently proposed 3D face reconstruction techniques 
and train an image translation network that can perform estimation of high quality 
(a) diffuse and specular albedo, and 
(b) diffuse and specular normals. 
This is made possible with a large training dataset of 200 faces 
acquired with high quality facial capture techniques
and a fast photorealistic differentiable rendering framework.
We demonstrate that it is possible to produce rendering-ready faces from arbitrary face images varying in pose, occlusions, etc., 
including black-and-white and drawn portraits. 
Our results exhibit unprecedented level of detail and realism in the reconstructions,
while preserving the identity of subjects in the input photographs.

\section*{Acknowledgments}
AL was supported by EPSRC Project DEFORM (EP/S010203/1) and
SM by an Imperial College DTA.
AG acknowledges funding by the EPSRC Early Career Fellowship (EP/N006259/1)
and SZ from a Google Faculty Fellowship 
and the EPSRC Fellowship DEFORM (EP/S010203/1).

\ifCLASSOPTIONcaptionsoff
  \newpage
\fi



\bibliographystyle{IEEEtran}
\bibliography{IEEEabrv,bibliography.bib}
%

%
\vspace{-1cm}
\begin{IEEEbiography}[{
    \includegraphics[width=1in,height=1in,clip,keepaspectratio]{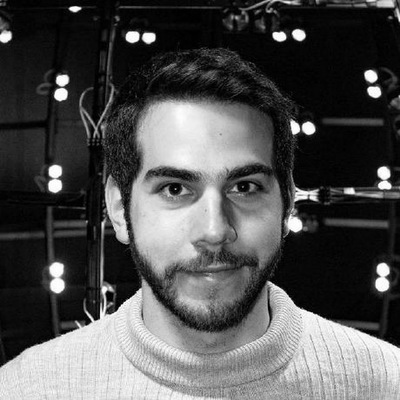}}]
    {Alexandros Lattas}
is a PhD candidate in the Department of Computing, Imperial College London,
under the supervision of Prof Stefanos Zafeiriou and Dr Abhijeet Ghosh.
He received his BSc in Management \& Technology (Software Engineering)
from the Athens University of Economics and Business (AUEB), Greece, in 2017.
He joined the department of computing at Imperial College London, 
in October 2017, where he pursued an MSc in Advanced Computing. 
His interests lie in the field of photorealistic 3D human modeling
with Deep Learning, 3D Computer Vision and Graphics.
\vspace{-1cm}
\end{IEEEbiography}

\begin{IEEEbiography}[{
    \includegraphics[width=1in,height=1in,clip,keepaspectratio]{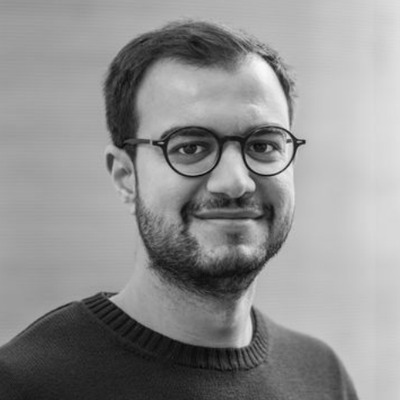}}]
    {Stylianos Moschoglou}
received his Diploma/MEng in Electrical and Computer Engineering 
from Aristotle University of Thessaloniki, Greece, in 2014. 
In 2015-16, he pursued an MSc in Computing (specialization Artificial Intelligence) 
at Imperial College London, U.K., 
where he completed his project under the supervision of Dr. Stefanos Zafeiriou.
He is currently a PhD student at the Department of Computing, Imperial College London, 
under the supervision of Dr. Stefanos Zafeiriou. 
His interests lie within the area of Machine Learning 
and in particular in Generative Adversarial Networks and Component Analysis.
\vspace{-1cm}
\end{IEEEbiography}

\begin{IEEEbiography}[{
    \includegraphics[width=1in,height=1in,clip,keepaspectratio]{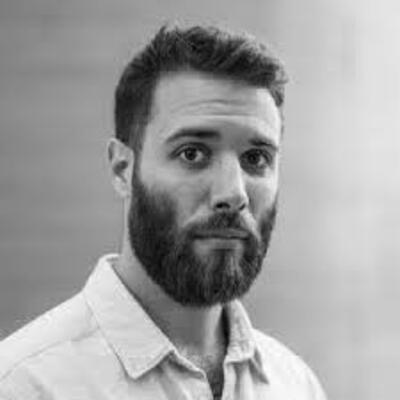}}]
    {Stylianos Ploumpis}
received the Diploma and MEng 
in Production Engineering \& Management 
from Democritus University of Thrace, Greece (D.U.T.H.), in 2013. 
He joined the department of computing at Imperial College London, 
in October 2015, where he pursued an MSc in Computing specializing in Machine Learning. 
Currently, he is a PhD candidate/Researcher at the Department of Computing at Imperial College, 
under the supervision of Dr.~Stefanos Zafeiriou. 
His research interests lie in the field of 3D Computer Vision, Pattern Recognition and Machine Learning.
\vspace{-1cm}
\end{IEEEbiography}

\begin{IEEEbiography}[{
    \includegraphics[width=1in,height=1in,clip,keepaspectratio]{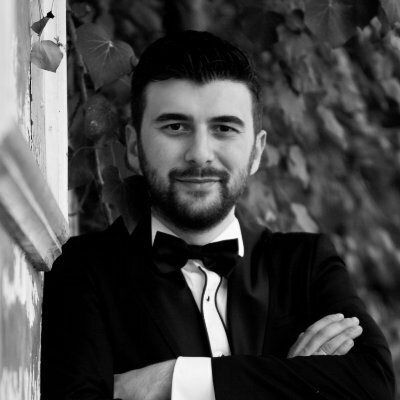}}]
    {Baris Gecer}
is a PhD. student in the Department of Computing, Imperial College London,
under the supervision of Dr. Stefanos Zafeiriou.
His main research interests are photorealistic 3D Face modeling 
and synthesis by Generative Adversarial Nets and Deep Learning. 
He obtained his M.S. degree from Bilkent University Computer Engineering department 
under the supervision of Prof. Selim Aksoy in 2016 
and obtained his undergraduate degree in Computer Engineering 
from Hacettepe University in 2014.
\vspace{-1cm}
\end{IEEEbiography}

\begin{IEEEbiography}[{
    \includegraphics[width=1in,height=1in,clip,keepaspectratio]{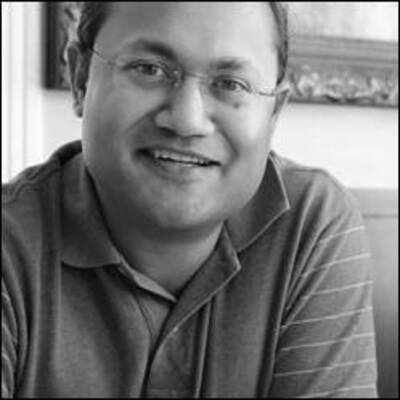}}]
    {Abhijeet Ghosh}
is a Reader (Sr. Associate Professor) in Graphics \& Imaging 
within the Department of Computing at Imperial College London, 
and an Adjunct Professor of Computer Science at NTNU, Norway.
He leads the Realistic Graphics and Imaging group 
and his current research interests include appearance modeling, 
and computational illumination and photography for graphics and vision. His research has been supported with a Royal Society Wolfson Research Merit Award, a Google Faculty Research Award, and an EPSRC Early Career Fellowship.
\vspace{-1cm}
\end{IEEEbiography}

\begin{IEEEbiography}[{
    \includegraphics[width=1in,height=1in,clip,keepaspectratio]{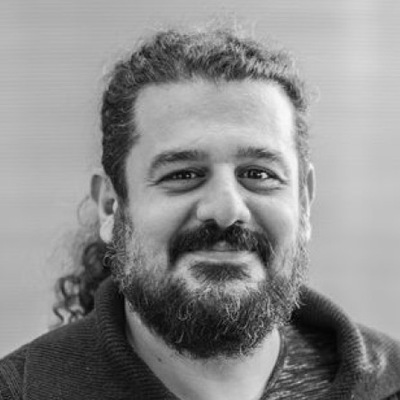}}]
    {Stefanos Zafeiriou}
is a Professor in Machine Learning and Computer Vision with the Dept.~of Computing, Imperial College London, London, U.K, and an EPSRC Early Career Research Fellow. Between 2016-2020 he was also a Distinguishing Research Fellow with the University of Oulu under Finish Distinguishing Professor Programme. He was a recipient of the Prestigious Junior Research Fellowships from Imperial College London in 2011. He was the recipient of the President’s Medal for Excellence in Research Supervision for 2016. He served Associate and Guest Editor in various journals including IEEE Trans. Pattern Analysis and Machine Intelligence, International Journal of Computer Vision, IEEE Transactions on Affective Computing, Computer Vision and Image Understanding, IEEE Transactions on Cybernetics the Image and Vision Computing Journal.
\end{IEEEbiography}

\end{document}


\title{
    Supplemental Materials
    for\\
    AvatarMe\textsuperscript{++}: Facial Shape and BRDF Inference \\
    with Photorealistic Rendering-Aware GANs
    }
    
\author{\small{
    Alexandros Lattas, \and
    Stylianos Moschoglou, \and
    Stylianos Ploumpis, \and \\
    Baris Gecer, \and
    Abhijeet Ghosh, \and
    Stefanos Zafeiriou}
}
\maketitle

\section{AvatarMe\textsuperscript{++} Generalization Results}
Training AvatarMe\textsuperscript{++} with stochastically varied rendering scene parameters,
makes the network domain-agnostic to an extend, depending on the degree of variation.
In this manner, the generalization of AvatarMe\textsuperscript{++} is significantly increased,
when compared to AvatarMe \cite{lattas_avatarme_2020}.
Below, we show the results of Fig.~11 of the main manuscript, in high resolution,
and compare them with the results of AvatarMe.

\begin{figure}[h]
    \centering
    \captionsetup[subfigure]{labelformat=empty}
    \subfloat{
        \includegraphics[width=0.32\linewidth]{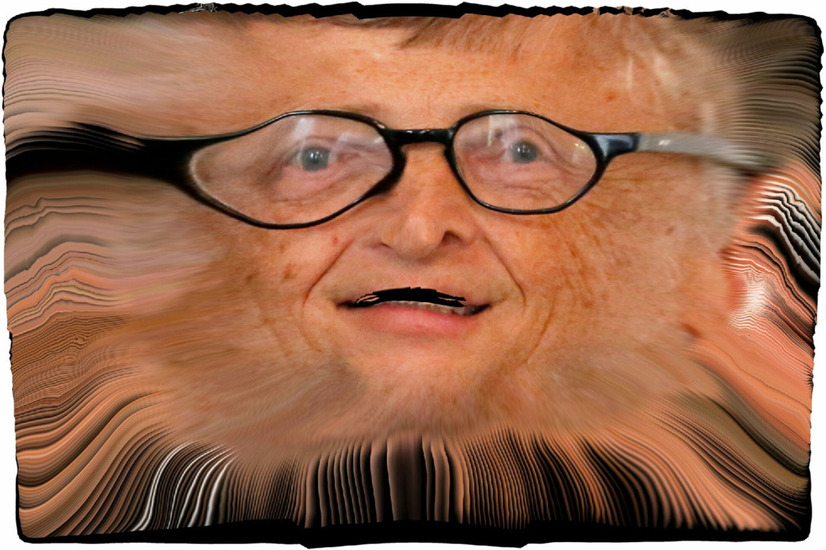}}
    \subfloat{
        \includegraphics[width=0.32\linewidth]{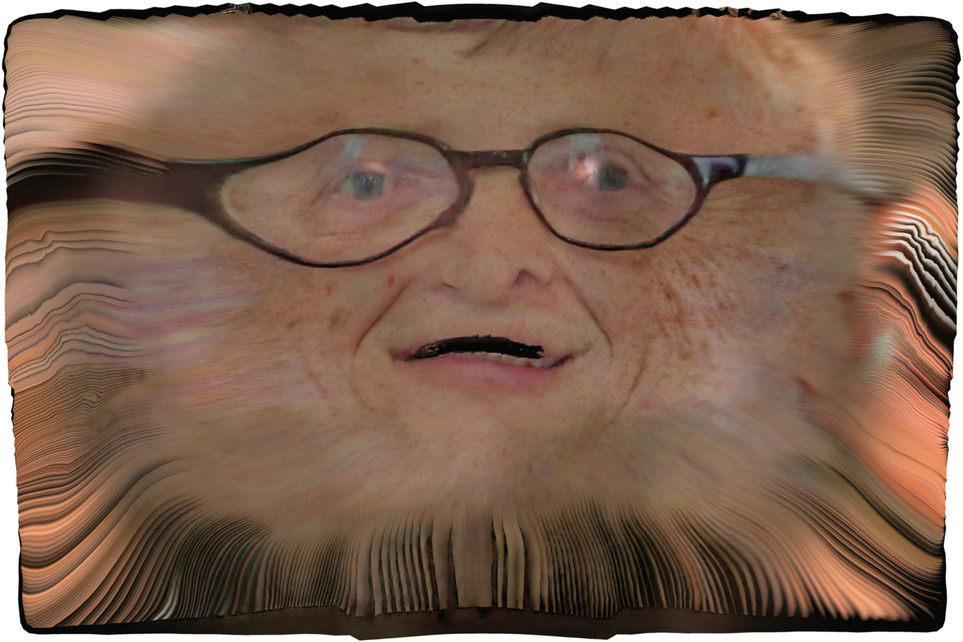}}
    \subfloat{
        \includegraphics[width=0.32\linewidth]{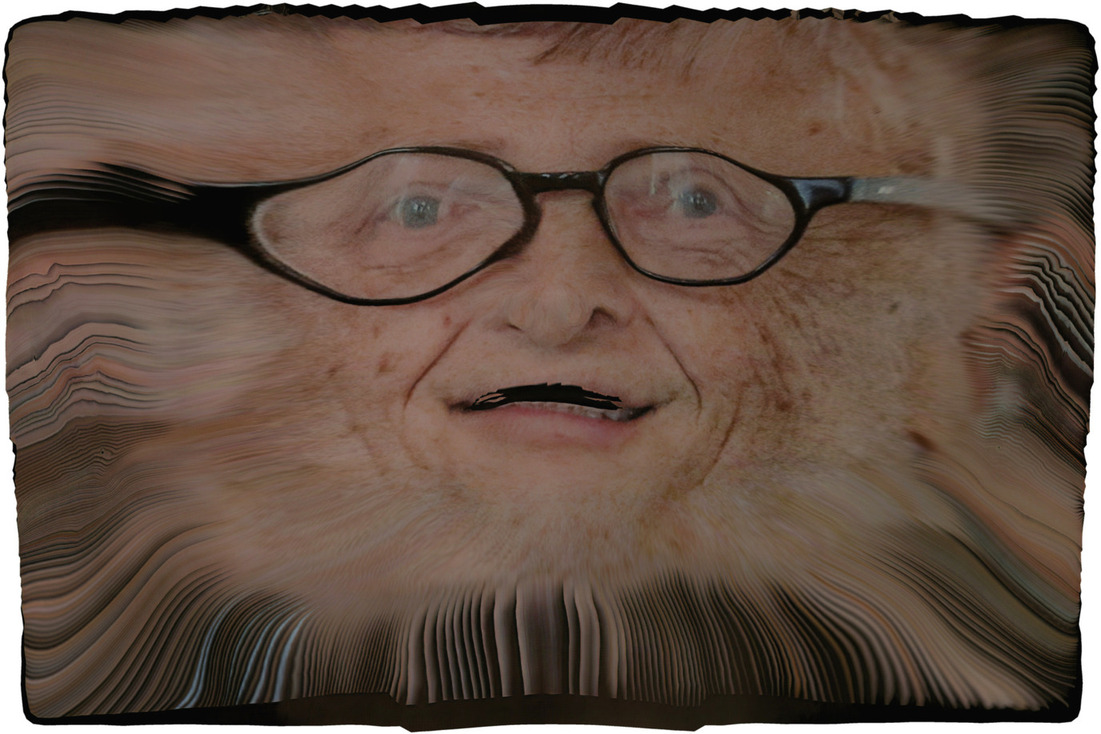}}
        
    \subfloat{
        \includegraphics[width=0.32\linewidth]{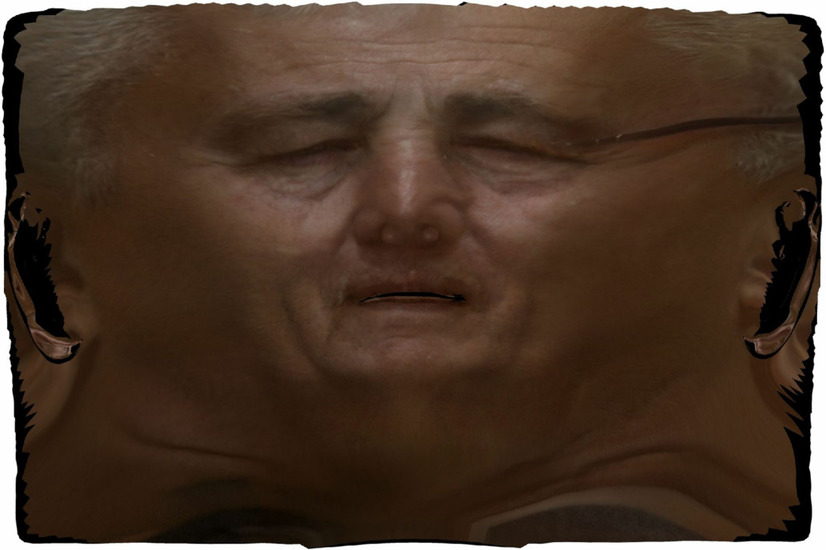}}
    \subfloat{
        \includegraphics[width=0.32\linewidth]{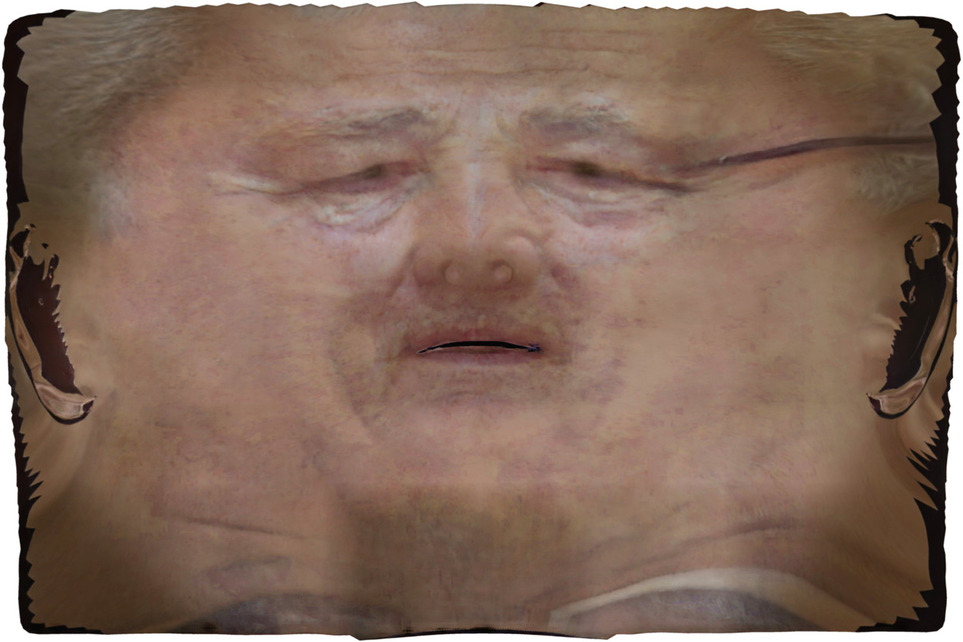}}
    \subfloat{
        \includegraphics[width=0.32\linewidth]{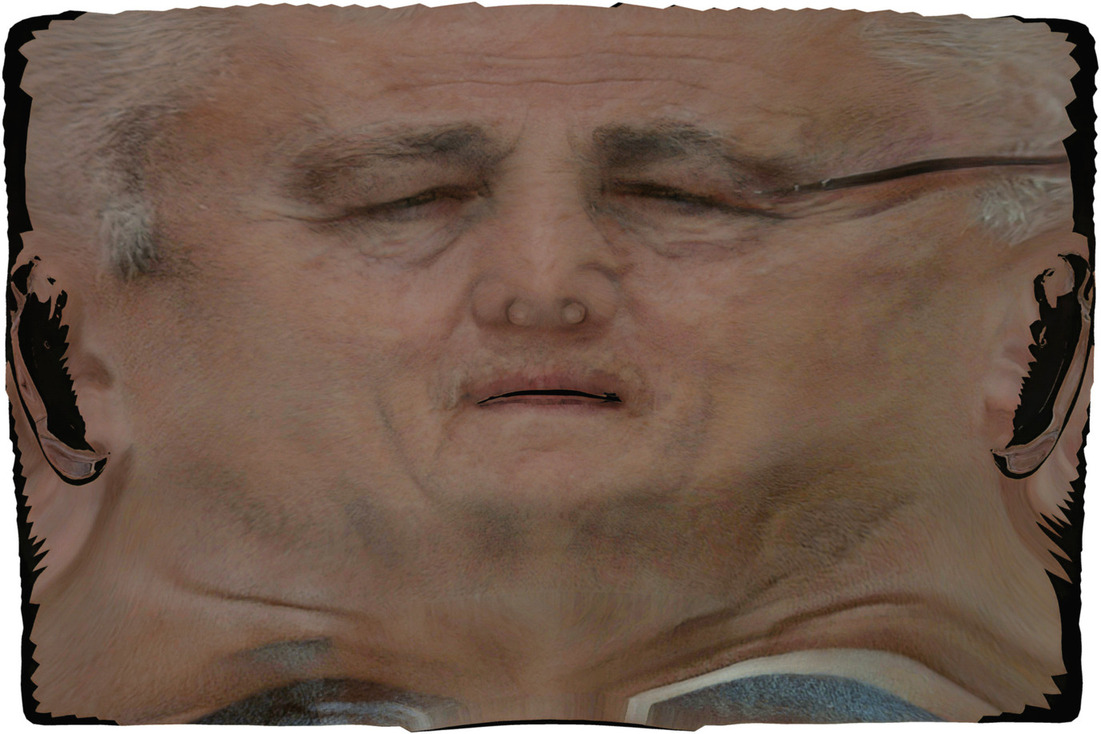}}
        
    \subfloat{
        \includegraphics[width=0.32\linewidth]{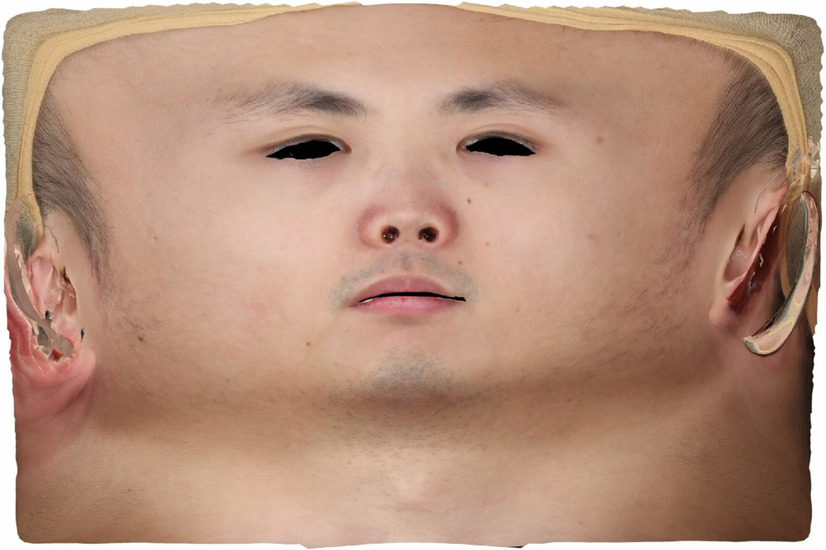}}
    \subfloat{
        \includegraphics[width=0.32\linewidth]{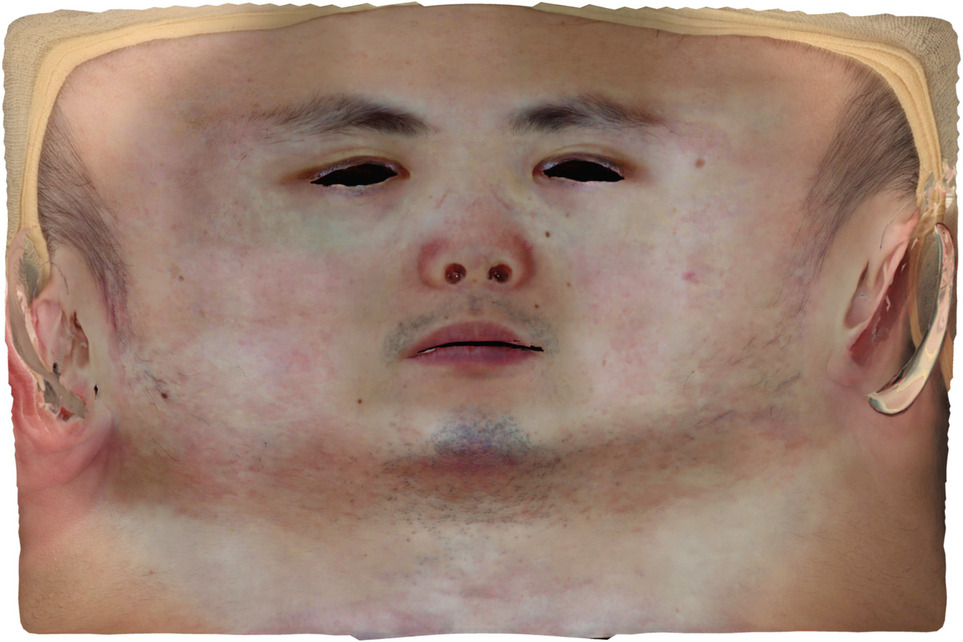}}
    \subfloat{
        \includegraphics[width=0.32\linewidth]{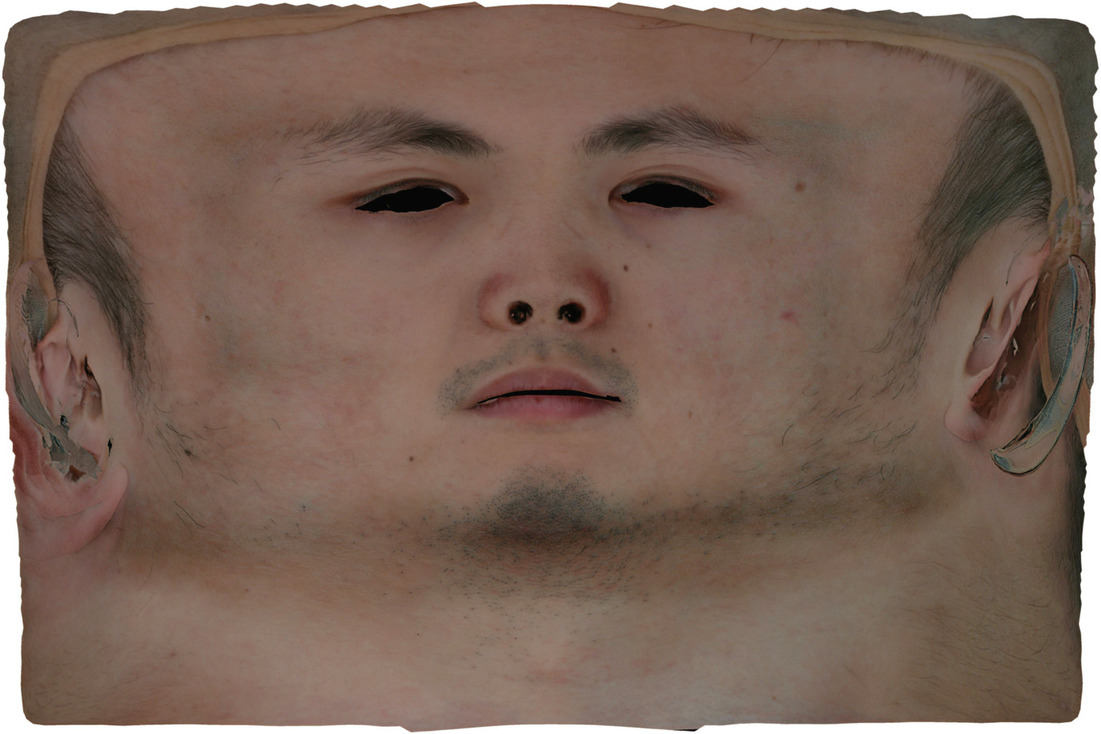}}
        
    \subfloat{
        \includegraphics[width=0.32\linewidth]{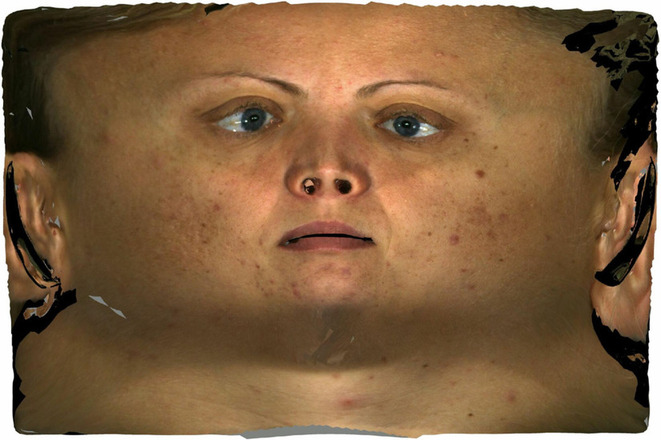}}
    \subfloat{
        \includegraphics[width=0.32\linewidth]{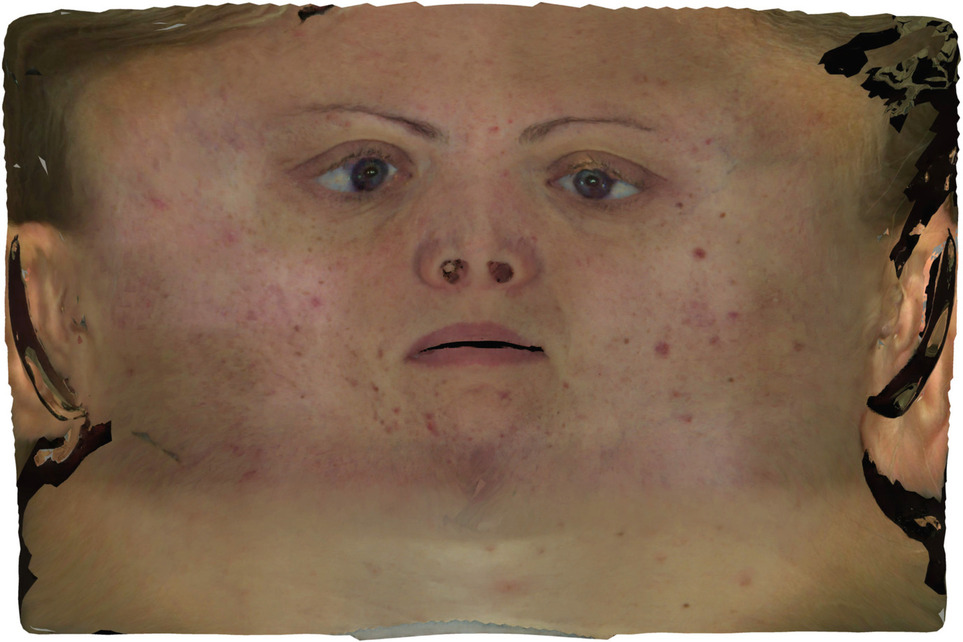}}
    \subfloat{
        \includegraphics[width=0.32\linewidth]{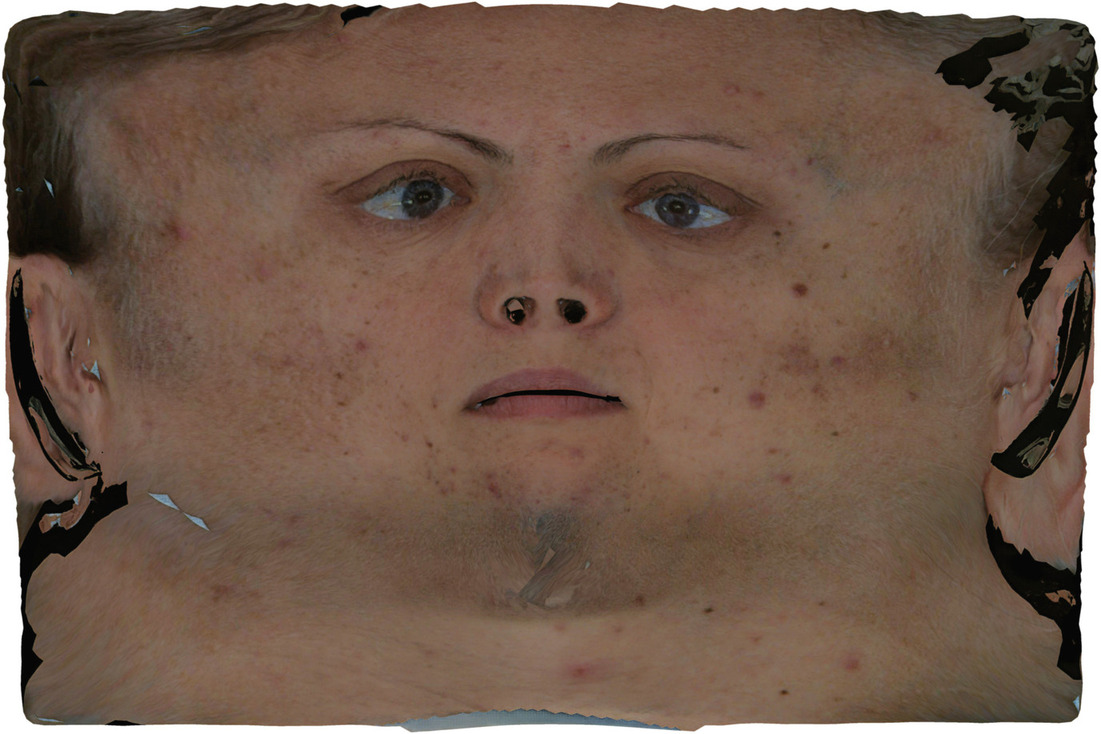}}
        
    \captionsetup[subfloat]{justification=centering}
    \subfloat[
        Input]{
        \includegraphics[width=0.32\linewidth]{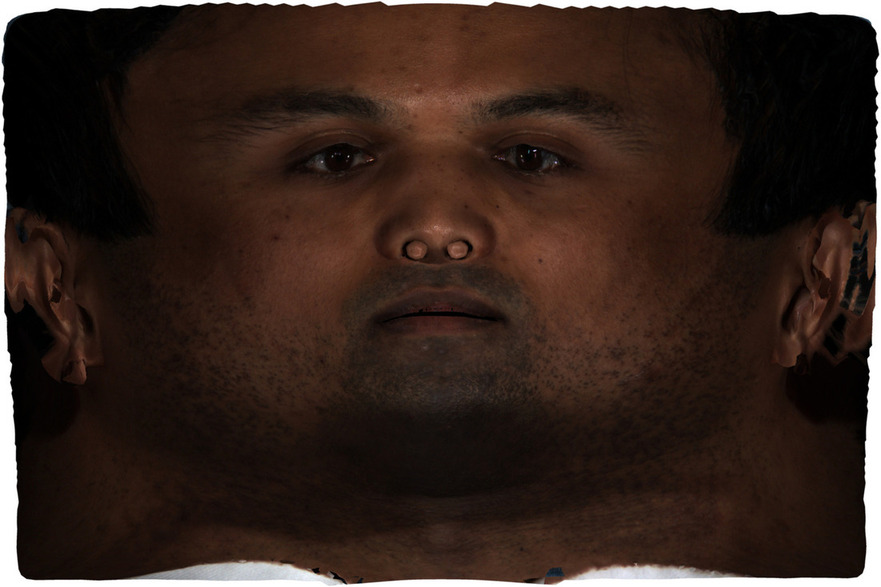}}
    \subfloat[Diffuse Albedo (AvatarMe)]{
        \includegraphics[width=0.32\linewidth]{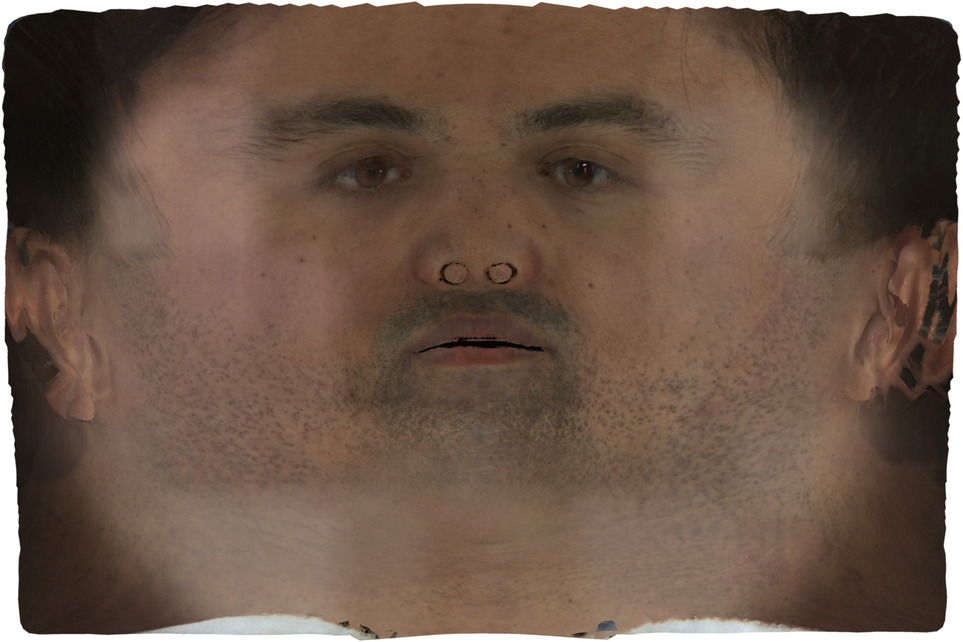}}
    \subfloat[Diffuse Albedo (AvatarMe\textsuperscript{++})]{
        \includegraphics[width=0.32\linewidth]{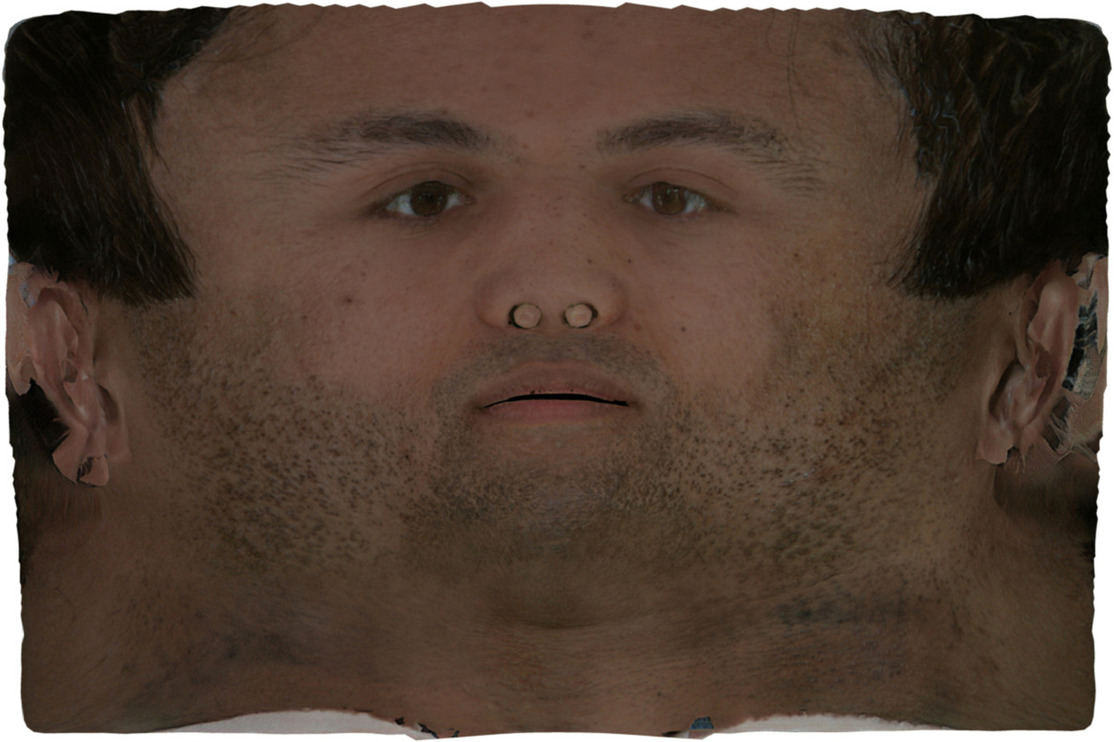}}
    
    \caption{
        Generalization of AvatarMe \cite{lattas_avatarme_2020}, compared to AvatarMe\textsuperscript{++}: Diffuse Albedo 
        from top to bottom:
        a) Reconstructed subject, with Facial Details Synthesis \cite{chen_photo-realistic_2019} proxy and texture,
        b) Reconstructed subject with OSTeC \cite{gecer2020ostec} fitting and texture completion,
        c) Captured subject from FaceScape \cite{yang_facescape_2020} dataset,
        d) Captured subject from Superface \cite{DBLP:conf/eccv/BerrettiBP12} dataset,
        e) Captured subject with a 3dMDface system (https://3dmd.com).
    }
    \label{fig:results_others_diffAlb}
\end{figure}

\begin{figure}[h]
    \centering
    \captionsetup[subfigure]{labelformat=empty}
    \subfloat{
        \includegraphics[width=0.32\linewidth]{fig/results/others/chen_ouruv_004_inp.jpg}}
    \subfloat{
        \includegraphics[width=0.32\linewidth]{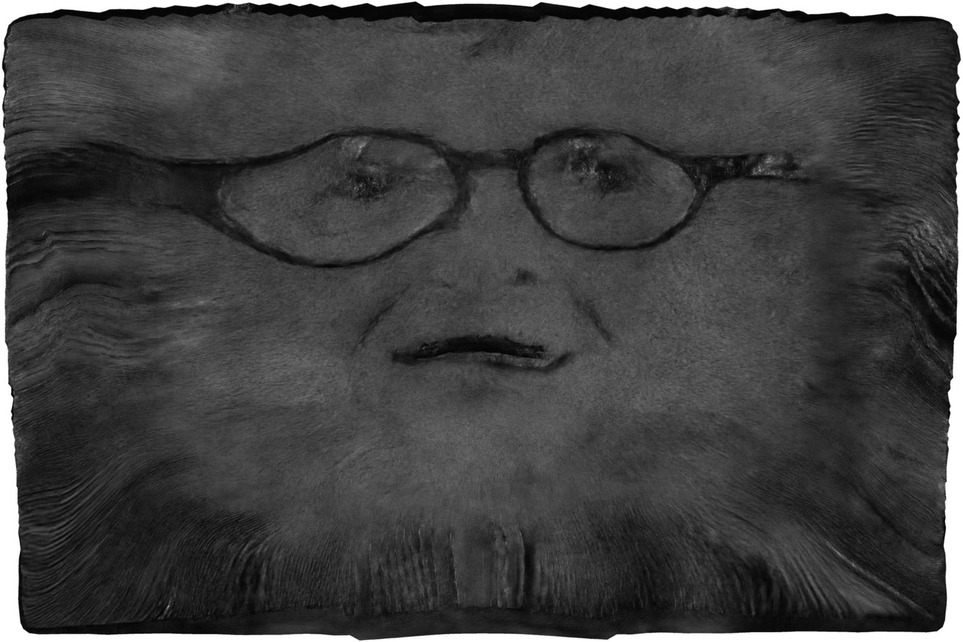}}
    \subfloat{
        \includegraphics[width=0.32\linewidth]{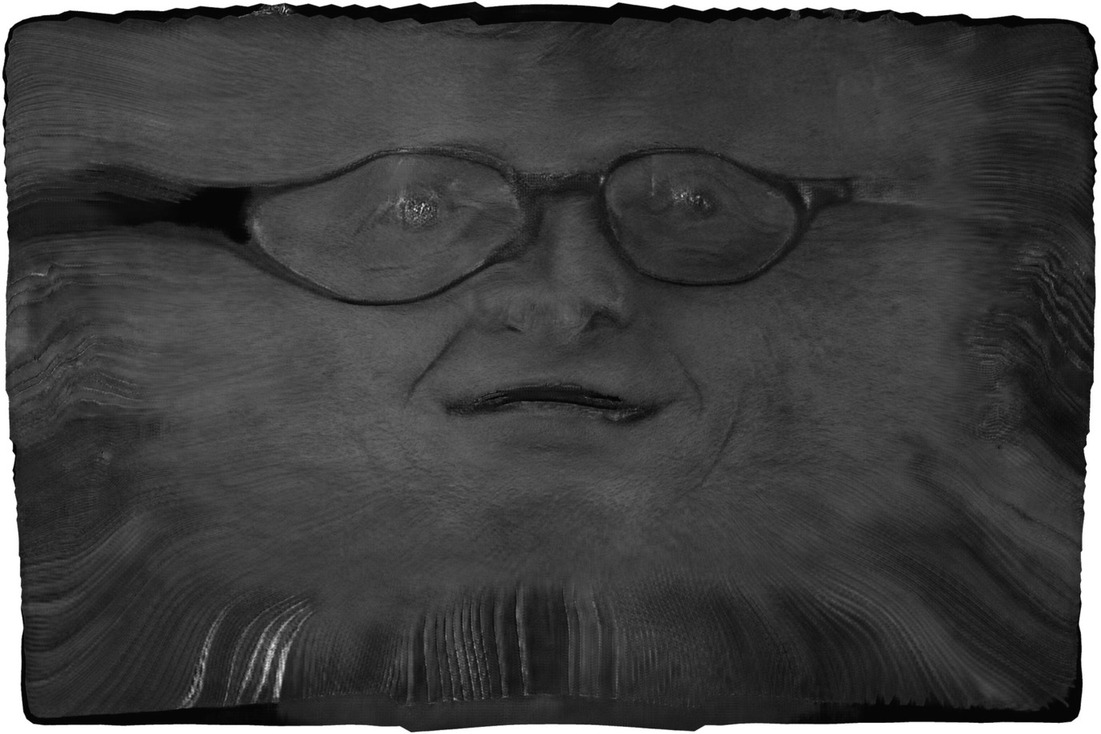}}
        
    \subfloat{
        \includegraphics[width=0.32\linewidth]{fig/results/others/ostec_im13_inp.jpg}}
    \subfloat{
        \includegraphics[width=0.32\linewidth]{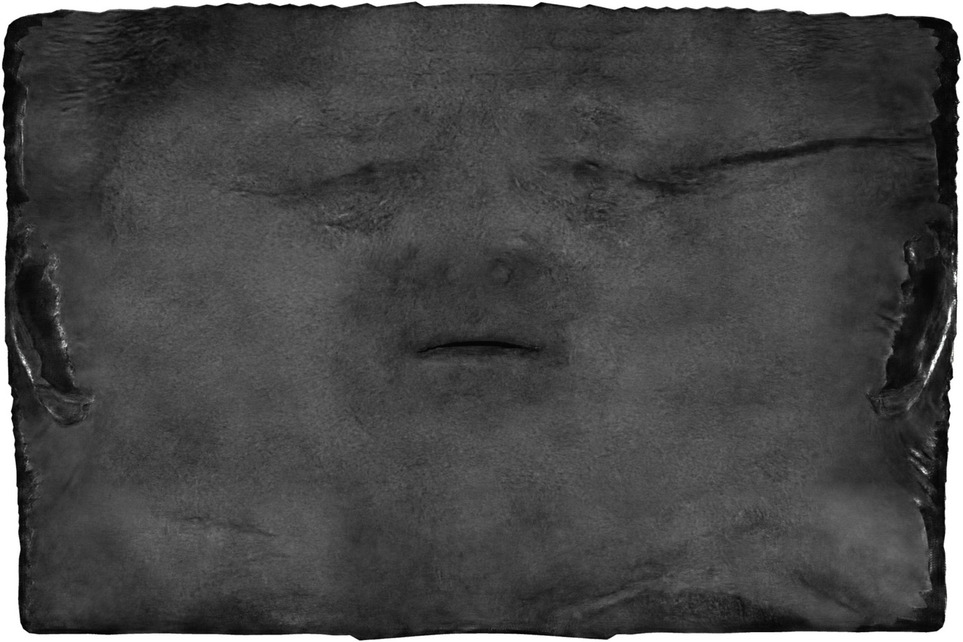}}
    \subfloat{
        \includegraphics[width=0.32\linewidth]{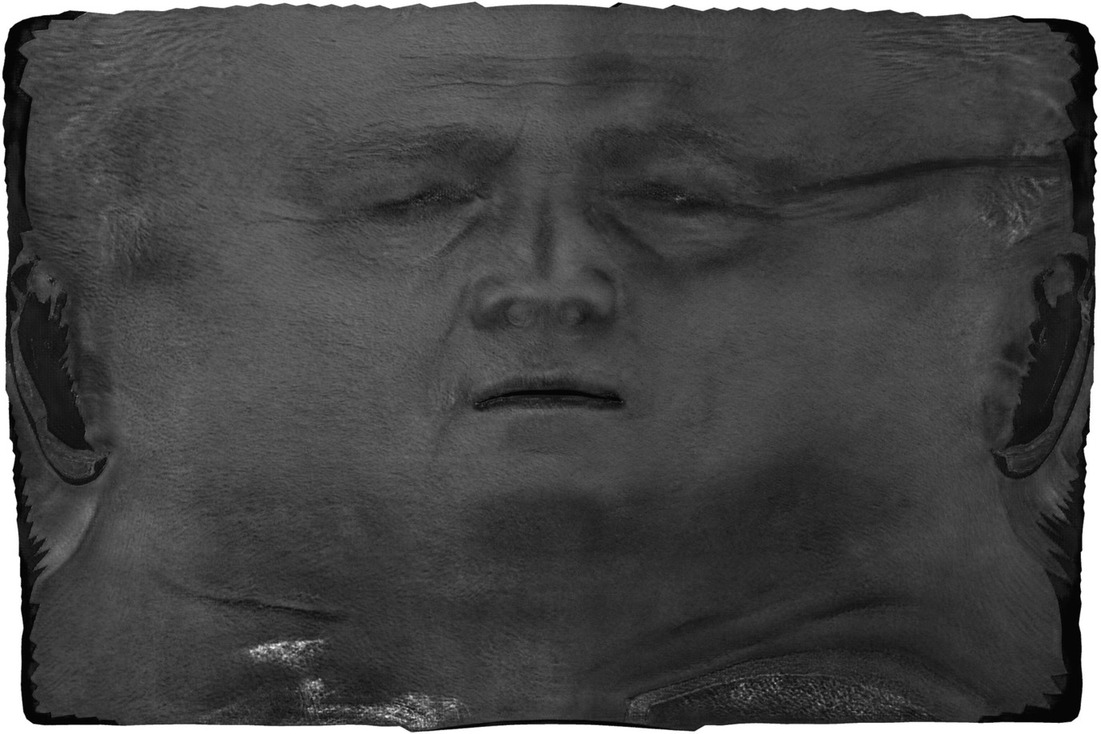}}
        
    \subfloat{
        \includegraphics[width=0.32\linewidth]{fig/results/others/facescape_inp.jpg}}
    \subfloat{
        \includegraphics[width=0.32\linewidth]{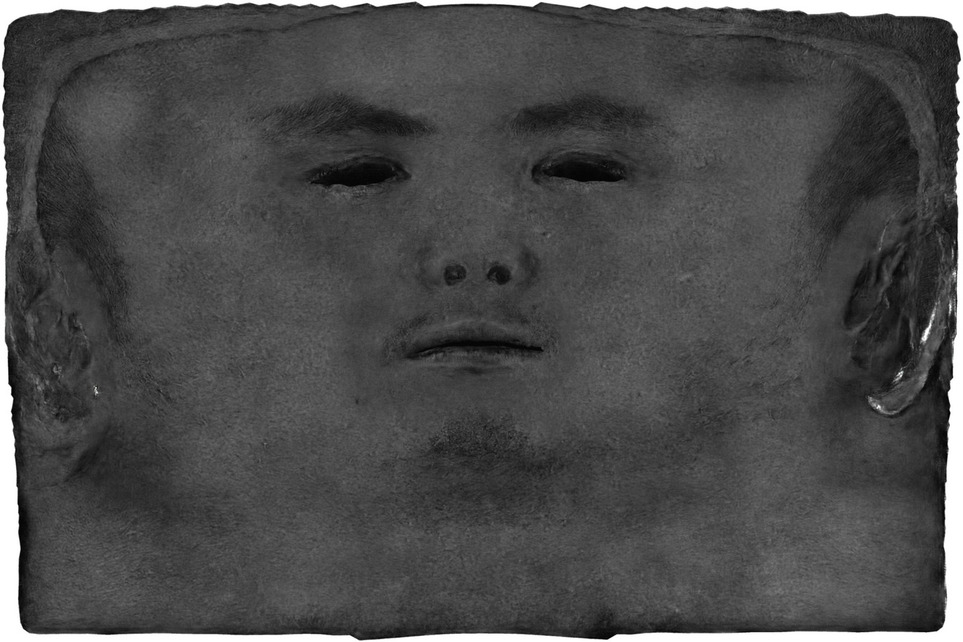}}
    \subfloat{
        \includegraphics[width=0.32\linewidth]{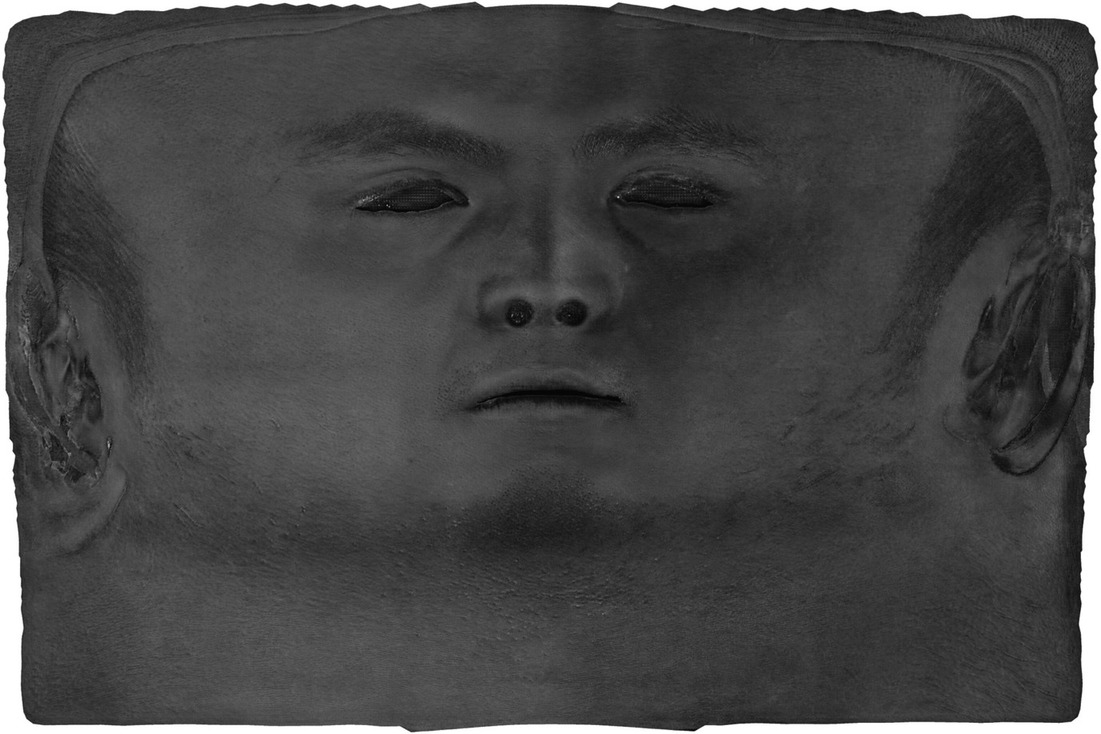}}
        
    \subfloat{
        \includegraphics[width=0.32\linewidth]{fig/results/others/sf_input.jpg}}
    \subfloat{
        \includegraphics[width=0.32\linewidth]{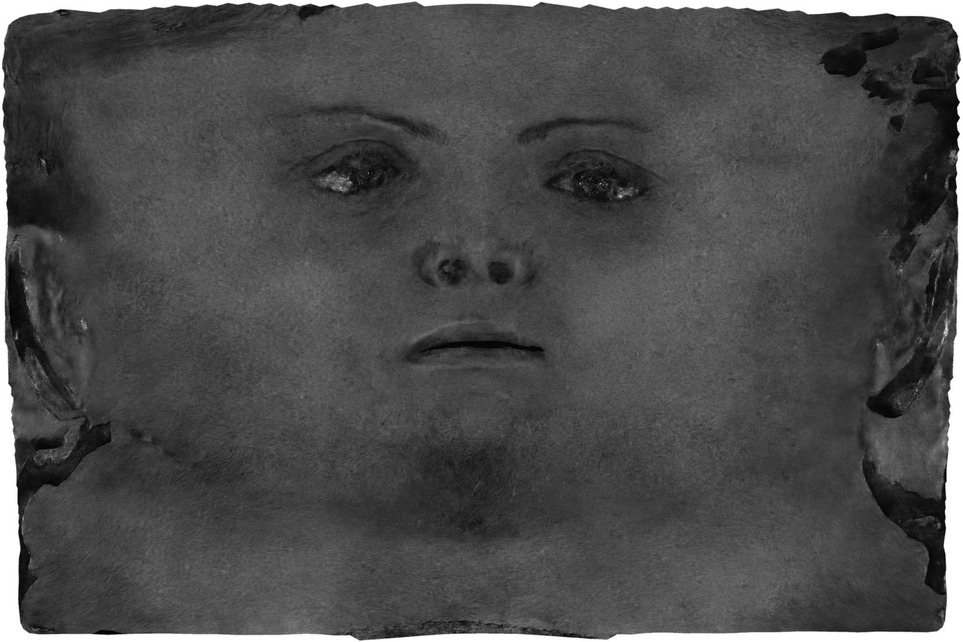}}
    \subfloat{
        \includegraphics[width=0.32\linewidth]{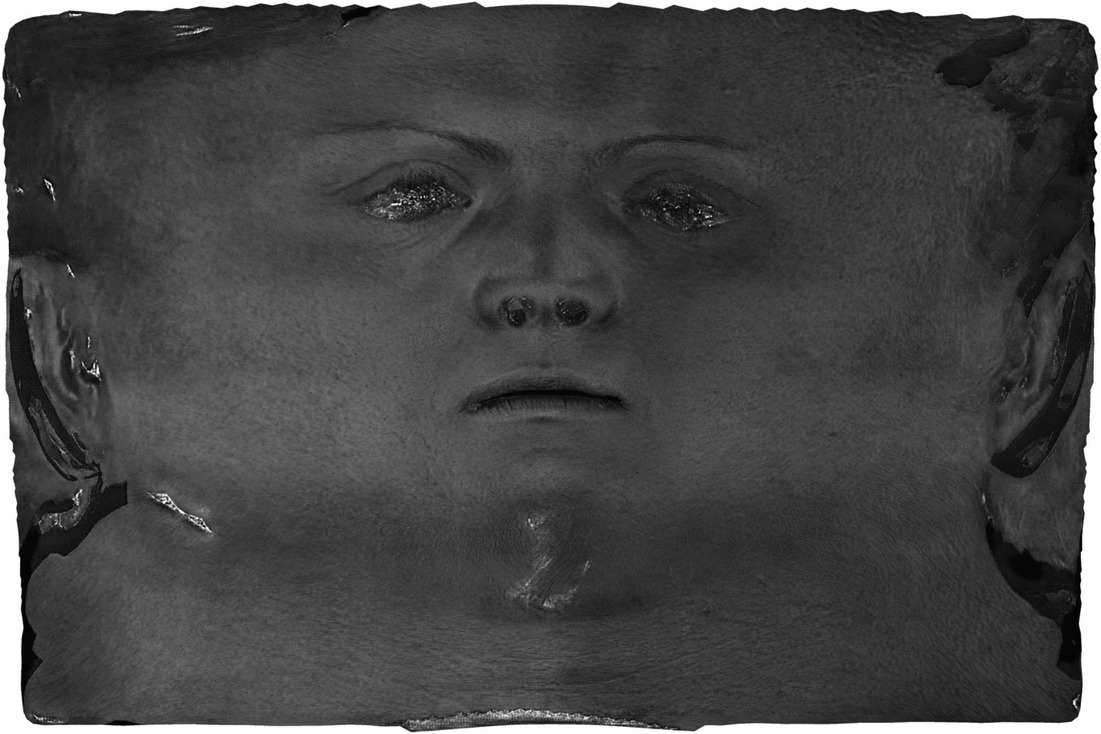}}
        
    \captionsetup[subfloat]{justification=centering}
    \subfloat[
        Input]{
        \includegraphics[width=0.32\linewidth]{fig/results/others/3dmd_alan_input.jpg}}
    \subfloat[Spec.~Alb. AvatarMe]{
        \includegraphics[width=0.32\linewidth]{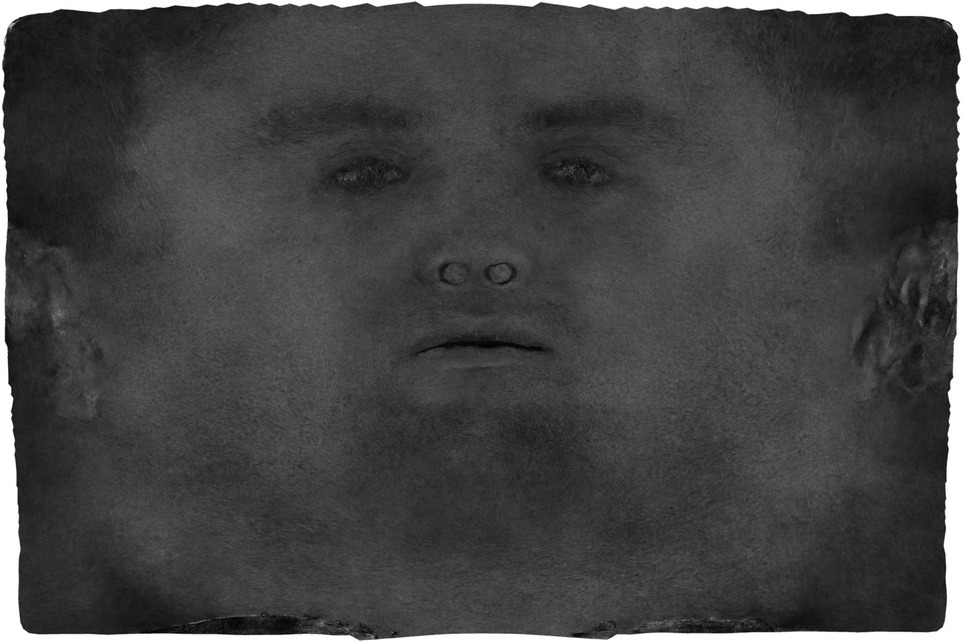}}
    \subfloat[Spec.~Alb. AvatarMe\textsuperscript{++}]{
        \includegraphics[width=0.32\linewidth]{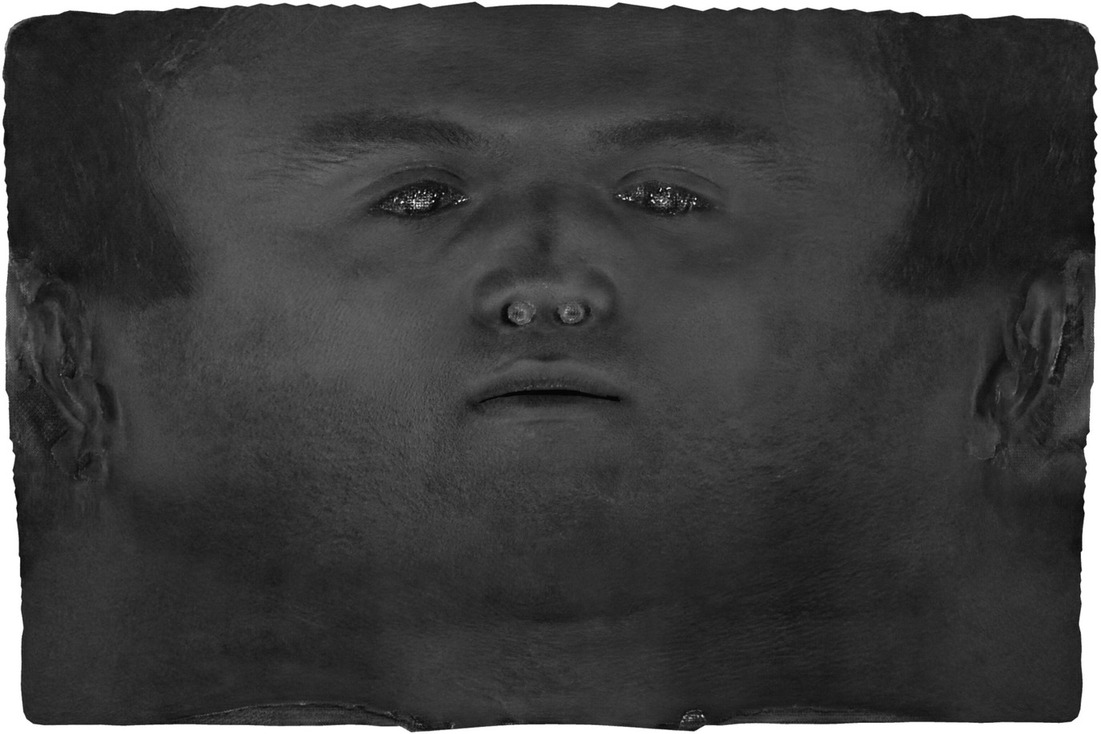}}
    
    \caption{
        Generalization of AvatarMe \cite{lattas_avatarme_2020}, compared to AvatarMe\textsuperscript{++}: Specular Albedo 
        from top to bottom:
        a) Reconstructed subject, with Facial Details Synthesis \cite{chen_photo-realistic_2019} proxy and texture,
        b) Reconstructed subject with OSTeC \cite{gecer2020ostec} fitting and texture completion,
        c) Captured subject from FaceScape \cite{yang_facescape_2020} dataset,
        d) Captured subject from Superface \cite{DBLP:conf/eccv/BerrettiBP12} dataset,
        e) Captured subject with a 3dMDface system (https://3dmd.com).
    }
    \label{fig:results_others_specAlb}
\end{figure}

\begin{figure}[h]
    \centering
    \captionsetup[subfigure]{labelformat=empty}
    \subfloat{
        \includegraphics[width=0.32\linewidth]{fig/results/others/chen_ouruv_004_inp.jpg}}
    \subfloat{
        \includegraphics[width=0.32\linewidth]{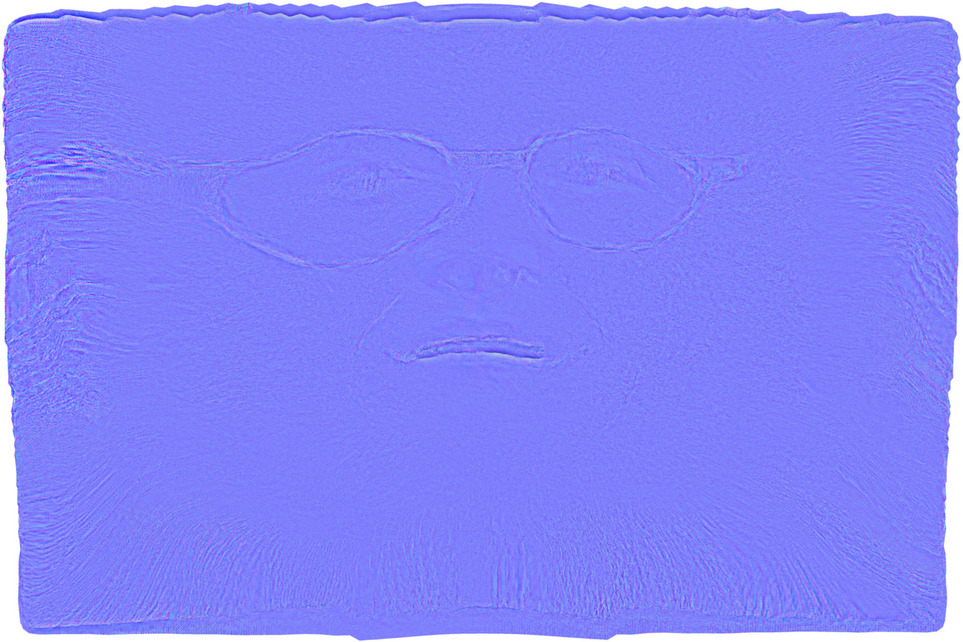}}
    \subfloat{
        \includegraphics[width=0.32\linewidth]{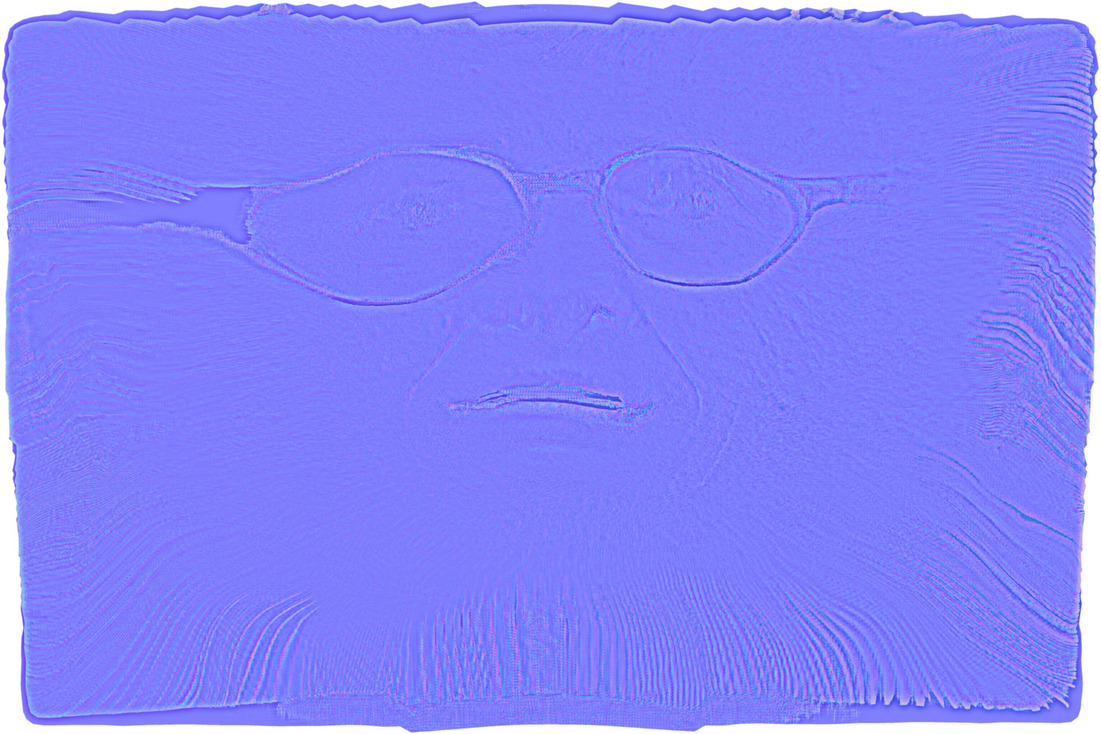}}
        
    \subfloat{
        \includegraphics[width=0.32\linewidth]{fig/results/others/ostec_im13_inp.jpg}}
    \subfloat{
        \includegraphics[width=0.32\linewidth]{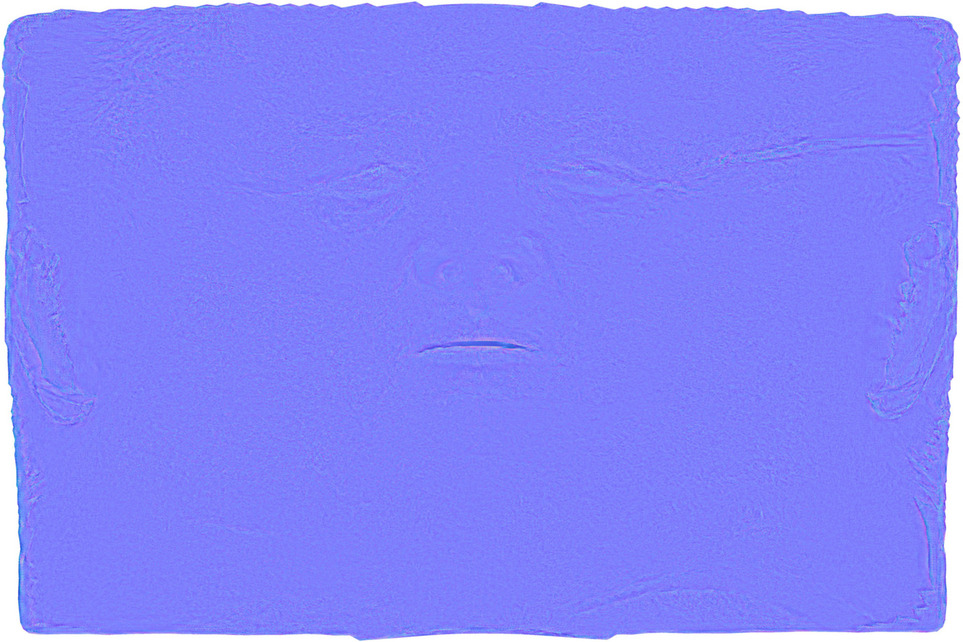}}
    \subfloat{
        \includegraphics[width=0.32\linewidth]{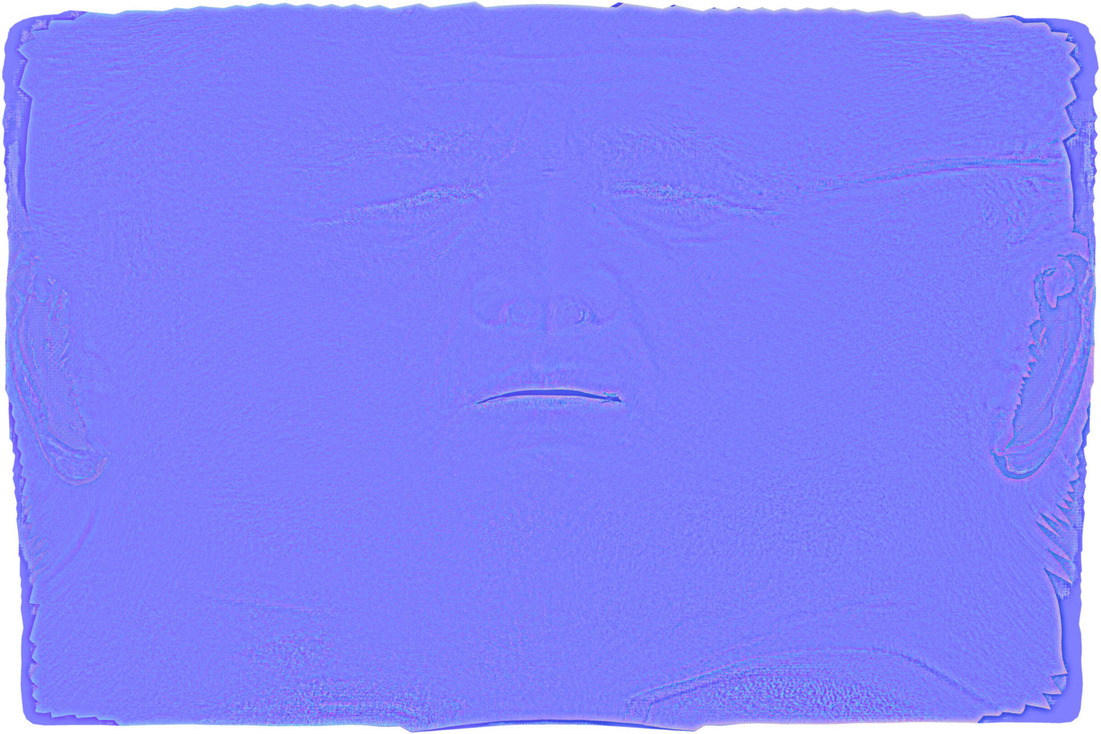}}
        
    \subfloat{
        \includegraphics[width=0.32\linewidth]{fig/results/others/facescape_inp.jpg}}
    \subfloat{
        \includegraphics[width=0.32\linewidth]{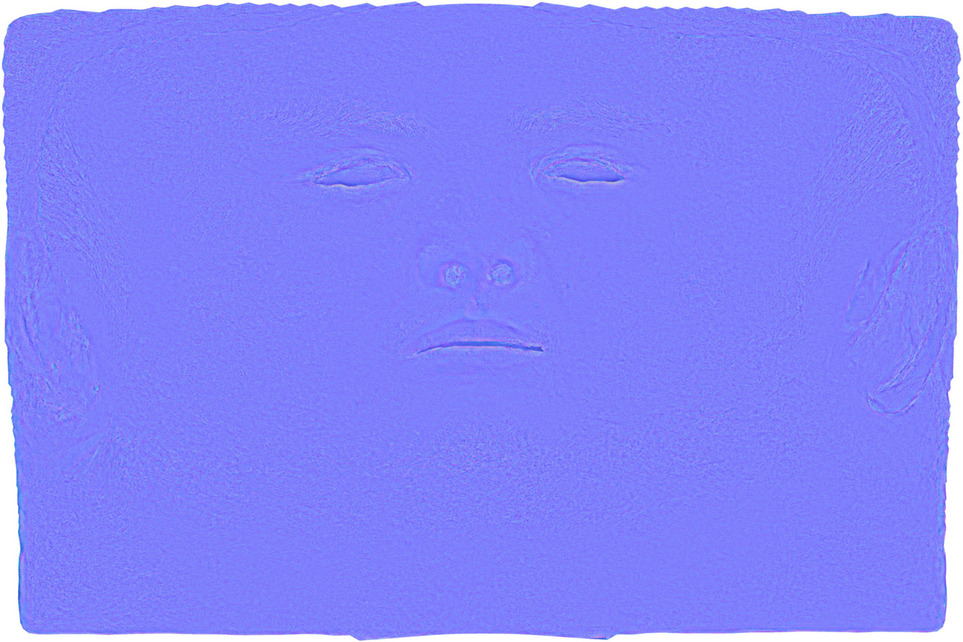}}
    \subfloat{
        \includegraphics[width=0.32\linewidth]{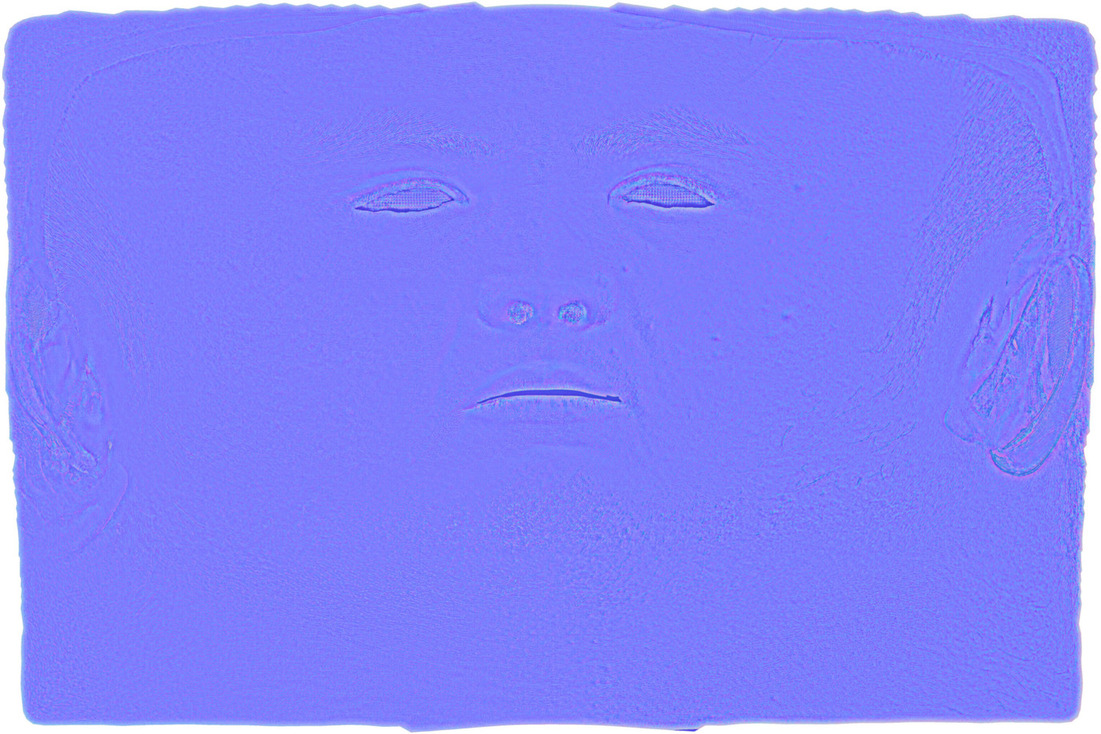}}   
        
    \subfloat{
        \includegraphics[width=0.32\linewidth]{fig/results/others/sf_input.jpg}}
    \subfloat{
        \includegraphics[width=0.32\linewidth]{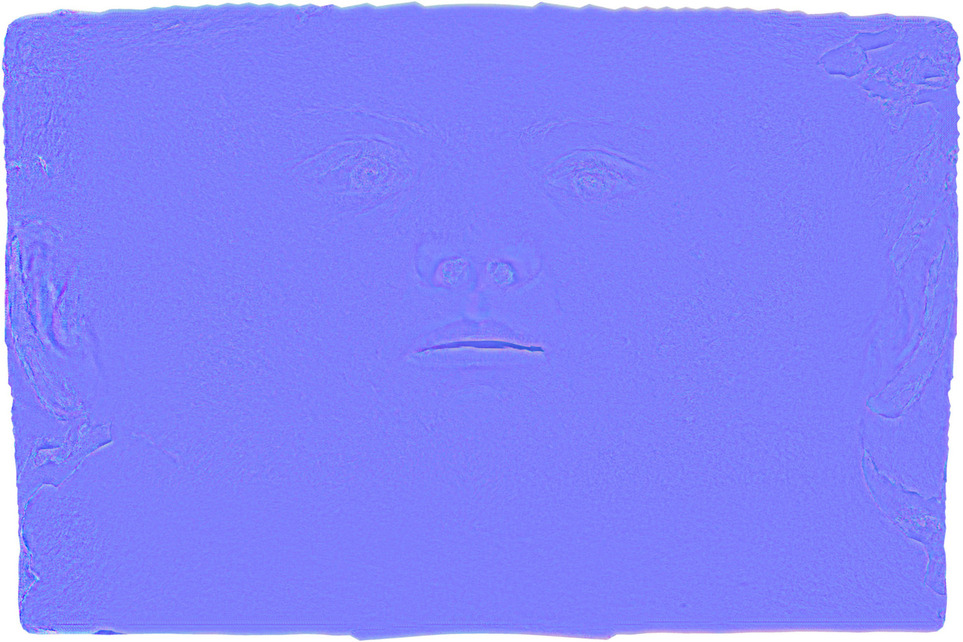}}
    \subfloat{
        \includegraphics[width=0.32\linewidth]{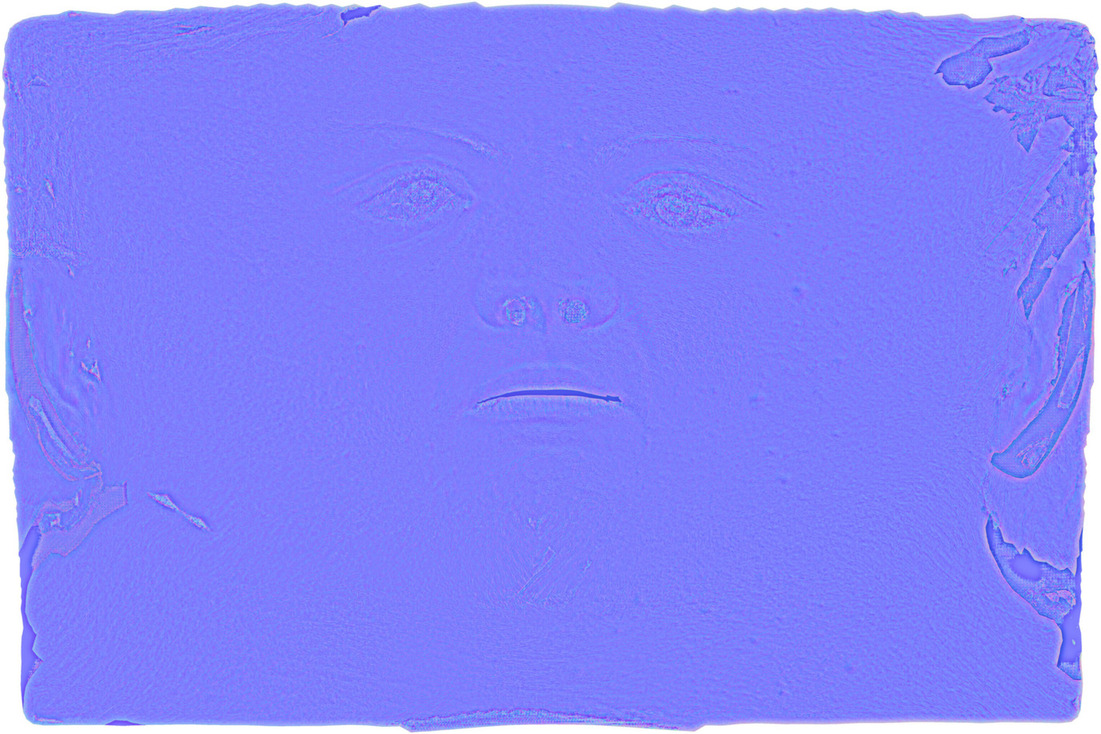}}
        
    \captionsetup[subfloat]{justification=centering}
    \subfloat[
        Input]{
        \includegraphics[width=0.32\linewidth]{fig/results/others/3dmd_alan_input.jpg}}
    \subfloat[Specular Normals (AvatarMe)]{
        \includegraphics[width=0.32\linewidth]{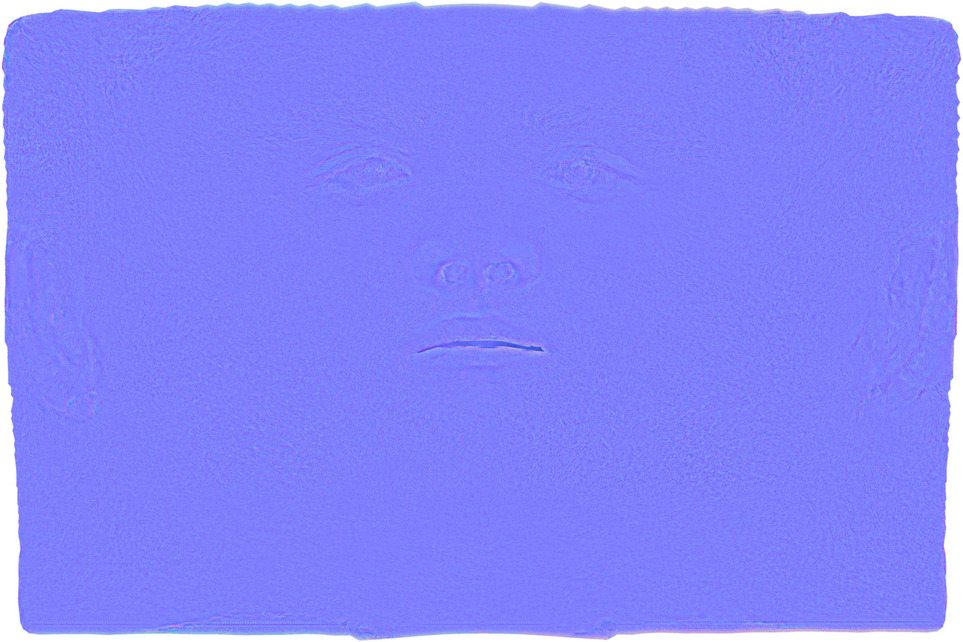}}
    \subfloat[Specular Normals (AvatarMe\textsuperscript{++})]{
        \includegraphics[width=0.32\linewidth]{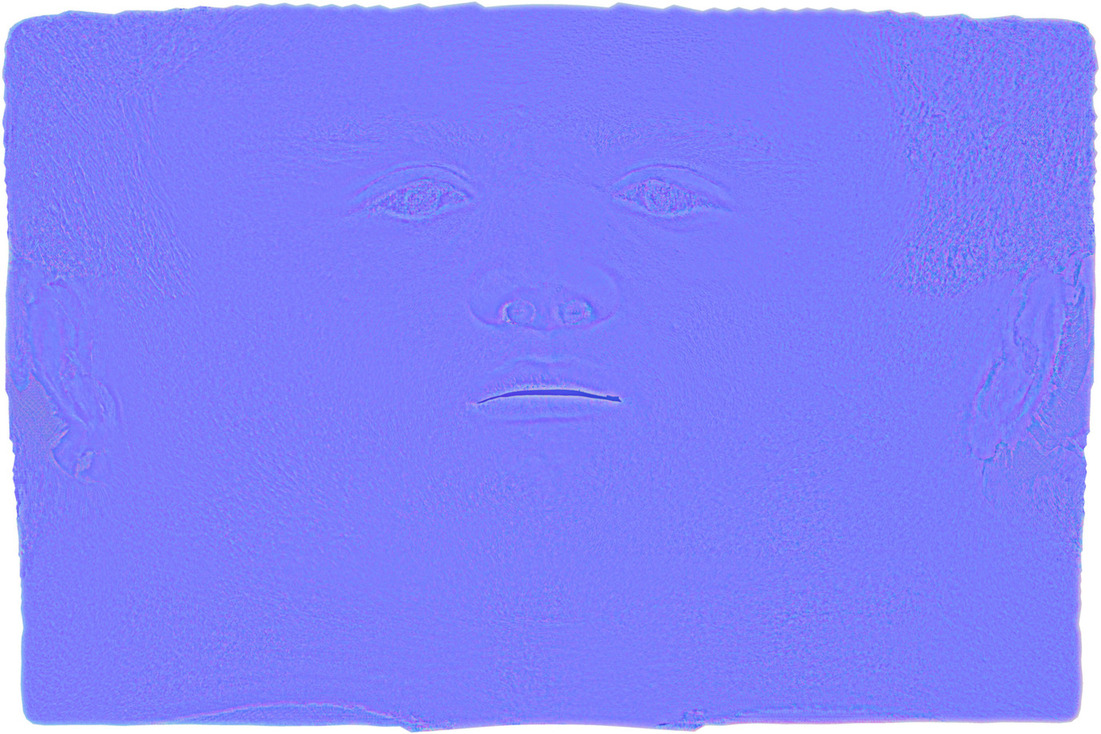}}
    
    \caption{
       Generalization of AvatarMe \cite{lattas_avatarme_2020}, compared to AvatarMe\textsuperscript{++}: Specular Normals (in tangent space) 
        from top to bottom:
        a) Reconstructed subject, with Facial Details Synthesis \cite{chen_photo-realistic_2019} proxy and texture,
        b) Reconstructed subject with OSTeC \cite{gecer2020ostec} fitting and texture completion,
        c) Captured subject from FaceScape \cite{yang_facescape_2020} dataset,
        d) Captured subject from Superface \cite{DBLP:conf/eccv/BerrettiBP12} dataset,
        e) Captured subject with a 3dMDface system (https://3dmd.com).
    }
    \label{fig:results_others_specNormals}
\end{figure}

Specifically, we acquire various textures with baked illumination
from different 3DMM fitting methods \cite{gecer2020ostec, chen_photo-realistic_2019},
datasets \cite{DBLP:conf/eccv/BerrettiBP12, yang_facescape_2020}
and the 3DMD capturing system.
We register them to the same template and transform the textures to our topology,
before feeding them to the final AvatarMe and AvatarMe\textsuperscript{++} networks,
used for the results in the main manuscript.
Please note that only a single network is trained for AvatarMe\textsuperscript{++}
and used for all examples on this document,
having a single target environment and the stochastic rendering loss,
as explained in our method section.
Fig.~\ref{fig:results_others_diffAlb} shows the comparison of generated diffuse albedo,
Fig.~\ref{fig:results_others_specAlb} shows the comparison of generated specular albedo and
Fig.~\ref{fig:results_others_specNormals} shows the comparison of generated specular normals.

\bibliographystyle{IEEEtran}
\bibliography{IEEEabrv,supplemental.bib}